\documentclass{article}

\usepackage{arxiv,times}

\usepackage[utf8]{inputenc} %
\usepackage[T1]{fontenc}    %
\usepackage{hyperref}       %
\usepackage{url}            %
\usepackage{booktabs}       %
\usepackage{amsfonts}       %
\usepackage{nicefrac}       %
\usepackage{microtype}      %
\usepackage{xcolor}         %

\usepackage{todonotes}

\usepackage{amsmath, amssymb, amsthm, thmtools}
\usepackage[nameinlink,capitalise]{cleveref}
\usepackage{bm}
\usepackage{tikz}
\usetikzlibrary {shapes.geometric}

\usepackage{amsmath,amsfonts,bm}

\def\eqref#1{equation~\ref{#1}}

\def\1{\bm{1}}

\def\rr{{\textnormal{r}}}

\def\vtheta{{\bm{\theta}}}

\def\ve{{\bm{e}}}

\def\vg{{\bm{g}}}

\def\vq{{\bm{q}}}

\def\vv{{\bm{v}}}
\def\vw{{\bm{w}}}
\def\vx{{\bm{x}}}

\def\vz{{\bm{z}}}

\def\mA{{\bm{A}}}
\def\mB{{\bm{B}}}
\def\mC{{\bm{C}}}
\def\mD{{\bm{D}}}
\def\mE{{\bm{E}}}

\def\mG{{\bm{G}}}
\def\mH{{\bm{H}}}
\def\mI{{\bm{I}}}

\def\mK{{\bm{K}}}

\def\mM{{\bm{M}}}

\def\mP{{\bm{P}}}
\def\mQ{{\bm{Q}}}
\def\mR{{\bm{R}}}
\def\mS{{\bm{S}}}
\def\mT{{\bm{T}}}
\def\mU{{\bm{U}}}
\def\mV{{\bm{V}}}
\def\mW{{\bm{W}}}
\def\mX{{\bm{X}}}

\def\mZ{{\bm{Z}}}

\def\mSigma{{\bm{\Sigma}}}

\DeclareMathAlphabet{\mathsfit}{\encodingdefault}{\sfdefault}{m}{sl}
\SetMathAlphabet{\mathsfit}{bold}{\encodingdefault}{\sfdefault}{bx}{n}
\newcommand{\tens}[1]{\bm{\mathsfit{#1}}}

\def\tQ{{\tens{Q}}}
\def\tR{{\tens{R}}}
\def\tS{{\tens{S}}}

\def\gL{{\mathcal{L}}}

\def\gN{{\mathcal{N}}}

\def\sN{{\mathbb{N}}}

\def\sP{{\mathbb{P}}}

\newcommand{\E}{\mathbb{E}}

\newcommand{\R}{\mathbb{R}}

\DeclareMathOperator{\Tr}{Tr}

\def\mGamma{{\bm{\Gamma}}}
\def\mDelta{{\bm{\Delta}}}

\def\vgamma{{\bm{\gamma}}}
\def\valpha{{\bm{\alpha}}}

\DeclareMathOperator{\Li}{Li}

\DeclareMathOperator*{\rank}{rank}
\newcommand{\oucp}{o_\sP^{\mathrm{ucp}}(1)}
\DeclareMathOperator*{\diag}{diag}

\newtheorem{theorem}{Theorem}[section]
\newtheorem{corollary}[theorem]{Corollary}

\newtheorem*{setup*}{Setup}
\newtheorem{lemma}[theorem]{Lemma}
\newtheorem{prop}[theorem]{Proposition}

\theoremstyle{definition}

\newtheorem{remark}[theorem]{Remark}

\title{%
A prism hierarchy of learning regimes in large linear autoencoders}
\author{%
  Eugene~Golikov\\
  Applied AI Institute\\
  Moscow, Russia \\
  \texttt{e.golikov@applied-ai.ru} \\
    \And
  Yaroslav~Gusev\\
  Applied AI Institute\\
  Moscow, Russia \\
  \texttt{i.gusev@applied-ai.ru} \\
    \And
  Dmitry Yarotsky\\
  Applied AI Institute \& \\ Steklov Mathematical Institute of Russian Academy of Sciences\\
  Moscow, Russia \\
  \texttt{yarotsky@gmail.com} \\
}

\begin{document}

\maketitle

\begin{abstract} 
Theoretical studies of machine learning models commonly consider different limiting regimes in which the learning dynamics of gradient descent becomes theoretically tractable. 
It is, however, desirable to have a systematically obtained picture of qualitatively different extreme learning regimes for a particular type of models. 
In this paper we propose such a picture for large weight-tied linear autoencoders characterized by input and latent dimensions, initialization magnitude, and training set size. 
This model is nonlinear in the weights and its gradient flow does not have a general theoretical solution. 
We show that at the level of the formal loss-expansion hierarchy, its extreme regimes are naturally associated with faces of a triangular prism. 
In particular, there are five basic extreme regimes associated with the 2-faces of the prism: (1) large-data, (2) small-data, (3) mean-field, (4) narrow-latent, and (5) free. 
For regimes (1,2,3,4), we derive and rigorously characterize the limiting train and population dynamics under gradient flow, rigorously prove convergence to these limits, and get good agreement with experimental results.
The resulting limits reveal spectral critical slowing, initialization-driven trapping away from the PCA optimum, and regimes in which population loss converges exponentially while training loss relaxes only algebraically.\end{abstract}

\section{Introduction}
\label{sec:intro}A common route to tractable learning dynamics is to take an asymptotic
limit in which model width, data size, or initialization scale in a
particular way. Different limits can simplify the same training problem
differently, including lazy-training linearizations
\citep{chizat2019lazy,jacot2018neural,lee2017deep}, mean-field limits
\citep{sirignano2020mean,rotskoff2018neural}, and solvable
linear-network dynamics
\citep{saxe2013exact,saxe2019mathematical,tu2024mixed}.
Such analyses usually select a regime first and then study its
dynamics. For a model with several scalable parameters, this leaves a
broader question:
\emph{What are its distinct extreme learning regimes, how are they
related, and what dynamics arise in each?}
Answering this directly is difficult because even a few hyperparameters
admit infinitely many joint large-system scalings.

We study this question for the weight-tied linear autoencoder
\citep{baldi1989neural,baldi2012autoencoders}. Although linear in its
input, the model is nonlinear in the weights and its gradient flow is
not generally explicitly solvable. Moreover, with a finite training
set, train and population performance can evolve differently. We let
the input dimension $p$, latent width $n$, training-set size $m$, and
initialization variance $\sigma^2$ scale jointly, and analyze both
train and population learning trajectories.

We identify five basic extreme regimes -- \emph{large-data},
\emph{small-data}, \emph{mean-field}, \emph{narrow-latent}, and
\emph{free} -- together with intermediate phases combining them. In four
of the five basic regimes we derive limiting train and population
dynamics and rigorously prove convergence on compact time intervals.
The resulting effective systems are qualitatively different: they range
from spectral and decoupled covariance dynamics to a nonlinear
alignment system and an infinite hierarchy of leakage moments. They
also reveal spectral critical slowing, initialization-dependent
trapping away from the PCA optimum, and parametrically different
relaxation rates of train and population losses.

To organize the possible scalings systematically, we build on the
coefficient-expansion viewpoint of \citet{yarotsky2026gradient}, in
which regimes are distinguished by dominant terms in formal time
expansions. We make the associated phase structure explicit through
leading-support signatures across expansion orders and extend the
classification jointly to train and population losses. For the
autoencoder, the resulting leading supports form triangular prisms,
whose normal fan gives a dual triangular-bipyramid phase diagram.
Thus the formal classification identifies the basic extreme regimes
before their trajectories are solved, while the limiting analyses show
that these regimes correspond to genuinely different learning
dynamics.

\paragraph{Our contributions.}
\begin{enumerate}

\item We derive and rigorously justify limiting train and population
dynamics in four basic extreme regimes:
\emph{large-data}, \emph{small-data}, \emph{mean-field}, and
\emph{narrow-latent} (\cref{sec:solutions}). The limits range from
explicit spectral flows to a nonlinear alignment system and an
infinite moment hierarchy. Numerical experiments on CIFAR-10 agree well
with the corresponding finite-dimensional predictions
(\cref{fig:exp-main,fig:narrow-latent}).

\item The limiting solutions expose several distinct mechanisms:
spectral critical slowing when relevant dimensions become comparable,
initialization-dependent alignment or trapping away from the PCA
optimum, and strong separation between train and population relaxation
rates. In particular, in narrow-latent alignment the population
residual can decay exponentially while the training residual decays
only algebraically (\cref{sec:solutions,sec:discussion}).

\item Building on the dominant-term expansion viewpoint of
\citet{yarotsky2026gradient}, we formalize the phase structure through
leading-support signatures and extend the classification jointly to
train and population losses. For the weight-tied linear autoencoder,
this yields a triangular-prism hierarchy, equivalently a dual
triangular-bipyramid phase diagram with five basic extreme regimes and
additional phases combining them
(\cref{sec:classification,fig:prism}).

\end{enumerate}

\section{Problem statement}
\label{sec:problem}

We consider a shallow linear weight-tied autoencoder with \emph{input dimension} $p$ and \emph{latent dimension} $n$:
$f_{\mU}(\vx) = \mA \vx$, where $\vx \in \R^p$ is the input, $\mA = \mU^\top \mU$ is the model linear map, and $\mU \in \R^{n \times p}$ is the weight matrix we train.
Consider a training set $\mX \in \R^{p \times m}$ of size $m$ and define the \emph{train} square loss for this model:
\begin{equation}\label{eq:loss_def}
  \widehat L(\mU)
  = \frac{\left\|\mX - \mU^\top \mU \mX\right\|_F^2}{2 p m}
  = \frac{\Tr\left[(\mI - \mA) \mS (\mI - \mA)\right]}{2p},
  \qquad
  \mS
  = \frac{1}{m} \mX \mX^\top \in \R^{p \times p}.
\end{equation}
In order to define the \emph{population} loss, we assume data isotropy:
\begin{equation}
  \mSigma:=\E[\vx\vx^\top]=\mI;
  \label{eq:pop-isotropy-main}
  \qquad
  L(\mU)
  = \E_{\vx} \left[\frac{\left\|\vx - \mU^\top \mU \vx\right\|^2}{2 p}\right]
  = \frac{\left\|\mI - \mA\right\|_F^2}{2 p}.
\end{equation}
We train our model with gradient flow with learning rate $\eta$:
\begin{equation}\label{eq:gf}
  \frac{d\mU}{dt}
  = -\eta \frac{\partial \widehat L(\mU)}{\partial\mU},
  \qquad
  \mU(0) 
  \sim \gN(0, \sigma^2 \mI_n \otimes \mI_p).
\end{equation}
Both $\eta$ and $\sigma^2$ may depend on $p$, $m$, $n$.
Define the loss values at time $t$: $\widehat L(t) = \widehat L(\mU(t))$, $L(t) = L(\mU(t))$.
We will be interested in their evolution in the limit of large $p,n,m$.
As we shall see in \cref{sec:solutions}, this evolution depends qualitatively on the \emph{joint} scaling of $p$, $n$, $m$, and $\sigma^2$, suggesting existence of several \emph{learning regimes} in the high-dimensional limit.
In \cref{sec:classification}, we identify five basic learning regimes, each associated with a face of a triangular prism.
In \cref{sec:solutions}, we derive explicit finite-time limiting dynamics in four of these regimes and rigorously prove convergence to these limits.
The corresponding solutions have very different nature, therefore deserve to be regarded as genuine learning regimes.

\section{A formal phase diagram of learning regimes (\ref{sec:appdiag})}
\label{sec:classification}

\paragraph{General framework.}
There is no established, canonical way to define a phase diagram for the
learning dynamics of a model with several jointly scaled hyperparameters. One can
identify particular solvable limits, but this does not by itself specify
which limits should count as different phases or whether other regimes have
been missed. We propose an operational definition based on the Taylor
coefficients of averaged learning observables at initialization. It builds
on the diagrammatic expansion method of \citet{yarotsky2026gradient}, but
makes the equivalence relation defining phases explicit and allows several
observables, such as train and population losses, to be classified jointly.

Here is the general setting. Fix averaged observables
$F_\ell(t;\vtheta)$ and positive hyperparameters
$\vtheta=(\theta_1,\ldots,\theta_d)$. Their Taylor
coefficients at initialization are the derivatives
$D_{\ell,s}(\vtheta)
=\left.\frac{d^s}{dt^s}F_\ell(t;\vtheta)\right|_{t=0}$.
We say that these observables are in the \emph{Laurent-polynomial
coefficient setting} if, at every positive order $s\geq1$,
\begin{equation}\label{eq:polynomial-setting}
D_{\ell,s}(\vtheta)
=\sum_{\vq\in Q_{\ell,s}}
c_{\ell,s,\vq}\vtheta^{\vq},
\qquad Q_{\ell,s}\subset\mathbb Z^d\text{ finite},\quad
c_{\ell,s,\vq}\ne0.
\end{equation}
Negative integer powers are permitted, as in factors $p^{-s}m^{-s}$
arising from loss normalization. We omit order zero so that adding a
constant to an observable does not change its learning phase. The
condition concerns derivatives at the chosen reference time and does not
assume that the infinite Taylor series converges.

Let $a\to\infty$ and scale the hyperparameters as
$\theta_i(a)\sim\kappa_i a^{\alpha_i}$ with $\kappa_i>0$. The
scaling vector $\valpha=(\alpha_1,\ldots,\alpha_d)$ assigns
growth exponent $\vq\cdot\valpha$ to the monomial
$\vtheta^{\vq}$. We define the \emph{leading support}
at order $s$ to be the set attaining the largest such exponent:
\begin{equation}\label{eq:phase-support}
S_{\ell,s}(\valpha)
=\operatorname*{Argmax}_{\vq\in Q_{\ell,s}}
\vq\cdot\valpha.
\end{equation}
Geometrically, this is the support on the face of
$\operatorname{conv}(Q_{\ell,s})$ exposed by $\valpha$.
We call the collection $(S_{\ell,s}(\valpha))_{\ell,s\geq1}$ the
\emph{leading-support signature}, and put two scaling vectors in the same
\emph{formal phase} exactly when their signatures agree. Agreement through
any fixed order $K$ need not imply agreement at higher orders, so a
finite truncation does not generally determine the phase. Because we
retain monomials rather than the value of their sum,
the definition also applies when tied leading terms cancel for special
choices of the constants $\kappa_i$.

We call a boundary between adjacent formal phases \emph{persistent} if their
leading supports differ at infinitely many orders $s$. Equivalently,
however many initial coefficients one discards, some later coefficients
still select different monomials. Such a difference cannot be attributed
to finitely many exceptional orders; it is an essential distinction
between the phases at the level of their formal expansions.

The map from continuous scaling vectors to leading-support signatures
turns the classification into a discrete combinatorial problem. 
For a fixed finite realization, an analytic gradient-flow trajectory is locally determined by its full Taylor series at initialization, and this local solution determines the trajectory wherever it can be analytically continued along real time.
This motivates considering the entire coefficient
sequence rather than only the initial slope. For averaged losses,
however, the formal series need not converge, and coefficient-wise
large-system limits need not commute with summation or analytic
continuation. Directly constructing and comparing limiting trajectories
can therefore be difficult even for the present linear autoencoder,
whereas each finite-order coefficient can be analyzed exactly. Our
signature diagram offers a systematic classification at that level and
suggests distinctions relevant to the dynamics, but we do not claim it
exhausts trajectory-level regimes. In particular, long-time behavior
may require separate analysis, as in \cref{sec:solutions}.

\paragraph{Autoencoder phase diagram.}
For the phase-diagram calculation, we specialize the model of \cref{sec:problem} to i.i.d. standard Gaussian training data.
Our observables are
the averaged train loss $F_{\rm tr}(t)=\mathbb E[\widehat L(t)]$ and
population loss $F_{\rm pop}(t)=\mathbb E[L(t)]$. The five positive
hyperparameters are $\vtheta=(p,n,m,\sigma,\eta)$, with scaling exponents
$\valpha=(\alpha_p,\alpha_n,\alpha_m,\alpha_\sigma,
\alpha_\eta)$ and $\alpha_p,\alpha_n,\alpha_m>0$. The learning rate
$\eta$ multiplies the whole gradient-flow vector field:
$F_\ell(t;p,n,m,\sigma,\eta)
=F_\ell(\eta t;p,n,m,\sigma,1)$.
Consequently, every order-$s$ coefficient has the same factor $\eta^s$;
$\alpha_\eta$ changes the clock but not the leading-support signature; thus $\eta$ does not affect the phase.

We will use three convenient ratios:
$\phi=p/m$ (input-to-sample), $\rho=n\sigma^2$ (relative initialization
magnitude), and $\psi=p/n$ (input-to-hidden). Their scaling exponents are
\begin{equation}\label{eq:phase-effective-exponents}
\vgamma=(\alpha_\phi,\alpha_\rho,\alpha_\psi)
=\bigl(\alpha_p-\alpha_m,\
\alpha_n+2\alpha_\sigma,\ \alpha_p-\alpha_n\bigr).
\end{equation}
Recall that the \emph{normal fan} of a polytope partitions scaling directions
according to which face maximizes a linear score on that polytope.

\begin{theorem}[Autoencoder phase diagram]\label{th:prism}
Assume that the training matrix $\mX \in \mathbb{R}^{p\times m}$ has
i.i.d. standard Gaussian entries, and that the initialization is Gaussian
as in \cref{eq:gf}. Then, at every order $s\geq1$, the averaged train and population losses have
Laurent-polynomial coefficients in $\vtheta$. Their joint
formal phase depends only on $\vgamma$ and is the normal fan of
a triangular prism. Equivalently, the ray from $\boldsymbol 0$ through
each nonzero $\vgamma$ meets the boundary of the dual triangular
bipyramid;
its phase is determined by the unique face whose relative interior
contains that intersection point. The bipyramid has five vertices, nine
edges, and six triangular faces; the balanced point
$\vgamma=\boldsymbol 0$ represents the remaining phase.
The five vertex rays are:
\begin{equation}\label{eq:phase-five-rays}
\begin{array}{ll}
\alpha_\phi<0,\quad \alpha_\rho=\alpha_\psi=0
  & \text{(``Large-data'')},\\
\alpha_\psi<0,\quad \alpha_\phi=\alpha_\rho=0
  & \text{(``Mean-field'')},\\
\alpha_\rho<0,\quad\alpha_\psi=-\alpha_\rho,\quad\alpha_\phi=0
  & \text{(``Narrow-latent'')},\\
\alpha_\phi>0,\quad \alpha_\rho=\alpha_\psi=0
  & \text{(``Small-data'')},\\
\alpha_\rho>0,\quad \alpha_\phi=\alpha_\psi=0
  & \text{(``Free'')}.
\end{array}
\end{equation}
Every transition between adjacent phases is persistent.
\end{theorem}

\begin{figure}[t]
\centering
\resizebox{\linewidth}{!}{%
\begin{tikzpicture}[
    line cap=round, line join=round,
    front edge/.style={draw=black!65,line width=0.65pt},
    back edge/.style={draw=black!55,densely dashed,line width=0.55pt},
    phase arrow/.style={->,line width=0.95pt,>=stealth},
    phase vertex/.style={circle,inner sep=1.7pt},
    large/.style={blue!70!black}, small/.style={orange!85!black},
    mean/.style={teal!75!black}, narrow/.style={violet!75!black},
    free/.style={red!75!black}
]
\begin{scope}[shift={(2.75cm,0cm)},
              x={(0.992cm,0cm)},y={(0.273cm,0.641cm)},
              z={(-0.210cm,0.833cm)}]
  \coordinate (O) at (-2,-1,-1); \coordinate (U) at (1,2,-1);
  \coordinate (V) at (1,-1,-1);  \coordinate (Ot) at (-2,-1,1);
  \coordinate (Ut) at (1,2,1);  \coordinate (Vt) at (1,-1,1);
  \fill[blue!5] (O)--(V)--(Vt)--(Ot)--cycle;
  \fill[blue!7] (Ot)--(Ut)--(Vt)--cycle;
  \fill[orange!6] (U)--(V)--(Vt)--(Ut)--cycle;
  \draw[back edge] (O)--(U);
  \draw[front edge] (O)--(V)--(U)--(Ut)--(Ot)--(O);
  \draw[front edge] (Ot)--(Vt)--(Ut);
  \draw[front edge] (Vt)--(V);

  \foreach \rr in {0,1,2} {
    \foreach \uu in {0,1,2,3} {
      \foreach \vv in {0,1,2,3} {
        \pgfmathtruncatemacro{\uvsum}{\uu+\vv}
        \ifnum\uvsum<4
          \ifnum\rr=0
            \filldraw[fill=white,draw=black!65,line width=0.35pt]
              ({\uu+\vv-2},{\uu-1},{\rr-1}) circle (1.0pt);
          \else
            \fill[black!85]
              ({\uu+\vv-2},{\uu-1},{\rr-1}) circle (1.05pt);
          \fi
        \fi
      }
    }
  }

  \draw[phase arrow,large] (0,0,1)--(0,0,1.75)
    node[above,font=\scriptsize,text=black] {Large-data};
  \draw[small,densely dashed,line width=0.9pt]
    (0,0,-1)--(0,0,-1.77);
  \draw[phase arrow,small] (0,0,-1.77)--(0,0,-2.35)
    node[below,font=\scriptsize,text=black] {Small-data};
  \draw[phase arrow,mean] (-.5,-1,0)--(-.5,-2.5,0)
    node[left,font=\scriptsize,text=black] {Mean-field};
  \draw[narrow,densely dashed,line width=0.9pt]
    (-.5,.5,0)--(-1.43,1.43,0);
  \draw[phase arrow,narrow] (-1.43,1.43,0)--(-1.7,1.7,0)
    node[above left,font=\scriptsize,text=black] {Narrow-latent};
  \draw[phase arrow,free] (1,.5,0)--(1.85,.5,0)
    node[right,font=\scriptsize,text=black] {Free};
\end{scope}

\begin{scope}[shift={(9.15cm,0cm)},
              x={(1.606cm,0cm)},y={(0.442cm,1.038cm)},
              z={(-0.340cm,1.349cm)}]
  \coordinate (L) at (0,0,1.5);   \coordinate (S) at (0,0,-1.2);
  \coordinate (M) at (0,-1.3,0);  \coordinate (N) at (-1.3,1.3,0);
  \coordinate (F) at (1.3,0,0);
  \fill[blue!8] (L)--(N)--(M)--cycle;
  \fill[blue!4] (L)--(M)--(F)--cycle;
  \fill[orange!9] (S)--(M)--(F)--cycle;
  \draw[black!45,->,line width=0.5pt] (-1.48,0,0)--(1.48,0,0)
    node[above right=-1pt,font=\scriptsize] {$\alpha_\rho$};
  \draw[black!45,->,line width=0.5pt] (0,-1.52,0)--(0,1.6,0)
    node[above right=-1pt,font=\scriptsize] {$\alpha_\psi$};
  \draw[black!45,->,line width=0.5pt] (0,0,1.5)--(0,0,-1.52)
    node[right=2pt,font=\scriptsize] {$\alpha_\phi$};
  \draw[black!45,densely dotted,line width=0.5pt] (0,0,0)--(N);
  \draw[back edge] (S)--(N);
  \draw[back edge] (N)--(F);
  \draw[front edge] (L)--(N)--(M)--(L)--(F)--(M)--(S)--(F);
  \node[phase vertex,fill=blue!70!black] at (L) {};
  \node[phase vertex,fill=orange!85!black] at (S) {};
  \node[phase vertex,fill=teal!75!black] at (M) {};
  \node[phase vertex,fill=violet!75!black] at (N) {};
  \node[phase vertex,fill=red!75!black] at (F) {};
  \node[circle,draw=black!75,fill=white,inner sep=1.3pt] at (0,0,0) {};
  \node[font=\scriptsize,above=4pt] at (L) {Large-data};
  \node[font=\scriptsize,below left=4pt] at (S) {Small-data};
  \node[font=\scriptsize,below left=3pt] at (M) {Mean-field};
  \node[font=\scriptsize,left=4pt] at (N) {Narrow-latent};
  \node[font=\scriptsize,below right=4pt] at (F) {Free};
  \node[font=\scriptsize,anchor=west,fill=white,inner sep=1pt]
    at (.1,.12,0) {Balanced};
\end{scope}
\end{tikzpicture}
}
\caption{The autoencoder phase diagram in scaling coordinates
$\vgamma=(\alpha_\phi,\alpha_\rho,\alpha_\psi)$ (\textbf{right}),
and the dual triangular prisms formed by leading monomial supports $\cup_{\valpha} S_{\text{tr},s(\boldsymbol(\alpha))}, \cup_{\valpha} S_{\text{pop},s}(\boldsymbol(\alpha)) $
at Taylor order $s=2$ (\textbf{left}). The five colored face normals correspond
to the five labeled vertex rays. Hollow dots on the lowest layer occur
only for the train loss; the prism for the population loss begins one
layer higher.}\label{fig:prism}
\end{figure}

\paragraph{Proof sketch.}
Repeated differentiation of gradient flow and Gaussian Wick contraction
give the Laurent-polynomial coefficients. For $s\geq2$, their potentially
leading monomials are indexed by integers
$u,v\geq0$, $u+v\leq s+1$, and $r\in\{0,\ldots,s\}$ for the train loss,
or $r\in\{1,\ldots,s\}$ for the population loss. Up to a term constant
across these monomials, the exponent for $\mathbf k=(u,v,r)$ is
$\mathbf k\cdot\mathbf w$, where
$\mathbf w=(\alpha_\rho+\alpha_\psi,\alpha_\rho,-\alpha_\phi)$.
Equivalently, the affine coordinates $\mathbf x=(u+v,u,r)$ turn this
exponent into $\alpha_\rho x_1+\alpha_\psi x_2-\alpha_\phi x_3$.
The index sets are triangular prisms, all with the same
normal fan (see Figure \ref{fig:prism}). Distinct phase cells select different monomials at every
$s\geq2$, yielding persistent transitions.
See the full proof in \cref{sec:appdiag}.
\qed

Figure \ref{fig:prism} shows the phase diagram.  \Cref{tab:regime-summary} expresses the five vertex regimes in terms of
the original scales of $m,p,n,\sigma^2$ and summarizes their limiting
dynamics (see \cref{sec:solutions}). The relations explain the names: ``Large-data'' and
``Small-data'' compare $m$ with $p$; ``Mean-field'' and ``Narrow-latent''
highlight different latent-width scalings; in ``Free'', the initialized
model dominates the target.

Other phases combine these basic extremes. An open bipyramid edge
interpolates between its two endpoint rays and selects the common edge
of their prism facets; an open triangular face combines three vertex
rays and selects their common prism vertex. At the balanced center, all
monomials of the corresponding prisms tie. The ``Large-data'' and ``Small-data'' regimes are opposite to each other; this is the only opposite pair among the five vertex regimes.

\section{Limiting solutions in extreme regimes (\ref{app:solutions})}
\label{sec:solutions}

\begin{figure}
  \centering
  \includegraphics[width=0.24\linewidth]{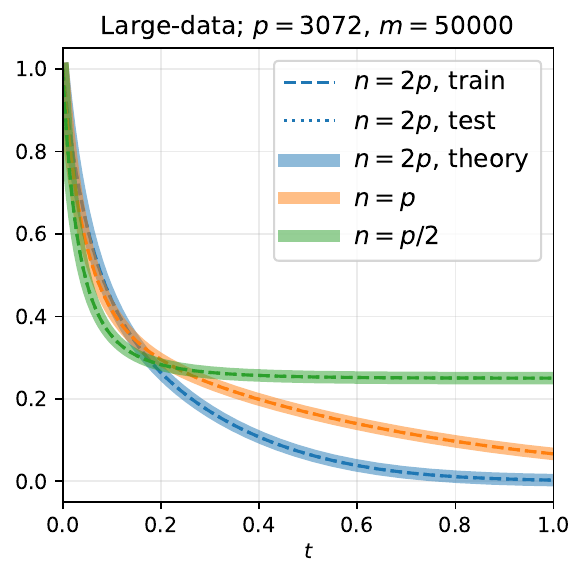}
  \includegraphics[width=0.24\linewidth]{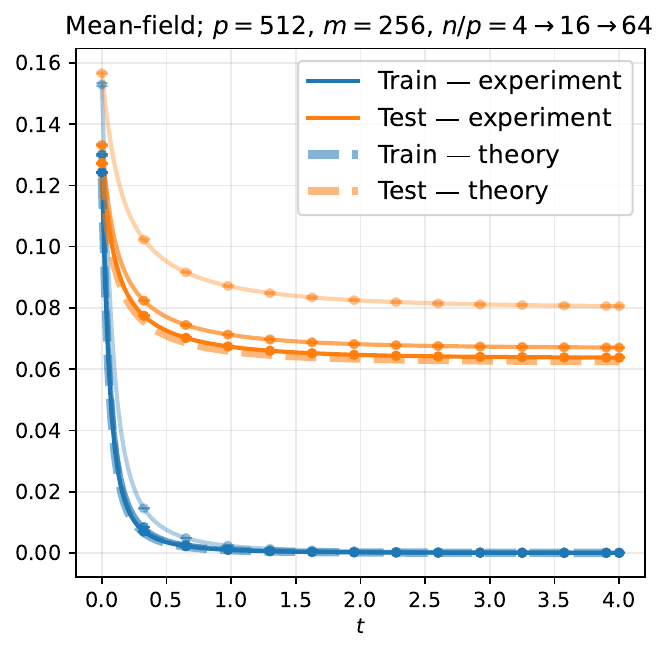}
  \includegraphics[width=0.24\linewidth]{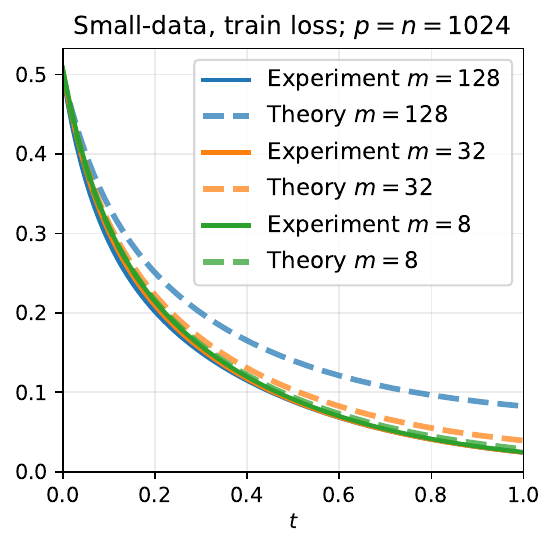}
  \includegraphics[width=0.24\linewidth]{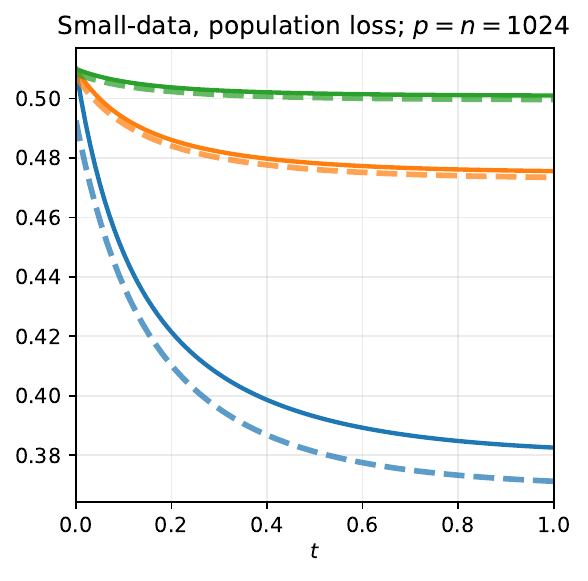}
  \caption{
    Empirical validation of the limiting predictions (\cref{sec:solutions}) across different regimes on subsets of CIFAR-10.
    See \ref{app:experiments} for experimental details.
    \emph{Narrow-latent} is isolated on \cref{fig:narrow-latent}.
    \label{fig:exp-main}
  } 
\end{figure}
\begin{table}[t]
  \caption{
    Five basic regimes, ordered by complexity from lowest (top row) to highest (bottom row).
  }
  \label{tab:regime-summary}
  \centering
  \footnotesize
  \setlength{\tabcolsep}{3.5pt}
  \renewcommand{\arraystretch}{1.15}
  \begin{tabular}{@{}llll@{}}
    \toprule
    \textbf{Regime}
    & \textbf{Scaling}
    & \textbf{Limiting dynamics}
    & \textbf{Solution status} \\
    \midrule

    Large-data
    &
    \begin{tabular}[t]{@{}l@{}}
      $m\gg p\asymp n\asymp \sigma^{-2}$;\\
      $\eta=p$
    \end{tabular}
    &
    \begin{tabular}[t]{@{}l@{}}
      Spectral logistic flow of $\mA$\\
      over MP law $\mu_{p/n}$
    \end{tabular}
    &
    \begin{tabular}[t]{@{}l@{}}
      Closed-form formula via\\
      MP generating function
    \end{tabular}
    \\
    \midrule

    Mean-field
    &
    \begin{tabular}[t]{@{}l@{}}
      $n\asymp\sigma^{-2}\gg p\asymp m$;\\
      $\eta=p$
    \end{tabular}
    &
    \begin{tabular}[t]{@{}l@{}}
      Decoupled empirical covariance\\
      eigenmodes; average over $\nu_{p/m}$
    \end{tabular}
    &
    \begin{tabular}[t]{@{}l@{}}
      Closed-form integral formula;\\
      large-time asymptotics
    \end{tabular}
    \\
    \midrule

    Narrow-latent
    &
    \begin{tabular}[t]{@{}l@{}}
      $n\ll \sigma^{-2}\asymp p\asymp m$;\\
      $\eta=p$
    \end{tabular}
    &
    \begin{tabular}[t]{@{}l@{}}
      Row-decoupled dynamics for\\
      row-norm $a$ and data alignment $r$
    \end{tabular}
    &
    \begin{tabular}[t]{@{}l@{}}
      Finite-dimensional implicit ODE;\\
      explicit asymptotics
    \end{tabular}
    \\
    \midrule

    Small-data
    &
    \begin{tabular}[t]{@{}l@{}}
      $m\ll p\asymp n\asymp \sigma^{-2}$;\\
      $\eta=m$
    \end{tabular}
    &
    \begin{tabular}[t]{@{}l@{}}
      Active/inactive block dynamics\\
      with leakage moments $a, \{\mu_k\}_{k\geq 1}$
    \end{tabular}
    &
    \begin{tabular}[t]{@{}l@{}}
      Infinite moment hierarchy;\\
      numerical truncation
    \end{tabular}
    \\
    \midrule

    Free
    &
    \begin{tabular}[t]{@{}l@{}}
        $\sigma^{-2}\ll n\asymp p\asymp m$;\\
        $\eta = \sigma^{-2}$
    \end{tabular}
    &
    \begin{tabular}[t]{@{}l@{}}
      Target term negligible; dynamics\\
      approximately deflate initial model
    \end{tabular}
    &
    \begin{tabular}[t]{@{}l@{}}
      Classified by the prism;\\
      trajectory solution open
    \end{tabular}
    \\
    \bottomrule
  \end{tabular}
\end{table}

We present explicit solutions for both train and population losses in regimes corresponding to four out of five two-dimensional faces of the prism.
These regimes are naturally arranged by the complexity of the final solution, as summarized in \cref{tab:regime-summary}.

As noted in \cref{sec:classification}, the learning rate $\eta$
only rescales time. We choose it so that the relevant matrix
dynamics evolve on times $t=O(1)$. Indeed, \cref{eq:gf} gives
$
\dot{\mU}
=\frac{\eta}{p}\mU(2\mS-\mS\mA-\mA\mS) $ with $ \mA=\mU^\top\mU.
$
In the large-data, mean-field, and narrow-latent regimes,
$\|\mS\|_{\mathrm{op}}$ and $\|\mA(0)\|_{\mathrm{op}}$ are of
order one, giving a characteristic rate $\eta/p$.
In the small-data regime, the nonzero eigenvalues of $\mS$
are of order $p/m$, giving rate $\eta/m$.
In the free regime, $\|\mA(0)\|_{\mathrm{op}}$ is of order
$p\sigma^2\gg1$, giving rate $\eta\sigma^2$.
The choices $\eta=p$, $m$, and $\sigma^{-2}$ in
\cref{tab:regime-summary} normalize these respective rates.

Recall the parameters $\phi=p/m, \rho=n\sigma^2, \psi=p/n$ that appeared in our phase diagram analysis. Each of the five basic extreme regimes admits a two-parameter refinement by varying these parameters or their combinations. 
For mean-field and narrow-latent results, we impose the following common spectral assumption.
For some deterministic \(K<\infty\) and deterministic probability measure \({\nu_\phi}\),
\begin{equation}
  \mathbb P\!\left(\|\mS\|_{\mathrm{op}}\le K\right)
  \longrightarrow1,
  \qquad
  \frac1{p}\sum_{i=1}^{p}\delta_{\lambda_i(\mS)}\Longrightarrow{\nu_\phi}
  \qquad\text{in prob.}
  \label{eq:spectral-assumptions-main}
\end{equation}
In the large-data regime these follow from the stronger near-isotropy assumption stated in \cref{sec:large-data}; the small-data regime requires a different normalization stated in \cref{sec:small-data}.
All the theorems in the present section prove convergence of losses and other relevant quantities, uniformly in probability on all compact time intervals.
We therefore adopt the following standard stochastic process notation:
\begin{equation*}
  \boxed{Z=\oucp}
  \quad \equiv \quad
  \boxed{\sup_{0 \leq t \leq T}\|Z(t)\| \to 0 \; \forall T \in \R_+ \; \text{under a given joint $p,n,m \to \infty$ limit.}}
\end{equation*}
For statements involving expected losses, we use a mild additional
uniform-integrability condition on the relevant normalized empirical
covariance.  Namely, for some $\delta>0$, we assume
\begin{equation}
    \sup_{N \in \sN}
    \E\|\mS_N^\sharp\|_{\mathrm{op}}^{1+\delta}<\infty,
    \label{eq:loss-UI-main}
\end{equation}
where
\begin{equation}
    \mS_N^\sharp
    :=
    \begin{cases}
        \mS_{p_N,m_N},
        & \text{in the mean-field, large-data, and narrow-latent regimes},\\
        \frac{m_N}{p_N} \mS_{p_N,m_N},
        & \text{in the small-data regime},
    \end{cases}
\end{equation}
where $(p_N,m_N)_{N=1}^\infty$ is a sequence corresponding to the relevant joint $p,m\to\infty$ limit.
For small data this is equivalently a moment bound on
$\bm{\Gamma}=p^{-1}\mX^\top\mX$.
Gaussian training data satisfy these uniform-integrability conditions by the
standard Gaussian singular-value bounds; see e.g. the proof of \cref{cor:mf-expectations}.

We check our solutions numerically on subsets of CIFAR-10: \cref{fig:exp-main}.
To approximate the isotropy assumption in \cref{eq:pop-isotropy-main}, we globally whiten CIFAR-10 using the full training split; \cref{app:experiments} describes the preprocessing and the remaining finite-sample anisotropy.
Note that the subset we actually train on still has non-trivial $\mS$.

\subsection{Mean-field (\ref{app:mean-field})}
\label{sec:mean-field}

In the corresponding limit, $\mA(0)$ becomes scalar, and the whole dynamics of $\mA(t)$ decouple into independent ones for each eigencomponent of $\mS$:
\begin{theorem}[Short version of \cref{thm:mf-finite-time} + \cref{cor:mf-expectations}]
\label{thm:mf-finite-time-main}
Consider $p,n,m \to \infty$ in such a way that $\psi \to 0$, while $\phi,\rho$ stay finite.
Then under \cref{eq:spectral-assumptions-main},
\begin{equation}
  \mA(\cdot) - a(\cdot;\mS) = \oucp,
  \qquad
  a(t;\lambda)
  := \frac{\rho}{\rho+(1-\rho)e^{-4\lambda t}}.
\end{equation}
Moreover, $\widehat L(\cdot) - \widehat \gL(\cdot) = \oucp$ and (under \cref{eq:pop-isotropy-main}) $L(\cdot) - \gL(\cdot) = \oucp$ with
\begin{equation}
  \gL(t)
  = \int \ell(\lambda t)\,d{\nu_\phi}(\lambda),
  \qquad
  \widehat\gL(t)
  = \int \lambda \ell(\lambda t)\,d{\nu_\phi}(\lambda),
  \qquad
  \ell(t)
  := \frac{(1-\rho)^2 e^{-8 t}}{2 \left(\rho + (1-\rho) e^{-4 t}\right)^2}.
\end{equation}
Under \cref{eq:loss-UI-main}, the corresponding loss convergence also holds in \(L^1\), and hence for expected losses.
\end{theorem}

\paragraph{Gaussian data.}
Although ${\nu_\phi}$ becomes the Marchenko-Pastur (MP) distribution, the above integrals cannot be evaluated analytically.
Nevertheless, their large-$t$ asymptotics of limit losses $\gL$ and $\widehat\gL$ are tractable.
Indeed, since $\ell(t)$ decays exponentially, and $\nu_\phi$ features a spectral gap for any $\phi \neq 1$ (i.e. there is a gap between zero and the bulk of the spectrum),
$\widehat\gL(t)$ also decays exponentially when $\phi \neq 1$.
When $\phi < 1$, this also applies to $\gL(t)$.
For $\phi > 1$, there is a point mass at zero, which does not affect $\widehat\gL(t)$ due to the factor $\lambda$, while it results in a remainder term for $\gL(t)$: $\gL(\infty) = \tfrac{(1-\rho)^2}{2} \left(1 - \tfrac{1}{\phi}\right)_+$.
When \(\rho=1\), \(a_t(\lambda)\equiv1\), so both limiting losses vanish identically.
Suppose $\rho \neq 1$; otherwise, $\gL = \widehat\gL \equiv 0$.
When $\phi = 1$, the spectral gap disappears; because of this, both $\widehat\gL(t)$ and $\gL(t)$ decay much slower:
\begin{equation}
  \gL(t) - \gL(\infty)
  \asymp \begin{cases}
    \frac{1}{t^{1/2}}, & \phi = 1, \\
    \frac{e^{-8 (1 - \sqrt\phi)^2 t}}{t^{3/2}}, & \phi \neq 1,
  \end{cases}
  \qquad\qquad
  \widehat\gL(t)
  \asymp \begin{cases}
    \frac{1}{t^{3/2}}, & \phi = 1, \\
    \frac{e^{-8 (1 - \sqrt\phi)^2 t}}{t^{3/2}}, & \phi \neq 1
  \end{cases}
\end{equation}
as $t \to \infty$, where the exact expressions are derived in \ref{sec:mean-field_large_time}.
As a result, if we fix large enough $t$, the training loss as a function of
$\phi$ develops a maximum near $\phi=1$, where the problem becomes critically
determined.  This has a close analogue in noiseless linear regression under
gradient flow.  There too, the training residual is controlled by the lower
edge of the empirical-covariance spectrum: for $\phi\neq1$ the nonzero
spectrum is separated from zero and the training loss relaxes exponentially,
whereas at $\phi=1$ the spectral gap closes and the relaxation becomes
algebraic.  Thus the finite-time peak near $\phi=1$ in both problems has the
same spectral origin: increasingly small nonzero covariance eigenvalues
produce increasingly slow learning modes, with a qualitative change of the
large-time decay exactly at the critical aspect ratio.
This dynamical critical slowing should be distinguished from the familiar
large generalization peak of noisy ridgeless regression, where amplification
of label noise near the interpolation threshold provides an additional
mechanism.

We emphasize that we \emph{do not} expect these asymptotics to hold exactly for \emph{finite} $p,n,m$ losses $L(t)$ and $\widehat L(t)$.
This is due to the fact that the above theorem proves convergence \emph{only on compact} time intervals.
Therefore we would expect that for large enough $p,n,m$, the above asymptotic formulas could be numerically close for large $t$, but not for infinite $t$.\subsection{Large-data (\ref{app:large-data})}
\label{sec:large-data}

Here, we adopt a stronger condition for data covariance: $\|\mS-\mI_p\|_{\mathrm{op}} \to 0$.
Since the empirical covariance becomes the identity, both train and population losses follow the same logistic curve:
\begin{theorem}[Short version of \cref{thm:ld-main}]
\label{thm:large-data-main}
Consider $p,n,m \to \infty$ in such a way that $\phi \to 0$, while $\psi,\rho$ stay finite.
Then under $\|\mS-\mI_p\|_{\mathrm{op}} \to 0$ in probability,
\begin{equation}
  \mA(\cdot)-\Theta(\cdot; \mA(0)) = \oucp,
  \qquad
  \Theta(t;s)
  := \frac{s}{s+(1-s)e^{-4t}},
\end{equation}
\begin{equation}\label{eq:losses-ld}
  L(\cdot) - \gL(\cdot) = \oucp,
  \qquad
  \widehat L(\cdot) - \gL(\cdot) = \oucp,
  \qquad
  \gL(t)
  = \frac{1}{2} \int (\Theta(t;\rho\lambda) - 1)^2 \, \mu_\psi(d\lambda),
\end{equation}
where $\mu_\psi$ is the MP law.
The above admits a closed formula:
\begin{equation}
  \gL(t)
  = \frac12 - \rho e^{4t} H_\psi\left(\rho (1-e^{4t})\right) + \frac{\rho^2 e^{8t}}{2} H_\psi'\left(\rho (1-e^{4t})\right),
\end{equation}
\begin{equation}
  H_\psi(q)
  = \frac{1-q\left(1+\psi\right) - \sqrt{1-2q\left(1+\psi\right) + q^2\left(1-\psi\right)^2}}{2 q^2 \psi}.
\end{equation}
Under \cref{eq:loss-UI-main}, the corresponding loss convergence also holds in \(L^1\), and hence for expected losses.
\end{theorem}
The spectrum of $\mA(0) / \rho$ follows a MP law $\mu_\psi$.
This causes a qualitative difference in limit loss $\gL(t)$ behavior for large $t$ between $\psi = 1$ and $\psi \neq 1$, similar to that in the mean-field regime discussed above.
The difference is again caused by the spectral gap present when $\psi \neq 1$, but in contrast to the mean-field regime, the loss decays exponentially in both scenarios, albeit with different exponents (see \ref{app:large-data-rigorous-large-time} for derivations):
\begin{equation}
  \gL(t) - \gL(\infty)
  \asymp \begin{cases}
    e^{-2t}, & \psi = 1,\\
    e^{-8t}, & \psi \neq 1,
  \end{cases}
  \qquad
  \gL(\infty)
  = \frac{p-\min(n,p)}{2p}
  = \frac{1}{2} \left(1 - \frac{1}{\psi}\right)_+,
\end{equation}
where the second formula follows from the fact that every nonzero eigenvalue of $\mA(0)$ converges to $1$, while zero ones stay zero, while $\rank \mA(0) = \min(n,p)$.
As in the mean-field limit, the same remark on large-$t$ asymptotics of $L(t)$ for finite $p,n,m$ applies here.

\subsection{Narrow-latent (\ref{app:narrow-latent})}
\label{sec:narrow-latent}

\begin{figure}
  \centering
  \includegraphics[width=0.96\textwidth]{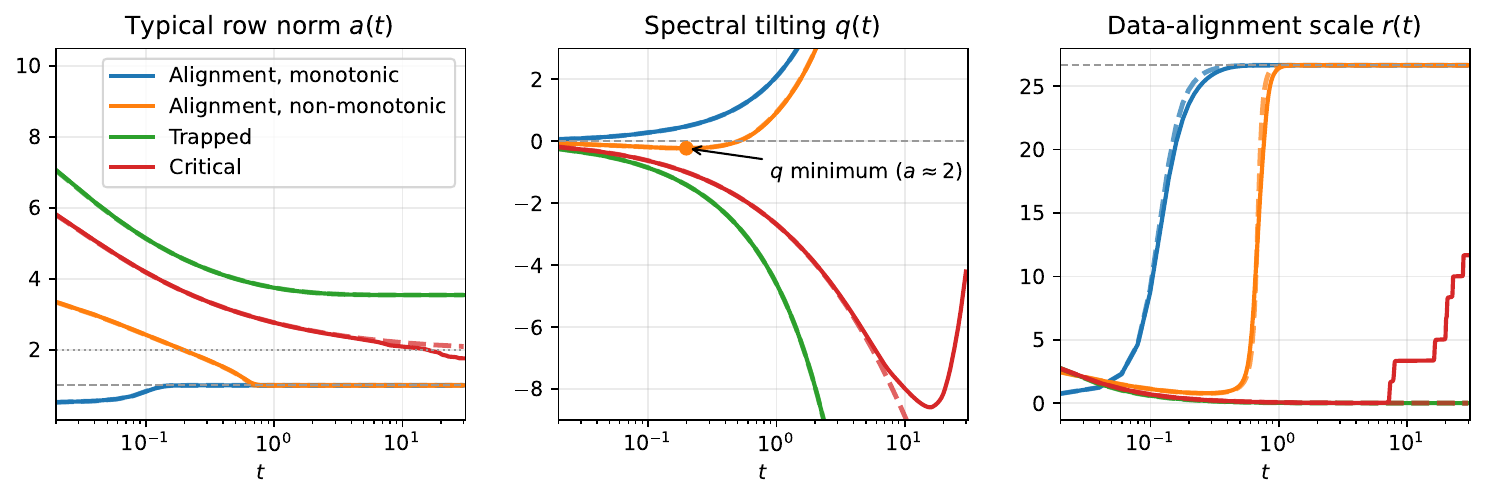}
  \includegraphics[width=0.32\textwidth]{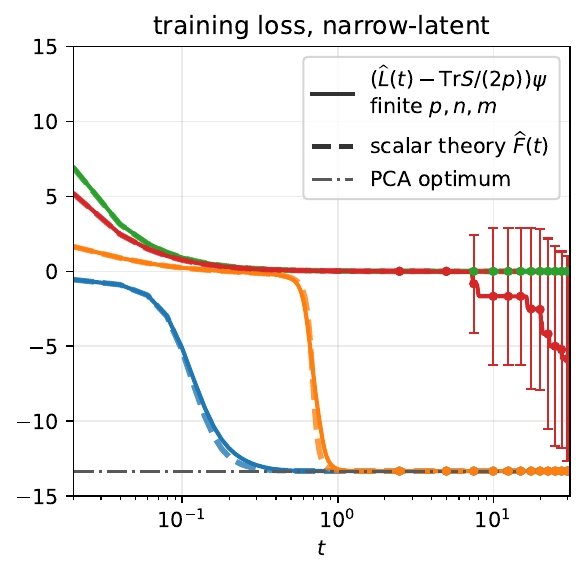}
  \includegraphics[width=0.32\textwidth]{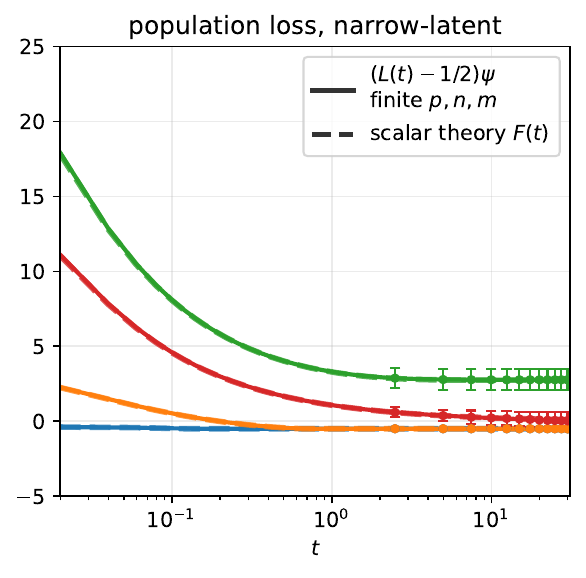}
  \includegraphics[width=0.32\textwidth]{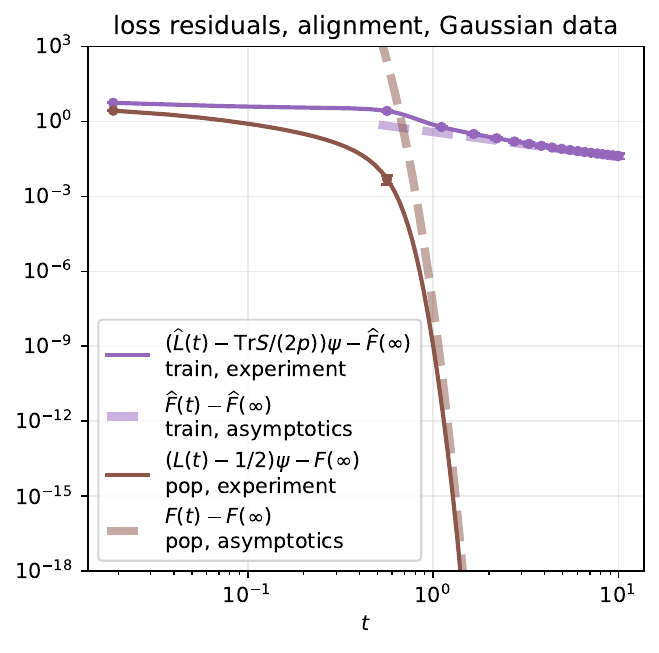}
  \caption{
    We measure quantities related to \emph{narrow-latent} and compare to their theoretical limits on a CIFAR-10 subset, see \cref{sec:narrow-latent} for details.
    The bottom-right plot checks \cref{eq:narrow-latent-alignment-loss-tails} and uses Gaussian data.
    See \ref{app:experiments} for implementation details.
    \label{fig:narrow-latent}
  }
\end{figure}

Define the exponential transform and its logarithmic derivative by
\begin{equation}
M_{\nu_\phi}(q):=\int e^{q\lambda}\,d{\nu_\phi}(\lambda),
\qquad
m_{\nu_\phi}(q):=\frac{M_{\nu_\phi}'(q)}{M_{\nu_\phi}(q)}.
\end{equation}
The theorem below establishes limits of latent-space Gramians $\mG = \mU \mU^\top$, $\mH = \mU \mS \mU^\top$.
They become scalar (converging, respectively, to $a$ and $r$ defined below), while the model progressively aligns with the data covariance spectrum (the evolution of $q$ defined below):
\begin{theorem}[Short version of \cref{thm:nl-general-nu} + \cref{cor:nl-expected-losses}]
\label{thm:nl-finite-time-main}
Consider $p,n,m \to \infty$ in such a way that $\psi \to \infty$, while $\phi,\psi\rho$ stay finite.
Then under \cref{eq:spectral-assumptions-main},
\begin{equation}
  \mG(\cdot) - a(\cdot) \mI = \oucp,
  \qquad
  \mH(\cdot) - r(\cdot) \mI = \oucp,
\end{equation}
\begin{equation}\label{eq:nl-train_loss-limit-main}
  \widehat L(\cdot) - \frac{\Tr\mS}{2p} = \oucp,
  \quad
  \left(\widehat L(\cdot) - \frac{\Tr\mS}{2p}\right) \psi - \widehat F(\cdot) = \oucp,
  \quad
  \widehat F(t)
  = r(t) \left(\frac{a(t)}{2} - 1\right),
\end{equation}
where $r(t) = a(t) m_{\nu_\phi}(q(t))$, and $a(t),q(t)$ are the solution of the ODE system below:
\begin{equation}\label{eq:proxy-nl-main}
  \dot a(t)=4a(t)m_{\nu_\phi}(q(t))(1-a(t)),
  \qquad
  \dot q(t)=2(2-a(t)),
  \qquad
  a(0)=\psi\rho,\qquad q(0)=0.
\end{equation}
Moreover, under the isotropy condition \cref{eq:pop-isotropy-main},
\begin{equation}\label{eq:nl-test_loss-limit-main}
  L(\cdot) - \frac{1}{2} = \oucp,
  \quad
  \left(L(\cdot) - \frac{1}{2}\right) \psi - F(\cdot) = \oucp,
  \quad
  F(t)
  = a(t) \left(\frac{a(t)}{2} - 1\right).
\end{equation}
Under \cref{eq:loss-UI-main}, the corresponding loss convergence also holds in \(L^1\), and hence for expected losses.
\end{theorem}The scalars have a direct interpretation.  Since the theorem above
shows \(\mG=\mU\mU^\top\approx a\mI\), the quantity \(a\) is the
asymptotic squared norm of a typical latent row of \(\mU\), while the
off-diagonal part of \(\mG\) vanishes.  Likewise
\(\mH=\mU\mS\mU^\top\approx r\mI\), so \(r\) is the
corresponding covariance-weighted row energy, or data-alignment scale.
Finally, \(q\) encodes spectral-tilting: \(q>0\) progressively favors large eigenvalue directions, whereas \(q<0\) favors the lower edge and, when $\phi > 1$, the null space of $\mS$.

The limiting ODE \cref{eq:proxy-nl-main} exhibits several qualitatively different behaviors, governed
by the initialization magnitude $\kappa := a(0) = \psi\rho = p\sigma^2$ and by the null-space mass $w_0 := \nu_\phi(\{0\})$:
\begin{prop}[Short version of \cref{prop:nl-large-time-alternatives}]
\label{prop:narrow-latent-phases}
Under $\nu_\phi \neq \delta_0$,
\begin{enumerate}
    \item \textbf{Monotonic alignment} (blue curves on \cref{fig:narrow-latent}).
    If $0<\kappa\leq 2$, then $q(t)$ is nondecreasing and $q(t)\to+\infty$, $a(t)\to1$.
    Thus the dynamics favor the upper edge of the data spectrum.
    \item \textbf{Non-monotonic alignment} (orange curves).
    Suppose $\kappa>2$ and $\tfrac{\kappa^2}{\kappa-1}w_0^2 < 4$.
    Then $q(t)$ initially decreases, so the dynamics move toward the lower spectral edge, but this motion eventually reverses. 
    In particular, $q(t)$ has a finite minimum, hence $q(t)\to+\infty$, $a(t)\to1$. 
    \item \textbf{Null-space trapping} (green curves).
    Suppose $\kappa>2$ and $\frac{\kappa^2}{\kappa-1}w_0^2 > 4$.
    The reversal never occurs: $q(t)\to-\infty$, $r(t)\to0$, $a(t)\to a_\ast>2$.
    The limiting dynamics become trapped in the null space of the empirical covariance rather than aligning with its principal directions.
    \item \textbf{Critical boundary} (red curves).
    At $\tfrac{\kappa^2}{\kappa-1}w_0^2 = 4$ one has $q(t)\to-\infty$, $a(t)\to2$, giving the boundary between alignment and strict trapping.
\end{enumerate}
\end{prop}
In particular, if $w_0=0$, trapping is impossible for any finite $\kappa$.
The proof, the first integral underlying the threshold above, and detailed
large-time asymptotics are given in \ref{app:narrow-large-time}.
For Gaussian data, $w_0 = \max\left(0,1-\frac1\phi\right)$,
so a nontrivial trapping transition can occur only when $\phi>1$, i.e. when
the empirical covariance has a macroscopic null space. In the alignment
cases, $q(t)\to+\infty$ concentrates the row-energy distribution near the
upper spectral edge. 
When the eigenvalues relevant to the rank-\(n\) PCA optimum converge to the same upper edge, as happens in the MP specialization, this recovers the PCA solution.

The trapped and critical branches also have a finite-dimensional optimization
interpretation. \ref{app:narrow-landscape} shows that null-space
critical points satisfying
$\mU\mU^\top\succ2\mI$ are nonglobal local minima of the tied empirical loss,
whereas those with
$\lambda_{\min}(\mU\mU^\top)=2$ are degenerate saddles: their Hessian is
positive semidefinite, but decreasing perturbations appear at higher order.
Thus the strict trapped limit $a_\ast>2$ lies inside the local-minimum region,
while the critical limit $a_\ast=2$ approaches its saddle boundary.
We underline that there is no contradiction with the result of \citet{kawaguchi2016deep} on the absence of non-global minima in linear nets as their result does not apply to networks with tied weights.

This also
suggests a plausible explanation for the late-time departure of the critical
finite-dimensional trajectories (red curves in \cref{fig:narrow-latent}) from
the limiting prediction.  Finite-size corrections keep the trajectory only
approximately on the critical separatrix, and near the corresponding
degenerate saddle such perturbations can eventually drive it away along a
descent direction.  

A second notable feature is that successful alignment does not imply the same
relaxation time for training and population performance. Under a regular
upper spectral edge, in either alignment regime the population correction
approaches its limiting value exponentially (up to a polynomial prefactor),
whereas the training correction decays only as $t^{-1}$; see
\ref{app:narrow-large-time}. For Gaussian data, the
MP square-root edge gives, for $\kappa\neq1$,
\begin{equation}
  F(t)-F(\infty)
  \asymp t^6e^{-8\lambda_+t},
  \quad
  \widehat F(t)-\widehat F(\infty)
  \sim \frac{3}{8t},
  \qquad
  F(\infty) = -\frac{1}{2},
  \quad
  \widehat F(\infty) = -\frac{\lambda_+}{2},
  \label{eq:narrow-latent-alignment-loss-tails}
\end{equation}
where $\lambda_+ = (1 + \sqrt\phi)^2$ is the upper MP spectral edge.
Once the row norms have nearly equilibrated, the remaining slow motion is primarily spectral alignment, which continues to change the training loss, because it is sensitive to the empirical covariance spectrum, but has no direct effect on the isotropic population loss at fixed row norm.
\cref{fig:narrow-latent} illustrates both
the alignment/trapping alternatives and this separation of time scales.
Further asymptotics for all four cases are given in \ref{app:narrow-large-time}.
\subsection{Small-data (\ref{app:small-data})}
\label{sec:small-data}

The relevant data assumption is the active-space near-isometry $\|\mGamma - \mI_m\|_{\mathrm{op}} \to 0$, where $\mGamma = \frac1{p}\mX^\top\mX$.
\begin{theorem}[Short version of \cref{thm:sd-main}]
Consider $p,n,m \to \infty$ in such a way that $\phi \to \infty$, while $\psi,\rho$ stay finite.
Then under $\|\mGamma - \mI_m\|_{\mathrm{op}} \to 0$ in probability, $\widehat L(\cdot) - \widehat\gL(\cdot) = \oucp$ and $L(\cdot) - \gL = \oucp$, where
\begin{equation}\label{eq:losses-mf}
  \gL
  = \frac{(1-\rho)^2 + \psi \rho^2}{2},
  \qquad
  \widehat\gL(t)
  = \frac{\left(1-a(t)\right)^2 + a(t) \mu_1(t)}{2}.
\end{equation}
where $(a,\mu_{k\geq 1})$ is the unique solution of a hierarchical system of ODEs specified in \cref{thm:sd-main}.
The $O(1/\phi)$ correction for $L$ also converges: $(L(\cdot)-L(0)) \phi - C(\cdot) = \oucp$, where $C(t)$ is given by an explicit functional involving $a$, $\mu_1$, and $\mu_2$.
Under \cref{eq:loss-UI-main}, the corresponding loss convergence also holds in \(L^1\), and hence for expected losses.
\end{theorem}

As for the other learning regimes, we validate the limit loss predictions of the above theorem numerically on CIFAR-10: \cref{fig:exp-main}, two last plots.
One can immediately verify from the exact definition of $C(t)$ in \cref{thm:sd-main} that $C'(t) \leq 0$ and $C'(0) < 0$, hence the limit population loss decreases monotonically.
Due to the lack of space, we do not discuss the physical meaning of $(a,\mu_{k\geq 1})$ in the main.
We also leave a detailed analysis of the limit ODE hierarchy for future work.

\section{Discussion, limitations, and future work}
\label{sec:discussion}

Our main contribution is a systematic classification of extreme learning
regimes for weight-tied linear autoencoders.  The diagrammatic expansion of
\cref{sec:classification} organizes dominant loss terms into a triangular-prism hierarchy,
whose five facets correspond to the large-data, small-data, mean-field,
narrow-latent, and free regimes.  \cref{sec:solutions} provides an independent
dynamical justification of this picture: four of these facets lead to
qualitatively different limiting systems, ranging from decoupled spectral
modes to a nonlinear alignment dynamics and an infinite hierarchy of leakage
moments.
Thus the prism provides not only a classification of formal loss
expansions, but also a map between asymptotic scalings and distinct
mechanisms of learning.

\paragraph{Spectral geometry and initialization dependence.}
A recurring theme is that these mechanisms are controlled by spectral
geometry.  Closing spectral gaps cause critical slowing at $p/m=1$ in the
mean-field regime and at $p/n=1$ in the large-data regime.  In the
narrow-latent regime, the spectral clock $q(t)$ instead dynamically tilts the
directions explored by a typical latent row.  This produces particularly
rich initialization dependence: for sufficiently small initialization the
model aligns with leading data directions and approaches PCA, while for
larger initialization $q(t)$ can first move toward the lower spectral edge
before reversing.  With a sufficiently large initialization and a nontrivial
data null space, the reversal never occurs and the limiting dynamics become
trapped, so that the training loss does not approach its global PCA
minimum.

\paragraph{Train--population separation.}
The solutions show that training and population losses can evolve
on parametrically different scales.  At the large-data facet the two losses
coincide asymptotically, whereas at the opposite small-data facet training
changes at order one while the leading population loss is frozen and its
first nontrivial change is only of order $m/p$.  In the narrow-latent
alignment branch the separation is dynamical: after row norms have nearly
equilibrated, the remaining slow motion is mainly spectral alignment, which
continues to affect training loss but is largely invisible to isotropic
population loss.  For Gaussian data this yields the striking asymptotic
separation \cref{eq:narrow-latent-alignment-loss-tails}.

\paragraph{Limitations and open problems.}
The prism classification is complete at the
level of dominant expansion monomials, but this does not by itself prove
completeness of all possible limiting trajectories.  The free facet remains a natural target for
analysis.  Moreover, our convergence theorems hold uniformly only on fixed
time intervals, whereas the large-time asymptotics are taken after the
high-dimensional limit; understanding the joint large-dimension/large-time
regime is therefore an important open problem.  Finally, the cleanest
population results rely on isotropy or related spectral assumptions.

\section*{AI Use Statement}

We used a popular LLM extensively as a research assistant in developing \cref{app:solutions}.
All formal results in \cref{app:solutions} were formulated and proved with substantial assistance from the LLM. 
This included assistance in identifying appropriate theorem statements, constructing proof strategies, working out technical arguments, and drafting the resulting proofs. 
The authors independently inspected, checked, and revised all such arguments before including them in the manuscript and take full responsibility for their correctness.
We also used an AI assistant to facilitate coding and as an auxiliary tool for polishing the manuscript.

No LLM was used in developing the prism classification. 
In particular, the diagrammatic construction, the classification of extreme regimes, and the results of \cref{sec:classification} and \cref{sec:appdiag} were developed independently by the authors.

\addcontentsline{toc}{section}{Bibliography}
\bibliography{references}
\bibliographystyle{apalike}

\appendix

\newpage
\tableofcontents

\newpage

\section{Related works}
\label{app:prior_works}
\subsection{Linear autoencoders, PCA, and optimization landscapes}

\paragraph{Prior work.}
Linear autoencoders provide one of the classical connections between neural
networks and principal component analysis. Early works characterized their
optimal solutions in terms of singular-value decomposition and principal
subspaces \citep{bourlard1988auto,baldi1989neural}, with later work
giving a broader analytical treatment of linear autoencoders
\citep{baldi2012autoencoders}. More recently, \citet{kunin2019loss} analyzed
the critical-point structure of regularized linear autoencoders and showed how
regularization resolves part of the non-identifiability of the latent
representation.

\paragraph{Our contribution.}
We complement the classical characterization of linear-autoencoder optima
with a dynamical and high-dimensional analysis of the weight-tied model from
random initialization. Our asymptotic setting simultaneously varies the input
dimension, latent dimension, training-set size, and initialization magnitude,
placing the usual undercomplete and overcomplete distinction inside a larger
family of learning regimes. Importantly, the PCA optimum need not describe
the outcome of the limiting gradient flow: in the narrow-latent regime,
sufficiently large initialization together with a nontrivial data null space
can lead to null-space trapping rather than alignment with the principal
subspace. We further relate this phenomenon to the finite-dimensional
landscape, showing that null-space critical points satisfying
$\mU\mU^\top\succ2\mI$ are nonglobal local minima of the tied empirical loss.

\subsection{Training and generalization dynamics}

\paragraph{Prior work.}
Exact learning dynamics have been derived for several linear or otherwise
tractable neural-network models. \citet{saxe2013exact,saxe2019mathematical}
obtained exact solutions for deep linear networks under suitably aligned
initializations. For linear autoencoders, \citet{pretorius2018learning}
integrated the gradient flow on a finite training set, again relying on an
aligned initialization. \citet{nguyen2019dynamics} studied gradient descent in
weight-sharing autoencoders for several structured generative models.
For nonlinear shallow autoencoders, \citet{nguyen2021analysis} and
\citet{refinetti2022dynamics} derived high-dimensional dynamics in,
respectively, wide mean-field and narrow-latent settings, using online SGD.

Related work has also studied train and population dynamics in supervised linear
models. In particular, \citet{lampinen2019analytic} and
\citet{advani2020high} derive analytically tractable generalization dynamics
for high-dimensional linear networks. Finally, \citet{cui2023high} study
finite-sample autoencoder generalization, but at the level of the empirical
risk minimizer rather than its training trajectory.

\paragraph{Our contribution.}
We derive limiting train and population trajectories for a finite, repeatedly
used training set and random isotropic initialization, obtaining rigorous
finite-time limits in four qualitatively different regimes. Unlike analyses
that assume an initialization already aligned with the data singular
directions, our narrow-latent limit resolves the evolution of this alignment
itself. It exhibits initialization-dependent outcomes: the limiting dynamics
may align with the leading empirical-covariance directions and approach PCA,
or, for sufficiently large initialization and a nontrivial null space, become
trapped away from the PCA optimum. The solutions also expose dynamical
phenomena not visible from the final estimator alone: train and population
losses can evolve on parametrically different time scales, and in the
small-data regime training changes by $O(1)$ while the population loss changes
only at order $m/p$.

\subsection{Large-size limits and learning-regime classification}

\paragraph{Prior work.}
Different infinite-size scalings can lead to qualitatively different neural
network dynamics. Classical examples include mean-field and NTK/lazy limits,
as well as intermediate or feature-learning scalings. Several works have
systematically compared or classified such limits
\citep{golikov2020towards,golikov2020dynamically,yang2021tensor_iv}.
\citet{woodworth2020kernel} emphasized in particular that initialization scale
can control transitions between kernel and rich regimes.
Most such classifications send network width to infinity while the remaining
problem dimensions are fixed.

Most closely related to our approach, \citet{yarotsky2026gradient} proposed a
diagram-expansion method in which learning regimes are identified from
dominant families of terms in the loss expansion, and applied it to tensor
factorization models.

\paragraph{Our contribution.}
We extend this classification problem to a setting in which the input
dimension $p$, latent dimension $n$, training-set size $m$, and initialization
scale $\sigma^2$ may all vary jointly. Extending the diagrammatic formalism
with data edges allows train and population losses to be treated separately.
For the weight-tied linear autoencoder, the resulting dominant monomials form
a finite hierarchy naturally represented by a triangular prism. Its five
facets give the large-data, small-data, mean-field, narrow-latent, and free
basic regimes, while edges and vertices represent combinations of these
extremes.

\subsection{High-dimensional generalization and spectral criticality}

\paragraph{Prior work.}
High-dimensional learning can exhibit sharp behavior near interpolation
thresholds. The double-descent literature highlighted anomalous generalization
near the point where model and sample dimensions become comparable
\citep{belkin2019reconciling}, while exact analyses of high-dimensional
ridgeless regression characterize this phenomenon using random-matrix
spectra \citep{hastie2022surprises}. Closely related dynamically,
\citet{advani2020high} showed how the spectrum of the empirical covariance
controls train and generalization-error trajectories in high-dimensional
linear models.

\paragraph{Our contribution.}
We identify related spectral critical phenomena in an unsupervised
reconstruction problem and show how they arise in several distinct learning
regimes. In the mean-field regime, the mechanism closely parallels noiseless
linear regression under gradient flow: as $p/m\to1$, the lower edge of the
nonzero empirical-covariance spectrum approaches zero, producing increasingly
slow learning modes and changing the large-time relaxation from exponential
to algebraic at the critical aspect ratio. This dynamical critical slowing
should be distinguished from the familiar generalization peak of noisy
ridgeless regression, where amplification of label noise near the
interpolation threshold provides an additional mechanism. In the large-data
regime, an analogous spectral slowdown occurs at $p/n=1$, now because the
lower edge of the random initial Gram spectrum reaches zero. The
narrow-latent solution reveals a further phenomenon: different spectral
features can control training and population relaxation, leading to
parametrically separated time scales. Thus spectral geometry controls not
only the limiting performance but also the time scales on which optimization
and population performance evolve.

\section{Proof of the autoencoder phase diagram (\cref{th:prism})} \label{sec:appdiag}

\begin{figure}
    \centering
    \begin{tikzpicture}[
        node distance=2cm,
        every node/.style={font=\sffamily\small},
        scale=0.25
    ]
        
        \tikzset{
            out node/.style={circle, draw=black, fill=cyan, minimum size=1mm, scale=.6},
            in node/.style={circle, draw=black, fill=orange, minimum size=1mm},
            hid node/.style={rectangle, draw=black, fill=brown, minimum size=1mm, scale=.7},
            data node/.style={diamond, draw=black, fill=gray, scale=.5},
            U/.style={red, very thick},
            X/.style={black, very thick, densely dashed}
        }

        \begin{scope}[shift={(-25,0)}]
            \node at (-2,1.7) {\large $D$};
        
            \node[out node] (0) at (0,2) {};
            \node[hid node] (-1) at (-2,0) {};
            \node[hid node] (+1) at (2,0) {};
            \node[out node] (2) at (0,-2) {};
            
            \draw[U] (0) to [bend right=15] (-1);
            \draw[U] (0) to [bend left=15] (+1);
            \draw[U] (2) to [bend right=-15] (-1);
            \draw[U] (2) to [bend left=-15] (+1);
        \end{scope}

        \begin{scope}[shift={(-20,0)}]
            \node at (-1.4,1.4) {\large $R$};
        
            \node[hid node] (0) at (0,1.5) {};
            \node[out node] (1) at (0,-1.5) {};
            
            \draw[U] (0) to [bend right=30] (1);
            \draw[U] (0) to [bend right=-30] (1);
        \end{scope}
        
        \begin{scope}[shift={(-12,0)}]
            \node at (-3,1.7) {\large $\widehat{D}$};
        
            \node[out node] (0) at (0,2) {};
            \node[hid node] (-1) at (-2,1) {};
            \node[hid node] (+1) at (2,1) {};
            \node[out node] (-2) at (-2,-1) {};
            \node[out node] (+2) at (2,-1) {};
            \node[data node] (3) at (0,-2) {};
            
            \draw[U] (0) to [bend right=15] (-1);
            \draw[U] (0) to [bend left=15] (+1);
            \draw[U] (-2) to [bend right=-15] (-1);
            \draw[U] (+2) to [bend left=-15] (+1);
            \draw[X] (-2) to [bend right=15] (3);
            \draw[X] (+2) to [bend left=15] (3);
        \end{scope}

        \begin{scope}[shift={(-6,0)}]
            \node at (-2,1.7) {\large $\widehat{R}$};
        
            \node[hid node] (0) at (0,2) {};
            \node[out node] (-1) at (-2,0) {};
            \node[out node] (+1) at (2,0) {};
            \node[data node] (2) at (0,-2) {};
            
            \draw[U] (0) to [bend right=15] (-1);
            \draw[U] (0) to [bend left=15] (+1);
            \draw[X] (2) to [bend right=-15] (-1);
            \draw[X] (2) to [bend left=-15] (+1);
        \end{scope}

        \begin{scope}[shift={(-1,0)}]
            \node at (-1.4,1.4) {\large $\widehat{F}$};
        
            \node[out node] (0) at (0,1.5) {};
            \node[data node] (1) at (0,-1.5) {};
            
            \draw[X] (0) to [bend right=30] (1);
            \draw[X] (0) to [bend right=-30] (1);
        \end{scope}

        \begin{scope}[shift={(1,0)}]
            \draw[thick] (1, -3) to (1, 3);
        \end{scope}

        \begin{scope}[shift={(6,0)}]
        
            \node[out node] (0) at (0,2) {};
            \node[hid node] (-1) at (-2,1) {};
            \node[hid node] (+1) at (2,1) {};
            \node[out node] (-2) at (-2,-1) {};
            \node[out node] (+2) at (2,-1) {};
            \node[data node] (3) at (0,-2) {};
            
            \draw[U] (0) to [bend right=15] (-1);
            \draw[U] (0) to [bend left=15] (+1);
            \draw[U] (-2) to [bend right=-15] (-1);
            \draw[U, ultra thick, dotted] (+2) to [bend left=-15] (+1);
            \draw[X] (-2) to [bend right=15] (3);
            \draw[X] (+2) to [bend left=15] (3);
        \end{scope}

        \begin{scope}[shift={(10,0)}]
        
            \node[hid node] (0) at (-1,1) {};
            \node[out node] (-1) at (1,1) {};
            \node[out node] (+1) at (-1,-1) {};
            \node[data node] (2) at (1,-1) {};
            
            \draw[U] (0) to [bend right=-15] (-1);
            \draw[U, ultra thick, dotted] (0) to [bend left=-15] (+1);
            \draw[X] (2) to [bend right=15] (-1);
            \draw[X] (2) to [bend left=15] (+1);
        \end{scope}

        \begin{scope}[shift={(17,0)}]
            \node at (-4,2.5) { merging};
            \node at (-4,0) {\huge $\leadsto$};
        
            \node[out node] (0) at (0,2) {};
            \node[hid node] (-1) at (-2,1) {};
            \node[hid node] (+1) at (2,1) {};
            \node[out node] (++1) at (4,1) {};
            \node[out node] (-2) at (-2,-1) {};
            \node[out node] (+2) at (2,-1) {};
            \node[data node] (++2) at (4,-1) {};
            \node[data node] (3) at (0,-2) {};
            
            \draw[U] (0) to [bend right=15] (-1);
            \draw[U] (0) to [bend left=15] (+1);
            \draw[U] (-2) to [bend right=-15] (-1);
            \draw[X] (-2) to [bend right=15] (3);
            \draw[X] (+2) to [bend left=15] (3);

            \draw[U] (+1) to [bend right=-15] (++1);
            \draw[X] (++2) to [bend right=15] (++1);
            \draw[X] (++2) to [bend left=15] (+2);
        \end{scope}

        \begin{scope}[shift={(-12,-6)}]
            \node[align=center] at (-9,0) {edge pairings \\ \& node contractions};
        
            \node[hid node] (0) at (0,2) {};
            \node[out node] (-1) at (-2,1) {};
            \node[out node] (+1) at (2,1) {};
            \node[data node] (-2) at (-2,-1) {};
            \node[data node] (+2) at (2,-1) {};
            \node[out node] (3) at (0,-2) {};
            
            \draw[U] (0) to [bend right=15] (-1);
            \draw[U] (0) to [bend left=15] (+1);
            \draw[X] (-2) to [bend right=-15] (-1);
            \draw[X] (+2) to [bend left=-15] (+1);
            \draw[X] (-2) to [bend right=15] (3);
            \draw[X] (+2) to [bend left=15] (3);
        \end{scope}

        \begin{scope}[shift={(-5,-6.5)}]
        
            \node at (-3,0.5) {\huge $\leadsto$};

            \node[hid node] (0) at (0,2) {};
            \node[out node] (1) at (0,0) {};
            \node[data node] (-2) at (-2,-1) {};
            \node[data node] (+2) at (2,-1) {};
            
            \draw[U, double] (0) -- (1);
            \draw[X, double] (-2) -- (1);
            \draw[X, double] (+2) -- (1);
        \end{scope}

        \begin{scope}[shift={(-1,-6)}]
        
            \node at (-1.5,0) { $+$};

            \node[hid node] (0) at (0,2) {};
            \node[out node] (1) at (0,0.65) {};
            \node[data node] (2) at (0,-0.65) {};
            \node[out node] (3) at (0,-2) {};
            
            \draw[U, double] (0) -- (1);
            \draw[X, double] (2) -- (1);
            \draw[X, double] (2) -- (3);
        \end{scope}

        \begin{scope}[shift={(2,-6)}]
        
            \node at (-1.5,0) {$+$};

            \node[hid node] (0) at (0,2) {};
            \node[out node] (1) at (0,0.65) {};
            \node[data node] (2) at (0,-2) {};
            
            \draw[U, double] (0) -- (1);
            \draw[X, double] (2) to [bend left=30] (1);
            \draw[X, double] (2) to [bend right=30] (1);

            \node at (10.5,0) { $= p n m^2 \sigma^2 + p^2 n m \sigma^2 + p n m \sigma^2$};
        \end{scope}

    \end{tikzpicture}
    \caption{Diagrams used to represent the averaged derivatives
$D_{\mathrm{pop},s}$ and $D_{\mathrm{tr},s}$. 
      \textbf{Top row: Left:} Base diagrams $D,R,\widehat D,\widehat R,\widehat F$ (Eqs. (\ref{eq:app-base-pop}), (\ref{eq:app-base-train})). Blue circles, brown squares, and grey diamonds denote $p$-, $n$-, and $m$-nodes, respectively.
      Red and dashed black edges denote $\mU$, and $\mX$, respectively. 
      \textbf{Right:} Merging two diagrams. For each pair of matched $\mU$-edges, the diagrams are attached and both edges removed. \textbf{Bottom row:} Three possible edge pairings with associated node contractions for the ring diagram $\Tr[U^\top U(XX^\top)^2]$. 
    }
    \label{fig:diagrams}
\end{figure}

\subsection{Laurent coefficients and diagram expansions}
\label{sec:applossexp}

We prove the coefficient assertion of \cref{th:prism} and introduce the
diagrammatic notation used in the next subsection. Throughout, $\mX$
has independent $\gN(0,1)$ entries and is independent of the Gaussian
initialization $\mU(0)$; expectations include both sources of randomness.
We write $L_{\mathrm{pop}}=L$ and $L_{\mathrm{tr}}=\widehat L$.

\begin{lemma}\label{lem:app-laurent-coefficients}
For every $s\geq1$, the averaged derivatives
$D_{\mathrm{pop},s}$ and $D_{\mathrm{tr},s}$ of
\cref{eq:polynomial-setting} are Laurent polynomials in
$(p,n,m,\sigma,\eta)$. Every monomial has degree $s$ in $\eta$ and
even degree in $\sigma$.
\end{lemma}

\begin{proof}
\noindent\emph{Base diagrams.}
Expanding the losses in traces gives
\begin{equation}\label{eq:app-loss-diagrams}
L=\frac12+\frac{H_{\mathrm{pop}}}{p},\qquad
\widehat L=\frac{\widehat F}{2pm}+\frac{T}{pm},
\end{equation}
where $H_{\mathrm{pop}}=\frac12D-R$,
$T=\frac12\widehat D-\widehat R$, and
\begin{align}
D&=\Tr[(\mU^\top\mU)^2], &
R&=\Tr[\mU^\top\mU],\label{eq:app-base-pop}\\
\widehat D&=\Tr[(\mU^\top\mU)^2\mX\mX^\top], &
\widehat R&=\Tr[\mU^\top\mU\mX\mX^\top], &
\widehat F&=\Tr[\mX\mX^\top].\label{eq:app-base-train}
\end{align}
A \emph{diagram} records the summation indices in such a trace:
vertices have types $p$, $n$, or $m$, and an edge labeled $\mU$
or $\mX$ represents an entry $U_{ai}$ or $X_{i\mu}$, respectively.
Thus $\mU$-edges join $n$- to $p$-vertices and $\mX$-edges join
$p$- to $m$-vertices. For example,
\begin{equation}\label{eq:app-rhat-indices}
\widehat R=\sum_{a=1}^{n}\sum_{i,j=1}^{p}\sum_{\mu=1}^{m}
U_{ai}U_{aj}X_{j\mu}X_{i\mu}.
\end{equation}
Its diagram is a four-edge ring with successive vertex types
$p,n,p,m$. The other traces in \cref{eq:app-base-pop,eq:app-base-train}
also give rings, possibly with more vertices or with no $m$-vertices.
The value of a diagram is the sum, over all assignments to its
vertices, of the product of its edge entries. These five base
diagrams and the operations below are illustrated in
\cref{fig:diagrams}.

\medskip\noindent\emph{Merging and differentiation.}
For functions represented by diagrams, write
$G_1\star G_2=\langle\nabla_{\mU}G_1,\nabla_{\mU}G_2\rangle$.
To evaluate this product graphically, choose one $\mU$-edge in
each diagram, remove the two chosen edges, identify their
$n$-endpoints and identify their $p$-endpoints, then sum the
resulting diagrams over all choices of edges. Each choice counts
with its multiplicity. A merger of two rings is again a ring;
the operation is bilinear, but need not be associative.

Since $\widehat F$ is independent of $\mU$, gradient flow yields
\begin{equation}\label{eq:app-derivative-recursion}
\frac{d^s L_\ell}{dt^s}
=-\eta\left\langle
\nabla_{\mU}\frac{d^{s-1}L_\ell}{dt^{s-1}},
\nabla_{\mU}\widehat L\right\rangle,
\qquad s\geq1.
\end{equation}
Set $G_{\mathrm{pop}}^{[0]}=H_{\mathrm{pop}}$,
$G_{\mathrm{tr}}^{[0]}=T$, and recursively
$G_\ell^{[s]}=G_\ell^{[s-1]}\star T$. Thus the order of the
successive mergers is fixed.
\cref{eq:app-loss-diagrams,eq:app-derivative-recursion} give
\begin{align}
D_{\mathrm{pop},s}
&=\frac{(-\eta)^s}{p^{s+1}m^s}
\mathbb E\bigl[G_{\mathrm{pop}}^{[s]}(\mU(0),\mX)\bigr],
\label{eq:app-pop-coefficient}\\
D_{\mathrm{tr},s}
&=\frac{(-\eta)^s}{p^{s+1}m^{s+1}}
\mathbb E\bigl[G_{\mathrm{tr}}^{[s]}(\mU(0),\mX)\bigr].
\label{eq:app-train-coefficient}
\end{align}

For later use, we note that the edge types can be arranged in two
contiguous arcs. Each of the four rings $D,R,\widehat D,\widehat R$
has a cyclic sequence of edge labels
of the form $\mathsf U^{2h}\mathsf X^{2k}$, where $\mathsf U$
and $\mathsf X$ denote edge types, $k=0$ for $D,R$, and
$k=1$ for $\widehat D,\widehat R$. If two rings have this form,
merge $\mU$-edges at opposite ends of their $\mX$-arcs (choosing
any $\mU$-edge when an $\mX$-arc is empty). The resulting ring
again has a contiguous $\mU$-arc and a contiguous $\mX$-arc:
the selected $\mU$-edges disappear and the two $\mX$-arcs join
around the cyclic boundary. Induction gives such a two-arc ring
for any selected sequence of base factors. This specifies the
merger choices used to establish realizability in
\cref{sec:app-pareto}.

\medskip\noindent\emph{Wick pairing.}
To average a diagram, pair its $\mU$-edges among themselves and
its $\mX$-edges among themselves. A paired pair of edges forces
their corresponding endpoints to have equal indices; it contributes
$\sigma^2$ for a pair of $\mU$-edges and $1$ for a pair of
$\mX$-edges. Each pair becomes one edge in the contracted graph.
If $N_p,N_n,N_m$ vertices of the respective types remain and the
original diagram has $N_U$ $\mU$-edges, the pairing contributes
\begin{equation}\label{eq:app-wick-monomial}
p^{N_p}n^{N_n}m^{N_m}\sigma^{N_U}.
\end{equation}
For example, the three pairings of the ring representing
$\Tr[\mU^\top\mU(\mX\mX^\top)^2]$ are shown in
\cref{fig:diagrams}. The contracted graph remains connected,
since it is obtained by identifying vertices of the original
connected ring.

There are finitely many mergers and pairings at each fixed $s$.
Every resulting term has the form \cref{eq:app-wick-monomial},
with an even $N_U$; the denominators in
\cref{eq:app-pop-coefficient,eq:app-train-coefficient} introduce
only integer negative powers of $p,m$. This proves the claimed
Laurent-polynomial form and the common factor $\eta^s$.

Finally, averaging may be interchanged with each fixed-order
derivative at $t=0$. Indeed, with $\mA=\mU^\top\mU$ and
$\mS=\mX\mX^\top/m$, the flow satisfies
\begin{equation}
\frac{d}{dt}\|\mU(t)\|_F^2
=\frac{4\eta}{p}\Tr[\mS(\mA-\mA^2)]
\leq\frac{\eta}{p}\Tr\mS,
\end{equation}
because $\mA-\mA^2\preceq\mI/4$. On a bounded positive time
interval, the weight norm and every fixed-order loss derivative
therefore admit polynomial bounds in the Gaussian initial
variables. Their finite moments justify differentiation under
the expectation.
\end{proof}

\subsection{Monomials that can be leading}
\label{sec:app-pareto}

Let $Q_{\ell,s}$ be the exponent set of $D_{\ell,s}$ in
\cref{eq:polynomial-setting}. Its five coordinates correspond,
in order, to $p,n,m,\sigma,\eta$. We say that
$\vq'\in Q_{\ell,s}$ \emph{dominates}
$\vq\in Q_{\ell,s}$ if
\begin{equation}
q'_\sigma=q_\sigma,\qquad q'_\eta=q_\eta,\qquad
q'_j\geq q_j\quad(j\in\{p,n,m\}),
\end{equation}
with at least one strict inequality. An exponent is
\emph{Pareto-optimal} if no other exponent in $Q_{\ell,s}$
dominates it; the set of all such exponents is the \emph{Pareto set}
$P_{\ell,s}$. We compare only equal powers of $\sigma$, since
$\alpha_\sigma$ may have either sign; the powers of $\eta$ are
automatically equal at each order $s$. Because
$\alpha_p,\alpha_n,\alpha_m>0$, dominance implies
$\vq'\cdot\valpha>
\vq\cdot\valpha$. Hence every leading
exponent belongs to $P_{\ell,s}$.

\begin{prop}[Leading monomials]\label{prop:app-leading-monomials}
For every $s\geq1$, the train and population Pareto sets consist
exactly of the vectors
\begin{equation}\label{eq:app-pareto-parametrization}
\vq_s(u,v,r)
=\bigl(u-r,\ v+1,\ r-s,\ 2(u+v+1),\ s\bigr),
\qquad u,v\geq0,\quad u+v\leq s+1,
\end{equation}
where $r\in\{0,\ldots,s\}$ for the train loss and
$r\in\{1,\ldots,s\}$ for the population loss. Every listed
vector occurs with a nonzero coefficient and belongs to the
leading support for some admissible scaling vector.
\end{prop}

\begin{proof}
Consider the expansions
\cref{eq:app-pop-coefficient,eq:app-train-coefficient}, and let
$s_D$ count the factors of type $D$ or $\widehat D$ among the
$s+1$ factors in a term. Put $\delta_{\mathrm{tr}}=1$ and
$\delta_{\mathrm{pop}}=0$. Any resulting ring has $s_D+1$
$n$-vertices, $s+\delta_\ell$ $m$-vertices,
$2(s_D+1)$ $\mU$-edges, and $2(s+\delta_\ell)$
$\mX$-edges. These counts hold regardless of the merger choices.

For a Wick pairing, denote by $N_p,N_n,N_m$ the numbers of
vertices after contraction. Set
\begin{equation}
v=N_n-1,\qquad u=s_D-v,\qquad r=N_m-\delta_\ell.
\end{equation}
The vertex counts give $u,v\geq0$, $u+v=s_D\leq s+1$, and
the stated ranges of $r$. The contracted graph is connected and
has $E=(s_D+1)+(s+\delta_\ell)$ edges, one for each paired
pair. Thus it has at most $E+1$ vertices, and
\begin{equation}\label{eq:app-tree-bound}
N_p\leq E+1-N_n-N_m=s+u-r+1.
\end{equation}
By \cref{eq:app-pop-coefficient,eq:app-train-coefficient}, the
exponent of this term is
\begin{equation}
\vq
=\bigl(N_p-s-1,\ v+1,\ r-s,\ 2(u+v+1),\ s\bigr).
\end{equation}
It is dominated by $\vq_s(u,v,r)$ if that vector
is realized.

We now construct a pairing with equality in
\cref{eq:app-tree-bound}. Use the two-arc merger choices described
in \cref{sec:applossexp}. In the resulting ring, the $\mX$-arc
contains $s+\delta_\ell$ $m$-vertices and
$s+\delta_\ell-1$ internal $p$-vertices. Pair the two
$\mX$-edges incident to each of $s-r$ selected internal
$p$-vertices. Each such pairing identifies neighboring
$m$-vertices and leaves its $p$-vertex as a petal. The remaining
$m$-vertices form $r+\delta_\ell$ groups, each with two
unpaired $\mX$-edges at its ends.

Likewise, the $\mU$-arc has $s_D+1$ $n$-vertices and $s_D$
internal $p$-vertices. Pair both $\mU$-edges at each of
$u=s_D-v$ selected internal $p$-vertices. This leaves $v+1$
groups of $n$-vertices and $u$ further petals. Finally, pair
the two unpaired edges incident to each remaining $m$- or
$n$-group. These pairings preserve the groups of $m$- and
$n$-vertices and identify all $p$-vertices of the residual
ring into one. The contracted diagram consequently has
\begin{equation}
N_m=r+\delta_\ell,\qquad N_n=v+1,\qquad
N_p=(s-r)+u+1=s+u-r+1,
\end{equation}
as required.

The coefficient of this monomial cannot vanish by cancellation.
For fixed $s_D$, every merger term has the same sign, since it
contains exactly $s+1-s_D$ factors of type $R$ or $\widehat R$;
the Wick weights and merger multiplicities are positive. Its
$\sigma$-exponent $2(s_D+1)$ fixes $s_D$, so terms with opposite
signs cannot contribute to the same exponent vector. Hence all
listed vectors occur.

Every unlisted exponent is dominated by a listed one, as shown
above. Listed vectors with distinct $s_D=u+v$ have different
$\sigma$-exponents; at fixed $s_D$, they satisfy
$q_p+q_n+q_m=s_D+1-s$ and are therefore mutually
non-dominating. This proves the claimed Pareto sets.

To see that all listed vectors can actually lead, choose
$\alpha_p=\alpha_n=\alpha_m=c>0$ and
$\alpha_\sigma=-c/2$, with arbitrary $\alpha_\eta$.
Every listed vector then has scaling exponent
$-sc+s\alpha_\eta$, and every dominated vector has a strictly
smaller exponent. Thus all listed vectors belong to the
leading support at this balanced scaling.
\end{proof}

\subsection{The normal fan and persistent transitions}
\label{sec:proofprism}

\begin{proof}[Proof of \cref{th:prism}]
The Laurent-polynomial assertion follows from
\cref{lem:app-laurent-coefficients}. By
\cref{prop:app-leading-monomials}, every exponent outside the
Pareto sets is strictly dominated for admissible
$\valpha$. It therefore cannot tie for the maximum,
and we may determine the leading supports using only the vectors
$\vq_s(u,v,r)$.

Recall that
$\alpha_\phi=\alpha_p-\alpha_m$,
$\alpha_\rho=\alpha_n+2\alpha_\sigma$, and
$\alpha_\psi=\alpha_p-\alpha_n$. Substituting
\cref{eq:app-pareto-parametrization} gives
\begin{align}
\vq_s(u,v,r)\cdot\valpha
&=\alpha_\rho-s\alpha_m+s\alpha_\eta
  +u(\alpha_\rho+\alpha_\psi)
  +v\alpha_\rho-r\alpha_\phi
  \notag\\
&=\alpha_\rho-s\alpha_m+s\alpha_\eta
  +\alpha_\rho x_1+\alpha_\psi x_2-\alpha_\phi x_3,
\label{eq:app-exposed-score}
\end{align}
where $\mathbf x=(x_1,x_2,x_3)=(u+v,u,r)$. The first three
terms on the right are common to all monomials at order $s$.
Thus the leading supports depend only on
$\vgamma=(\alpha_\phi,\alpha_\rho,\alpha_\psi)$;
in particular, $\alpha_\eta$ does not affect them.

In the $\mathbf x$ coordinates, the convex hulls of the train and
population Pareto sets are
\begin{align}
\Pi_{\mathrm{tr},s}
&=\bigl\{\mathbf x:
  0\leq x_2\leq x_1\leq s+1,\quad 0\leq x_3\leq s\bigr\},
  \label{eq:app-train-prism}\\
\Pi_{\mathrm{pop},s}
&=\bigl\{\mathbf x:
  0\leq x_2\leq x_1\leq s+1,\quad 1\leq x_3\leq s\bigr\}.
  \label{eq:app-pop-prism}
\end{align}
The Pareto monomials correspond to all lattice points in these
polytopes. By \cref{eq:app-exposed-score}, their leading supports
are the lattice points on the faces exposed by the linear score
$\alpha_\rho x_1+\alpha_\psi x_2-\alpha_\phi x_3$.

For $s\geq2$, both polytopes are triangular prisms with the same
normal fan. At $s=1$, the train polytope is still a prism, whereas
the population polytope has $x_3=1$ throughout. Its support
depends only on $\alpha_\rho,\alpha_\psi$ and introduces no
additional distinctions beyond those made by the train support.
Consequently, the joint formal phases are the cells of the
common prism normal fan.

The five prism facets are $x_3=\max x_3$,
$x_3=\min x_3$, $x_2=0$, $x_1=x_2$, and
$x_1=s+1$. Their outward normal rays, expressed in
$\vgamma$ coordinates, are respectively the
``Large-data'', ``Small-data'', ``Mean-field'',
``Narrow-latent'', and ``Free'' rays in
\cref{eq:phase-five-rays}. For example, the outward score
normal to $x_1=x_2$ is $-x_1+x_2$, which gives
$\alpha_\rho<0$ and $\alpha_\psi=-\alpha_\rho$.

Every direction $\vgamma$ is attainable with
$\alpha_p,\alpha_n,\alpha_m>0$. Indeed, given
$\vgamma$, choose $M$ sufficiently large and set
\begin{equation}
\alpha_p=M,\qquad
\alpha_m=M-\alpha_\phi,\qquad
\alpha_n=M-\alpha_\psi,\qquad
\alpha_\sigma=\frac{\alpha_\rho-\alpha_n}{2}.
\end{equation}
Hence the admissibility conditions do not remove any cell of
the normal fan.

To obtain the dual description, translate either
full-dimensional prism so that the origin lies in its interior.
Its polar, formed using the dual score coordinates
$\boldsymbol\lambda=(\alpha_\rho,\alpha_\psi,-\alpha_\phi)$,
is a triangular bipyramid. Under the invertible change from
$\boldsymbol\lambda$ to $\vgamma$, its five vertices
lie on the rays in \cref{eq:phase-five-rays}. Directions in the
relative interior of each normal cone expose one fixed prism
face; equivalently, their rays meet the relative interior of
one fixed face of the bipyramid. This proves the stated phase
description, including its five vertices, nine edges, and six
triangular faces. At $\vgamma=\boldsymbol 0$, every
Pareto monomial is leading, giving the balanced phase.

Finally, distinct cells of this normal fan expose different
faces of the train prism at every $s\geq2$. Their train
leading supports therefore differ at infinitely many orders.
In particular, every transition between adjacent formal phases
is persistent. This conclusion concerns supports of formal
coefficients and does not assume that tied monomials have a
nonzero sum after particular scaling constants are substituted.
\end{proof}

\section{Solutions of extreme regimes (\cref{sec:solutions})}
\label{app:solutions}

\subsection{Common notation, stability principles, and preliminary results}
\label{app:common-preliminaries}

This subsection fixes notation used throughout \cref{app:solutions} and isolates the
finite-dimensional facts shared by the four solved regimes.  The common
object is the empirical covariance \(\mS\), while the state of the model can
be represented either by the input-space Gram matrix \(\mA=\mU^\top\mU\) or,
in the narrow-latent regime, by the latent-space Gramians built from
\(\mU\).  The proofs below separate the deterministic stability of these
matrix ODEs from the random-matrix statements used to identify the relevant
proxy trajectories.

Throughout \cref{app:solutions}, \(N\in\mathbb N\) denotes a generic asymptotic
sequence index.  Thus the dimensions and parameters are sequences
\begin{equation}
  p=p_N,
  \qquad
  n=n_N,
  \qquad
  m=m_N,
  \qquad
  \sigma=\sigma_N,
  \label{eq:common-sequence-index}
\end{equation}
and the subscript \(N\) is suppressed whenever this causes no ambiguity.
In the mean-field and large-data subsections we sometimes use \(p\) itself
as the sequence index, which is only a notational shorthand for the same
asymptotic convention.

All four regimes use the same Gaussian weight initialization, independent
of the training matrix:
\begin{equation}
  \mU_N(0)=\sigma_N\mZ_N,
  \qquad
  (\mZ_N)_{\alpha i}\stackrel{\mathrm{i.i.d.}}{\sim}\mathcal N(0,1),
  \qquad
  \mZ_N\ \text{independent of }\mX_N.
  \label{eq:common-Gaussian-init}
\end{equation}
The regime-specific scalings below specify the relevant rescaled variance,
for example \(n_N\sigma_N^2\) or \(p_N\sigma_N^2\), rather than redefining
the initialization.

For a training matrix \(\mX\in\mathbb R^{p\times m}\), weight matrix
\(\mU\in\mathbb R^{n\times p}\), and population covariance
\(\mSigma=\E[\vx\vx^\top]\), define
\begin{equation}
  \mS:=\frac{\mX\mX^\top}{m},
  \qquad
  \mA:=\mU^\top\mU,
  \qquad
  \mE:=\mI_p-\mA,
  \qquad
  \gamma:=\frac{\eta}{p}.
  \label{eq:common-notation}
\end{equation}
The empirical and population losses are
\begin{equation}
  \widehat L(\mU)
  =\frac{1}{2p}\Tr(\mE\mS\mE),
  \qquad
  L(\mU)
  =\frac{1}{2p}\Tr(\mE\mSigma\mE).
  \label{eq:common-losses}
\end{equation}
In the isotropic case \(\mSigma=\mI_p\), this reduces to
\begin{equation}
  L(\mU)=\frac{1}{2p}\Tr(\mE^2).
  \label{eq:losses_as_E}
\end{equation}

\paragraph{Common proof architecture.}
All four solved regimes are perturbative stability arguments around a proxy
that becomes exactly solvable in the corresponding extreme limit.  There
are two recurring mechanisms.  In the mean-field regime, the perturbation
is on the state side: \(\mA(0)\) is close to a scalar matrix, and stability
with respect to the initial Gram matrix transfers the explicit commuting
solution to the true trajectory.  The narrow-latent regime uses the same
state-side idea after passing to latent-space Gramians: a scalarized proxy is
self-consistent because weighted Gaussian Gramians concentrate, and a
finite-time stability estimate transfers the proxy to the exact flow.  In
the large-data regime, the perturbation is on the data side:
\(\mS\) is close to \(\mI_p\), so stability with respect to \(\mS\) reduces
the dynamics to the population-covariance flow.  In the small-data regime,
a rotation to the training span turns the normalized sample Gram matrix
\(\bm{\Gamma}=p^{-1}\mX^\top\mX\) into the data-side parameter; the exactly
orthogonal case \(\bm{\Gamma}=\mI_m\) produces a closed hierarchy of matrix
sequences \(\tQ\), and stability with respect to \(\bm{\Gamma}\) transfers that
hierarchy to near-isometric data.

Thus each proof has the same three layers: identify an exactly tractable
proxy, prove that either the state or the data geometry is close to the
proxy in operator norm, and propagate that closeness on every fixed time
interval by a dimension-free stability estimate.  The random-matrix input
is regime-specific; the deterministic ODE facts are collected next.

\paragraph{Standing spectral assumptions on the data.}
The four proofs use the same basic spectral regularity principle, after the
normalization appropriate to the regime. For a sequence of positive
semidefinite data operators
$\mS_N\in\mathbb R^{p_N\times p_N}$, write
\begin{equation}
  \mu_{\mS_N}
  :=
  \frac1{p_N}
  \sum_{i=1}^{p_N}
  \delta_{\lambda_i(\mS_N)}.
  \label{eq:common-esd}
\end{equation}
The two standing spectral hypotheses are that, for some deterministic
$K<\infty$ and deterministic probability measure $\nu$,
\begin{equation}
  \mathbb P\!\left(
    \|\mS_N\|_{\mathrm{op}}\le K
  \right)
  \longrightarrow1,
  \label{eq:common-spectrum-tight}
\end{equation}
and
\begin{equation}
  \mu_{\mS_N}
  \Longrightarrow
  \nu
  \qquad\text{in probability},
  \label{eq:common-esd-convergence}
\end{equation}
where the second convergence is weak convergence of probability measures.
Together,
\cref{eq:common-spectrum-tight}--\cref{eq:common-esd-convergence}
imply
\begin{equation}
  \operatorname{supp}\nu\subset[0,K].
  \label{eq:common-nu-support}
\end{equation}
The operator-norm bound is the input needed for dimension-free finite-time
ODE estimates; the spectral limit is needed only when a random finite-size
spectral functional is replaced by a deterministic limiting formula.

In the mean-field and narrow-latent regimes these assumptions are imposed
directly on the empirical covariance
\begin{equation}
  \mS_N=\frac1{m_N}\mX_N\mX_N^\top.
  \label{eq:common-S-sequence}
\end{equation}
In the large-data regime we assume the stronger near-isotropy condition
\begin{equation}
  \|\mS_N-\mI_{p_N}\|_{\mathrm{op}}
  \xrightarrow{\mathbb P}0.
  \label{eq:common-data-near-isometry}
\end{equation}
This automatically implies \cref{eq:common-spectrum-tight} and
\cref{eq:common-esd-convergence}, with
\begin{equation}
  \nu=\delta_1.
  \label{eq:common-near-isometry-nu}
\end{equation}

The small-data regime requires one normalization caveat.  There
\(\phi_N=p_N/m_N\to\infty\), so the unscaled empirical covariance
\(\mS_N=\mX_N\mX_N^\top/m_N\) has operator norm of order \(\phi_N\) and is
not bounded.  The dynamically relevant normalized data operator is
\begin{equation}
  \widetilde{\mS}_N
  :=\frac1{\phi_N}\mS_N
  =\frac1{p_N}\mX_N\mX_N^\top,
  \qquad
  \bm{\Gamma}_N
  :=\frac1{p_N}\mX_N^\top\mX_N.
  \label{eq:common-small-data-normalization}
\end{equation}
The nonzero eigenvalues of \(\widetilde{\mS}_N\) and
\(\bm{\Gamma}_N\) coincide.  The small-data assumption is the active-space
near-isometry
\begin{equation}
  \|\bm{\Gamma}_N-\mI_{m_N}\|_{\mathrm{op}}
  \xrightarrow{\mathbb P}0.
  \label{eq:common-data-Gram-near-isometry}
\end{equation}
Thus the active spectral law converges to \(\delta_1\) and
\(\|\widetilde{\mS}_N\|_{\mathrm{op}}\to1\) in probability.  If one
instead uses the ambient \(p_N\times p_N\) empirical spectral distribution
of \(\widetilde{\mS}_N\), its limit is \(\delta_0\), because
\(m_N/p_N\to0\) and only \(m_N\) eigenvalues can be nonzero.  
Thus the spectral-limit convention is as follows: in the mean-field and
narrow-latent regimes the normalized data operator may have a general
limiting law $\nu$ (denoted $\nu_\phi$ in the corresponding regime
statements); in the large-data regime the limiting law is $\delta_1$; and
in the small-data regime the active law is $\delta_1$ (while the ambient
law of the normalized operator is $\delta_0$).  
Equivalently, all four regimes use a bounded normalized data
operator and a limiting spectral law; only the normalization differs in the
small-data corner.

\begin{prop}[Common gradient-flow identities]
\label{prop:common-gradient-flow}
For every fixed symmetric \(\mS\succeq0\), the gradient flow of the empirical
loss in \cref{eq:common-losses} is
\begin{equation}
  \dot{\mU}
  =\gamma\mU\left[\mS(\mI_p-\mA)+(\mI_p-\mA)\mS\right].
  \label{eq:common-U-flow}
\end{equation}
Equivalently, if
\begin{equation}
  \mathcal G_{\mS}(\mA)
  :=\mS(\mI_p-\mA)+(\mI_p-\mA)\mS
  =2\mS-\mS\mA-\mA\mS,
  \label{eq:common-G-def}
\end{equation}
then \(\dot{\mU}=\gamma\mU\mathcal G_{\mS}(\mA)\).
\end{prop}

\begin{proof}
Since \(d\mE=-d\mA\),
\begin{equation}
  d\widehat L
  =-\frac{1}{2p}\Tr\!\left[d\mA(\mS\mE+\mE\mS)\right].
  \label{eq:common-dL-dA}
\end{equation}
Using
\begin{equation}
  d\mA=d\mU^\top\mU+\mU^\top d\mU
  \label{eq:common-dA}
\end{equation}
and the symmetry of \(\mS\mE+\mE\mS\) gives
\begin{equation}
  \frac{\partial\widehat L}{\partial\mU}
  =-\frac1p\mU(\mS\mE+\mE\mS).
  \label{eq:common-gradient}
\end{equation}
Multiplying by the learning rate \(\eta\) yields
\cref{eq:common-U-flow}.
\end{proof}

\begin{lemma}[Closed Gram dynamics and a priori bound]
\label{lem:common-gram-flow}
Let \(\mU\) solve \cref{eq:common-U-flow} with \(\gamma\ge0\), and set
\(\mA=\mU^\top\mU\).  Then \(\mA\) satisfies the closed matrix ODE
\begin{equation}
  \dot{\mA}=\gamma\mathcal F_{\mS}(\mA),
  \qquad
  \mathcal F_{\mS}(\mA)
  :=2(\mS\mA+\mA\mS)-\mS\mA^2-\mA^2\mS-2\mA\mS\mA.
  \label{eq:common-A-flow}
    \end{equation}
If \(\mA(0)\succeq0\), the solution exists globally, remains positive
semidefinite, and obeys
\begin{equation}
  0\preceq\mA(t),
  \qquad
  \|\mA(t)\|_{\mathrm{op}}
  =\|\mU(t)\|_{\mathrm{op}}^2
  \le R_0:=\max\{1,\|\mA(0)\|_{\mathrm{op}}\},
  \qquad t\ge0.
  \label{eq:common-apriori}
      \end{equation}
\end{lemma}

\begin{proof}
Set
\begin{equation}
  \mathcal G_{\mS}(\mA)
  :=\mS(\mI_p-\mA)+(\mI_p-\mA)\mS
  =2\mS-\mS\mA-\mA\mS.
  \label{eq:common-G-proof-def}
\end{equation}
Since \(\mA\) and \(\mS\) are symmetric,
\(\mathcal G_{\mS}(\mA)\) is symmetric as well.  Differentiating
\(\mA=\mU^\top\mU\) and using
\(\dot{\mU}=\gamma\mU\mathcal G_{\mS}(\mA)\) gives
\begin{equation}
\begin{aligned}
  \dot{\mA}
  &=\dot{\mU}^\top\mU+\mU^\top\dot{\mU}\\
  &=\gamma\left[
      \mathcal G_{\mS}(\mA)\mA
      +\mA\mathcal G_{\mS}(\mA)
    \right].
\end{aligned}
  \label{eq:common-A-differentiation}
\end{equation}
Expanding the two products in
\cref{eq:common-A-differentiation} yields
\begin{equation}
\begin{aligned}
  \dot{\mA}
  =\gamma\bigl[
    2(\mS\mA+\mA\mS)
    -\mS\mA^2
    -\mA^2\mS
    -2\mA\mS\mA
  \bigr],
\end{aligned}
  \label{eq:common-A-flow-expanded-proof}
\end{equation}
which is exactly \cref{eq:common-A-flow}.

Positive semidefiniteness is automatic along the \(\mU\)-flow, since
\begin{equation}
  \mA(t)=\mU(t)^\top\mU(t)\succeq0.
  \label{eq:common-A-positive}
\end{equation}
It remains to prove the dimension-free upper bound.  Define
\begin{equation}
  \alpha(t):=\lambda_{\max}(\mA(t)).
  \label{eq:common-alpha}
\end{equation}
For a differentiable curve of real symmetric matrices, the upper right
Dini derivative of the largest eigenvalue is
\begin{equation}
  D^+\alpha(t)
  =\max_{\substack{
      \vv\in E_{\max}(\mA(t))\\
      \|\vv\|_2=1
    }}
    \vv^\top\dot{\mA}(t)\vv,
  \label{eq:common-eigenvalue-derivative}
\end{equation}
where \(E_{\max}(\mA)\) denotes the top eigenspace of \(\mA\).
For completeness, we verify this formula.  Let \(h_j\downarrow0\) be a
sequence realizing the upper right limsup, and let \(\vv_j\) be a unit top
eigenvector of \(\mA(t+h_j)\).  Passing to a subsequence if necessary,
compactness of the unit sphere gives
\begin{equation}
  \vv_j\longrightarrow\vv,
  \qquad
  \vv\in E_{\max}(\mA(t)).
  \label{eq:common-top-vector-limit}
\end{equation}
Since
\begin{equation}
  \vv_j^\top\mA(t)\vv_j\le\alpha(t),
  \label{eq:common-Rayleigh-upper-at-t}
\end{equation}
we have
\begin{equation}
\begin{aligned}
  \limsup_{h\downarrow0}
  \frac{\alpha(t+h)-\alpha(t)}{h}
  &\le
  \lim_{j\to\infty}
  \vv_j^\top
  \frac{\mA(t+h_j)-\mA(t)}{h_j}
  \vv_j\\
  &=\vv^\top\dot{\mA}(t)\vv.
\end{aligned}
  \label{eq:common-eigenvalue-upper}
\end{equation}
Conversely, for any fixed unit
\(\vv\in E_{\max}(\mA(t))\), the variational characterization of the
largest eigenvalue gives
\begin{equation}
  \alpha(t+h)-\alpha(t)
  \ge
  \vv^\top\bigl(\mA(t+h)-\mA(t)\bigr)\vv.
  \label{eq:common-Rayleigh-lower}
\end{equation}
Therefore
\begin{equation}
  \liminf_{h\downarrow0}
  \frac{\alpha(t+h)-\alpha(t)}{h}
  \ge
  \vv^\top\dot{\mA}(t)\vv.
  \label{eq:common-eigenvalue-lower}
\end{equation}
Taking the maximum over the top eigenspace proves
\cref{eq:common-eigenvalue-derivative}.

Now fix a unit \(\vv\in E_{\max}(\mA(t))\), so that
\begin{equation}
  \mA(t)\vv=\alpha(t)\vv.
  \label{eq:common-top-eigenvector}
\end{equation}
Using \cref{eq:common-A-differentiation} and the symmetry of
\(\mathcal G_{\mS}(\mA)\),
\begin{equation}
  \vv^\top\dot{\mA}\vv
  =2\gamma\alpha(t)\,
  \vv^\top\mathcal G_{\mS}(\mA)\vv.
  \label{eq:common-top-derivative-first}
\end{equation}
Moreover,
\begin{equation}
\begin{aligned}
  \vv^\top\mathcal G_{\mS}(\mA)\vv
  &=\vv^\top\mS(\mI_p-\mA)\vv
    +\vv^\top(\mI_p-\mA)\mS\vv\\
  &=2(1-\alpha(t))\vv^\top\mS\vv.
\end{aligned}
  \label{eq:common-top-derivative-second}
\end{equation}
Combining
\cref{eq:common-eigenvalue-derivative},
\cref{eq:common-top-derivative-first}, and
\cref{eq:common-top-derivative-second} yields
\begin{equation}
  D^+\alpha(t)
  =\max_{\substack{
      \vv\in E_{\max}(\mA(t))\\
      \|\vv\|_2=1
    }}
    4\gamma\alpha(t)(1-\alpha(t))\vv^\top\mS\vv.
  \label{eq:common-alpha-derivative}
\end{equation}
Because \(\gamma\ge0\) and \(\mS\succeq0\), the right-hand side is
nonpositive whenever \(\alpha(t)\ge1\).  The Dini-derivative comparison
principle therefore prevents \(\alpha\) from crossing upward through the
level \(\max\{1,\alpha(0)\}\), and hence
\begin{equation}
  \alpha(t)\le\max\{1,\alpha(0)\}
  \qquad
  \text{for every }t\ge0.
  \label{eq:common-alpha-barrier}
\end{equation}
Since \(\mA(t)\succeq0\), its operator norm equals its largest eigenvalue,
so
\begin{equation}
  \|\mA(t)\|_{\mathrm{op}}
  \le
  R_0:=\max\{1,\|\mA(0)\|_{\mathrm{op}}\}.
  \label{eq:common-A-bound-proof}
\end{equation}
Finally,
\begin{equation}
  \|\mU(t)\|_{\mathrm{op}}^2
  =\|\mU(t)^\top\mU(t)\|_{\mathrm{op}}
  =\|\mA(t)\|_{\mathrm{op}}
  \le R_0.
  \label{eq:common-U-bound-proof}
\end{equation}
In every fixed finite dimension the vector field in
\cref{eq:common-U-flow} is polynomial, hence locally Lipschitz, in the
entries of \(\mU\).  A finite maximal existence time could therefore occur
only through blow-up of \(\|\mU(t)\|_{\mathrm{op}}\), which is excluded by
\cref{eq:common-U-bound-proof}.  Thus the solution exists globally, and
\cref{eq:common-apriori} follows.
\end{proof}

\begin{lemma}[Finite-time stability on bounded sets]
\label{lem:common-ode-stability}
Let \(Y\) and \(\overline Y\) be trajectories in a finite-dimensional
normed space satisfying
\begin{equation}
  \dot Y=F(Y;\theta)+R(t),
  \qquad
  \dot{\overline Y}=F(\overline Y;\overline\theta),
  \label{eq:common-abstract-stability-odes}
\end{equation}
and suppose both trajectories stay in a set on which
\begin{equation}
  \|F(Y;\theta)-F(\widetilde Y;\overline\theta)\|
  \le L\|Y-\widetilde Y\|+M\|\theta-\overline\theta\|.
  \label{eq:common-abstract-Lipschitz}
\end{equation}
Then, for every \(t\ge0\),
\begin{equation}
  \|Y(t)-\overline Y(t)\|
  \le e^{Lt}\left[
  \|Y(0)-\overline Y(0)\|
  +\int_0^t\left(M\|\theta-\overline\theta\|+\|R(s)\|\right)ds
  \right].
  \label{eq:common-abstract-stability-bound}
\end{equation}
\end{lemma}

\begin{proof}
Set
\begin{equation}
    E(t):=Y(t)-\overline Y(t),
    \qquad
    \Delta\theta:=\theta-\overline\theta,
    \qquad
    d(t):=\|E(t)\|.
    \label{eq:common-stability-error-def}
\end{equation}
Subtracting the two equations in
\cref{eq:common-abstract-stability-odes} and integrating from \(0\)
to \(t\) gives
\begin{equation}
\begin{aligned}
    E(t)
    ={}&
    E(0)
    +
    \int_0^t
    \left[
        F(Y(s);\theta)
        -
        F(\overline Y(s);\overline\theta)
    \right]ds
    +
    \int_0^t R(s)\,ds .
\end{aligned}
\label{eq:common-stability-error-integral}
\end{equation}
Since both trajectories remain in the set on which
\cref{eq:common-abstract-Lipschitz} holds, we may apply that bound
at \(Y(s)\) and \(\overline Y(s)\).  Taking norms in
\cref{eq:common-stability-error-integral} therefore yields
\begin{equation}
    d(t)
    \le
    d(0)
    +
    L\int_0^t d(s)\,ds
    +
    \int_0^t
    \left(
        M\|\Delta\theta\|
        +
        \|R(s)\|
    \right)ds .
    \label{eq:common-stability-error-ineq}
\end{equation}
Define
\begin{equation}
    g(s)
    :=
    M\|\Delta\theta\|+\|R(s)\|
\label{eq:common-stability-g-def}
\end{equation}
and
\begin{equation}
    h(t)
    :=
    d(0)
    +
    \int_0^t g(s)\,ds
    +
    L\int_0^t d(s)\,ds .
    \label{eq:common-stability-h-def}
\end{equation}
Then \cref{eq:common-stability-error-ineq} gives
\begin{equation}
    d(t)\le h(t).
    \label{eq:common-stability-dh}
\end{equation}
Moreover, \(h\) is absolutely continuous and, for almost every \(t\),
\begin{equation}
    h'(t)
    =
    g(t)+Ld(t)
    \le
    g(t)+Lh(t).
    \label{eq:common-stability-h-differential}
\end{equation}
Multiplying by \(e^{-Lt}\), we obtain
\begin{equation}
    \frac{d}{dt}
    \left(
        e^{-Lt}h(t)
    \right)
    \le
    e^{-Lt}g(t)
    \qquad
    \text{for almost every }t.
    \label{eq:common-stability-integrating-factor}
\end{equation}
Integrating once more and using \(h(0)=d(0)\) gives
\begin{equation}
    h(t)
    \le
    e^{Lt}
    \left[
        d(0)
        +
        \int_0^t e^{-Ls}g(s)\,ds
    \right].
    \label{eq:common-stability-h-bound}
\end{equation}
Combining \cref{eq:common-stability-dh} and
\cref{eq:common-stability-h-bound} yields the slightly sharper estimate
\begin{equation}
\begin{aligned}
    \|Y(t)-\overline Y(t)\|
    \le{}&
    e^{Lt}\|Y(0)-\overline Y(0)\|
    \\
    &+
    \int_0^t
    e^{L(t-s)}
    \left(
        M\|\theta-\overline\theta\|
        +
        \|R(s)\|
    \right)ds .
\end{aligned}
\label{eq:common-abstract-stability-sharp}
\end{equation}
Finally, since \(L\ge0\) and hence
\(e^{L(t-s)}\le e^{Lt}\) for \(0\le s\le t\),
\cref{eq:common-abstract-stability-sharp} implies
\begin{equation}
    \|Y(t)-\overline Y(t)\|
    \le
    e^{Lt}
    \left[
        \|Y(0)-\overline Y(0)\|
        +
        \int_0^t
        \left(
            M\|\theta-\overline\theta\|
            +
            \|R(s)\|
        \right)ds
    \right],
    \label{eq:common-abstract-stability-bound-proof}
\end{equation}
which is \cref{eq:common-abstract-stability-bound}.
\end{proof}

\begin{lemma}[Stability with respect to the Gram state]
\label{lem:common-A-stability}
Let \(\mA(t)\) and \(\mB(t)\) solve
\begin{equation}
  \dot{\mA}=\gamma\mathcal F_{\mS}(\mA),
  \qquad
  \dot{\mB}=\gamma\mathcal F_{\mS}(\mB),
  \label{eq:common-two-A-flows}
\end{equation}
with positive semidefinite initial conditions.  Set
\begin{equation}
  K_{\mS}:=\|\mS\|_{\mathrm{op}},
  \qquad
  R:=\max\{1,\|\mA(0)\|_{\mathrm{op}},\|\mB(0)\|_{\mathrm{op}}\}.
  \label{eq:common-A-stability-constants}
\end{equation}
Then
\begin{equation}
  \|\mA(t)-\mB(t)\|_{\mathrm{op}}
  \le
  \exp\!\left(\gamma K_{\mS}(4+8R)t\right)
  \|\mA(0)-\mB(0)\|_{\mathrm{op}}.
  \label{eq:common-A-stability-bound}
  \end{equation}
\end{lemma}

\begin{proof}
\cref{lem:common-gram-flow} bounds both trajectories by \(R\).  If
\(\mD=\mA-\mB\), then
\begin{equation}
  \|\mathcal F_{\mS}(\mA)-\mathcal F_{\mS}(\mB)\|_{\mathrm{op}}
  \le K_{\mS}(4+8R)\|\mD\|_{\mathrm{op}}.
  \label{eq:common-F-A-Lipschitz}
\end{equation}
Indeed,
\begin{equation}
  \mA^2-\mB^2=\mA\mD+\mD\mB,
  \qquad
  \mA\mS\mA-\mB\mS\mB
  =\mD\mS\mA+\mB\mS\mD,
  \label{eq:common-matrix-differences}
\end{equation}
and the bound follows by submultiplicativity.  Applying
\cref{lem:common-ode-stability} with no parameter perturbation and no
residual gives \cref{eq:common-A-stability-bound}.
\end{proof}

\begin{prop}[Stability with respect to the data covariance]
\label{prop:common-S-stability}
Let \(\mS,\mT\succeq0\), let \(\mA_{\mS}\) and \(\mA_{\mT}\) solve
\begin{equation}
  \dot{\mA}_{\mS}=\gamma\mathcal F_{\mS}(\mA_{\mS}),
  \qquad
  \dot{\mA}_{\mT}=\gamma\mathcal F_{\mT}(\mA_{\mT}),
  \qquad
  \mA_{\mS}(0)=\mA_{\mT}(0)=\mA_0\succeq0.
  \label{eq:common-S-stability-flows}
\end{equation}
Set
\begin{equation}
  R:=\max\{1,\|\mA_0\|_{\mathrm{op}}\},
  \qquad
  K:=\max\{\|\mS\|_{\mathrm{op}},\|\mT\|_{\mathrm{op}}\}.
  \label{eq:common-S-stability-constants}
\end{equation}
Then, for every \(T_0<\infty\),
\begin{equation}
  \sup_{0\le t\le T_0}
  \|\mA_{\mS}(t)-\mA_{\mT}(t)\|_{\mathrm{op}}
  \le C_{\gamma,T_0,K,R}\|\mS-\mT\|_{\mathrm{op}},
  \label{eq:common-S-stability-bound}
\end{equation}
where
\begin{equation}
  C_{\gamma,T_0,K,R}
  :=(4R+4R^2)
  \frac{e^{\gamma K(4+8R)T_0}-1}{K(4+8R)},
  \label{eq:common-S-stability-constant}
\end{equation}
with the continuous value \(\gamma T_0(4R+4R^2)\) when \(K=0\).
In particular, if \(\gamma=1\), \(\mT=\mI_p\), and
\(\|\mS-\mI_p\|_{\mathrm{op}}\le1\), then
\begin{equation}
  \sup_{0\le t\le T_0}
  \|\mA_{\mS}(t)-\mA_{\mI}(t)\|_{\mathrm{op}}
  \le K_{T_0,R}\|\mS-\mI_p\|_{\mathrm{op}},
  \label{eq:ld-stability-bound}
\end{equation}
where one may take
\begin{equation}
  K_{T_0,R}
  :=(4R+4R^2)\frac{e^{(8+16R)T_0}-1}{8+16R}.
  \label{eq:ld-KTR}
\end{equation}
\end{prop}

\begin{proof}
On the ball \(\|\mA\|_{\mathrm{op}}\le R\),
\begin{equation}
  \|\mathcal F_{\mS}(\mA)-\mathcal F_{\mS}(\mB)\|_{\mathrm{op}}
  \le K(4+8R)\|\mA-\mB\|_{\mathrm{op}},
  \label{eq:common-S-state-Lipschitz}
\end{equation}
while direct subtraction of the two covariance parameters gives
\begin{equation}
  \|\mathcal F_{\mS}(\mA)-\mathcal F_{\mT}(\mA)\|_{\mathrm{op}}
  \le(4R+4R^2)\|\mS-\mT\|_{\mathrm{op}}.
  \label{eq:common-F-S-Kipschitz}
\end{equation}
\cref{lem:common-gram-flow} keeps both trajectories in this ball.
\cref{lem:common-ode-stability}, applied with parameter
\(\theta=\mS\), yields \cref{eq:common-S-stability-bound}; the final
specialization uses \(K\le2\).
\end{proof}

\begin{lemma}[Stability of population and train losses]
\label{lem:common-loss-stability}
Let \(\mA,\mB\succeq0\) with
\begin{equation}
  \|\mA\|_{\mathrm{op}}\le R,
  \qquad
  \|\mB\|_{\mathrm{op}}\le R.
  \label{eq:common-loss-stability-assumption}
\end{equation}
For isotropic population covariance, define
\begin{equation}
  L_p(\mA):=\frac1{2p}\Tr(\mI_p-\mA)^2,
  \qquad
  \widehat L_{p,\mS}(\mA)
  :=\frac1{2p}\Tr\!\left[\mS(\mI_p-\mA)^2\right].
  \label{eq:common-loss-functionals}
\end{equation}
Then
\begin{equation}
  |L_p(\mA)-L_p(\mB)|
  \le(1+R)\|\mA-\mB\|_{\mathrm{op}},
  \label{eq:common-pop-loss-Lipschitz}
  \end{equation}
and, for \(\mS\succeq0\),
\begin{equation}
  |\widehat L_{p,\mS}(\mA)-\widehat L_{p,\mS}(\mB)|
  \le\|\mS\|_{\mathrm{op}}(1+R)
  \|\mA-\mB\|_{\mathrm{op}}.
  \label{eq:common-train-loss-Lipschitz}
\end{equation}
Moreover,
\begin{equation}
  |\widehat L_{p,\mS}(\mA)-L_p(\mA)|
  \le\frac12(1+R)^2\|\mS-\mI_p\|_{\mathrm{op}}.
  \label{eq:ld-train-pop-bound}
\end{equation}
Consequently, under the specialization in
\cref{prop:common-S-stability},
\begin{equation}
  \sup_{0\le t\le T_0}
  |L_p(\mA_{\mS}(t))-L_p(\mA_{\mI}(t))|
  \le(1+R)K_{T_0,R}\|\mS-\mI_p\|_{\mathrm{op}},
  \label{eq:ld-pop-proxy-bound}
\end{equation}
and
\begin{equation}
  \sup_{0\le t\le T_0}
  |\widehat L_{p,\mS}(\mA_{\mS}(t))-L_p(\mA_{\mI}(t))|
  \le\left[\frac12(1+R)^2+(1+R)K_{T_0,R}\right]
  \|\mS-\mI_p\|_{\mathrm{op}}.
  \label{eq:ld-train-proxy-bound}
\end{equation}
\end{lemma}

\begin{proof}
Set \(\mE_{\mA}=\mI_p-\mA\) and
\(\mE_{\mB}=\mI_p-\mB\).  Then
\begin{equation}
  \mE_{\mA}^2-\mE_{\mB}^2
  =\mE_{\mA}(\mE_{\mA}-\mE_{\mB})
  +(\mE_{\mA}-\mE_{\mB})\mE_{\mB},
  \label{eq:common-E-square-difference}
\end{equation}
so
\begin{equation}
  \|\mE_{\mA}^2-\mE_{\mB}^2\|_{\mathrm{op}}
  \le2(1+R)\|\mA-\mB\|_{\mathrm{op}}.
  \label{eq:common-E-square-bound}
\end{equation}
The first two inequalities follow from
\(|\Tr\mM|\le p\|\mM\|_{\mathrm{op}}\).  For the covariance perturbation,
\begin{equation}
  \widehat L_{p,\mS}(\mA)-L_p(\mA)
  =\frac1{2p}\Tr\!\left[(\mS-\mI_p)(\mI_p-\mA)^2\right].
  \label{eq:common-train-pop-difference}
\end{equation}
Since \(\mQ:=(\mI_p-\mA)^2\succeq0\),
\begin{equation}
  |\Tr(\mM\mQ)|\le\|\mM\|_{\mathrm{op}}\Tr\mQ,
  \qquad
  \Tr\mQ\le p(1+R)^2,
  \label{eq:common-positive-trace-bound}
\end{equation}
which proves \cref{eq:ld-train-pop-bound}.  The last two inequalities
combine the preceding estimates with
\cref{prop:common-S-stability}.
\end{proof}

\begin{lemma}[Commuting logistic flow]
\label{lem:common-commuting-flow}
Assume \(\mA_0,\mS\succeq0\) commute, and let
\begin{equation}
  \dot{\mA}=\gamma\mathcal F_{\mS}(\mA),
  \qquad
  \mA(0)=\mA_0.
  \label{eq:common-commuting-IVP}
\end{equation}
Then \(\mA(t)\) continues to commute with \(\mS\).  In a common orthonormal
eigenbasis,
\begin{equation}
  \mS=\mV\operatorname{diag}(\lambda_1,\ldots,\lambda_p)\mV^\top,
  \qquad
  \mA_0=\mV\operatorname{diag}(a_1(0),\ldots,a_p(0))\mV^\top,
  \label{eq:common-simultaneous-diagonalization}
\end{equation}
the eigenvalues solve
\begin{equation}
  \dot a_i(t)=4\gamma\lambda_i a_i(t)(1-a_i(t)),
  \qquad
  a_i(t)=\frac{a_i(0)}{a_i(0)+(1-a_i(0))e^{-4\gamma\lambda_i t}}.
  \label{eq:common-logistic-modes}
  \end{equation}
Zero eigenvalues of \(\mA_0\) remain zero.
\end{lemma}

\begin{proof}
On commuting matrices,
\begin{equation}
  \mathcal F_{\mS}(\mA)=4\mS\mA(\mI_p-\mA).
  \label{eq:common-F-commuting}
\end{equation}
Therefore the common eigenbasis does not rotate, and the matrix ODE reduces
to the scalar logistic equations in \cref{eq:common-logistic-modes}.
Separation of variables gives the stated formula, with the cases
\(a_i(0)\in\{0,1\}\) obtained by continuity.  Uniqueness of the matrix ODE
completes the argument.
\end{proof}

\subsection{Mean-field (\cref{sec:mean-field})}
\label{app:mean-field}

We now specialize the common setup of \cref{app:common-preliminaries} to the mean-field
corner.  All occurrences of \(\mS_p,\mA_p,\mE_p\), the train and population
losses, and the gradient flow use the notation and identities of
\cref{eq:common-notation}--\cref{eq:common-U-flow}; in particular the
learning-rate scaling \(\eta=p\) gives \(\gamma=1\).  We write
\(L_p(t):=L(\mU_p(t))\) and
\(\widehat L_p(t):=\widehat L(\mU_p(t))\).

The regime-specific assumptions are
\begin{equation}
  p,n,m\longrightarrow\infty,
  \qquad
  \frac pn\longrightarrow0,
  \qquad
  \frac{p}{m} \to \phi \in (0,\infty),
  \qquad
  \rho_p:=n\sigma_p^2\longrightarrow\rho\in(0,\infty).
  \label{eq:mf-scaling}
\end{equation}
These are the only mean-field-specific dimension and variance scalings.
For the population-loss statement we first impose isotropy,
\begin{equation}
  \mSigma_p:=\E[\vx\vx^\top]=\mI_p,
  \label{eq:mf-pop-isotropy}
\end{equation}
and return to anisotropic population covariance at the end of the
subsection.

The data covariance \(\mS_p\) is assumed to satisfy the two standing
spectral hypotheses \cref{eq:common-spectrum-tight} and
\cref{eq:common-esd-convergence}, with limiting law denoted by \({\nu_\phi}\).  The
operator-norm bound gives finite-time stability, while
\(\mu_{\mS_p}\Rightarrow{\nu_\phi}\) identifies the deterministic spectral averages
in the losses.

For $\lambda\ge0$ and $t\ge0$, define
\begin{equation}
a_t(\lambda)
:=
\frac{\rho}
{\rho+(1-\rho)e^{-4\lambda t}},
\label{eq:mf-at-def}
\end{equation}
and
\begin{equation}
e_t(\lambda)
:=
1-a_t(\lambda)
=
\frac{1-\rho}
{\rho e^{4\lambda t}+1-\rho}.
\label{eq:mf-et-def}
\end{equation}
We also define
\begin{equation}
\ell_t(\lambda)
:=
\frac12 e_t(\lambda)^2
=
\frac12
\left(
\frac{1-\rho}
{\rho e^{4\lambda t}+1-\rho}
\right)^2.
\label{eq:mf-ellt-def}
\end{equation}

\paragraph{Main theorem.}
\begin{theorem}[Finite-time mean-field limit]
\label{thm:mf-finite-time}
Assume \cref{eq:mf-scaling} and the common spectral hypotheses \cref{eq:common-spectrum-tight}--\cref{eq:common-esd-convergence}. Then, for
every fixed $T<\infty$,
\begin{equation}
\sup_{0\le t\le T}
\left\|
\mA_p(t)-a_t(\mS_p)
\right\|_{\mathrm{op}}
\xrightarrow[p\to\infty]{\mathbb P}0,
\label{eq:mf-main-matrix-convergence}
\end{equation}
where $a_t(\mS_p)$ is defined by functional calculus.

Under the population-isotropy assumption \cref{eq:mf-pop-isotropy}, it
follows that
\begin{equation}
\sup_{0\le t\le T}
\left|
L_p(t)
-
\frac12
\int e_t(\lambda)^2\,d{\nu_\phi}(\lambda)
\right|
\xrightarrow[p\to\infty]{\mathbb P}0,
\label{eq:mf-main-pop-convergence}
\end{equation}
and
\begin{equation}
\sup_{0\le t\le T}
\left|
\widehat L_p(t)
-
\frac12
\int
\lambda e_t(\lambda)^2\,d{\nu_\phi}(\lambda)
\right|
\xrightarrow[p\to\infty]{\mathbb P}0.
\label{eq:mf-main-train-convergence}
\end{equation}
\end{theorem}

\paragraph{Proof plan.}
First, Gaussian singular-value concentration shows that the random initial
Gram matrix is close to $\rho\mI_p$ in operator norm.  Second,
\cref{lem:common-A-stability} propagates this initial perturbation on
every fixed time interval.  Third, the scalar initial condition commutes with
$\mS_p$, so \cref{lem:common-commuting-flow} gives the proxy by
functional calculus.  Finally, \cref{lem:common-loss-stability}
transfers the matrix approximation to the two losses, and weak spectral
convergence of $\mS_p$ identifies their deterministic limits.

\paragraph{Common deterministic input.}
The closed Gram dynamics, global a priori bound, commuting logistic solution,
and finite-time stability with respect to the Gram state are exactly
Lemmas~\ref{lem:common-gram-flow}, \ref{lem:common-commuting-flow}, and
\ref{lem:common-A-stability} from \cref{app:common-preliminaries}.\paragraph{Concentration of the initial Gram matrix.}

\begin{lemma}
\label{lem:mf-init-concentration}
Under the standing initialization in \cref{eq:common-Gaussian-init} and \cref{eq:mf-scaling},
\begin{equation}
\|\mA_p(0)-\rho \mI_p\|_{\mathrm{op}}
\xrightarrow[p\to\infty]{\mathbb P}0.
\label{eq:mf-init-op-convergence}
\end{equation}
More precisely,
\begin{equation}
\|\mA_p(0)-\rho \mI_p\|_{\mathrm{op}}
=
O_{\mathbb P}\!\left(\sqrt{\frac pn}\right)
+
|\rho_p-\rho|.
\label{eq:mf-init-rate}
\end{equation}
\end{lemma}

\begin{proof}
By the standing initialization \cref{eq:common-Gaussian-init},
\begin{equation}
\mA_p(0)
=
\frac{\rho_p}{n}\mZ_p^\top \mZ_p.
\label{eq:mf-A0-Wishart}
\end{equation}
A standard Gaussian singular-value estimate states that if $\mZ$ is an
$n\times p$ matrix with i.i.d.\ standard Gaussian entries, then, for every
$r\ge0$,
\begin{equation}
\mathbb P
\left(
s_{\max}(\mZ)>\sqrt n+\sqrt p+r
\right)
\le
e^{-r^2/2},
\label{eq:mf-gaussian-smax}
\end{equation}
and, whenever $\sqrt n\ge\sqrt p+r$,
\begin{equation}
\mathbb P
\left(
s_{\min}(\mZ)<\sqrt n-\sqrt p-r
\right)
\le
e^{-r^2/2}.
\label{eq:mf-gaussian-smin}
\end{equation}
Taking $r=\sqrt p$ shows that, with probability at least
$1-2e^{-p/2}$ for all sufficiently large $p$,
\begin{equation}
1-2\sqrt{\frac pn}
\le
\frac{s_{\min}(\mZ_p)}{\sqrt n}
\le
\frac{s_{\max}(\mZ_p)}{\sqrt n}
\le
1+2\sqrt{\frac pn}.
\label{eq:mf-singular-sandwich}
\end{equation}
Consequently,
\begin{equation}
\left\|
\frac1n \mZ_p^\top \mZ_p-\mI_p
\right\|_{\mathrm{op}}
=
O_{\mathbb P}
\left(
\sqrt{\frac pn}
\right).
\label{eq:mf-Wishart-near-I}
\end{equation}
Using \cref{eq:mf-A0-Wishart},
\begin{equation}
\|\mA_p(0)-\rho \mI_p\|_{\mathrm{op}}
\le
\rho_p
\left\|
\frac1n\mZ_p^\top \mZ_p-\mI_p
\right\|_{\mathrm{op}}
+
|\rho_p-\rho|.
\label{eq:mf-A0-final-bound}
\end{equation}
\Cref{eq:mf-scaling,eq:mf-Wishart-near-I,eq:mf-A0-final-bound} prove the claims.
\end{proof}

\paragraph{Proof of the matrix trajectory limit.}

\begin{proof}[Proof of \cref{eq:mf-main-matrix-convergence}]
Let
\begin{equation}
\bar \mA_p(t):=a_t(\mS_p).
\label{eq:mf-Abar-def}
\end{equation}
By \cref{lem:common-commuting-flow}, $\bar \mA_p$ is the exact solution of
the same Gram ODE as $\mA_p$, with initial condition
\begin{equation}
\bar \mA_p(0)=\rho \mI_p.
\label{eq:mf-Abar-initial}
\end{equation}
Fix $T<\infty$. By \cref{lem:mf-init-concentration},
\begin{equation}
\|\mA_p(0)-\rho \mI_p\|_{\mathrm{op}}
\xrightarrow{\mathbb P}0.
\label{eq:mf-init-again}
\end{equation}
In particular, for any deterministic
$R_0>\max\{1,\rho\}$,
\begin{equation}
\mathbb P
\bigl(
\|\mA_p(0)\|_{\mathrm{op}}\le R_0
\bigr)
\longrightarrow1.
\label{eq:mf-A0-bounded-event}
\end{equation}
Intersecting this event with the event in
\cref{eq:common-spectrum-tight}, \cref{lem:common-A-stability} gives
\begin{equation}
\sup_{0\le t\le T}
\|\mA_p(t)-\bar \mA_p(t)\|_{\mathrm{op}}
\le
\exp\!\bigl(K(4+8R_0)T\bigr)
\|\mA_p(0)-\rho \mI_p\|_{\mathrm{op}}.
\label{eq:mf-trajectory-bound}
\end{equation}
The right-hand side converges to zero in probability by
\cref{eq:mf-init-again}, while the event on which the estimate holds has
probability tending to one. This proves
\cref{eq:mf-main-matrix-convergence}.
\end{proof}

The estimate \cref{eq:mf-trajectory-bound} also makes explicit why the
result is a finite-time statement. For fixed $T$ the stability factor is
harmless, but the argument by itself gives no useful control when $T$ is
allowed to diverge with the dimensions.

\paragraph{Transfer from the Gram trajectory to the losses.}
The needed population- and train-loss Lipschitz estimates are collected in
\cref{lem:common-loss-stability}.

Applying \cref{lem:common-loss-stability} to $\mA=\mA_p(t)$ and
$\mB=\bar \mA_p(t)$, and using \cref{eq:mf-main-matrix-convergence}, gives
\begin{equation}
\sup_{0\le t\le T}
\left|
L_p(t)
-
\frac1{2p}
\Tr
\bigl[
\mI_p-a_t(\mS_p)
\bigr]^2
\right|
\xrightarrow{\mathbb P}0,
\label{eq:mf-pop-quenched}
\end{equation}
and
\begin{equation}
\sup_{0\le t\le T}
\left|
\widehat L_p(t)
-
\frac1{2p}
\Tr
\left[
\mS_p
\bigl(
\mI_p-a_t(\mS_p)
\bigr)^2
\right]
\right|
\xrightarrow{\mathbb P}0.
\label{eq:mf-train-quenched}
\end{equation}
Since $a_t(\mS_p)$ is a spectral function of $\mS_p$, the comparison losses
can be written exactly as
\begin{equation}
\frac1{2p}
\Tr
\bigl[
\mI_p-a_t(\mS_p)
\bigr]^2
=
\int
\ell_t(\lambda)\,
d\mu_{\mS_p}(\lambda),
\label{eq:mf-pop-esd-exact}
\end{equation}
and
\begin{equation}
\frac1{2p}
\Tr
\left[
\mS_p
\bigl(
\mI_p-a_t(\mS_p)
\bigr)^2
\right]
=
\int
\lambda\ell_t(\lambda)\,
d\mu_{\mS_p}(\lambda).
\label{eq:mf-train-esd-exact}
\end{equation}
Thus, before taking any random-matrix limit, the full finite-dimensional
dynamics have already been reduced, up to an error tending to zero
uniformly on $[0,T]$, to spectral averages over the empirical covariance
$\mS_p$.

\paragraph{Uniform convergence of the spectral averages.}

\begin{lemma}
\label{lem:mf-uniform-spectral}
Under \cref{eq:common-spectrum-tight} and
\cref{eq:common-esd-convergence}, for every fixed $T<\infty$,
\begin{equation}
\sup_{0\le t\le T}
\left|
\int
\ell_t(\lambda)\,
d\mu_{\mS_p}(\lambda)
-
\int
\ell_t(\lambda)\,
d{\nu_\phi}(\lambda)
\right|
\xrightarrow{\mathbb P}0,
\label{eq:mf-uniform-pop-spectral}
\end{equation}
and
\begin{equation}
\sup_{0\le t\le T}
\left|
\int
\lambda\ell_t(\lambda)\,
d\mu_{\mS_p}(\lambda)
-
\int
\lambda\ell_t(\lambda)\,
d{\nu_\phi}(\lambda)
\right|
\xrightarrow{\mathbb P}0.
\label{eq:mf-uniform-train-spectral}
\end{equation}
\end{lemma}

\begin{proof}
We prove a slightly more general elementary statement. Let
\begin{equation}
h:
[0,T]\times[0,K]
\longrightarrow
\mathbb R
\label{eq:mf-h-domain}
\end{equation}
be continuous. Since its domain is compact, $h$ is uniformly continuous.
Hence, for every $\varepsilon>0$, there exists $\delta>0$ such that
\begin{equation}
|t-s|<\delta
\quad\Longrightarrow\quad
\sup_{\lambda\in[0,K]}
|h(t,\lambda)-h(s,\lambda)|
<
\frac{\varepsilon}{3}.
\label{eq:mf-h-uniform-cont}
\end{equation}
Choose a finite $\delta$-net $t_1,\ldots,t_N$ of $[0,T]$. For every
fixed $j$, weak convergence in probability in
\cref{eq:common-esd-convergence} gives
\begin{equation}
\int
h(t_j,\lambda)\,
d\mu_{\mS_p}(\lambda)
-
\int
h(t_j,\lambda)\,
d{\nu_\phi}(\lambda)
\xrightarrow{\mathbb P}0.
\label{eq:mf-grid-convergence}
\end{equation}
Since $N$ is finite,
\begin{equation}
\max_{1\le j\le N}
\left|
\int
h(t_j,\lambda)\,
d\mu_{\mS_p}(\lambda)
-
\int
h(t_j,\lambda)\,
d{\nu_\phi}(\lambda)
\right|
\xrightarrow{\mathbb P}0.
\label{eq:mf-grid-max}
\end{equation}
On the event $\|\mS_p\|_{\mathrm{op}}\le K$, for arbitrary $t\in[0,T]$
choose $t_j$ with $|t-t_j|<\delta$. Then
\begin{equation}
\begin{aligned}
\left|
\int
h(t,\lambda)\,
d(\mu_{\mS_p}-{\nu_\phi})(\lambda)
\right|
\le{}&
\int
|h(t,\lambda)-h(t_j,\lambda)|
\,d\mu_{\mS_p}(\lambda)
\\
&+
\left|
\int
h(t_j,\lambda)\,
d(\mu_{\mS_p}-{\nu_\phi})(\lambda)
\right|
\\
&+
\int
|h(t_j,\lambda)-h(t,\lambda)|
\,d{\nu_\phi}(\lambda).
\end{aligned}
\label{eq:mf-net-triangle}
\end{equation}
The first and third terms in \cref{eq:mf-net-triangle} are each less
than $\varepsilon/3$ by \cref{eq:mf-h-uniform-cont}, while the middle
term is controlled uniformly over the finite net by
\cref{eq:mf-grid-max}. Since the event
$\|\mS_p\|_{\mathrm{op}}\le K$ has probability tending to one, it follows
that
\begin{equation}
\sup_{0\le t\le T}
\left|
\int
h(t,\lambda)\,
d\mu_{\mS_p}(\lambda)
-
\int
h(t,\lambda)\,
d{\nu_\phi}(\lambda)
\right|
\xrightarrow{\mathbb P}0.
\label{eq:mf-general-uniform-spectral}
\end{equation}
Taking
\begin{equation}
h(t,\lambda)=\ell_t(\lambda)
\label{eq:mf-h-pop}
\end{equation}
and
\begin{equation}
h(t,\lambda)=\lambda\ell_t(\lambda)
\label{eq:mf-h-train}
\end{equation}
proves \cref{eq:mf-uniform-pop-spectral} and
\cref{eq:mf-uniform-train-spectral}.
\end{proof}

Combining \cref{eq:mf-pop-quenched},
\cref{eq:mf-train-quenched}, and
\cref{lem:mf-uniform-spectral} proves
\cref{eq:mf-main-pop-convergence} and
\cref{eq:mf-main-train-convergence}, and hence completes the proof of
\cref{thm:mf-finite-time}.

\begin{corollary}[Convergence of averaged losses]
\label{cor:mf-expectations}
Assume the hypotheses of \cref{thm:mf-finite-time}, including
population isotropy, and suppose in addition that, for some $\delta>0$,
\begin{equation}
    \sup_p
    \E\!\left[
        \|\mS_p\|_{\mathrm{op}}^{1+\delta}
    \right]
    <\infty.
    \label{eq:mf-Sp-UI}
\end{equation}
Then, for every fixed $T<\infty$,
\begin{equation}
    \E
    \left[
        \sup_{0\le t\le T}
        |L_p(t)-\mathcal L(t)|
    \right]
    \longrightarrow0,
    \label{eq:mf-pop-L1}
\end{equation}
and
\begin{equation}
    \E
    \left[
        \sup_{0\le t\le T}
        |\widehat L_p(t)-\widehat{\mathcal L}(t)|
    \right]
    \longrightarrow0,
    \label{eq:mf-train-L1}
\end{equation}
where
\begin{equation}
    \mathcal L(t)
    =
    \frac12
    \int
    e_t(\lambda)^2
    \,d\nu_\phi(\lambda),
    \qquad
    \widehat{\mathcal L}(t)
    =
    \frac12
    \int
    \lambda e_t(\lambda)^2
    \,d\nu_\phi(\lambda).
    \label{eq:mf-general-loss-limits}
\end{equation}
In particular, for every fixed $t\ge0$,
\begin{equation}
    \E[L_p(t)]
    \longrightarrow
    \mathcal L(t),
    \qquad
    \E[\widehat L_p(t)]
    \longrightarrow
    \widehat{\mathcal L}(t).
    \label{eq:mf-expectation-fixed-t}
\end{equation}
\end{corollary}

\begin{proof}
Set
\begin{equation}
    M_p
    :=
    \max\{1,\|\mA_p(0)\|_{\mathrm{op}}\}.
    \label{eq:mf-Mp-def}
\end{equation}
By \cref{lem:common-gram-flow},
\begin{equation}
    \sup_{t\ge0}
    \|\mA_p(t)\|_{\mathrm{op}}
    \le M_p.
    \label{eq:mf-A-Mp}
\end{equation}
Consequently,
\begin{equation}
    \sup_{t\ge0} L_p(t)
    \le
    \frac12(1+M_p)^2,
    \label{eq:mf-pop-dominator}
\end{equation}
and
\begin{equation}
    \sup_{t\ge0}\widehat L_p(t)
    \le
    \frac12
    \|\mS_p\|_{\mathrm{op}}
    (1+M_p)^2.
    \label{eq:mf-train-dominator}
\end{equation}

The Gaussian initialization and the mean-field scaling $p/n\to0$ imply,
by the same Gaussian singular-value estimate used in
\cref{lem:mf-init-concentration}, that for every fixed $q<\infty$,
\begin{equation}
    \sup_p \E[M_p^q]<\infty.
    \label{eq:mf-Mp-moments}
\end{equation}
Choose $0<\varepsilon<\delta$. Since the initialization is independent of
the training data, \cref{eq:mf-Sp-UI} and
\cref{eq:mf-Mp-moments} imply
\begin{equation}
    \sup_p
    \E
    \left[
        \left(
            \|\mS_p\|_{\mathrm{op}}
            (1+M_p)^2
        \right)^{1+\varepsilon}
    \right]
    <\infty.
    \label{eq:mf-train-dominator-moment}
\end{equation}
Together with
\cref{eq:mf-pop-dominator}--\cref{eq:mf-train-dominator},
this gives uniform integrability of the population and train losses on
every compact time interval. The deterministic limiting trajectories
$\mathcal L$ and $\widehat{\mathcal L}$ are bounded on $[0,T]$.
Their uniform convergence in probability follows from
\cref{thm:mf-finite-time}. Hence the two suprema in
\cref{eq:mf-pop-L1}--\cref{eq:mf-train-L1} are uniformly integrable
and converge to zero in probability, which upgrades the convergence to
$L^1$.
\end{proof}

\paragraph{Marchenko--Pastur specialization.}
Suppose now that the training data are isotropic Gaussian,
\begin{equation}
    (\mX_p)_{ij}
    \overset{\mathrm{i.i.d.}}{\sim}
    \mathcal N(0,1),
    \qquad
    \frac{p}{m}\longrightarrow\phi\in(0,\infty).
    \label{eq:mf-gaussian-data}
\end{equation}
Then the empirical spectral measure converges almost surely to the
Marchenko--Pastur law,
\begin{equation}
    \mu_{\mS_p}
    \Longrightarrow
    \mu_\phi,
    \label{eq:mf-MP-convergence}
\end{equation}
and
\begin{equation}
    \|\mS_p\|_{\mathrm{op}}
    \longrightarrow
    (1+\sqrt{\phi})^2
    \qquad\text{almost surely}.
    \label{eq:mf-top-edge}
\end{equation}
Here
\begin{equation}
    d\mu_\phi(\lambda)
    =
    \left(1-\frac1\phi\right)_+
    \delta_0(d\lambda)
    +
    \frac{
        \sqrt{(\lambda_+-\lambda)(\lambda-\lambda_-)}
    }{
        2\pi\phi\lambda
    }
    \mathbf 1_{[\lambda_-,\lambda_+]}(\lambda)\,d\lambda,
    \label{eq:mf-MP-law}
\end{equation}
with
\begin{equation}
    \lambda_\pm
    =
    (1\pm\sqrt{\phi})^2.
    \label{eq:mf-MP-edges}
\end{equation}
Thus the standing spectral assumptions hold with
$\nu_\phi=\mu_\phi$. Moreover, the standard Gaussian singular-value
bounds give, for every fixed $q<\infty$,
\begin{equation}
    \sup_p
    \E\!\left[
        \|\mS_p\|_{\mathrm{op}}^q
    \right]
    <\infty,
    \label{eq:mf-Sp-moments}
\end{equation}
so the additional hypothesis of
\cref{cor:mf-expectations} is also satisfied.

Consequently, both the convergence in probability of
\cref{thm:mf-finite-time} and the $L^1$ convergence of
\cref{cor:mf-expectations} hold with
\begin{equation}
    \mathcal L(t)
    =
    \frac12
    \int
    \left(
        \frac{1-\rho}
        {\rho e^{4\lambda t}+1-\rho}
    \right)^2
    d\mu_\phi(\lambda),
    \label{eq:mf-MP-pop-limit}
\end{equation}
and
\begin{equation}
    \widehat{\mathcal L}(t)
    =
    \frac12
    \int
    \lambda
    \left(
        \frac{1-\rho}
        {\rho e^{4\lambda t}+1-\rho}
    \right)^2
    d\mu_\phi(\lambda).
    \label{eq:mf-MP-train-limit}
\end{equation}
These are the Marchenko--Pastur specializations of the general
mean-field formulas.\paragraph{Remark on non-isotropic population covariance.}

We finally indicate how the result changes if the population covariance
is not the identity. Let
\begin{equation}
\mSigma_p
:=
\E[\vx\vx^\top]
\succeq0
\label{eq:mf-Sigma-def}
\end{equation}
be arbitrary. The empirical gradient flow \cref{eq:common-U-flow} depends
only on the training covariance $\mS_p$ and is therefore unchanged.
Consequently, under the standing initialization \cref{eq:common-Gaussian-init}, \cref{eq:mf-scaling}, and
\cref{eq:common-spectrum-tight}, the matrix trajectory conclusion
\begin{equation}
\sup_{0\le t\le T}
\|\mA_p(t)-a_t(\mS_p)\|_{\mathrm{op}}
\xrightarrow{\mathbb P}0
\label{eq:mf-aniso-matrix}
\end{equation}
continues to hold without any population-isotropy assumption.

The train loss is also unchanged and is still controlled solely by the
ordinary empirical spectral distribution of $\mS_p$. The population loss,
however, is now
\begin{equation}
L_p^{(\Sigma)}(t)
=
\frac{1}{2p}
\Tr
\left[
(\mI_p-\mA_p(t))
\mSigma_p
(\mI_p-\mA_p(t))
\right]
=
\frac{1}{2p}
\Tr
\left[
\mSigma_p
(\mI_p-\mA_p(t))^2
\right].
\label{eq:mf-aniso-pop-loss}
\end{equation}
Thus ordinary spectral convergence of $\mS_p$ is no longer sufficient to
determine the population limit.

To make the required additional information explicit, let
\begin{equation}
\mS_p
=
\sum_{i=1}^p
\lambda_{i,p}
q_{i,p}q_{i,p}^\top
\label{eq:mf-aniso-S-spectral}
\end{equation}
be an orthonormal spectral decomposition, and define the
$\mSigma_p$-weighted empirical spectral measure
\begin{equation}
\omega_p
:=
\frac1p
\sum_{i=1}^p
\bigl(
q_{i,p}^\top\mSigma_pq_{i,p}
\bigr)
\delta_{\lambda_{i,p}}.
\label{eq:mf-weighted-measure}
\end{equation}
Equivalently, for every bounded measurable function $f$,
\begin{equation}
\int
f(\lambda)\,
d\omega_p(\lambda)
=
\frac1p
\Tr
\bigl[
\mSigma_p f(\mS_p)
\bigr].
\label{eq:mf-weighted-functional}
\end{equation}
Suppose, in addition to the assumptions used for
\cref{eq:mf-aniso-matrix}, that
\begin{equation}
\frac1p
\Tr\mSigma_p
=
O(1),
\label{eq:mf-Sigma-trace-bound}
\end{equation}
and that, for a deterministic finite measure $\omega$ supported on
$[0,K]$,
\begin{equation}
\omega_p
\Longrightarrow
\omega
\qquad
\text{in probability}.
\label{eq:mf-weighted-convergence}
\end{equation}
Then
\begin{equation}
\sup_{0\le t\le T}
\left|
L_p^{(\Sigma)}(t)
-
\frac12
\int
e_t(\lambda)^2\,
d\omega(\lambda)
\right|
\xrightarrow{\mathbb P}0.
\label{eq:mf-aniso-pop-limit}
\end{equation}

Indeed, for $\mA,\mB\succeq0$ with
\begin{equation}
\|\mA\|_{\mathrm{op}},
\|\mB\|_{\mathrm{op}}
\le R,
\label{eq:mf-aniso-AB-bound}
\end{equation}
one has
\begin{equation}
\left|
L_p^{(\Sigma)}(\mA)
-
L_p^{(\Sigma)}(\mB)
\right|
\le
\frac{
\Tr\mSigma_p
}{p}
(1+R)
\|\mA-\mB\|_{\mathrm{op}}.
\label{eq:mf-aniso-loss-stability}
\end{equation}
To see this, set $\mE_{\mA}=\mI_p-\mA$ and $\mE_{\mB}=\mI_p-\mB$. Since $\mSigma_p\succeq0$,
\begin{equation}
\left|
\Tr
\left[
\mSigma_p
(\mE_{\mA}^2-\mE_{\mB}^2)
\right]
\right|
\le
\Tr(\mSigma_p)
\,
\|\mE_{\mA}^2-\mE_{\mB}^2\|_{\mathrm{op}},
\label{eq:mf-aniso-trace-bound}
\end{equation}
while
\begin{equation}
\|\mE_{\mA}^2-\mE_{\mB}^2\|_{\mathrm{op}}
\le
2(1+R)
\|\mA-\mB\|_{\mathrm{op}}.
\label{eq:mf-aniso-E-bound}
\end{equation}
\Cref{eq:mf-aniso-trace-bound,eq:mf-aniso-E-bound} give
\cref{eq:mf-aniso-loss-stability} after multiplication by $1/(2p)$.

For the comparison trajectory,
\begin{equation}
\frac{1}{2p}
\Tr
\left[
\mSigma_p
\bigl(
\mI_p-a_t(\mS_p)
\bigr)^2
\right]
=
\frac12
\int
e_t(\lambda)^2\,
d\omega_p(\lambda).
\label{eq:mf-aniso-weighted-comparison}
\end{equation}
The convergence in \cref{eq:mf-aniso-pop-limit} then follows by combining
\cref{eq:mf-aniso-matrix},
\cref{eq:mf-Sigma-trace-bound},
\cref{eq:mf-aniso-loss-stability}, and a finite-net argument identical
to the proof of \cref{lem:mf-uniform-spectral}.

When $\mSigma_p=\mI_p$, the weights in
\cref{eq:mf-weighted-measure} are all equal to one, and hence
\begin{equation}
\omega_p=\mu_{\mS_p}.
\label{eq:mf-isotropic-measures-equal}
\end{equation}
Thus \cref{eq:mf-aniso-pop-limit} reduces to
\cref{thm:mf-finite-time}. In the genuinely anisotropic case, by
contrast, the ordinary limiting eigenvalue distribution ${\nu_\phi}$ of $\mS_p$
does not in general determine $\omega$: the latter also records the
alignment between the eigenvectors of the empirical covariance and the
population covariance $\mSigma_p$. Thus anisotropy leaves the leading
mean-field matrix dynamics unchanged but introduces an additional weighted
spectral object in the population loss.

The results of this subsection concern the dimension limit on fixed
compact time intervals. They do not assert uniformity as $T\to\infty$ and
therefore do not, by themselves, justify interchange of the
large-dimension and large-time limits analyzed separately in
\cref{sec:mean-field_large_time}.

\subsubsection{Large time behavior: Marchenko--Pastur data covariance distribution}
\label{sec:mean-field_large_time}

For every positive empirical covariance eigenvalue \(\lambda>0\), we have
\begin{equation}
    a_\lambda(t)
    =
    \frac{\rho}
    {
        \rho+(1-\rho)e^{-4\lambda t}
    }
    \to 1
    \qquad
    \text{as }t\to\infty.
\end{equation}
Therefore
\begin{equation}
    \ell(\lambda t)\to 0
    \qquad
    \text{for every }\lambda>0.
\end{equation}

If \(\phi\le 1\), the limiting empirical covariance has no atom at zero, and hence $\lim_{t\to\infty}L(t)=0$.
If \(\phi>1\), then \(\mu_\phi\) has an atom at \(\lambda=0\) of mass $1-\frac1\phi$.
Those zero-eigenvalue directions are never trained by the empirical loss, so their reconstruction factor remains equal to its initialization value \(\rho\). Hence
\begin{equation}
    \ell(0)
    =
    \frac12(1-\rho)^2.
\end{equation}
Therefore
\begin{equation}
    \gL(\infty)
    =
    \frac12(1-\rho)^2
    \left(1-\frac1\phi\right)_+.
\end{equation}

On the other hand, the empirical train loss contains an additional factor of \(\lambda\). Therefore the zero-eigenvalue directions do not contribute to the empirical loss, and
\begin{equation}
    \lim_{t\to\infty}\widehat\gL(t)=0.
\end{equation}

We now compute the exact train loss asymptotics for large $t$.
For $\phi = 1$, since $\lambda_- = 0$ and $\lambda_+ = 4$, using a Laplace approximation,
\begin{equation}
  \begin{aligned}
    \widehat\gL(t)
    &= \frac12 \int_0^4 \lambda \left(\frac{1-\rho}{\rho e^{4\lambda t}+1-\rho}\right)^2 \frac{\sqrt{(4-\lambda) \lambda}}{2\pi\lambda} \, d\lambda
    \\&= \frac{1}{4 \pi} \int_0^4 \left(\frac{1-\rho}{\rho e^{4\lambda t}+1-\rho}\right)^2 \sqrt{(4-\lambda) \lambda} \, d\lambda
    \\&\sim \frac{1}{2 \pi} \int_0^\infty \left(\frac{1-\rho}{\rho e^{4\lambda t}+1-\rho}\right)^2 \sqrt{\lambda} \, d\lambda
  \end{aligned}
\end{equation}
as $t\to\infty$.
This integral could be evaluated explicitly:
\begin{equation}
    \widehat\gL(t)
    \sim \frac{\mathrm{Li}_{1/2}\left(\frac{\rho-1}{\rho}\right) - \mathrm{Li}_{3/2}\left(\frac{\rho-1}{\rho}\right)}{32 \sqrt{\pi} t^{3/2}},
    \qquad
    t\to\infty,
    \qquad
    \phi = 1.
\end{equation}
In contrast, when $\phi \neq 1$, $\lambda_- > 0$, which yields
\begin{equation}
  \begin{aligned}
    \widehat\gL(t)
    &= \frac12 \int_{\lambda_-}^{\lambda_+} \lambda \left(\frac{1-\rho}{\rho e^{4\lambda t}+1-\rho}\right)^2 \frac{\sqrt{(\lambda_+-\lambda)(\lambda-\lambda_-)}}{2\pi\phi\lambda} \, d\lambda
    \\&= \frac{1}{4 \pi \phi} \int_{\lambda_-}^{\lambda_+} \left(\frac{1-\rho}{\rho e^{4\lambda t}+1-\rho}\right)^2 \sqrt{(\lambda_+-\lambda)(\lambda-\lambda_-)} \, d\lambda
    \\&\sim \frac{\sqrt{\lambda_+ - \lambda_-}}{4 \pi \phi} \left(\frac{1-\rho}{\rho}\right)^2 \int_{\lambda_-}^{\lambda_+} e^{-8 \lambda t} \sqrt{\lambda-\lambda_-} \, d\lambda
  \end{aligned}
\end{equation}
as $t\to\infty$.
This integral could be evaluated explicitly:
\begin{equation}
    \widehat\gL(t)
    \sim \left(\frac{1-\rho}{\rho}\right)^2 \frac{e^{-8 (1 - \sqrt\phi)^2 t}}{64 \sqrt{2\pi} \phi^{3/4} t^{3/2}},
    \qquad
    t\to\infty,
    \qquad
    \phi \neq 1.
\end{equation}
Putting all together,
\begin{equation}
  \boxed{
    \widehat\gL(t)
    \sim \frac{1}{64 \sqrt{2\pi} \phi^{3/4} t^{3/2}} \begin{cases}
      2 \sqrt{2} \left[\Li_{1/2} - \Li_{3/2}\right]\left(\frac{\rho-1}{\rho}\right), & \phi = 1, \\
      \left(\frac{1-\rho}{\rho}\right)^2 e^{-8 (1 - \sqrt\phi)^2 t}, & \phi \neq 1
    \end{cases}
    \qquad
    t\to\infty.
  }
\end{equation}

We now perform a similar computation for the population loss.
For $\phi = 1$, again using a Laplace approximation,
\begin{equation}
  \begin{aligned}
    \gL(t) - \gL(\infty)
    &= \frac12 \int_0^4 \left(\frac{1-\rho}{\rho e^{4\lambda t}+1-\rho}\right)^2 \frac{\sqrt{(4-\lambda) \lambda}}{2\pi\lambda} \, d\lambda
    \\&= \frac{1}{4 \pi} \int_0^4 \left(\frac{1-\rho}{\rho e^{4\lambda t}+1-\rho}\right)^2 \sqrt{\frac{4-\lambda}{\lambda}} \, d\lambda
    \\&\sim \frac{1}{2 \pi} \int_0^\infty \left(\frac{1-\rho}{\rho e^{4\lambda t}+1-\rho}\right)^2 \, \frac{d\lambda}{\sqrt{\lambda}}
    \\&= \frac{\mathrm{Li}_{-1/2}\left(\frac{\rho-1}{\rho}\right) - \mathrm{Li}_{1/2}\left(\frac{\rho-1}{\rho}\right)}{4 \sqrt{\pi} t^{1/2}}
  \end{aligned}
\end{equation}
as $t\to\infty$.
When $\phi \neq 1$, $\lambda_- > 0$, which yields similarly to the train loss case,
\begin{equation}
  \begin{aligned}
    \gL(t) - \gL(\infty)
    &= \frac12 \int_{\lambda_-}^{\lambda_+} \left(\frac{1-\rho}{\rho e^{4\lambda t}+1-\rho}\right)^2 \frac{\sqrt{(\lambda_+-\lambda)(\lambda-\lambda_-)}}{2\pi\phi\lambda} \, d\lambda
    \\&= \frac{1}{4 \pi \phi} \int_{\lambda_-}^{\lambda_+} \left(\frac{1-\rho}{\rho e^{4\lambda t}+1-\rho}\right)^2 \sqrt{(\lambda_+-\lambda)(\lambda-\lambda_-)} \, \frac{d\lambda}{\lambda}
    \\&\sim \frac{\sqrt{\lambda_+ - \lambda_-}}{4 \pi \phi} \left(\frac{1-\rho}{\rho}\right)^2 \int_{\lambda_-}^{\lambda_+} e^{-8 \lambda t} \sqrt{\lambda-\lambda_-} \, \frac{d\lambda}{\lambda}
    \\&= \left(\frac{1-\rho}{\rho}\right)^2 \frac{e^{-8 (1 - \sqrt\phi)^2 t}}{64 \sqrt{2\pi} \phi^{3/4} \left(1 - \sqrt\phi\right)^2 t^{3/2}}
  \end{aligned}
\end{equation}
as $t\to\infty$.
Putting all together,
\begin{equation}
  \boxed{
    \gL(t) - \gL(\infty)
    \sim \frac{1}{32 \sqrt\pi \phi^{3/4}} \begin{cases}
      \frac{8}{t^{1/2}} \left[\Li_{-1/2} - \Li_{1/2}\right]\left(\frac{\rho-1}{\rho}\right), & \phi = 1, \\
      \left(\frac{1-\rho}{\rho}\right)^2 \frac{e^{-8 \left(1 - \sqrt\phi\right)^2 t}}{\left(1 - \sqrt\phi\right)^2 \left(2 t\right)^{3/2}}, & \phi \neq 1
    \end{cases}
    \qquad
    t\to\infty.
  }
\end{equation}

\subsection{Large-data (\cref{sec:large-data})}
\label{app:large-data}

We now specialize the common setup of \cref{app:common-preliminaries} to the large-data corner.
The common empirical covariance \(\mS_p\), Gram state \(\mA_p\), error
\(\mE_p\), losses, and gradient flow will not be redefined here.  We take
\(\eta=p\), so \(\gamma=1\), and assume
\begin{equation}
  \phi_p:=\frac pm\longrightarrow0,
  \qquad
  \psi_p:=\frac pn\longrightarrow\psi\in(0,\infty),
  \qquad
  \rho_p:=n\sigma_p^2\longrightarrow\rho\in(0,\infty).
  \label{eq:ld-scaling}
\end{equation}
The training samples are independent copies of a centered isotropic vector,
\begin{equation}
  \E\vx=0,
  \qquad
  \E[\vx\vx^\top]=\mI_p,
  \label{eq:ld-isotropy}
\end{equation}
The data hypothesis is the common near-isotropy condition
\cref{eq:common-data-near-isometry}, now applied to \(\mS_p\).  For
convenience set
\begin{equation}
  \varepsilon_p:=\|\mS_p-\mI_p\|_{\mathrm{op}}.
  \label{eq:ld-cov-error}
\end{equation}
Thus \(\varepsilon_p\to0\) in probability; in particular the standing
spectral assumptions of \cref{app:common-preliminaries} hold automatically with
\(\nu=\delta_1\).  The covariance concentration is the data-side input that
replaces the empirical trajectory by the \(\mS=\mI_p\) proxy and identifies
both train and population losses with the same logistic limiting curve.  We
write
\(L_p(t):=L(\mU_p(t))\) and
\(\widehat L_p(t):=\widehat L(\mU_p(t))\).

For $t\ge0$ and $s\ge0$, define the scalar logistic flow map
\begin{equation}
    \Theta_t(s)
    :=
    \frac{s}{s+(1-s)e^{-4t}}
    =
    \frac{se^{4t}}{1+s(e^{4t}-1)}.
    \label{eq:ld-theta}
\end{equation}
For a positive semidefinite matrix $\mB$, $\Theta_t(\mB)$ denotes the
corresponding functional calculus.

We also recall the Marchenko--Pastur distribution with aspect ratio
$\psi$:
\begin{equation}
    \mu_\psi(d\lambda)
    =
    \left(1-\frac1\psi\right)_+\delta_0(d\lambda)
    +
    \frac{
      \sqrt{(\lambda_+-\lambda)(\lambda-\lambda_-)}
    }{2\pi\psi\lambda}
    \mathbf 1_{\{\lambda\in[\lambda_-,\lambda_+]\}}\,d\lambda,
    \label{eq:ld-mp-law}
\end{equation}
where
\begin{equation}
    \lambda_-=(1-\sqrt\psi)^2,
    \qquad
    \lambda_+=(1+\sqrt\psi)^2.
    \label{eq:ld-mp-edges}
\end{equation}
Define the deterministic limiting loss
\begin{equation}
    \mathcal L_{\psi,\rho}(t)
    :=
    \frac12
    \int
    \bigl(\Theta_t(\rho\lambda)-1\bigr)^2
    \,\mu_\psi(d\lambda).
    \label{eq:ld-limit-loss}
\end{equation}

\paragraph{Main theorem.}
\begin{theorem}[Large-data limit]
\label{thm:ld-main}
Assume \cref{eq:ld-scaling}, \cref{eq:ld-isotropy}, and the common
near-isotropy condition
\cref{eq:common-data-near-isometry}.  Let $\mU_p(t)$ be the gradient-flow
trajectory for the empirical loss \cref{eq:common-losses} with
learning rate \(\eta=p\), and let
$\mA_p(t)=\mU_p(t)^\top \mU_p(t)$.  Then, for every $T<\infty$,
\begin{equation}
    \sup_{0\le t\le T}
    \left\|
      \mA_p(t)-\Theta_t(\mA_p(0))
    \right\|_{\mathrm{op}}
    \xrightarrow[p\to\infty]{\mathbb P}0.
    \label{eq:ld-main-trajectory}
\end{equation}
Moreover,
\begin{equation}
    \sup_{0\le t\le T}
    \left|
      L_p(t)-\mathcal L_{\psi,\rho}(t)
    \right|
    \xrightarrow[p\to\infty]{\mathbb P}0,
    \label{eq:ld-main-pop-conv}
\end{equation}
and
\begin{equation}
    \sup_{0\le t\le T}
    \left|
      \widehat L_p(t)-\mathcal L_{\psi,\rho}(t)
    \right|
    \xrightarrow[p\to\infty]{\mathbb P}0.
    \label{eq:ld-main-train-conv}
\end{equation}
If, in addition,
\begin{equation}
    \E\|\mS-\mI_p\|_{\mathrm{op}}\longrightarrow0
    \label{eq:ld-L1-cov}
\end{equation}
and the family $\{\|\mS\|_{\mathrm{op}}\}_p$ has a uniformly bounded
moment of some order strictly larger than one, then
\begin{equation}
    \E
    \sup_{0\le t\le T}
    \left|
      L_p(t)-\mathcal L_{\psi,\rho}(t)
    \right|
    \longrightarrow0
    \label{eq:ld-main-pop-L1}
\end{equation}
and
\begin{equation}
    \E
    \sup_{0\le t\le T}
    \left|
      \widehat L_p(t)-\mathcal L_{\psi,\rho}(t)
    \right|
    \longrightarrow0.
    \label{eq:ld-main-train-L1}
\end{equation}
In particular, the expected train and population losses converge
uniformly on every compact time interval to the same deterministic
curve \cref{eq:ld-limit-loss}.
\end{theorem}

\paragraph{Proof plan.}
The proof is a data-side stability argument.  First, the common Gram-flow
results reduce the empirical trajectory to the proxy with $\mS=\mI_p$,
with an error controlled by $\|\mS-\mI_p\|_{\mathrm{op}}$.  Second, the
assumed covariance concentration (or the sub-Gaussian sufficient condition
proved below) makes this perturbation vanish.  Third, the proxy is the
logistic matrix flow of \cref{lem:common-commuting-flow}.  Finally,
the Marchenko--Pastur law for the Gaussian initialization identifies the
limiting proxy loss; \cref{lem:common-loss-stability} transfers this
limit to both train and population losses.  The final uniform-integrability
argument gives the stated convergence of expectations.

\paragraph{Exact finite-dimensional dynamics and deterministic stability.}
With $\eta=p$, \cref{prop:common-gradient-flow} and
\cref{lem:common-gram-flow} give the exact $\mU$- and $\mA$-flows and
the dimension-free global bound.  For a positive semidefinite covariance
$\mS$, write $\mA_{\mS}(t)$ for the solution of the common Gram ODE with
initial state $\mA_0$, and write $\mA_{\mI}(t)$ for the corresponding
solution with covariance $\mI_p$.  \cref{prop:common-S-stability}
gives
\cref{eq:ld-stability-bound} with the constant \cref{eq:ld-KTR} whenever
$\|\mS-\mI_p\|_{\mathrm{op}}\le1$.  The population- and train-loss transfers
are exactly \cref{eq:ld-pop-proxy-bound}--\cref{eq:ld-train-proxy-bound}
from \cref{lem:common-loss-stability}.

\paragraph{A sufficient condition for empirical covariance concentration.}

The deterministic argument above only needs
\cref{eq:common-data-near-isometry}.  We now verify this assumption for
a broad class of non-Gaussian data.

For a real random variable $Y$, define the sub-Gaussian and
sub-exponential Orlicz norms by
\begin{equation}
    \|Y\|_{\psi_2}
    :=
    \inf\left\{
      s>0:
      \E\exp(Y^2/s^2)\le2
    \right\},
    \qquad
    \|Y\|_{\psi_1}
    :=
    \inf\left\{
      s>0:
      \E\exp(|Y|/s)\le2
    \right\}.
    \label{eq:ld-orlicz}
\end{equation}
Assume that the isotropic data vectors satisfy the uniform
sub-Gaussian condition
\begin{equation}
    \sup_p
    \sup_{\|\vv\|_2=1}
    \|\langle \vx^{(p)},\vv\rangle\|_{\psi_2}
    \le K
    \label{eq:ld-subgaussian}
\end{equation}
for a constant $K<\infty$ independent of dimension.  No independence
between the coordinates of $\vx^{(p)}$ is required.

\begin{prop}[Covariance concentration]
\label{prop:ld-covariance-concentration}
Under \cref{eq:ld-isotropy} and
\cref{eq:ld-subgaussian}, there exist constants
$C_K,c_K>0$ depending only on $K$ such that, for every $u\ge0$,
\begin{equation}
    \mathbb P\left(
      \|\mS-\mI_p\|_{\mathrm{op}}
      >
      C_K
      \left[
        \sqrt{\frac{p+u}{m}}
        +
        \frac{p+u}{m}
      \right]
    \right)
    \le
    2e^{-c_Ku}.
    \label{eq:ld-cov-tail}
\end{equation}
In particular, if $p/m\to0$, then
\begin{equation}
    \|\mS-\mI_p\|_{\mathrm{op}}
    \xrightarrow{\mathbb P}0.
    \label{eq:ld-cov-in-prob}
\end{equation}
Furthermore, for every fixed $q\in[1,\infty)$,
\begin{equation}
    \left(
      \E
      \|\mS-\mI_p\|_{\mathrm{op}}^q
    \right)^{1/q}
    \le
    C_{K,q}
    \left(
      \sqrt{\frac pm}+\frac pm
    \right),
    \label{eq:ld-cov-Lq}
\end{equation}
for all sufficiently large $p$.  In particular,
\cref{eq:ld-L1-cov} holds.
\end{prop}

\begin{proof}
Fix $\vv\in\mathbb S^{p-1}$.  By isotropy,
\begin{equation}
    \vv^\top(\mS-\mI_p)\vv
    =
    \frac1m
    \sum_{j=1}^m
    \left(
      \langle \vx_j,\vv\rangle^2-1
    \right).
    \label{eq:ld-fixed-v}
\end{equation}
For
\begin{equation}
    Y_j(\vv)
    :=
    \langle \vx_j,\vv\rangle^2-1,
    \label{eq:ld-Yj}
\end{equation}
we have $\E Y_j(\vv)=0$.  The elementary implication
``sub-Gaussian squared is sub-exponential'' gives
\begin{equation}
    \|Y_j(\vv)\|_{\psi_1}
    \le
    C K^2.
    \label{eq:ld-square-subexp}
\end{equation}
For completeness, before centering one has directly from
\cref{eq:ld-orlicz}
\begin{equation}
    \|\langle \vx_j,\vv\rangle^2\|_{\psi_1}
    =
    \|\langle \vx_j,\vv\rangle\|_{\psi_2}^2
    \le K^2,
    \label{eq:ld-square-orlicz}
\end{equation}
and centering a sub-exponential random variable changes its
$\psi_1$ norm by at most a universal multiplicative constant,
yielding \cref{eq:ld-square-subexp}.

The scalar Bernstein inequality for independent centered
sub-exponential variables therefore gives universal constants
$c,C>0$ such that
\begin{equation}
    \mathbb P\left(
      |\vv^\top(\mS-\mI_p)\vv|>t
    \right)
    \le
    2\exp\left[
      -cm
      \min\left\{
        \frac{t^2}{K^4},
        \frac{t}{K^2}
      \right\}
    \right].
    \label{eq:ld-fixed-v-Bernstein}
\end{equation}

Let $\mathcal N$ be a $1/4$-net of the Euclidean unit sphere.
A volumetric argument gives
\begin{equation}
    |\mathcal N|\le9^p.
    \label{eq:ld-net-size}
\end{equation}
For every symmetric $p\times p$ matrix $\mM$,
\begin{equation}
    \|\mM\|_{\mathrm{op}}
    \le
    2\max_{\vv\in\mathcal N}|\vv^\top \mM\vv|.
    \label{eq:ld-net-operator-bound}
\end{equation}
Indeed, choose a unit vector $\vx$ for which
$|\vx^\top \mM\vx|=\|\mM\|_{\mathrm{op}}$, and choose
$\vv\in\mathcal N$ with $\|\vx-\vv\|_2\le1/4$.  Then
\begin{equation}
    |\vx^\top \mM\vx-\vv^\top \mM\vv|
    \le
    |(\vx-\vv)^\top \mM\vx|
    +
    |\vv^\top \mM(\vx-\vv)|
    \le
    \frac12\|\mM\|_{\mathrm{op}},
    \label{eq:ld-net-proof}
\end{equation}
which implies \cref{eq:ld-net-operator-bound}.

Combining \cref{eq:ld-fixed-v-Bernstein},
\cref{eq:ld-net-size}, and a union bound gives
\begin{equation}
    \mathbb P\left(
      \|\mS-\mI_p\|_{\mathrm{op}}>2t
    \right)
    \le
    2\exp\left[
      p\log9
      -
      cm
      \min\left\{
        \frac{t^2}{K^4},
        \frac{t}{K^2}
      \right\}
    \right].
    \label{eq:ld-net-union}
\end{equation}
Taking
\begin{equation}
    t
    =
    C_K
    \left[
      \sqrt{\frac{p+u}{m}}
      +
      \frac{p+u}{m}
    \right]
    \label{eq:ld-tail-choice}
\end{equation}
with $C_K$ sufficiently large proves
\cref{eq:ld-cov-tail} after a harmless adjustment of constants.

When $p/m\to0$, the right-hand side in brackets in
\cref{eq:ld-tail-choice} converges to zero for each fixed $u$, which
proves \cref{eq:ld-cov-in-prob}.  Finally, integrating the tail
bound \cref{eq:ld-cov-tail} yields, for every fixed $q\ge1$,
\begin{equation}
    \E\|\mS-\mI_p\|_{\mathrm{op}}^q
    \le
    C_{K,q}
    \left[
      \left(\frac pm\right)^{q/2}
      +
      \left(\frac pm\right)^q
    \right],
    \label{eq:ld-tail-integration}
\end{equation}
which is equivalent, after changing the constant, to
\cref{eq:ld-cov-Lq}.
\end{proof}

\begin{remark}[Isotropy alone does not imply the large-data reduction]
\label{rem:ld-isotropy-not-enough}
The concentration assumption \cref{eq:common-data-near-isometry} cannot
be deduced from isotropy and $p/m\to0$ alone.  For example, let
$J$ be uniform on $\{1,\ldots,p\}$, let $\epsilon$ be an independent
Rademacher variable, and set
\begin{equation}
    \vx=\sqrt p\,\epsilon e_J.
    \label{eq:ld-sparse-example}
\end{equation}
Then
\begin{equation}
    \E[\vx\vx^\top]=\mI_p.
    \label{eq:ld-sparse-isotropic}
\end{equation}
Take
\begin{equation}
    m=p\log\log p.
    \label{eq:ld-sparse-m}
\end{equation}
Then $p/m\to0$, but if $N_0$ is the number of coordinates that are
never sampled, one has
\begin{equation}
    \E N_0
    =
    p\left(1-\frac1p\right)^m
    \sim
    \frac{p}{\log p}
    \longrightarrow\infty.
    \label{eq:ld-unsampled-mean}
\end{equation}
Moreover, if $I_j$ is the indicator that coordinate $j$ is
unsampled, then for $j\ne k$,
\begin{equation}
    \operatorname{Cov}(I_j,I_k)
    =
    \left(1-\frac2p\right)^m
    -
    \left(1-\frac1p\right)^{2m}
    \le0.
    \label{eq:ld-unsampled-cov}
\end{equation}
Consequently,
\begin{equation}
    \operatorname{Var}(N_0)
    \le
    \E N_0,
    \qquad
    \mathbb P(N_0=0)
    \le
    \frac{1}{\E N_0}
    \longrightarrow0.
    \label{eq:ld-unsampled-prob}
\end{equation}
With probability tending to one there is therefore a coordinate
$\ve_j$ for which $Se_j=0$, and on that event
\begin{equation}
    \|\mS-\mI_p\|_{\mathrm{op}}\ge1.
    \label{eq:ld-sparse-failure}
\end{equation}
Thus some concentration or regularity assumption beyond mere
second-moment isotropy is indispensable.
\end{remark}

\paragraph{Exact solution of the population-covariance proxy.}
\cref{lem:common-commuting-flow}, specialized to $\mS=\mI_p$, gives
\begin{equation}
    \mA_{\mI}(t)
    =\Theta_t(\mA_0),
    \label{eq:ld-logistic-matrix}
\end{equation}
where $\Theta_t$ is defined in \cref{eq:ld-theta}.  Thus the dynamical part
of the large-data approximation is already deterministic once
$\|\mS-\mI_p\|_{\mathrm{op}}\to0$.

\paragraph{The initialization spectral limit.}

Under the standing initialization \cref{eq:common-Gaussian-init},
\begin{equation}
    \mA_p(0)
    =
    \mU(0)^\top \mU(0)
    =
    \rho_p \mW_p,
    \qquad
    \mW_p:=\frac1nZ^\top \mZ.
    \label{eq:ld-Wishart-init}
\end{equation}
We use the following standard random-matrix facts.  If
$p/n\to\psi\in(0,\infty)$ and $\mZ$ has independent standard Gaussian
entries, then almost surely
\begin{equation}
    \mu_{\mW_p}
    :=
    \frac1p\sum_{i=1}^p\delta_{\lambda_i(\mW_p)}
    \Longrightarrow
    \mu_\psi,
    \label{eq:ld-MP-convergence}
\end{equation}
and
\begin{equation}
    \lambda_{\max}(\mW_p)
    \longrightarrow
    (1+\sqrt\psi)^2.
    \label{eq:ld-Wishart-upper-edge}
\end{equation}
The first statement is the Marchenko--Pastur theorem, and the second
is the corresponding extreme-eigenvalue theorem.  In particular,
\begin{equation}
    \|\mA_p(0)\|_{\mathrm{op}}
    =
    O_{\mathbb P}(1).
    \label{eq:ld-A0-tight}
\end{equation}

The Gaussian operator-norm tail bound
\begin{equation}
    \mathbb P\left(
      \|\mZ\|_{\mathrm{op}}
      >
      \sqrt n+\sqrt p+s
    \right)
    \le
    2e^{-s^2/2}
    \label{eq:ld-gaussian-op-tail}
\end{equation}
also implies that, for every fixed $q<\infty$,
\begin{equation}
    \sup_p
    \E
    \|\mA_p(0)\|_{\mathrm{op}}^q
    <
    \infty.
    \label{eq:ld-A0-moments}
\end{equation}

We next turn the pointwise Marchenko--Pastur convergence into uniform
convergence of the loss curve.

\begin{lemma}[Uniform spectral convergence of the proxy loss]
\label{lem:ld-uniform-MP}
Let
\begin{equation}
    L_p^{I}(t)
    :=
    \frac1{2p}
    \Tr
    \bigl(\mI_p-\Theta_t(\mA_p(0))\bigr)^2.
    \label{eq:ld-proxy-loss}
\end{equation}
Then, for every $T<\infty$,
\begin{equation}
    \sup_{0\le t\le T}
    |L_p^{I}(t)-\mathcal L_{\psi,\rho}(t)|
    \longrightarrow0
    \label{eq:ld-proxy-uniform}
\end{equation}
almost surely along the Gaussian initialization sequence.
\end{lemma}

\begin{proof}
Using \cref{eq:ld-Wishart-init},
\begin{equation}
    L_p^{I}(t)
    =
    \frac1p
    \sum_{i=1}^p
    g_p(t,\lambda_i(\mW_p)),
    \label{eq:ld-proxy-spectral-sum}
\end{equation}
where
\begin{equation}
    g_p(t,\lambda)
    :=
    \frac12
    \bigl[
      \Theta_t(\rho_p\lambda)-1
    \bigr]^2.
    \label{eq:ld-gp}
\end{equation}
Similarly, define
\begin{equation}
    g(t,\lambda)
    :=
    \frac12
    \bigl[
      \Theta_t(\rho\lambda)-1
    \bigr]^2.
    \label{eq:ld-g}
\end{equation}
By \cref{eq:ld-Wishart-upper-edge}, almost surely there is a finite
$M$ such that, for all sufficiently large $p$,
\begin{equation}
    \operatorname{spec}(\mW_p)\subset[0,M].
    \label{eq:ld-spectrum-compact}
\end{equation}
The map
\begin{equation}
    (t,\lambda,r)
    \longmapsto
    \frac12
    [\Theta_t(r\lambda)-1]^2
    \label{eq:ld-joint-continuity-map}
\end{equation}
is continuous on every compact subset of
$[0,\infty)^3$.  Since $\rho_p\to\rho$, it follows that
\begin{equation}
    \sup_{\substack{0\le t\le T\\0\le\lambda\le M}}
    |g_p(t,\lambda)-g(t,\lambda)|
    \longrightarrow0.
    \label{eq:ld-gp-to-g-uniform}
\end{equation}

It remains to prove
\begin{equation}
    \sup_{0\le t\le T}
    \left|
      \int g(t,\lambda)\,\mu_{\mW_p}(d\lambda)
      -
      \int g(t,\lambda)\,\mu_\psi(d\lambda)
    \right|
    \longrightarrow0.
    \label{eq:ld-uniform-weak-goal}
\end{equation}
The function $g$ is uniformly continuous on
$[0,T]\times[0,M]$.  Therefore, for every $\delta>0$, there exist
times $t_1,\ldots,t_N\in[0,T]$ such that for every
$t\in[0,T]$ there is an index $j$ with
\begin{equation}
    \sup_{0\le\lambda\le M}
    |g(t,\lambda)-g(t_j,\lambda)|
    \le\delta.
    \label{eq:ld-time-net}
\end{equation}
For each fixed $j$, weak convergence
\cref{eq:ld-MP-convergence} gives
\begin{equation}
    \int g(t_j,\lambda)\,\mu_{\mW_p}(d\lambda)
    \longrightarrow
    \int g(t_j,\lambda)\,\mu_\psi(d\lambda).
    \label{eq:ld-finite-time-MP}
\end{equation}
Since the collection of $t_j$ is finite,
\cref{eq:ld-time-net} and \cref{eq:ld-finite-time-MP} imply
\begin{equation}
    \limsup_{p\to\infty}
    \sup_{0\le t\le T}
    \left|
      \int g(t,\lambda)\,\mu_{\mW_p}(d\lambda)
      -
      \int g(t,\lambda)\,\mu_\psi(d\lambda)
    \right|
    \le2\delta.
    \label{eq:ld-time-net-limsup}
\end{equation}
Letting $\delta\downarrow0$ proves
\cref{eq:ld-uniform-weak-goal}.  Together with
\cref{eq:ld-gp-to-g-uniform}, this proves
\cref{eq:ld-proxy-uniform}.
\end{proof}

\paragraph{Proof of \cref{thm:ld-main}.}

Let
\begin{equation}
    R_p
    :=
    \max\{1,\|\mA_p(0)\|_{\mathrm{op}}\}.
    \label{eq:ld-Rp}
\end{equation}
By \cref{eq:ld-A0-tight}, the sequence $(R_p)_p$ is tight.
Fix $\delta>0$.  We can therefore choose a deterministic
$R<\infty$ such that
\begin{equation}
    \limsup_{p\to\infty}
    \mathbb P(R_p>R)
    \le\delta.
    \label{eq:ld-R-tight-choice}
\end{equation}
On the event
\begin{equation}
    \{R_p\le R\}\cap\{\varepsilon_p\le1\},
    \label{eq:ld-good-event}
\end{equation}
\cref{prop:common-S-stability} and
\cref{lem:common-commuting-flow} give
\begin{equation}
    \sup_{0\le t\le T}
    \|\mA_p(t)-\Theta_t(\mA_p(0))\|_{\mathrm{op}}
    \le
    K_{T,R}\varepsilon_p.
    \label{eq:ld-path-good-event}
\end{equation}
Since $\varepsilon_p\to0$ in probability, while $\delta$ is
arbitrary, \cref{eq:ld-path-good-event} proves
\cref{eq:ld-main-trajectory}.

Similarly, \cref{lem:common-loss-stability} gives on the same event
\begin{equation}
    \sup_{0\le t\le T}
    |L_p(t)-L_p^I(t)|
    \le
    (1+R)K_{T,R}\varepsilon_p.
    \label{eq:ld-pop-to-proxy}
\end{equation}
Combining this with \cref{lem:ld-uniform-MP} proves
\cref{eq:ld-main-pop-conv}.  Also,
\begin{equation}
    \sup_{0\le t\le T}
    |\widehat L_p(t)-L_p(t)|
    \le
    \frac12(1+R_p)^2\varepsilon_p,
    \label{eq:ld-train-pop-random}
\end{equation}
and therefore \cref{eq:ld-main-train-conv} follows.

It remains to justify the $L^1$ assertions.  Lemma
\ref{lem:common-gram-flow} gives
\begin{equation}
    \sup_{t\ge0}L_p(t)
    \le
    \frac12(1+R_p)^2.
    \label{eq:ld-pop-UI-bound}
\end{equation}
Likewise,
\begin{equation}
    \sup_{t\ge0}\widehat L_p(t)
    \le
    \frac12
    \|\mS\|_{\mathrm{op}}
    (1+R_p)^2.
    \label{eq:ld-train-UI-bound}
\end{equation}
By \cref{eq:ld-A0-moments}, $(R_p)_p$ has uniformly bounded moments
of every fixed order.  Under the additional assumption in
\cref{thm:ld-main}, the right-hand sides of
\cref{eq:ld-pop-UI-bound} and \cref{eq:ld-train-UI-bound} are
uniformly integrable.  Therefore the convergence in probability of
the suprema in \cref{eq:ld-main-pop-conv} and
\cref{eq:ld-main-train-conv} upgrades to $L^1$ convergence, proving
\cref{eq:ld-main-pop-L1} and \cref{eq:ld-main-train-L1}.

In particular, under the sub-Gaussian hypothesis
\cref{eq:ld-subgaussian}, Proposition
\ref{prop:ld-covariance-concentration} verifies all the required
data-side assumptions.

\paragraph{Closed-form evaluation of the limiting loss.}

We now evaluate \cref{eq:ld-limit-loss}.  Define
\begin{equation}
    q_t
    :=
    \rho(1-e^{4t})
    \le0.
    \label{eq:ld-qt}
\end{equation}
From \cref{eq:ld-theta},
\begin{equation}
    \Theta_t(\rho\lambda)
    =
    \frac{\rho\lambda e^{4t}}
    {1-q_t\lambda}.
    \label{eq:ld-theta-q}
\end{equation}
Hence
\begin{equation}
    \begin{aligned}
    \bigl(\Theta_t(\rho\lambda)-1\bigr)^2
    &=
    1
    -
    2\rho e^{4t}
    \frac{\lambda}{1-q_t\lambda}
    \\
    &\quad+
    \rho^2e^{8t}
    \frac{\lambda^2}{(1-q_t\lambda)^2}.
    \end{aligned}
    \label{eq:ld-square-q}
\end{equation}
Introduce
\begin{equation}
    H_\psi(q)
    :=
    \int
    \frac{\lambda}{1-q\lambda}
    \,\mu_\psi(d\lambda),
    \qquad q\le0.
    \label{eq:ld-H}
\end{equation}
Since the Marchenko--Pastur distribution has compact support on its
nonzero part and $1-q\lambda\ge1$ for $q\le0$, differentiation under
the integral gives
\begin{equation}
    H_\psi'(q)
    =
    \int
    \frac{\lambda^2}{(1-q\lambda)^2}
    \,\mu_\psi(d\lambda).
    \label{eq:ld-H-prime}
\end{equation}
Substitution into \cref{eq:ld-limit-loss} yields
\begin{equation}
    \boxed{
    \mathcal L_{\psi,\rho}(t)
    =
    \frac12
    -
    \rho e^{4t}H_\psi(q_t)
    +
    \frac{\rho^2e^{8t}}{2}H_\psi'(q_t),
    \qquad
    q_t=\rho(1-e^{4t}).
    }
    \label{eq:ld-closed-loss-H}
\end{equation}

For completeness, we also derive the explicit expression for
$H_\psi$.  Let
\begin{equation}
    M_\psi(q)
    :=
    \int
    \frac{1}{1-q\lambda}
    \,\mu_\psi(d\lambda).
    \label{eq:ld-M}
\end{equation}
The Marchenko--Pastur Stieltjes-transform equation, expressed in the
variable $q$, is
\begin{equation}
    \psi qM_\psi(q)^2
    -
    \bigl[1+(\psi-1)q\bigr]M_\psi(q)
    +
    1
    =
    0.
    \label{eq:ld-M-quadratic}
\end{equation}
The solution satisfying $M_\psi(0)=1$ is
\begin{equation}
    M_\psi(q)
    =
    \frac{
      1+(\psi-1)q
      -
      \sqrt{
        1-2(1+\psi)q+(1-\psi)^2q^2
      }
    }{
      2\psi q
    },
    \label{eq:ld-M-explicit}
\end{equation}
with the value at $q=0$ understood by continuity.  Since
\begin{equation}
    M_\psi(q)
    =
    1+qH_\psi(q),
    \label{eq:ld-M-H-relation}
\end{equation}
we obtain
\begin{equation}
    \boxed{
    H_\psi(q)
    =
    \frac{
      1-(1+\psi)q
      -
      \sqrt{
        1-2(1+\psi)q+(1-\psi)^2q^2
      }
    }{
      2\psi q^2
    }.
    }
    \label{eq:ld-H-explicit}
\end{equation}
The branch of the square root is the one for which
$H_\psi$ is regular at the origin.  Expanding
\cref{eq:ld-H-explicit} at $q=0$ gives
\begin{equation}
    H_\psi(0)=1,
    \qquad
    H_\psi'(0)=1+\psi.
    \label{eq:ld-H-origin}
\end{equation}
\Cref{eq:ld-closed-loss-H,eq:ld-H-explicit} reproduce the closed large-data formula in
the main text.

\subsubsection{Large-time behavior}
\label{app:large-data-rigorous-large-time}

We emphasize that the results in this subsection concern the
large-time behavior of the already established proportional-limit
curve $\mathcal L_{\psi,\rho}(t)$.  Thus the order of limits is
\begin{equation}
    \lim_{t\to\infty}
    \lim_{\substack{p,n,m\to\infty\\
                    p/n\to\psi,\ p/m\to0}}
    L_p(t),
    \label{eq:ld-order-limits}
\end{equation}
and the compact-time estimates proved above do not by themselves
justify interchanging these two limits.

For every $s>0$,
\begin{equation}
    \lim_{t\to\infty}\Theta_t(s)=1,
    \qquad
    \Theta_t(0)=0.
    \label{eq:ld-theta-infty}
\end{equation}
The Marchenko--Pastur measure has an atom at zero of mass
$(1-\psi^{-1})_+$.  Dominated convergence in
\cref{eq:ld-limit-loss} therefore yields
\begin{equation}
    \boxed{
    \mathcal L_{\psi,\rho}(\infty)
    =
    \frac12
    \left(1-\frac1\psi\right)_+.
    }
    \label{eq:ld-final-loss}
\end{equation}

The same conclusion can also be read directly from the finite
dimensional population-covariance flow.  If
$a_i(0)>0$, then \cref{eq:common-logistic-modes} implies
\begin{equation}
    a_i(t)-1
    =
    \frac{
      (a_i(0)-1)e^{-4t}
    }{
      a_i(0)+(1-a_i(0))e^{-4t}
    }.
    \label{eq:ld-mode-error}
\end{equation}
If the positive initial eigenvalues are fixed and bounded away from
zero, then
\begin{equation}
    \begin{aligned}
    L_p^I(t)-L_p^I(\infty)
    &=
    \frac{e^{-8t}}{2p}
    \sum_{i:\,a_i(0)>0}
    \left(
      \frac{1-a_i(0)}{a_i(0)}
    \right)^2
    \\
    &\quad+
    O(e^{-12t}).
    \end{aligned}
    \label{eq:ld-finite-dim-late}
\end{equation}
The prefactor can, however, become singular when the smallest
positive initial eigenvalues tend to zero.  This is precisely what
happens at the critical proportional aspect ratio $\psi=1$.

\paragraph{Noncritical aspect ratio.}

Assume $\psi\ne1$ and put
\begin{equation}
    \epsilon:=e^{-4t}.
    \label{eq:ld-epsilon}
\end{equation}
For every $s>0$,
\begin{equation}
    1-\Theta_t(s)
    =
    \frac{(1-s)\epsilon}
    {s+(1-s)\epsilon}.
    \label{eq:ld-one-minus-theta}
\end{equation}
For $\psi\ne1$, the positive part of the Marchenko--Pastur support
is separated from zero:
\begin{equation}
    \inf\{
      \lambda>0:
      \lambda\in\operatorname{supp}\mu_\psi
    \}
    =
    (1-\sqrt\psi)^2
    >
    0.
    \label{eq:ld-MP-gap}
\end{equation}
Consequently, uniformly over the positive support,
\begin{equation}
    \bigl(1-\Theta_t(\rho\lambda)\bigr)^2
    =
    \epsilon^2
    \left(
      \frac{1-\rho\lambda}{\rho\lambda}
    \right)^2
    +
    O(\epsilon^3).
    \label{eq:ld-noncritical-expansion}
\end{equation}
The zero atom, when $\psi>1$, contributes exactly
$\mathcal L_{\psi,\rho}(\infty)$.  Therefore
\begin{equation}
    \boxed{
    \mathcal L_{\psi,\rho}(t)
    -
    \mathcal L_{\psi,\rho}(\infty)
    =
    C_{\psi,\rho}e^{-8t}
    +
    O(e^{-12t}),
    \qquad
    \psi\ne1,
    }
    \label{eq:ld-noncritical-rate}
\end{equation}
where
\begin{equation}
    C_{\psi,\rho}
    :=
    \frac12
    \int_{\lambda>0}
    \left(
      \frac{1-\rho\lambda}{\rho\lambda}
    \right)^2
    \mu_\psi(d\lambda).
    \label{eq:ld-noncritical-constant}
\end{equation}

\paragraph{Critical aspect ratio.}

We now assume $\psi=1$.  In this case
\begin{equation}
    \mu_1(d\lambda)
    =
    \frac{1}{2\pi}
    \sqrt{\frac{4-\lambda}{\lambda}}
    \mathbf 1_{\{0<\lambda<4\}}
    \,d\lambda.
    \label{eq:ld-MP-critical-density}
\end{equation}
Since there is no atom at zero,
\begin{equation}
    \mathcal L_{1,\rho}(\infty)=0.
    \label{eq:ld-critical-final}
\end{equation}
Using \cref{eq:ld-one-minus-theta} with
$s=\rho\lambda$ gives
\begin{equation}
    \mathcal L_{1,\rho}(t)
    =
    \frac{1}{4\pi}
    \int_0^4
    \left[
      \frac{
        \epsilon(1-\rho\lambda)
      }{
        \rho\lambda+(1-\rho\lambda)\epsilon
      }
    \right]^2
    \sqrt{\frac{4-\lambda}{\lambda}}
    \,d\lambda.
    \label{eq:ld-critical-integral}
\end{equation}

Choose a fixed
\begin{equation}
    0<\delta<
    \min\left\{
      4,\frac{1}{2\rho}
    \right\}.
    \label{eq:ld-critical-delta}
\end{equation}
On $\lambda\in[\delta,4]$, for all sufficiently small $\epsilon$,
the denominator in \cref{eq:ld-critical-integral} is bounded below
by a positive constant depending only on $\delta$ and $\rho$.
Therefore
\begin{equation}
    \int_\delta^4
    \left[
      \frac{
        \epsilon(1-\rho\lambda)
      }{
        \rho\lambda+(1-\rho\lambda)\epsilon
      }
    \right]^2
    \sqrt{\frac{4-\lambda}{\lambda}}
    \,d\lambda
    =
    O(\epsilon^2).
    \label{eq:ld-critical-large-lambda}
\end{equation}
Thus the leading contribution comes from $\lambda$ of order
$\epsilon$.  In the integral over $(0,\delta)$, set
\begin{equation}
    \lambda
    =
    \frac{\epsilon y}{\rho}.
    \label{eq:ld-critical-rescale}
\end{equation}
Then
\begin{equation}
    \frac{
      \epsilon(1-\rho\lambda)
    }{
      \rho\lambda+(1-\rho\lambda)\epsilon
    }
    =
    \frac{1-\epsilon y}
    {1+y-\epsilon y},
    \label{eq:ld-critical-ratio-rescaled}
\end{equation}
and a direct substitution yields
\begin{equation}
    \begin{aligned}
    \frac{1}{\sqrt\epsilon}
    \mathcal L_{1,\rho}^{(0,\delta)}(t)
    &=
    \frac{1}{4\pi\sqrt\rho}
    \int_0^{\rho\delta/\epsilon}
    \left(
      \frac{1-\epsilon y}
      {1+y-\epsilon y}
    \right)^2
    \\
    &\qquad\qquad\times
    \sqrt{
      4-\frac{\epsilon y}{\rho}
    }\,
    y^{-1/2}\,dy.
    \end{aligned}
    \label{eq:ld-critical-rescaled-integral}
\end{equation}
For every fixed $y>0$, the integrand in
\cref{eq:ld-critical-rescaled-integral} converges to
\begin{equation}
    \frac{2y^{-1/2}}{(1+y)^2}.
    \label{eq:ld-critical-pointwise}
\end{equation}
Moreover, because $\rho\lambda\le\rho\delta<1/2$ on the region
under consideration, for sufficiently small $\epsilon$ the
integrand in \cref{eq:ld-critical-rescaled-integral} is bounded by
\begin{equation}
    C
    \frac{y^{-1/2}}{(1+y)^2},
    \label{eq:ld-critical-dominator}
\end{equation}
where the right-hand side is integrable on $(0,\infty)$.  Dominated
convergence therefore gives
\begin{equation}
    \lim_{t\to\infty}
    \frac{\mathcal L_{1,\rho}(t)}
    {\sqrt\epsilon}
    =
    \frac{1}{2\pi\sqrt\rho}
    \int_0^\infty
    \frac{y^{-1/2}}{(1+y)^2}\,dy.
    \label{eq:ld-critical-limit-integral}
\end{equation}
The remaining integral is a beta integral:
\begin{equation}
    \int_0^\infty
    \frac{y^{-1/2}}{(1+y)^2}\,dy
    =
    B\left(\frac12,\frac32\right)
    =
    \frac{\pi}{2}.
    \label{eq:ld-critical-beta}
\end{equation}
Since $\sqrt\epsilon=e^{-2t}$, we conclude that
\begin{equation}
    \boxed{
    \mathcal L_{1,\rho}(t)
    \sim
    \frac{e^{-2t}}{4\sqrt\rho},
    \qquad
    t\to\infty.
    }
    \label{eq:ld-critical-rate}
\end{equation}
Thus the critical proportional limit has a slower $e^{-2t}$ decay,
even though at any fixed finite dimension the idealized
population-covariance flow with full-rank initialization eventually
has the $e^{-8t}$ loss decay in
\cref{eq:ld-finite-dim-late}.  The discrepancy is caused by the
hard edge of the Marchenko--Pastur spectrum and is an explicit
noncommutation of the large-dimension and large-time limits.

\paragraph{Possible extension to anisotropic data.}

The deterministic part of the argument extends directly to
non-isotropic data.  Suppose instead that
\begin{equation}
    \E[\vx\vx^\top]
    =
    \mSigma_p,
    \qquad
    \mSigma_p\succeq0,
    \qquad
    \sup_p\|\mSigma_p\|_{\mathrm{op}}<\infty,
    \label{eq:ld-aniso-cov}
\end{equation}
and assume the empirical covariance satisfies
\begin{equation}
    \|\mS-\mSigma_p\|_{\mathrm{op}}
    \xrightarrow{\mathbb P}0.
    \label{eq:ld-aniso-concentration}
\end{equation}
The exact population loss then becomes
\begin{equation}
    L_{\mSigma_p}(\mA)
    =
    \frac{1}{2p}
    \Tr
    \bigl[
      \mSigma_p(\mI_p-\mA)^2
    \bigr].
    \label{eq:ld-aniso-loss}
\end{equation}
Repeating \cref{prop:common-S-stability}, with $\mI_p$ replaced
by $\mSigma_p$, shows that on every fixed time interval the empirical
trajectory is close to the deterministic population-covariance
flow
\begin{equation}
    \dot \mA_\Sigma
    =
    2(\mSigma_p\mA_\Sigma+\mA_\Sigma\mSigma_p)
    -
    \mSigma_p\mA_\Sigma^2
    -
    \mA_\Sigma^2\mSigma_p
    -
    2\mA_\Sigma\mSigma_p\mA_\Sigma.
    \label{eq:ld-aniso-flow}
\end{equation}
Indeed, on bounded operator-norm sets the same calculation gives
\begin{equation}
    \sup_{0\le t\le T}
    \|\mA_{\mS}(t)-\mA_\Sigma(t)\|_{\mathrm{op}}
    \le
    C_{T,R,\sup_p\|\mSigma_p\|_{\mathrm{op}}}
    \|\mS-\mSigma_p\|_{\mathrm{op}}.
    \label{eq:ld-aniso-stability}
\end{equation}
Thus the large-data empirical-to-population reduction remains
straightforward.

What is special about isotropy is not the concentration step but the
subsequent solvability.  For $\mSigma_p=\mI_p$,
\cref{eq:ld-aniso-flow} collapses to the scalar spectral logistic
flow \cref{eq:ld-logistic-matrix}.  For a general covariance,
$\mA_\Sigma(t)$ need not commute with $\mSigma_p$, and the eigenvectors
of $\mA_\Sigma(t)$ may rotate.

If the initial Gram matrix happens to commute with the covariance,
\begin{equation}
    [\mA_0,\mSigma_p]=0,
    \label{eq:ld-aniso-commute}
\end{equation}
then the two matrices can be diagonalized simultaneously:
\begin{equation}
    \mSigma_p
    =
    \operatorname{diag}(\lambda_1,\ldots,\lambda_p),
    \qquad
    \mA_0
    =
    \operatorname{diag}(a_1(0),\ldots,a_p(0)).
    \label{eq:ld-aniso-diagonal}
\end{equation}
\Cref{eq:ld-aniso-flow} then reduces to
\begin{equation}
    \dot a_i(t)
    =
    4\lambda_i a_i(t)(1-a_i(t)),
    \label{eq:ld-aniso-scalar}
\end{equation}
with solution
\begin{equation}
    a_i(t)
    =
    \Theta_{\lambda_i t}(a_i(0)).
    \label{eq:ld-aniso-scalar-solution}
\end{equation}
For generic Gaussian initialization in the proportional regime
$p/n\to\psi>0$, however, $\mA_0$ does not commute with a deterministic
non-scalar $\mSigma_p$.  The large-dimensional limit is then
genuinely noncommutative.  One possible route is to combine the
deterministic stability estimate \cref{eq:ld-aniso-stability} with
asymptotic freeness of the orthogonally invariant Wishart
initialization and $\mSigma_p$, leading to an operator-valued version
of \cref{eq:ld-aniso-flow}.  Establishing and analyzing that limit
requires additional arguments and lies beyond the isotropic
large-data result proved here.

\subsection{Narrow-latent (\cref{sec:narrow-latent})}
\label{app:narrow-latent}

This subsection uses the common notation, losses, and gradient-flow identity
from \cref{app:common-preliminaries} and records only the additional objects specific to the
narrow-latent reduction.  We take \(\eta=p\), hence \(\gamma=1\), and
\begin{equation}
  p=p_N\longrightarrow\infty,
  \qquad
  n=n_N\longrightarrow\infty,
  \qquad
  m=m_N\longrightarrow\infty,
\end{equation}
\begin{equation}
  \frac np\longrightarrow0,
  \qquad
  \phi_N = \frac{p_N}{m_N} \to \phi \in (0,\infty),
  \qquad
  \kappa_N:=p_N\sigma_N^2\longrightarrow\kappa\in(0,\infty).
  \label{eq:nl-scaling}
\end{equation}
Thus
\begin{equation}
  \psi_N:=\frac{p_N}{n_N}\longrightarrow\infty,
  \qquad
  \rho_N:=n_N\sigma_N^2,
  \qquad
  \kappa_N=\psi_N\rho_N.
  \label{eq:nl-psi}
\end{equation}
The training samples are centered and isotropic,
\begin{equation}
  \E\vx=0,
  \qquad
  \E[\vx\vx^\top]=\mI_p.
  \label{eq:nl-isotropy}
\end{equation}

The empirical covariance \(\mS_N\) is the common matrix
\cref{eq:common-S-sequence}.  We assume the standing spectral conditions
\cref{eq:common-spectrum-tight} and \cref{eq:common-esd-convergence}, with
limiting law denoted by \({\nu_\phi}\).
Spectral boundedness controls the matrix ODE
on compact time intervals and the weak spectral limit determines the scalar
limiting coefficients.

The regime-specific state variables are the two latent-space Gramians
\begin{equation}
  \mG_N(t):=\mU_N(t)\mU_N(t)^\top,
  \qquad
  \mH_N(t):=\mU_N(t)\mS_N\mU_N(t)^\top.
  \label{eq:nl-GH}
\end{equation}
Specializing the common flow \cref{eq:common-U-flow} to \(\gamma=1\) gives
the exact latent-space form
\begin{equation}
  \dot{\mU}_N
  =2\mU_N\mS_N-\mH_N\mU_N-\mG_N\mU_N\mS_N.
  \label{eq:nl-flow-GH}
\end{equation}
By \cref{eq:common-spectrum-tight}--\cref{eq:common-esd-convergence}, the
limiting law satisfies \(\operatorname{supp}{\nu_\phi}\subset[0,K]\).

Define the exponential transform by
\begin{equation}
M_{\nu_\phi}(q):=\int e^{q\lambda}\,d{\nu_\phi}(\lambda).
\label{eq:nl-Mnu}
\end{equation}
and its logarithmic derivative by
\begin{equation}
m_{\nu_\phi}(q):=\frac{M_{\nu_\phi}'(q)}{M_{\nu_\phi}(q)}.
\label{eq:nl-mnu}
\end{equation}
The limiting scalar dynamics are
\begin{equation}
\begin{aligned}
\dot a(t)&=4a(t)m_{\nu_\phi}(q(t))(1-a(t)),\\
\dot q(t)&=2(2-a(t)),\\
a(0)&=\kappa,\qquad q(0)=0,
\end{aligned}
\label{eq:nl-limit-ODE}
\end{equation}
and
\begin{equation}
r(t):=a(t)m_{\nu_\phi}(q(t)).
\label{eq:nl-limit-r}
\end{equation}
These three scalars have a direct interpretation.  Since the theorem below
shows \(\mG_N=\mU_N\mU_N^\top\approx a\mI_n\), the quantity \(a(t)\) is the
asymptotic squared norm of a typical latent row of \(\mU_N(t)\), while the
off-diagonal part of \(\mG_N\) vanishes.  Likewise
\(\mH_N=\mU_N\mS_N\mU_N^\top\approx r\mI_n\), so \(r(t)\) is the
corresponding covariance-weighted row energy, or data-alignment scale.  Their
ratio has the spectral meaning
\begin{equation}
  \frac{r(t)}{a(t)}
  =m_{\nu_\phi}(q(t))
  =\int \lambda\,d{\nu_\phi}_{q(t)}(\lambda),
  \label{eq:nl-ar-interpretation}
\end{equation}
where \(\nu_{\phi,q}\) is the exponentially tilted law defined below.  Finally,
\(q(t)\) is the spectral-tilting clock.  In the proxy solution one has
\(q=2R\) and a spectral component with data eigenvalue \(\lambda\) is
multiplied by \(e^{q\lambda/2}\); hence positive \(q\) progressively favors
large-eigenvalue directions, whereas negative \(q\) favors the lower edge
and, when present, the null space.

\paragraph{Main theorem.}
\begin{theorem}[Narrow-latent limit for a general spectral measure]
\label{thm:nl-general-nu}
Assume \cref{eq:nl-scaling} and the common spectral hypotheses \cref{eq:common-spectrum-tight}--\cref{eq:common-esd-convergence}.
Let \(a,q,r\) be defined by
\cref{eq:nl-limit-ODE} and
\cref{eq:nl-limit-r}.
Then, for every fixed \(T<\infty\),
\begin{equation}
\sup_{0\le t\le T}
\|
\mG_N(t)-a(t)\mI_n
\|_{\mathrm{op}}
\xrightarrow{\mathbb P}0
\label{eq:nl-G-general-limit}
\end{equation}
and
\begin{equation}
\sup_{0\le t\le T}
\|
\mH_N(t)-r(t)\mI_n
\|_{\mathrm{op}}
\xrightarrow{\mathbb P}0.
\label{eq:nl-H-general-limit}
\end{equation}
\end{theorem}

\paragraph{Proof plan.}

\paragraph{Step 1: express the losses through \(\mG\) and \(\mH\).}
Population isotropy implies
\begin{equation}
L(\mU)
=
\frac1{2p}
\Tr(\mI_p-\mU^\top \mU)^2 .
\label{eq:nl-plan-pop}
\end{equation}
Using cyclicity of trace,
\begin{equation}
L(\mU)
=
\frac12
-
\frac1p\Tr\mG
+
\frac1{2p}\Tr\mG^2 .
\label{eq:nl-plan-pop-G}
\end{equation}
Similarly, the empirical loss satisfies
\begin{equation}
\widehat L(\mU)
=
\frac1{2p}\Tr\mS
-
\frac1p\Tr\mH
+
\frac1{2p}\Tr(\mG\mH).
\label{eq:nl-plan-train-GH}
\end{equation}
Consequently, the losses are determined at the \(1/\psi_N\) scale
once \(\mG\) and \(\mH\) are known to leading order.

\paragraph{Step 2: identify the desired scalarization.}
We shall construct scalar functions \(a_N(t)\) and \(r_N(t)\) and
prove, uniformly on every fixed interval \(0\le t\le T\),
\begin{equation}
\mG(t)
=
a_N(t)\mI_n+o_{\mathbb P}(1)
\quad\text{in operator norm},
\label{eq:nl-plan-G-scalar}
\end{equation}
and
\begin{equation}
\mH(t)
=
r_N(t)\mI_n+o_{\mathbb P}(1)
\quad\text{in operator norm}.
\label{eq:nl-plan-H-scalar}
\end{equation}
This is the rigorous formulation of asymptotic row orthogonality and
self-averaging in the narrow-latent regime.

\paragraph{Step 3: construct a scalarized proxy dynamics.}
If one formally replaces
\begin{equation}
\mG(t)\approx a_N(t)\mI_n,
\qquad
\mH(t)\approx r_N(t)\mI_n
\label{eq:nl-plan-substitution}
\end{equation}
in \cref{eq:nl-flow-GH}, then one obtains
\begin{equation}
\dot{\overline \mU}
=
(2-a_N)\overline \mU \mS-r_N\overline \mU .
\label{eq:nl-plan-proxy-ode}
\end{equation}
This equation can be solved explicitly:
\begin{equation}
\overline \mU(t)
=
e^{-B_N(t)}\mU(0)e^{R_N(t)\mS},
\label{eq:nl-plan-proxy}
\end{equation}
where
\begin{equation}
\dot R_N(t)=2-a_N(t),
\qquad
\dot B_N(t)=r_N(t).
\label{eq:nl-plan-clocks}
\end{equation}
The central dynamical task is then to prove
\begin{equation}
\sup_{0\le t\le T}
\|\mU(t)-\overline \mU(t)\|_{\mathrm{op}}
\xrightarrow{\mathbb P}0.
\label{eq:nl-plan-U-close}
\end{equation}

\paragraph{Step 4: prove self-consistency of the proxy.}
For the explicit proxy \(\overline \mU\), define
\begin{equation}
\overline \mG
=
\overline \mU\overline \mU^\top,
\qquad
\overline \mH
=
\overline \mU \mS\overline \mU^\top .
\label{eq:nl-plan-GbarHbar}
\end{equation}
Conditional on \(\mS\), these are weighted Gaussian sample-covariance
matrices.  Since \(n/p\to0\), we shall prove
\begin{equation}
\sup_{0\le t\le T}
\|\overline \mG(t)-a_N(t)\mI_n\|_{\mathrm{op}}
\xrightarrow{\mathbb P}0
\label{eq:nl-plan-Gbar-close}
\end{equation}
and
\begin{equation}
\sup_{0\le t\le T}
\|\overline \mH(t)-r_N(t)\mI_n\|_{\mathrm{op}}
\xrightarrow{\mathbb P}0.
\label{eq:nl-plan-Hbar-close}
\end{equation}
These estimates imply that the proxy solves the exact matrix ODE up
to a vanishing residual.  \cref{lem:common-ode-stability} then proves
\cref{eq:nl-plan-U-close}.  \Crefrange{eq:nl-plan-Gbar-close}{eq:nl-plan-U-close} together yield the desired scalarization
\cref{eq:nl-plan-G-scalar}--\cref{eq:nl-plan-H-scalar}.

Thus, logically, Step 4 supplies the estimate that closes the
comparison introduced in Step 3.

\paragraph{Step 5: pass from the empirical scalar dynamics to a
deterministic spectral limit.}
The finite-\(N\) functions \(a_N,r_N\) depend on the empirical
spectrum of \(\mS_N\).  If
\begin{equation}
\mu_{\mS_N}
=
\frac1p\sum_{j=1}^{p}
\delta_{\lambda_j(\mS_N)}
\xrightarrow{\mathbb P}{\nu_\phi}
\label{eq:nl-plan-mu-to-nu}
\end{equation}
for some deterministic probability measure \({\nu_\phi}\), then we shall
prove
\begin{equation}
\sup_{0\le t\le T}|a_N(t)-a(t)|
\xrightarrow{\mathbb P}0
\label{eq:nl-plan-a-limit}
\end{equation}
and
\begin{equation}
\sup_{0\le t\le T}|r_N(t)-r(t)|
\xrightarrow{\mathbb P}0,
\label{eq:nl-plan-r-limit}
\end{equation}
where \(a,r\) are determined by the exponential transform of
\({\nu_\phi}\).  No Marchenko--Pastur assumption is required for this step.

Combining all five steps gives the chain
\begin{equation}
\mG,\mH
\;\approx\; \overline \mG,\overline \mH 
\;\approx\; a_N\mI_n,r_N\mI_n
\;\approx\; a\mI_n,r\mI_n,
\label{eq:nl-plan-chain}
\end{equation}
which inserted into
\cref{eq:nl-plan-pop-G}--\cref{eq:nl-plan-train-GH}
produces the limiting loss formulas.

We now carry out these steps.

\subsubsection{Exact identities and a priori bounds}
\label{app:narrow-exact}

The common Gram-flow bound, \cref{lem:common-gram-flow}, applies to
\cref{eq:common-U-flow} with $\gamma=1$ and gives global existence together
with the dimension-free estimate \cref{eq:common-apriori}.

The loss identities announced in Step 1 are also exact.

\begin{lemma}[Exact loss representation]
\label{lem:nl-loss-identities}
Under the isotropy assumption \cref{eq:nl-isotropy},
\begin{equation}
L(\mU)
=
\frac12
-
\frac1p\Tr\mG
+
\frac1{2p}\Tr\mG^2 .
\label{eq:nl-pop-exact-G}
\end{equation}
For every data matrix \(\mX\),
\begin{equation}
\widehat L(\mU)
=
\frac1{2p}\Tr\mS
-
\frac1p\Tr\mH
+
\frac1{2p}\Tr(\mG\mH).
\label{eq:nl-train-exact-GH}
\end{equation}
\end{lemma}

\begin{proof}
Isotropy gives
\begin{equation}
\begin{aligned}
L(\mU)
&=
\frac1{2p}
\E
\left[
\vx^\top(\mI_p-\mU^\top \mU)^2\vx
\right]
\\
&=
\frac1{2p}
\Tr(\mI_p-\mU^\top \mU)^2 .
\end{aligned}
\label{eq:nl-pop-isotropy-proof}
\end{equation}
Expanding and using
\begin{equation}
\Tr(\mU^\top \mU)
=
\Tr(\mU\mU^\top)
=
\Tr\mG
\label{eq:nl-tr-A-G}
\end{equation}
and
\begin{equation}
\Tr(\mU^\top \mU)^2
=
\Tr(\mU\mU^\top)^2
=
\Tr\mG^2
\label{eq:nl-tr-A2-G2}
\end{equation}
proves \cref{eq:nl-pop-exact-G}.

For the empirical loss,
\begin{equation}
\widehat L(\mU)
=
\frac1{2p}
\Tr
\left[
(\mI_p-\mU^\top \mU)\mS(\mI_p-\mU^\top \mU)
\right].
\label{eq:nl-train-expand-start}
\end{equation}
Expanding the product and using cyclicity of trace,
\begin{equation}
\Tr(\mU^\top\mU\mS)
=
\Tr(\mU\mS\mU^\top)
=
\Tr\mH
\label{eq:nl-tr-AS-H}
\end{equation}
and
\begin{equation}
\Tr(\mU^\top \mU\mS\mU^\top \mU)
=
\Tr(\mU\mU^\top \mU\mS\mU^\top)
=
\Tr(\mG\mH),
\label{eq:nl-tr-ASA-GH}
\end{equation}
which proves \cref{eq:nl-train-exact-GH}.
\end{proof}

\subsubsection{The empirical scalar system and proxy dynamics}
\label{app:narrow-proxy}

We now implement Steps 2 and 3.

For a fixed realization of \(\mS\), define
\begin{equation}
M_N(q)
=
M_{\mS}(q)
=
\frac1p\Tr\!\left(e^{q\mS}\right)
\label{eq:nl-MN}
\end{equation}
and
\begin{equation}
m_N(q)
=
\frac{M_N'(q)}{M_N(q)}
=
\frac{
\Tr(\mS e^{q\mS})
}{
\Tr(e^{q\mS})
}.
\label{eq:nl-mN}
\end{equation}
The scalar pair \((a_N,q_N)\) is defined by
\begin{equation}
\begin{aligned}
\dot a_N(t)
&=
4a_N(t)m_N(q_N(t))(1-a_N(t)),
\\
\dot q_N(t)
&=
2(2-a_N(t)),
\\
a_N(0)
&=
\kappa_N,
\qquad
q_N(0)=0 .
\end{aligned}
\label{eq:nl-finite-scalar}
\end{equation}
We then set
\begin{equation}
r_N(t)
=
a_N(t)m_N(q_N(t)),
\label{eq:nl-rN}
\end{equation}
\begin{equation}
R_N(t)
=
\frac12q_N(t),
\label{eq:nl-RN}
\end{equation}
and
\begin{equation}
B_N(t)
=
\int_0^t r_N(s)\,ds.
\label{eq:nl-BN}
\end{equation}

\begin{lemma}[Scalar identities and bounds]
\label{lem:nl-scalar-identities}
On the event
\begin{equation}
\|\mS\|_{\mathrm{op}}\le K,
\label{eq:nl-S-K-event}
\end{equation}
the solution of \cref{eq:nl-finite-scalar} exists globally and
satisfies
\begin{equation}
\min\{1,\kappa_N\}
\le
a_N(t)
\le
\max\{1,\kappa_N\},
\qquad
t\ge0.
\label{eq:nl-a-barrier}
\end{equation}
Moreover,
\begin{equation}
a_N(t)
=
\kappa_N
e^{-2B_N(t)}
M_N(q_N(t))
\label{eq:nl-a-selfconsistent}
\end{equation}
and
\begin{equation}
r_N(t)
=
\kappa_N
e^{-2B_N(t)}
M_N'(q_N(t)).
\label{eq:nl-r-selfconsistent}
\end{equation}
For every fixed \(T<\infty\), there exists a deterministic
\(C_T<\infty\) such that, with probability tending to one,
\begin{equation}
\sup_{0\le t\le T}
\left(
|a_N(t)|
+
|q_N(t)|
+
|r_N(t)|
+
|B_N(t)|
\right)
\le C_T.
\label{eq:nl-scalar-uniform-bound}
\end{equation}
\end{lemma}

\begin{proof}
Since \(\mS\succeq0\),
\begin{equation}
0\le m_N(q)\le\|\mS\|_{\mathrm{op}}\le K.
\label{eq:nl-mN-bound}
\end{equation}
Hence
\begin{equation}
a_N<1
\quad\Longrightarrow\quad
\dot a_N\ge0,
\qquad
a_N>1
\quad\Longrightarrow\quad
\dot a_N\le0,
\label{eq:nl-a-direction}
\end{equation}
which proves \cref{eq:nl-a-barrier}.

Define
\begin{equation}
\Phi_N(t)
=
\log a_N(t)
+
2B_N(t)
-
\log M_N(q_N(t)).
\label{eq:nl-PhiN}
\end{equation}
Then
\begin{equation}
\begin{aligned}
\dot\Phi_N(t)
&=
4m_N(q_N)(1-a_N)
+
2a_Nm_N(q_N)
\\
&\quad
-
2(2-a_N)m_N(q_N)
\\
&=
0.
\end{aligned}
\label{eq:nl-PhiN-dot}
\end{equation}
Since
\begin{equation}
M_N(0)=1,
\qquad
\Phi_N(0)=\log\kappa_N,
\label{eq:nl-PhiN-zero}
\end{equation}
we obtain \cref{eq:nl-a-selfconsistent}.  Multiplying by
\(m_N=M_N'/M_N\) gives \cref{eq:nl-r-selfconsistent}.

Finally, \cref{eq:nl-a-barrier} and
\cref{eq:nl-mN-bound} imply
\begin{equation}
|\dot q_N(t)|
\le
2\left(
2+\max\{1,\kappa_N\}
\right)
\label{eq:nl-qN-bound}
\end{equation}
and
\begin{equation}
0\le r_N(t)
\le
K\max\{1,\kappa_N\}.
\label{eq:nl-rN-bound}
\end{equation}
Integration gives \cref{eq:nl-scalar-uniform-bound}.
\end{proof}

Define the proxy trajectory
\begin{equation}
\overline \mU_N(t)
=
e^{-B_N(t)}
\mU(0)e^{R_N(t)\mS}.
\label{eq:nl-proxy}
\end{equation}
Differentiation gives
\begin{equation}
\dot{\overline \mU}_N
=
(2-a_N)\overline \mU_N\mS
-
r_N\overline \mU_N .
\label{eq:nl-proxy-flow}
\end{equation}
Thus \(\overline \mU_N\) is exactly the dynamics obtained by replacing
the true Gram matrices \(\mG,\mH\) by \(a_N\mI_n,r_N\mI_n\).

\subsubsection{Self-consistency of the proxy and stability of the exact flow}
\label{app:narrow-self-consistency}

We now carry out Step 4.  Conditional on \(\mS\), the matrices in this
subsection are functions only of the independent Gaussian
initialization \(\mZ\).

We first prove a uniform concentration result.

\begin{lemma}[Uniform weighted Gaussian Gram concentration]
\label{lem:nl-weighted-gram}
Let \(\mZ_N\in\mathbb R^{n_N\times p_N}\) have independent standard
Gaussian entries, and suppose
\begin{equation}
\frac{n_N}{p_N}\longrightarrow0 .
\label{eq:nl-gram-np}
\end{equation}
Let \(\mC_N(t)\), \(0\le t\le T\), be deterministic symmetric matrices
satisfying
\begin{equation}
\sup_N\sup_{0\le t\le T}
\|\mC_N(t)\|_{\mathrm{op}}
\le K
\label{eq:nl-C-bound}
\end{equation}
and
\begin{equation}
\sup_N
\sup_{\substack{s,t\in[0,T]\\s\ne t}}
\frac{
\|\mC_N(t)-\mC_N(s)\|_{\mathrm{op}}
}{
|t-s|
}
\le K_1.
\label{eq:nl-C-Lipschitz}
\end{equation}
Then
\begin{equation}
\sup_{0\le t\le T}
\left\|
\frac1{p_N}
\mZ_N\mC_N(t)\mZ_N^\top
-
\frac{\Tr\mC_N(t)}{p_N}\mI_{n_N}
\right\|_{\mathrm{op}}
\xrightarrow{\mathbb P}0.
\label{eq:nl-weighted-gram-conv}
\end{equation}
\end{lemma}

\begin{proof}
Fix first \(t\) and write
\begin{equation}
\mD_N
=
\frac1p\mZ\mC\mZ^\top
-
\frac{\Tr\mC}{p}\mI_n.
\label{eq:nl-D-fixed}
\end{equation}
For a deterministic unit vector \(\vv\in\mathbb R^n\),
\begin{equation}
\mZ^\top \vv
\overset{d}=\vg,
\qquad
\vg\sim N(0,\mI_p),
\label{eq:nl-Zv-gaussian}
\end{equation}
and therefore
\begin{equation}
\vv^\top \mD_N\vv
=
\frac1p
\left(
\vg^\top \mC\vg-\Tr\mC
\right).
\label{eq:nl-quadratic-reduction}
\end{equation}
Writing the eigenvalues of \(\mC\) as
\(\lambda_1,\ldots,\lambda_p\),
\begin{equation}
\vg^\top \mC\vg-\Tr\mC
=
\sum_{i=1}^{p}
\lambda_i(\vg_i^2-1).
\label{eq:nl-quadratic-diagonal}
\end{equation}
For
\begin{equation}
|s|\le\frac1{4K},
\label{eq:nl-s-range}
\end{equation}
the Gaussian moment-generating function and the elementary inequality
for \(-u-\frac12\log(1-2u)\) give
\begin{equation}
\log
\E
\exp
\left[
s(\vg^\top \mC\vg-\Tr\mC)
\right]
\le
2s^2\|\mC\|_F^2
\le
2s^2pK^2 .
\label{eq:nl-quadratic-mgf}
\end{equation}
A Chernoff bound therefore yields a constant \(c>0\) such that
\begin{equation}
\mathbb P
\left(
|\vg^\top \mC\vg-\Tr\mC|>p\varepsilon
\right)
\le
2\exp
\left[
-cp
\min
\left\{
\frac{\varepsilon^2}{K^2},
\frac{\varepsilon}{K}
\right\}
\right].
\label{eq:nl-quadratic-tail}
\end{equation}

Let \(\mathcal N\) be a \(1/4\)-net of the unit sphere
\(\mathbb S^{n-1}\) satisfying
\begin{equation}
|\mathcal N|
\le
9^n.
\label{eq:nl-net-size}
\end{equation}
For every symmetric matrix \(\mD\),
\begin{equation}
\|\mD\|_{\mathrm{op}}
\le
2\max_{\vv\in\mathcal N}
|\vv^\top \mD\vv|.
\label{eq:nl-net-op}
\end{equation}
Hence
\begin{equation}
\mathbb P
\left(
\|\mD_N\|_{\mathrm{op}}>2\varepsilon
\right)
\le
2
\exp
\left[
n\log9-c_\varepsilon p
\right],
\label{eq:nl-fixed-time-bound}
\end{equation}
which tends to zero because \(n/p\to0\).

To obtain uniformity in time, define
\begin{equation}
\mD_N(t)
=
\frac1p\mZ\mC_N(t)\mZ^\top
-
\frac{\Tr\mC_N(t)}p\mI_n .
\label{eq:nl-Dt}
\end{equation}
For \(s,t\in[0,T]\),
\begin{equation}
\|\mD_N(t)-\mD_N(s)\|_{\mathrm{op}}
\le
\left(
\frac{\|\mZ\|_{\mathrm{op}}^2}{p}+1
\right)
K_1|t-s|.
\label{eq:nl-D-time-Lipschitz}
\end{equation}
The preceding fixed-matrix argument applied to \(\mC=\mI_p\) gives
\begin{equation}
\frac{\|\mZ\|_{\mathrm{op}}^2}{p}
=
O_{\mathbb P}(1).
\label{eq:nl-Z-op-bound}
\end{equation}
We now make the passage from fixed time to uniform time explicit.  Fix
\(\varepsilon>0\) and \(\delta>0\).  By
\cref{eq:nl-Z-op-bound}, there exists a deterministic \(M<\infty\) such
that, for all sufficiently large \(N\),
\begin{equation}
\mathbb P\left(
  \left(
    \frac{\|\mZ\|_{\mathrm{op}}^2}{p}+1
  \right)K_1>M
\right)
<\delta.
\label{eq:nl-random-Lipschitz-truncation}
\end{equation}
Choose a deterministic mesh
\begin{equation}
0=t_0<t_1<\cdots<t_J=T,
\qquad
\max_{0\le j<J}(t_{j+1}-t_j)
\le\frac{\varepsilon}{4M}.
\label{eq:nl-deterministic-time-mesh}
\end{equation}
The number \(J\) depends only on \(T,M,\varepsilon\), and in particular is
independent of \(N\).  A union bound and
\cref{eq:nl-fixed-time-bound} therefore give
\begin{equation}
\mathbb P\left(
  \max_{0\le j\le J}\|\mD_N(t_j)\|_{\mathrm{op}}
  >\frac{\varepsilon}{2}
\right)
\longrightarrow0.
\label{eq:nl-mesh-union-bound}
\end{equation}
On the complement of the event in
\cref{eq:nl-random-Lipschitz-truncation}, every \(t\in[0,T]\) has a mesh
point \(t_j\) for which \cref{eq:nl-D-time-Lipschitz} implies
\begin{equation}
\|\mD_N(t)-\mD_N(t_j)\|_{\mathrm{op}}
\le\frac{\varepsilon}{4}.
\label{eq:nl-mesh-interpolation}
\end{equation}
Consequently,
\begin{equation}
\limsup_{N\to\infty}
\mathbb P\left(
  \sup_{0\le t\le T}\|\mD_N(t)\|_{\mathrm{op}}
  >\varepsilon
\right)
\le\delta.
\label{eq:nl-uniform-time-probability}
\end{equation}
Since \(\delta>0\) is arbitrary, this proves
\cref{eq:nl-weighted-gram-conv}.
\end{proof}

For the proxy trajectory,
\begin{equation}
\overline \mG_N
=
\overline \mU_N\overline \mU_N^\top
=
\frac1p
\mZ\mC_{G,N}(t)\mZ^\top,
\label{eq:nl-Gbar}
\end{equation}
where
\begin{equation}
\mC_{G,N}(t)
=
\kappa_N
e^{-2B_N(t)}
e^{q_N(t)\mS}.
\label{eq:nl-CG}
\end{equation}
By \cref{eq:nl-a-selfconsistent},
\begin{equation}
\frac1p
\Tr\mC_{G,N}(t)
=
a_N(t).
\label{eq:nl-CG-trace}
\end{equation}
Similarly,
\begin{equation}
\overline \mH_N
=
\overline \mU_N\mS\overline \mU_N^\top
=
\frac1p
\mZ\mC_{H,N}(t)\mZ^\top,
\label{eq:nl-Hbar}
\end{equation}
where
\begin{equation}
\mC_{H,N}(t)
=
\kappa_N
e^{-2B_N(t)}
\mS e^{q_N(t)\mS},
\label{eq:nl-CH}
\end{equation}
and
\begin{equation}
\frac1p
\Tr\mC_{H,N}(t)
=
r_N(t).
\label{eq:nl-CH-trace}
\end{equation}

On the event \(\|\mS\|_{\mathrm{op}}\le K\),
\cref{lem:nl-scalar-identities} implies uniform bounds on
\(q_N,B_N,a_N,r_N\) on every fixed time interval.  Consequently,
for every fixed \(T\),
\begin{equation}
\sup_{0\le t\le T}
\left(
\|\mC_{G,N}(t)\|_{\mathrm{op}}
+
\|\mC_{H,N}(t)\|_{\mathrm{op}}
\right)
\le C_T
\label{eq:nl-CGH-bound}
\end{equation}
with probability tending to one.  Differentiating
\cref{eq:nl-CG}--\cref{eq:nl-CH} and using the same bounds gives
\begin{equation}
\sup_{\substack{s,t\in[0,T]\\s\ne t}}
\frac{
\|\mC_{G,N}(t)-\mC_{G,N}(s)\|_{\mathrm{op}}
+
\|\mC_{H,N}(t)-\mC_{H,N}(s)\|_{\mathrm{op}}
}{
|t-s|
}
\le C_T
\label{eq:nl-CGH-Lipschitz}
\end{equation}
with probability tending to one.  Conditional on \(\mS\), we may
therefore apply \cref{lem:nl-weighted-gram} to obtain
\begin{equation}
\sup_{0\le t\le T}
\|\overline \mG_N(t)-a_N(t)\mI_n\|_{\mathrm{op}}
\xrightarrow{\mathbb P}0
\label{eq:nl-Gbar-scalar}
\end{equation}
and
\begin{equation}
\sup_{0\le t\le T}
\|\overline \mH_N(t)-r_N(t)\mI_n\|_{\mathrm{op}}
\xrightarrow{\mathbb P}0.
\label{eq:nl-Hbar-scalar}
\end{equation}

We now return to the comparison introduced in Step 3.

\begin{theorem}[Quenched narrow-latent approximation]
\label{thm:nl-quenched}
Assume \cref{eq:nl-scaling} and \cref{eq:common-spectrum-tight}.
Then, for every fixed \(T<\infty\),
\begin{equation}
\sup_{0\le t\le T}
\|\mU_N(t)-\overline \mU_N(t)\|_{\mathrm{op}}
\xrightarrow{\mathbb P}0.
\label{eq:nl-U-proxy-conv}
\end{equation}
Consequently,
\begin{equation}
\sup_{0\le t\le T}
\|\mG_N(t)-a_N(t)\mI_n\|_{\mathrm{op}}
\xrightarrow{\mathbb P}0
\label{eq:nl-G-scalar-main}
\end{equation}
and
\begin{equation}
\sup_{0\le t\le T}
\|\mH_N(t)-r_N(t)\mI_n\|_{\mathrm{op}}
\xrightarrow{\mathbb P}0.
\label{eq:nl-H-scalar-main}
\end{equation}
\end{theorem}

\begin{proof}
Define the exact vector field
\begin{equation}
F_\mS(\mV)
=
2\mV\mS-(\mV\mS\mV^\top)\mV-(\mV\mV^\top)\mV\mS.
\label{eq:nl-vector-field}
\end{equation}
By \cref{eq:nl-proxy-flow},
\begin{equation}
\begin{aligned}
F_\mS(\overline \mU_N)
-
\dot{\overline \mU}_N
&=
-
(\overline \mH_N-r_N\mI_n)\overline \mU_N
\\
&\quad
-
(\overline \mG_N-a_N\mI_n)\overline \mU_N\mS.
\end{aligned}
\label{eq:nl-proxy-residual}
\end{equation}
Thus
\begin{equation}
\begin{aligned}
\|
F_\mS(\overline \mU_N)
-
\dot{\overline \mU}_N
\|_{\mathrm{op}}
&\le
\|\overline \mH_N-r_N\mI_n\|_{\mathrm{op}}
\|\overline \mU_N\|_{\mathrm{op}}
\\
&\quad
+
\|\overline \mG_N-a_N\mI_n\|_{\mathrm{op}}
\|\overline \mU_N\|_{\mathrm{op}}
\|\mS\|_{\mathrm{op}}.
\end{aligned}
\label{eq:nl-residual-bound}
\end{equation}
Since
\begin{equation}
\|\overline \mU_N(t)\|_{\mathrm{op}}^2
=
\|\overline \mG_N(t)\|_{\mathrm{op}},
\label{eq:nl-Ubar-Gbar}
\end{equation}
\Cref{eq:nl-Gbar-scalar} and the uniform boundedness of
\(a_N\) imply
\begin{equation}
\sup_{0\le t\le T}
\|\overline \mU_N(t)\|_{\mathrm{op}}
=
O_{\mathbb P}(1).
\label{eq:nl-Ubar-bound}
\end{equation}
Hence
\begin{equation}
\varepsilon_N(T)
:=
\sup_{0\le t\le T}
\|
F_\mS(\overline \mU_N(t))
-
\dot{\overline \mU}_N(t)
\|_{\mathrm{op}}
\xrightarrow{\mathbb P}0.
\label{eq:nl-residual-small}
\end{equation}

It remains to prove stability of the exact ODE.  Suppose
\begin{equation}
\|\mV\|_{\mathrm{op}}\le M,
\qquad
\|\mW\|_{\mathrm{op}}\le M,
\qquad
\|\mS\|_{\mathrm{op}}\le K.
\label{eq:nl-Lipschitz-domain}
\end{equation}
Then
\begin{equation}
\|\mV\mV^\top-\mW\mW^\top\|_{\mathrm{op}}
\le
2M\|\mV-\mW\|_{\mathrm{op}}
\label{eq:nl-G-Lipschitz}
\end{equation}
and
\begin{equation}
\|\mV\mS\mV^\top-\mW\mS\mW^\top\|_{\mathrm{op}}
\le
2KM\|\mV-\mW\|_{\mathrm{op}}.
\label{eq:nl-H-Lipschitz}
\end{equation}
Expanding \cref{eq:nl-vector-field} therefore gives
\begin{equation}
\|F_\mS(\mV)-F_\mS(\mW)\|_{\mathrm{op}}
\le
\left(
2K+6KM^2
\right)
\|\mV-\mW\|_{\mathrm{op}}.
\label{eq:nl-F-Lipschitz}
\end{equation}
Set
\begin{equation}
\mE_N(t)
=
\mU_N(t)-\overline \mU_N(t)
\label{eq:nl-error}
\end{equation}
and introduce the proxy residual
\begin{equation}
\mR_N(t)
:=
F_{\mS}(\overline\mU_N(t))-\dot{\overline\mU}_N(t).
\label{eq:nl-proxy-residual-def}
\end{equation}
By definition of \(\varepsilon_N(T)\),
\begin{equation}
\sup_{0\le t\le T}\|\mR_N(t)\|_{\mathrm{op}}
\le\varepsilon_N(T).
\label{eq:nl-proxy-residual-sup}
\end{equation}
Moreover, the exact and proxy trajectories start from the same matrix, so
\begin{equation}
\mE_N(0)=0.
\label{eq:nl-error-zero}
\end{equation}
Subtracting their differential equations gives the exact error equation
\begin{equation}
\dot{\mE}_N(t)
=
F_{\mS}(\mU_N(t))-F_{\mS}(\overline\mU_N(t))
+\mR_N(t).
\label{eq:nl-error-ode}
\end{equation}

To make the bounded-set argument explicit, fix the deterministic spectral
bound \(K\) from \cref{eq:common-spectrum-tight} and choose \(M<\infty\).
Consider
\begin{equation}
\mathcal A_N(K,M)
:=
\left\{
\|\mS_N\|_{\mathrm{op}}\le K,
\ \sup_{0\le t\le T}\|\mU_N(t)\|_{\mathrm{op}}\le M,
\ \sup_{0\le t\le T}\|\overline\mU_N(t)\|_{\mathrm{op}}\le M
\right\}.
\label{eq:nl-stability-event}
\end{equation}
The standing spectral bound controls the first term.  For the exact
trajectory, \cref{lem:common-gram-flow} gives
\begin{equation}
\sup_{0\le t\le T}\|\mU_N(t)\|_{\mathrm{op}}^2
\le
\max\left\{1,\|\mU_N(0)\|_{\mathrm{op}}^2\right\},
\label{eq:nl-exact-U-bound-for-stability}
\end{equation}
and the common Gaussian initialization with \(n/p\to0\) makes the right-hand
side tight.  
The proxy term is tight by \cref{eq:nl-Ubar-bound}.  Hence \(M\) may be
chosen so that
\begin{equation}
\mathbb P\bigl(\mathcal A_N(K,M)\bigr)\longrightarrow1.
\label{eq:nl-stability-event-prob}
\end{equation}

On \(\mathcal A_N(K,M)\), the Lipschitz estimate
\cref{eq:nl-F-Lipschitz} holds for the two trajectories at every
\(t\in[0,T]\).  Integrating \cref{eq:nl-error-ode}, using
\cref{eq:nl-proxy-residual-sup}, and taking operator norms therefore gives
\begin{equation}
\|\mE_N(t)\|_{\mathrm{op}}
\le
C_T
\int_0^t
\|\mE_N(s)\|_{\mathrm{op}}\,ds
+
t\varepsilon_N(T),
\qquad
0\le t\le T,
\label{eq:nl-error-integral}
\end{equation}
where one may take \(C_T=2K+6KM^2\).  Gronwall's inequality now yields
\begin{equation}
\sup_{0\le t\le T}
\|\mE_N(t)\|_{\mathrm{op}}
\le
Te^{C_TT}\varepsilon_N(T)
\label{eq:nl-error-gronwall-bound}
\end{equation}
on \(\mathcal A_N(K,M)\).  Combining
\cref{eq:nl-stability-event-prob,eq:nl-residual-small,eq:nl-error-gronwall-bound}
gives
\begin{equation}
\sup_{0\le t\le T}
\|\mE_N(t)\|_{\mathrm{op}}
\xrightarrow{\mathbb P}0,
\label{eq:nl-error-gronwall}
\end{equation}
which proves \cref{eq:nl-U-proxy-conv}.

Finally,
\begin{equation}
\|
\mU_N\mU_N^\top
-
\overline \mU_N\overline \mU_N^\top
\|_{\mathrm{op}}
\le
\left(
\|\mU_N\|_{\mathrm{op}}
+
\|\overline \mU_N\|_{\mathrm{op}}
\right)
\|
\mU_N-\overline \mU_N
\|_{\mathrm{op}}
\label{eq:nl-G-transfer}
\end{equation}
and
\begin{equation}
\begin{aligned}
&
\|
\mU_N\mS \mU_N^\top
-
\overline \mU_N\mS\overline \mU_N^\top
\|_{\mathrm{op}}
\\
&\qquad\le
\|\mS\|_{\mathrm{op}}
\left(
\|\mU_N\|_{\mathrm{op}}
+
\|\overline \mU_N\|_{\mathrm{op}}
\right)
\|
\mU_N-\overline \mU_N
\|_{\mathrm{op}}.
\end{aligned}
\label{eq:nl-H-transfer}
\end{equation}
Combining these estimates with
\cref{eq:nl-Gbar-scalar}--\cref{eq:nl-Hbar-scalar}
proves
\cref{eq:nl-G-scalar-main}--\cref{eq:nl-H-scalar-main}.
\end{proof}

\cref{thm:nl-quenched} completes Steps 2--4 of the roadmap.
It is a quenched statement: conditional on a data covariance \(\mS\)
with bounded spectrum, the random Gaussian initialization produces
asymptotically scalar hidden-space Gram matrices.  No limiting
spectral distribution, and in particular no Marchenko--Pastur law,
has yet been used.

\subsubsection{Passage to a general limiting spectral measure}
\label{app:narrow-general-spectral}

We now implement Step 5 in detail.

The deterministic limiting spectral measure ${\nu_\phi}$ and the weak-convergence assumption are those stated at the beginning of this subsection.
\begin{equation}
M_{\nu_\phi}'(q)
=
\int
\lambda e^{q\lambda}\,d{\nu_\phi}(\lambda),
\label{eq:nl-Mnu-prime}
\end{equation}
\begin{equation}
M_{\nu_\phi}''(q)
=
\int
\lambda^2 e^{q\lambda}\,d{\nu_\phi}(\lambda),
\label{eq:nl-Mnu-second}
\end{equation}

We first show that the empirical transforms converge uniformly on
compact \(q\)-intervals.

\begin{lemma}[Uniform convergence of the spectral transforms]
\label{lem:nl-transform-convergence}
For every fixed \(Q<\infty\),
\begin{equation}
\sup_{|q|\le Q}
|M_N(q)-M_{\nu_\phi}(q)|
\xrightarrow{\mathbb P}0,
\label{eq:nl-M-uniform}
\end{equation}
\begin{equation}
\sup_{|q|\le Q}
|M_N'(q)-M_{\nu_\phi}'(q)|
\xrightarrow{\mathbb P}0,
\label{eq:nl-Mprime-uniform}
\end{equation}
\begin{equation}
\sup_{|q|\le Q}
|M_N''(q)-M_{\nu_\phi}''(q)|
\xrightarrow{\mathbb P}0,
\label{eq:nl-Msecond-uniform}
\end{equation}
and
\begin{equation}
\sup_{|q|\le Q}
|m_N(q)-m_{\nu_\phi}(q)|
\xrightarrow{\mathbb P}0.
\label{eq:nl-m-uniform}
\end{equation}
\end{lemma}

\begin{proof}
Fix \(Q<\infty\) and work on the event
\begin{equation}
\mathcal E_N
=
\{\|\mS_N\|_{\mathrm{op}}\le K\}.
\label{eq:nl-EN}
\end{equation}
Its probability tends to one.

For every fixed \(q\in[-Q,Q]\), the functions
\begin{equation}
f_{0,q}(\lambda)=e^{q\lambda},
\qquad
f_{1,q}(\lambda)=\lambda e^{q\lambda},
\qquad
f_{2,q}(\lambda)=\lambda^2e^{q\lambda}
\label{eq:nl-fjq}
\end{equation}
are bounded and continuous on \([0,K]\).  Weak convergence therefore
gives
\begin{equation}
M_N(q)
\xrightarrow{\mathbb P}
M_{\nu_\phi}(q),
\label{eq:nl-M-pointwise}
\end{equation}
\begin{equation}
M_N'(q)
\xrightarrow{\mathbb P}
M_{\nu_\phi}'(q),
\label{eq:nl-Mprime-pointwise}
\end{equation}
and
\begin{equation}
M_N''(q)
\xrightarrow{\mathbb P}
M_{\nu_\phi}''(q).
\label{eq:nl-Msecond-pointwise}
\end{equation}

On \(\mathcal E_N\), the mean-value theorem gives
\begin{equation}
|M_N(q)-M_N(\widetilde q)|
\le
Ke^{QK}|q-\widetilde q|,
\label{eq:nl-M-Lipschitz}
\end{equation}
\begin{equation}
|M_N'(q)-M_N'(\widetilde q)|
\le
K^2e^{QK}|q-\widetilde q|,
\label{eq:nl-Mprime-Lipschitz}
\end{equation}
and
\begin{equation}
|M_N''(q)-M_N''(\widetilde q)|
\le
K^3e^{QK}|q-\widetilde q|.
\label{eq:nl-Msecond-Lipschitz}
\end{equation}
The same bounds hold for the limiting transforms.

Fix \(\varepsilon>0\) and choose a deterministic finite grid
\begin{equation}
-Q=q_0<q_1<\cdots<q_J=Q
\label{eq:nl-q-grid}
\end{equation}
whose mesh is sufficiently small that the Lipschitz errors in
\cref{eq:nl-M-Lipschitz}--\cref{eq:nl-Msecond-Lipschitz}
are all at most \(\varepsilon/4\).  For every \(q\in[-Q,Q]\), let
\(q_j\) be a nearest grid point.  Then, for example,
\begin{equation}
\begin{aligned}
|M_N(q)-M_{\nu_\phi}(q)|
&\le
|M_N(q)-M_N(q_j)|
\\
&\quad
+
|M_N(q_j)-M_{\nu_\phi}(q_j)|
\\
&\quad
+
|M_{\nu_\phi}(q_j)-M_{\nu_\phi}(q)|.
\end{aligned}
\label{eq:nl-grid-decomp}
\end{equation}
Thus
\begin{equation}
\sup_{|q|\le Q}
|M_N(q)-M_{\nu_\phi}(q)|
\le
\max_{0\le j\le J}
|M_N(q_j)-M_{\nu_\phi}(q_j)|
+
\frac{\varepsilon}{2}
\label{eq:nl-grid-bound}
\end{equation}
on \(\mathcal E_N\).  Since the grid is finite,
\cref{eq:nl-M-pointwise} and a union bound imply
\cref{eq:nl-M-uniform}.  The same argument proves
\cref{eq:nl-Mprime-uniform} and
\cref{eq:nl-Msecond-uniform}.

To control the ratio, note that for \(|q|\le Q\),
\begin{equation}
M_N(q)\ge e^{-QK}
\label{eq:nl-MN-lower}
\end{equation}
on \(\mathcal E_N\), and
\begin{equation}
M_{\nu_\phi}(q)\ge e^{-QK}.
\label{eq:nl-Mnu-lower}
\end{equation}
Therefore
\begin{equation}
\begin{aligned}
|m_N(q)-m_{\nu_\phi}(q)|
&\le
e^{QK}
|M_N'(q)-M_{\nu_\phi}'(q)|
\\
&\quad
+
Ke^{3QK}
|M_N(q)-M_{\nu_\phi}(q)|.
\end{aligned}
\label{eq:nl-ratio-bound}
\end{equation}
Taking the supremum and using
\cref{eq:nl-M-uniform}--\cref{eq:nl-Mprime-uniform}
proves \cref{eq:nl-m-uniform}.
\end{proof}

The quantity \(m_{\nu_\phi}\) has a useful interpretation.  Define the
exponentially tilted probability measure
\begin{equation}
d\nu_{\phi,q}(\lambda)
=
\frac{
e^{q\lambda}
}{
M_{\nu_\phi}(q)
}
\,d{\nu_\phi}(\lambda).
\label{eq:nl-tilted-nu}
\end{equation}
Then
\begin{equation}
m_{\nu_\phi}(q)
=
\int\lambda\,d\nu_{\phi,q}(\lambda).
\label{eq:nl-mnu-mean}
\end{equation}
Furthermore,
\begin{equation}
m_{\nu_\phi}'(q)
=
\frac{
M_{\nu_\phi}''(q)M_{\nu_\phi}(q)-M_{\nu_\phi}'(q)^2
}{
M_{\nu_\phi}(q)^2
}
=
\operatorname{Var}_{\nu_{\phi,q}}(\lambda)
\ge0.
\label{eq:nl-mnu-variance}
\end{equation}
Hence
\begin{equation}
0\le m_{\nu_\phi}(q)\le K
\label{eq:nl-mnu-range}
\end{equation}
and
\begin{equation}
0\le m_{\nu_\phi}'(q)\le K^2.
\label{eq:nl-mnu-Lipschitz}
\end{equation}

We first obtain a common compact set containing the empirical and
limiting trajectories.

\begin{lemma}[Compactness of the scalar trajectories]
\label{lem:nl-scalar-compactness}
For every fixed \(T<\infty\), there exist deterministic constants
\(a_->0\), \(a_+<\infty\), and \(Q_T<\infty\) such that, with
probability tending to one,
\begin{equation}
a_N(t)\in[a_-,a_+],
\qquad
|q_N(t)|\le Q_T,
\qquad
0\le t\le T,
\label{eq:nl-finite-compact}
\end{equation}
while deterministically
\begin{equation}
a(t)\in[a_-,a_+],
\qquad
|q(t)|\le Q_T,
\qquad
0\le t\le T.
\label{eq:nl-limit-compact}
\end{equation}
\end{lemma}

\begin{proof}
Since \(\kappa_N\to\kappa\in(0,\infty)\), there exist deterministic
constants
\begin{equation}
0<\kappa_-<\kappa_+<\infty
\label{eq:nl-kappa-compact}
\end{equation}
such that
\begin{equation}
\kappa_-\le\kappa_N\le\kappa_+
\label{eq:nl-kappa-bounds}
\end{equation}
for all sufficiently large \(N\).

The barrier argument used in
\cref{lem:nl-scalar-identities} gives
\begin{equation}
\min\{1,\kappa_N\}
\le
a_N(t)
\le
\max\{1,\kappa_N\}
\label{eq:nl-aN-compact}
\end{equation}
and
\begin{equation}
\min\{1,\kappa\}
\le
a(t)
\le
\max\{1,\kappa\}.
\label{eq:nl-a-compact}
\end{equation}
Thus \(a_-\) and \(a_+\) may be chosen independently of \(N\).

Moreover,
\begin{equation}
|\dot q_N(t)|
\le
2(2+a_+)
\label{eq:nl-qNdot-compact}
\end{equation}
and
\begin{equation}
|\dot q(t)|
\le
2(2+a_+).
\label{eq:nl-qdot-compact}
\end{equation}
Since both \(q_N(0)\) and \(q(0)\) vanish, one may take
\begin{equation}
Q_T
=
2(2+a_+)T.
\label{eq:nl-QT}
\end{equation}
\end{proof}

We can now derive an explicit perturbation estimate.

\begin{prop}[Convergence of the scalar dynamics]
\label{prop:nl-scalar-convergence}
For every fixed \(T<\infty\),
\begin{equation}
\sup_{0\le t\le T}
\left(
|a_N(t)-a(t)|
+
|q_N(t)-q(t)|
\right)
\xrightarrow{\mathbb P}0.
\label{eq:nl-scalar-convergence}
\end{equation}
\end{prop}

\begin{proof}
Let \(Q_T\) be as in
\cref{lem:nl-scalar-compactness}, and define
\begin{equation}
\delta_N(T)
=
\sup_{|q|\le Q_T}
|m_N(q)-m_{\nu_\phi}(q)|.
\label{eq:nl-deltaNT}
\end{equation}
By \cref{lem:nl-transform-convergence},
\begin{equation}
\delta_N(T)
\xrightarrow{\mathbb P}0.
\label{eq:nl-deltaNT-zero}
\end{equation}

Set
\begin{equation}
\Delta a_N(t)
=
a_N(t)-a(t)
\label{eq:nl-Delta-a}
\end{equation}
and
\begin{equation}
\Delta q_N(t)
=
q_N(t)-q(t).
\label{eq:nl-Delta-q}
\end{equation}
Subtracting the \(q\)-equations gives
\begin{equation}
\dot{\Delta q}_N(t)
=
-2\Delta a_N(t),
\label{eq:nl-Delta-q-dot}
\end{equation}
hence
\begin{equation}
|\Delta q_N(t)|
\le
2
\int_0^t
|\Delta a_N(s)|\,ds.
\label{eq:nl-Delta-q-integral}
\end{equation}

Define
\begin{equation}
F_N(a,q)
=
4a(1-a)m_N(q)
\label{eq:nl-FN}
\end{equation}
and
\begin{equation}
F(a,q)
=
4a(1-a)m_{\nu_\phi}(q).
\label{eq:nl-F}
\end{equation}
Then
\begin{equation}
\dot{\Delta a}_N(t)
=
F_N(a_N(t),q_N(t))
-
F(a(t),q(t)).
\label{eq:nl-Delta-a-dot}
\end{equation}
Adding and subtracting \(F(a_N,q_N)\),
\begin{equation}
\begin{aligned}
|\dot{\Delta a}_N|
&\le
|F_N(a_N,q_N)-F(a_N,q_N)|
\\
&\quad
+
|F(a_N,q_N)-F(a,q)|.
\end{aligned}
\label{eq:nl-Delta-a-split}
\end{equation}

Since \(a_N\in[a_-,a_+]\), there exists \(C_a<\infty\) such that
\begin{equation}
|a_N(1-a_N)|
\le C_a.
\label{eq:nl-Ca}
\end{equation}
Therefore
\begin{equation}
|F_N(a_N,q_N)-F(a_N,q_N)|
\le
4C_a\delta_N(T).
\label{eq:nl-spectral-field-error}
\end{equation}

The derivatives of the limiting vector field are
\begin{equation}
\partial_aF(a,q)
=
4(1-2a)m_{\nu_\phi}(q)
\label{eq:nl-partial-a}
\end{equation}
and
\begin{equation}
\partial_qF(a,q)
=
4a(1-a)m_{\nu_\phi}'(q).
\label{eq:nl-partial-q}
\end{equation}
Using \cref{eq:nl-mnu-range},
\cref{eq:nl-mnu-Lipschitz}, and the compactness of the \(a\)
coordinate, there is a deterministic \(C_T<\infty\) such that
\begin{equation}
|F(a_N,q_N)-F(a,q)|
\le
C_T
\left(
|\Delta a_N|+|\Delta q_N|
\right).
\label{eq:nl-F-difference}
\end{equation}
Hence
\begin{equation}
|\dot{\Delta a}_N(t)|
\le
4C_a\delta_N(T)
+
C_T
\left(
|\Delta a_N(t)|+|\Delta q_N(t)|
\right).
\label{eq:nl-Delta-a-bound}
\end{equation}

Set
\begin{equation}
D_N(t)
=
|\Delta a_N(t)|+|\Delta q_N(t)|.
\label{eq:nl-DN}
\end{equation}
Since
\begin{equation}
D_N(0)
=
|\kappa_N-\kappa|,
\label{eq:nl-DN-zero}
\end{equation}
integration of
\cref{eq:nl-Delta-q-dot} and
\cref{eq:nl-Delta-a-bound} gives
\begin{equation}
D_N(t)
\le
|\kappa_N-\kappa|
+
4C_aT\delta_N(T)
+
C_T'
\int_0^tD_N(s)\,ds
\label{eq:nl-DN-integral}
\end{equation}
for a deterministic constant \(C_T'<\infty\).  Therefore
\begin{equation}
\sup_{0\le t\le T}D_N(t)
\le
e^{C_T'T}
\left[
|\kappa_N-\kappa|
+
4C_aT\delta_N(T)
\right].
\label{eq:nl-DN-gronwall}
\end{equation}
The right-hand side tends to zero in probability, proving
\cref{eq:nl-scalar-convergence}.
\end{proof}

The remaining scalar quantities converge as well.
Define
\begin{equation}
R(t)
=
\frac12q(t),
\label{eq:nl-limit-R}
\end{equation}
and
\begin{equation}
B(t)
=
\int_0^t r(s)\,ds.
\label{eq:nl-limit-B}
\end{equation}
\begin{prop}[Convergence of \(r_N,R_N,B_N\)]
\label{prop:nl-clock-convergence}
For every fixed \(T<\infty\),
\begin{equation}
\sup_{0\le t\le T}
|r_N(t)-r(t)|
\xrightarrow{\mathbb P}0,
\label{eq:nl-r-convergence}
\end{equation}
\begin{equation}
\sup_{0\le t\le T}
|R_N(t)-R(t)|
\xrightarrow{\mathbb P}0,
\label{eq:nl-R-convergence}
\end{equation}
and
\begin{equation}
\sup_{0\le t\le T}
|B_N(t)-B(t)|
\xrightarrow{\mathbb P}0.
\label{eq:nl-B-convergence}
\end{equation}
\end{prop}

\begin{proof}
From
\begin{equation}
r_N
=
a_Nm_N(q_N),
\qquad
r
=
am_{\nu_\phi}(q),
\label{eq:nl-r-def-pair}
\end{equation}
we have
\begin{equation}
\begin{aligned}
|r_N-r|
&\le
K|a_N-a|
+
a_+|m_N(q_N)-m_{\nu_\phi}(q_N)|
\\
&\quad
+
a_+|m_{\nu_\phi}(q_N)-m_{\nu_\phi}(q)|.
\end{aligned}
\label{eq:nl-r-error}
\end{equation}
By
\cref{eq:nl-deltaNT},
\cref{eq:nl-mnu-Lipschitz}, and
\cref{prop:nl-scalar-convergence},
\begin{equation}
|r_N-r|
\le
K|a_N-a|
+
a_+\delta_N(T)
+
a_+K^2|q_N-q|,
\label{eq:nl-r-error-final}
\end{equation}
which proves \cref{eq:nl-r-convergence}.

Since
\begin{equation}
R_N=\frac12q_N,
\qquad
R=\frac12q,
\label{eq:nl-R-pair}
\end{equation}
\Cref{eq:nl-R-convergence} follows immediately.

Finally,
\begin{equation}
|B_N(t)-B(t)|
\le
\int_0^t
|r_N(s)-r(s)|\,ds
\label{eq:nl-B-error}
\end{equation}
and hence
\begin{equation}
\sup_{0\le t\le T}
|B_N(t)-B(t)|
\le
T
\sup_{0\le t\le T}
|r_N(t)-r(t)|,
\label{eq:nl-B-error-sup}
\end{equation}
which proves \cref{eq:nl-B-convergence}.
\end{proof}

The limiting system admits the self-consistent representation
\begin{equation}
a(t)
=
\kappa
e^{-2B(t)}
M_{\nu_\phi}(q(t))
\label{eq:nl-a-limit-selfconsistent}
\end{equation}
and
\begin{equation}
r(t)
=
\kappa
e^{-2B(t)}
M_{\nu_\phi}'(q(t)).
\label{eq:nl-r-limit-selfconsistent}
\end{equation}
Indeed, if
\begin{equation}
\Phi(t)
=
\log a(t)
+
2B(t)
-
\log M_{\nu_\phi}(q(t)),
\label{eq:nl-Phi-limit}
\end{equation}
then
\begin{equation}
\begin{aligned}
\dot\Phi(t)
&=
4m_{\nu_\phi}(q)(1-a)
+
2am_{\nu_\phi}(q)
\\
&\quad
-
2(2-a)m_{\nu_\phi}(q)
\\
&=
0.
\end{aligned}
\label{eq:nl-Phi-limit-dot}
\end{equation}
Since
\begin{equation}
\Phi(0)=\log\kappa,
\label{eq:nl-Phi-limit-zero}
\end{equation}
\crefrange{eq:nl-a-limit-selfconsistent}{eq:nl-r-limit-selfconsistent} follow.

Equivalently, using \(q=2R\),
\begin{equation}
a(t)
=
\kappa
e^{-2B(t)}
M_{\nu_\phi}(2R(t)),
\label{eq:nl-a-limit-RB}
\end{equation}
\begin{equation}
r(t)
=
\kappa
e^{-2B(t)}
M_{\nu_\phi}'(2R(t)),
\label{eq:nl-r-limit-RB}
\end{equation}
\begin{equation}
\dot R(t)
=
2-a(t),
\label{eq:nl-Rdot-limit}
\end{equation}
and
\begin{equation}
\dot B(t)
=
r(t).
\label{eq:nl-Bdot-limit}
\end{equation}

Combining the quenched theorem with the scalar convergence proves the main
finite-time result stated at the beginning of this subsection.

\begin{proof}[Proof of \cref{thm:nl-general-nu}]
By \cref{thm:nl-quenched},
\begin{equation}
\sup_{0\le t\le T}
\|
\mG_N(t)-a_N(t)\mI_n
\|_{\mathrm{op}}
\xrightarrow{\mathbb P}0.
\label{eq:nl-G-quenched-used}
\end{equation}
Hence
\begin{equation}
\begin{aligned}
\sup_{0\le t\le T}
\|
\mG_N(t)-a(t)\mI_n
\|_{\mathrm{op}}
&\le
\sup_{0\le t\le T}
\|
\mG_N(t)-a_N(t)\mI_n
\|_{\mathrm{op}}
\\
&\quad
+
\sup_{0\le t\le T}
|a_N(t)-a(t)|.
\end{aligned}
\label{eq:nl-G-limit-triangle}
\end{equation}
The two terms tend to zero by
\cref{thm:nl-quenched} and
\cref{prop:nl-scalar-convergence}, respectively.
This proves \cref{eq:nl-G-general-limit}.  The same argument using
\cref{eq:nl-H-scalar-main} and
\cref{eq:nl-r-convergence} proves
\cref{eq:nl-H-general-limit}.
\end{proof}

\cref{thm:nl-general-nu} is the precise implementation of the
five-step roadmap stated at the beginning of the section.  The
dynamics first produces a data-dependent scalar approximation
\(a_N,r_N\); only afterward is the empirical spectral measure
replaced by its deterministic limit \({\nu_\phi}\).

\subsubsection{Consequences for population and train losses}
\label{app:narrow-losses}

We now return to Step 1 and substitute the scalarization obtained
above into the exact loss identities.

\begin{corollary}[Population loss]
\label{cor:nl-pop-loss}
Under the assumptions of
\cref{thm:nl-general-nu}, for every fixed \(T<\infty\),
\begin{equation}
\sup_{0\le t\le T}
\left|
\psi_N
\left(
L_N(t)-\frac12
\right)
-
a(t)
\left(
\frac{a(t)}2-1
\right)
\right|
\xrightarrow{\mathbb P}0.
\label{eq:nl-pop-limit}
\end{equation}
\end{corollary}

\begin{proof}
From \cref{eq:nl-pop-exact-G},
\begin{equation}
\psi_N
\left(
L_N-\frac12
\right)
=
-\frac1n\Tr\mG_N
+
\frac1{2n}\Tr\mG_N^2.
\label{eq:nl-pop-scaled-exact}
\end{equation}
\Cref{eq:nl-G-general-limit} implies
\begin{equation}
\left|
\frac1n\Tr\mG_N-a
\right|
\le
\|\mG_N-a\mI_n\|_{\mathrm{op}}
\xrightarrow{\mathbb P}0
\label{eq:nl-trG-limit}
\end{equation}
uniformly on \(0\le t\le T\).  Furthermore,
\begin{equation}
\|\mG_N^2-a^2\mI_n\|_{\mathrm{op}}
\le
\|\mG_N-a\mI_n\|_{\mathrm{op}}
\left(
\|\mG_N\|_{\mathrm{op}}+|a|
\right),
\label{eq:nl-G2-bound}
\end{equation}
so
\begin{equation}
\frac1n\Tr\mG_N^2
\xrightarrow{\mathbb P}
a^2
\label{eq:nl-trG2-limit}
\end{equation}
uniformly on compact time intervals.  Substitution into
\cref{eq:nl-pop-scaled-exact} gives
\cref{eq:nl-pop-limit}.
\end{proof}

\begin{corollary}[Train loss]
\label{cor:nl-train-loss}
Under the assumptions of
\cref{thm:nl-general-nu}, for every fixed \(T<\infty\),
\begin{equation}
\sup_{0\le t\le T}
\left|
\psi_N
\left(
\widehat L_N(t)
-
\frac1{2p}\Tr\mS_N
\right)
-
r(t)
\left(
\frac{a(t)}2-1
\right)
\right|
\xrightarrow{\mathbb P}0.
\label{eq:nl-train-limit-centered}
\end{equation}
If in addition
\begin{equation}
\frac{
\Tr\mS_N-p
}{
n
}
\xrightarrow{\mathbb P}0,
\label{eq:nl-trace-rate}
\end{equation}
then
\begin{equation}
\sup_{0\le t\le T}
\left|
\psi_N
\left(
\widehat L_N(t)-\frac12
\right)
-
r(t)
\left(
\frac{a(t)}2-1
\right)
\right|
\xrightarrow{\mathbb P}0.
\label{eq:nl-train-limit-half}
\end{equation}
\end{corollary}

\begin{proof}
From \cref{eq:nl-train-exact-GH},
\begin{equation}
\psi_N
\left(
\widehat L_N
-
\frac1{2p}\Tr\mS_N
\right)
=
-\frac1n\Tr\mH_N
+
\frac1{2n}\Tr(\mG_N\mH_N).
\label{eq:nl-train-scaled-exact}
\end{equation}
\Cref{eq:nl-H-general-limit} gives
\begin{equation}
\frac1n\Tr\mH_N
\xrightarrow{\mathbb P}
r.
\label{eq:nl-trH-limit}
\end{equation}
Also,
\begin{equation}
\|\mG_N\mH_N-ar\mI_n\|_{\mathrm{op}}
\le
\|\mG_N-a\mI_n\|_{\mathrm{op}}\|\mH_N\|_{\mathrm{op}}
+
|a|
\|\mH_N-r\mI_n\|_{\mathrm{op}},
\label{eq:nl-GH-bound}
\end{equation}
so
\begin{equation}
\frac1n
\Tr(\mG_N\mH_N)
\xrightarrow{\mathbb P}
ar.
\label{eq:nl-trGH-limit}
\end{equation}
This proves \cref{eq:nl-train-limit-centered}.  Finally,
\cref{eq:nl-trace-rate} is equivalent to
\begin{equation}
\psi_N
\left(
\frac1{2p}\Tr\mS_N
-
\frac12
\right)
\xrightarrow{\mathbb P}0,
\label{eq:nl-baseline-half}
\end{equation}
which yields \cref{eq:nl-train-limit-half}.
\end{proof}

For the expected losses, isotropy fixes the empirical baseline:
\begin{equation}
\E
\left[
\frac1p\Tr\mS_N
\right]
=
1.
\label{eq:nl-expected-trace}
\end{equation}

\begin{corollary}[Expected losses, uniformly on compact time intervals]
\label{cor:nl-expected-losses}
Under the assumptions of \cref{thm:nl-general-nu} and the
isotropy assumption \cref{eq:nl-isotropy}, for every fixed
\(T<\infty\),
\begin{equation}
\E
\left[
\sup_{0\le t\le T}
\left|
\psi_N
\left(
L_N(t)-\frac12
\right)
-
a(t)
\left(
\frac{a(t)}2-1
\right)
\right|
\right]
\longrightarrow
0.
\label{eq:nl-expected-pop-uniform-L1}
\end{equation}
If, in addition, for some \(\delta>0\),
\begin{equation}
\sup_N
\E
\|\mS_N\|_{\mathrm{op}}^{1+\delta}
<
\infty,
\label{eq:nl-UI-S}
\end{equation}
then
\begin{equation}
\E
\left[
\sup_{0\le t\le T}
\left|
\psi_N
\left(
\widehat L_N(t)
-
\frac1{2p}\Tr\mS_N
\right)
-
r(t)
\left(
\frac{a(t)}2-1
\right)
\right|
\right]
\longrightarrow
0.
\label{eq:nl-expected-train-uniform-L1}
\end{equation}
Consequently,
\begin{equation}
\sup_{0\le t\le T}
\left|
\psi_N
\left(
\E L_N(t)-\frac12
\right)
-
a(t)
\left(
\frac{a(t)}2-1
\right)
\right|
\longrightarrow
0,
\label{eq:nl-expected-pop}
\end{equation}
and, under \cref{eq:nl-UI-S},
\begin{equation}
\sup_{0\le t\le T}
\left|
\psi_N
\left(
\E\widehat L_N(t)-\frac12
\right)
-
r(t)
\left(
\frac{a(t)}2-1
\right)
\right|
\longrightarrow
0.
\label{eq:nl-expected-train}
\end{equation}
\end{corollary}

\begin{proof}
Set
\begin{equation}
R_N
:=
\max
\left\{
1,\|\mU_N(0)\|_{\mathrm{op}}^2
\right\}.
\label{eq:nl-UI-R}
\end{equation}
\cref{lem:common-gram-flow} gives, for all \(t\ge0\),
\begin{equation}
\|\mG_N(t)\|_{\mathrm{op}}
=
\|\mU_N(t)\|_{\mathrm{op}}^2
\le
R_N.
\label{eq:nl-G-UI-bound}
\end{equation}
Moreover, the standing Gaussian initialization
\cref{eq:common-Gaussian-init}, the narrow-latent scaling
\cref{eq:nl-scaling}, and the Gaussian singular-value bound
\cref{eq:mf-gaussian-smax} imply that, for every fixed \(q<\infty\),
\begin{equation}
\sup_N
\E R_N^q
<
\infty.
\label{eq:nl-UI-R-moments}
\end{equation}
Indeed, \(\mU_N(0)=\sigma_N\mZ_N\), while
\(p_N\sigma_N^2\to\kappa\) and \(n_N/p_N\to0\), so integration of
the Gaussian operator-norm tail gives uniformly bounded moments of
\(\|\mU_N(0)\|_{\mathrm{op}}\) of every fixed order.

For the population loss, the exact identity
\cref{eq:nl-pop-scaled-exact} and \cref{eq:nl-G-UI-bound} give,
uniformly in \(t\ge0\),
\begin{equation}
\left|
\psi_N
\left(
L_N(t)-\frac12
\right)
\right|
\le
\|\mG_N(t)\|_{\mathrm{op}}
+
\frac12
\|\mG_N(t)\|_{\mathrm{op}}^2
\le
R_N+\frac12R_N^2.
\label{eq:nl-pop-UI}
\end{equation}
The deterministic function
\(
a(t)(a(t)/2-1)
\)
is bounded on every compact time interval.  Hence
\cref{eq:nl-UI-R-moments} and \cref{eq:nl-pop-UI} imply that the
random variables
\begin{equation}
\sup_{0\le t\le T}
\left|
\psi_N
\left(
L_N(t)-\frac12
\right)
-
a(t)
\left(
\frac{a(t)}2-1
\right)
\right|
\label{eq:nl-pop-uniform-error}
\end{equation}
form a uniformly integrable family.  By
\cref{eq:nl-pop-limit}, they converge to zero in probability.
Vitali's theorem therefore yields
\cref{eq:nl-expected-pop-uniform-L1}.

For the train loss, the exact identity
\cref{eq:nl-train-scaled-exact} gives
\begin{equation}
\left|
\psi_N
\left(
\widehat L_N(t)
-
\frac1{2p}\Tr\mS_N
\right)
\right|
\le
\|\mH_N(t)\|_{\mathrm{op}}
\left(
1+\frac12\|\mG_N(t)\|_{\mathrm{op}}
\right).
\label{eq:nl-train-UI-first}
\end{equation}
Since
\begin{equation}
\|\mH_N(t)\|_{\mathrm{op}}
=
\|\mU_N(t)\mS_N\mU_N(t)^\top\|_{\mathrm{op}}
\le
\|\mS_N\|_{\mathrm{op}}
\|\mG_N(t)\|_{\mathrm{op}},
\label{eq:nl-H-UI-bound}
\end{equation}
we obtain from \cref{eq:nl-G-UI-bound}
\begin{equation}
\left|
\psi_N
\left(
\widehat L_N(t)
-
\frac1{2p}\Tr\mS_N
\right)
\right|
\le
\|\mS_N\|_{\mathrm{op}}
R_N
\left(
1+\frac12R_N
\right).
\label{eq:nl-train-UI}
\end{equation}
By the standing initialization
\cref{eq:common-Gaussian-init}, \(\mU_N(0)\) is independent of
\(\mX_N\), and hence \(R_N\) is independent of \(\mS_N\).
Consequently, \cref{eq:nl-UI-S} and
\cref{eq:nl-UI-R-moments} imply
\begin{equation}
\begin{aligned}
&
\E
\left[
\|\mS_N\|_{\mathrm{op}}
R_N
\left(
1+\frac12R_N
\right)
\right]^{1+\delta}
\\
&\qquad=
\E\|\mS_N\|_{\mathrm{op}}^{1+\delta}
\,
\E
\left[
R_N
\left(
1+\frac12R_N
\right)
\right]^{1+\delta}
\le
C,
\end{aligned}
\label{eq:nl-train-UI-moment}
\end{equation}
where \(C<\infty\) is independent of \(N\).  Since
\(
r(t)(a(t)/2-1)
\)
is bounded on \(0\le t\le T\), it follows that the random variables
\begin{equation}
\sup_{0\le t\le T}
\left|
\psi_N
\left(
\widehat L_N(t)
-
\frac1{2p}\Tr\mS_N
\right)
-
r(t)
\left(
\frac{a(t)}2-1
\right)
\right|
\label{eq:nl-train-uniform-error}
\end{equation}
are uniformly integrable.  Their convergence to zero in probability
is exactly \cref{eq:nl-train-limit-centered}.  Vitali's theorem
therefore gives \cref{eq:nl-expected-train-uniform-L1}.

Finally,
\begin{equation}
\begin{aligned}
&
\sup_{0\le t\le T}
\left|
\psi_N
\left(
\E L_N(t)-\frac12
\right)
-
a(t)
\left(
\frac{a(t)}2-1
\right)
\right|
\\
&\qquad\le
\E
\left[
\sup_{0\le t\le T}
\left|
\psi_N
\left(
L_N(t)-\frac12
\right)
-
a(t)
\left(
\frac{a(t)}2-1
\right)
\right|
\right],
\end{aligned}
\label{eq:nl-expected-pop-transfer}
\end{equation}
so \cref{eq:nl-expected-pop} follows from
\cref{eq:nl-expected-pop-uniform-L1}.  Similarly,
\cref{eq:nl-expected-trace} gives
\begin{equation}
\E
\left[
\frac1{2p}\Tr\mS_N
\right]
=
\frac12,
\label{eq:nl-expected-baseline-half}
\end{equation}
and therefore
\begin{equation}
\begin{aligned}
&
\sup_{0\le t\le T}
\left|
\psi_N
\left(
\E\widehat L_N(t)-\frac12
\right)
-
r(t)
\left(
\frac{a(t)}2-1
\right)
\right|
\\
&\qquad\le
\E
\left[
\sup_{0\le t\le T}
\left|
\psi_N
\left(
\widehat L_N(t)
-
\frac1{2p}\Tr\mS_N
\right)
-
r(t)
\left(
\frac{a(t)}2-1
\right)
\right|
\right].
\end{aligned}
\label{eq:nl-expected-train-transfer}
\end{equation}
\cref{eq:nl-expected-train} now follows from
\cref{eq:nl-expected-train-uniform-L1}.
\end{proof}
\subsubsection{Large-time behavior of the limiting system}
\label{app:narrow-large-time}

We now analyze the deterministic limiting system after taking the
high-dimensional limit.  All statements in this subsection concern the
order of limits
\begin{equation}
N\longrightarrow\infty
\quad\text{first},
\qquad
t\longrightarrow\infty
\quad\text{second}.
\label{eq:nl-order-limits}
\end{equation}
No uniform-in-time approximation of the finite-dimensional trajectory
is asserted.

Recall that
\begin{equation}
\dot a(t)
=
4a(t)m_{\nu_\phi}(q(t))(1-a(t)),
\qquad
\dot q(t)
=
2(2-a(t)),
\qquad
a(0)=\kappa,
\qquad
q(0)=0.
\label{eq:nl-large-system}
\end{equation}
We write
\begin{equation}
w_0
:=
\nu_\phi(\{0\})
\label{eq:nl-w0}
\end{equation}
for the null-space mass and define the scaled limiting loss corrections
\begin{equation}
F_{\mathrm{pop}}(t)
:=
a(t)\left(\frac{a(t)}2-1\right)
\label{eq:nl-Fpop}
\end{equation}
and
\begin{equation}
F_{\mathrm{tr}}(t)
:=
r(t)\left(\frac{a(t)}2-1\right),
\qquad
r(t)=a(t)m_{\nu_\phi}(q(t)).
\label{eq:nl-Ftr}
\end{equation}

We will use the following regular-edge conventions.  At the upper edge,
let
\begin{equation}
\lambda_+
:=
\sup\operatorname{supp}\nu_\phi>0
\label{eq:nl-upper-edge}
\end{equation}
and, whenever upper-edge asymptotics are invoked, assume that
\begin{equation}
\frac{d\nu_\phi}{d\lambda}(\lambda)
=
c_+
(\lambda_+-\lambda)^{\alpha_+}
\left(
1+O(\lambda_+-\lambda)
\right),
\qquad
\alpha_+>-1,
\qquad
c_+>0,
\label{eq:nl-upper-edge-density}
\end{equation}
in a neighborhood of \(\lambda_+\).  Set
\begin{equation}
\beta_+
:=
\alpha_++1,
\qquad
A_+
:=
c_+\Gamma(\beta_+).
\label{eq:nl-upper-edge-constants}
\end{equation}

For lower-edge asymptotics, write
\begin{equation}
\nu_\phi
=
w_0\delta_0+\nu_{\phi,\mathrm c},
\qquad
0<w_0<1,
\label{eq:nl-lower-decomposition}
\end{equation}
let
\begin{equation}
\lambda_-
:=
\inf\operatorname{supp}\nu_{\phi,\mathrm c}
\ge0,
\label{eq:nl-lower-edge}
\end{equation}
and assume that near \(\lambda_-\),
\begin{equation}
\frac{d\nu_{\phi,\mathrm c}}{d\lambda}(\lambda)
=
c_-
(\lambda-\lambda_-)^{\alpha_-}
\left(
1+O(\lambda-\lambda_-)
\right),
\qquad
\alpha_->-1,
\qquad
c_->0.
\label{eq:nl-lower-edge-density}
\end{equation}
Set
\begin{equation}
\beta_-
:=
\alpha_-+1,
\qquad
A_-
:=
c_-\Gamma(\beta_-).
\label{eq:nl-lower-edge-constants}
\end{equation}

\begin{prop}[Large-time alternatives and asymptotics]
\label{prop:nl-large-time-alternatives}
Assume $\nu_\phi\neq\delta_0$, let
$w_0=\nu_\phi(\{0\})$, and, for $\kappa>2$, set
\begin{equation}
    C_\kappa:=\frac{\kappa^2}{\kappa-1}.
    \label{eq:nl-Ckappa}
\end{equation}

\begin{enumerate}

\item
\textbf{Monotonic alignment.}
If $0<\kappa\le2$, then $q$ is nondecreasing and
\begin{equation}
    q(t)\to+\infty,
    \qquad
    a(t)\to1.
    \label{eq:nl-alignment-small-kappa}
\end{equation}

\item
\textbf{Non-monotonic alignment.}
If $\kappa>2$ and
\begin{equation}
    C_\kappa w_0^2<4,
    \label{eq:nl-alignment-condition}
\end{equation}
then $q$ initially decreases, has a unique minimum at $a=2$, and
subsequently increases, with
\begin{equation}
    q(t)\to+\infty,
    \qquad
    a(t)\to1.
    \label{eq:nl-alignment-large-kappa}
\end{equation}

In either alignment case, under the regular upper-edge assumption
\cref{eq:nl-upper-edge-density},
\begin{equation}
\begin{gathered}
    q(t)=2t+O(1),
    \qquad
    r(t)=\lambda_+-\frac{\beta_+}{2t}+O(t^{-2}),
    \\
    |1-a(t)|
    \asymp
    t^{2\beta_+}e^{-4\lambda_+t}
    \quad(\kappa\neq1),
    \qquad
    a(t)\equiv1\quad(\kappa=1),
    \\
    F_{\mathrm{pop}}(\infty)=-\frac12,
    \qquad
    F_{\mathrm{tr}}(\infty)=-\frac{\lambda_+}{2},
    \\
    F_{\mathrm{pop}}(t)-F_{\mathrm{pop}}(\infty)
    \asymp
    t^{4\beta_+}e^{-8\lambda_+t}
    \quad(\kappa\neq1),
    \qquad
    F_{\mathrm{tr}}(t)-F_{\mathrm{tr}}(\infty)
    \sim\frac{\beta_+}{4t}.
\end{gathered}
\label{eq:nl-alignment-summary}
\end{equation}

\item
\textbf{Strict null-space trapping.}
If $\kappa>2$ and
\begin{equation}
    C_\kappa w_0^2>4,
    \label{eq:nl-trapping-condition}
\end{equation}
then
\begin{equation}
    q(t)\to-\infty,
    \qquad
    a(t)\to a_*>2,
    \qquad
    r(t)\to0,
    \qquad
    \frac{a_*^2}{a_*-1}=C_\kappa w_0^2.
    \label{eq:nl-astar}
\end{equation}
Set $v_*:=2(a_*-2)$. Under the regular lower-edge assumption
\cref{eq:nl-lower-edge-density},
\begin{equation}
\begin{array}{c|c|c|c}
 & q(t) & a(t)-a_* & r(t) \\ \hline
\lambda_->0
&
-v_*t+O(1)
&
\asymp t^{-\beta_-}e^{-\lambda_-v_*t}
&
\asymp t^{-\beta_-}e^{-\lambda_-v_*t}
\\[1mm]
\lambda_-=0
&
-v_*t+o(t)
&
\asymp t^{-\beta_-}
&
\asymp t^{-(\beta_-+1)}
\end{array}
\label{eq:nl-trap-r-asympt}
\end{equation}
and
\begin{equation}
  \begin{gathered}
    F_{\mathrm{pop}}(\infty)
    =
    a_*\left(\frac{a_*}{2}-1\right),
    \qquad
    F_{\mathrm{tr}}(\infty)=0,\\
    F_{\mathrm{pop}}(t)-F_{\mathrm{pop}}(\infty)
    \asymp a(t)-a_*,
    \qquad
    F_{\mathrm{tr}}(t)\asymp r(t).
  \end{gathered}
  \label{eq:nl-trap-loss-summary}
\end{equation}

\item
\textbf{Critical boundary.}
If $\kappa>2$ and
\begin{equation}
    C_\kappa w_0^2=4,
    \label{eq:nl-critical-condition}
\end{equation}
then $a(t)>2$ at every finite time and
$q(t)\to-\infty$, $a(t)\to2$, $r(t)\to0$.
If \cref{eq:nl-lower-edge-density} holds and $\lambda_->0$, then
\begin{equation}
\begin{gathered}
    q(t)
    =
    -\frac{2}{\lambda_-}\log t
    +
    \frac{\beta_-}{\lambda_-}\log\log t
    +O(1),
    \qquad
    a(t)-2\sim\frac{1}{\lambda_-t},
    \qquad
    r(t)\sim\frac{1}{4\lambda_-t^2},
    \\
    F_{\mathrm{pop}}(t)\sim\frac{1}{\lambda_-t},
    \qquad
    F_{\mathrm{tr}}(t)\sim\frac{1}{8\lambda_-^2t^3}.
\end{gathered}
\label{eq:nl-critical-positive-summary}
\end{equation}
If instead $\lambda_-=0$, then
\begin{equation}
\begin{gathered}
    -q(t)\asymp t^{2/(\beta_-+2)},
    \qquad
    a(t)-2
    \asymp
    F_{\mathrm{pop}}(t)
    \asymp
    t^{-\beta_-/(\beta_-+2)},
    \\
    r(t)
    \asymp
    t^{-2(\beta_-+1)/(\beta_-+2)},
    \qquad
    F_{\mathrm{tr}}(t)
    \asymp
    t^{-(3\beta_-+2)/(\beta_-+2)}.
\end{gathered}
\label{eq:nl-critical-zero-summary}
\end{equation}

\end{enumerate}

In particular,
\begin{equation}
    w_0=0
    \quad\Longrightarrow\quad
    q(t)\to+\infty,
    \qquad
    a(t)\to1
    \qquad
    \text{for every }\kappa>0.
    \label{eq:nl-no-null-trap}
\end{equation}
The proof below gives the explicit leading constants, phase shifts, and
the corresponding asymptotics of $B(t)$.
\end{prop}

\begin{proof}
For every finite \(q\), the assumption
\(\nu_\phi\ne\delta_0\) implies
\begin{equation}
    m_{\nu_\phi}(q)>0.
    \label{eq:nl-large-m-positive}
\end{equation}
Hence
\begin{equation}
    a(t)<1
    \Longrightarrow
    \dot a(t)>0,
    \qquad
    a(t)>1
    \Longrightarrow
    \dot a(t)<0.
    \label{eq:nl-large-a-monotone}
\end{equation}

For \(\kappa\ne1\), eliminating time from
\cref{eq:nl-large-system} gives
\begin{equation}
    \frac{2-a}{a(1-a)}\,da
    =
    2m_{\nu_\phi}(q)\,dq.
    \label{eq:nl-first-integral-differential}
\end{equation}
Since
\begin{equation}
    \frac{2-a}{a(1-a)}
    =
    \frac2a+\frac1{1-a},
    \label{eq:nl-partial-fractions}
\end{equation}
and
\begin{equation}
    m_{\nu_\phi}(q)
    =
    \frac{d}{dq}\log M_{\nu_\phi}(q),
    \label{eq:nl-m-log-derivative}
\end{equation}
integration from \(a(0)=\kappa\), \(q(0)=0\) yields
\begin{equation}
    \frac{a(t)^2}{|1-a(t)|}
    =
    \frac{\kappa^2}{|1-\kappa|}
    M_{\nu_\phi}(q(t))^2.
    \label{eq:nl-first-integral}
\end{equation}
For \(\kappa=1\), uniqueness in
\cref{eq:nl-large-system} gives
\begin{equation}
    a(t)\equiv1.
    \label{eq:nl-kappa-one}
\end{equation}

\paragraph{Qualitative alternatives.}
If \(0<\kappa<2\), then \(a(t)<2\) for every \(t\), and hence
\begin{equation}
    \dot q(t)>0.
    \label{eq:nl-q-positive-small}
\end{equation}
Since \(m_{\nu_\phi}\) is nondecreasing by
\cref{eq:nl-mnu-variance},
\begin{equation}
    m_{\nu_\phi}(q(t))
    \ge
    m_{\nu_\phi}(0)
    =
    \int\lambda\,d\nu_\phi(\lambda)
    >
    0.
    \label{eq:nl-m-positive-alignment}
\end{equation}
Together with \cref{eq:nl-large-a-monotone}, this forces
\(a(t)\to1\). If \(q\) were bounded above, then
\(\dot q(t)\to2\), a contradiction. Thus
\cref{eq:nl-alignment-small-kappa} holds. The case
\(\kappa=2\) follows because
\begin{equation}
    \dot a(0)
    =
    -8m_{\nu_\phi}(0)
    <
    0,
    \label{eq:nl-kappa-two-derivative}
\end{equation}
so the trajectory immediately enters \(1<a<2\).

Now suppose \(\kappa>2\). While \(a>2\),
\cref{eq:nl-first-integral} takes the form
\begin{equation}
    F(a(t))
    =
    C_\kappa M_{\nu_\phi}(q(t))^2,
    \qquad
    F(a):=\frac{a^2}{a-1}.
    \label{eq:nl-F-branch}
\end{equation}
For \(a\ge2\),
\begin{equation}
    F(2)=4,
    \qquad
    F'(a)
    =
    \frac{a(a-2)}{(a-1)^2}
    \ge0,
    \label{eq:nl-F-properties}
\end{equation}
with strict inequality for \(a>2\), while
\begin{equation}
    M_{\nu_\phi}(q)
    \longrightarrow
    w_0
    \qquad
    \text{as }q\to-\infty.
    \label{eq:nl-M-minus-infinity}
\end{equation}

Under \cref{eq:nl-alignment-condition}, there is a unique finite
\(q_*\) satisfying
\begin{equation}
    C_\kappa M_{\nu_\phi}(q_*)^2=4.
    \label{eq:nl-alignment-crossing-value}
\end{equation}
At the corresponding time \(\tau\),
\begin{equation}
    a(\tau)=2,
    \qquad
    \dot a(\tau)
    =
    -8m_{\nu_\phi}(q_*)
    <
    0.
    \label{eq:nl-alignment-crossing}
\end{equation}
Thus \(q\) has a unique minimum at \(\tau\); afterward
\(1<a<2\), and the preceding argument proves
\cref{eq:nl-alignment-large-kappa}.

Under \cref{eq:nl-trapping-condition}, the right-hand side of
\cref{eq:nl-F-branch} remains strictly larger than \(4\), so
\(a(t)>2\) and \(q(t)\) is strictly decreasing. If \(q(t)\) had a
finite limit \(q_{\mathrm f}\), then
\cref{eq:nl-F-branch} would imply
\begin{equation}
    a(t)\longrightarrow a_{\mathrm f}>2,
    \label{eq:nl-trap-finite-a-contradiction}
\end{equation}
and hence
\begin{equation}
    \dot q(t)
    \longrightarrow
    2(2-a_{\mathrm f})<0,
    \label{eq:nl-trap-finite-q-contradiction}
\end{equation}
contradicting convergence of \(q\). Therefore \(q\to-\infty\), and
taking the limit in \cref{eq:nl-F-branch} proves
\cref{eq:nl-astar}. For later use, set
\begin{equation}
    D_*:=C_\kappa w_0^2>4,
    \label{eq:nl-Dstar}
\end{equation}
so that the relevant root of \cref{eq:nl-astar} is
\begin{equation}
    a_*
    =
    \frac{
        D_*+\sqrt{D_*(D_*-4)}
    }{2},
    \label{eq:nl-astar-explicit}
\end{equation}
and define
\begin{equation}
    v_*:=2(a_*-2)>0.
    \label{eq:nl-trap-speed}
\end{equation}

At the critical value \cref{eq:nl-critical-condition}, for every
finite \(q\),
\begin{equation}
    M_{\nu_\phi}(q)>w_0.
    \label{eq:nl-M-strict-above-atom}
\end{equation}
Hence \cref{eq:nl-F-branch} gives \(a(t)>2\) for every finite
\(t\). The same finite-limit contradiction shows that
\(q(t)\to-\infty\), after which \(F(a(t))\to4\), and therefore
\(a(t)\to2\). Since \(m_{\nu_\phi}(q)\to0\) as \(q\to-\infty\),
also \(r(t)\to0\).

Finally, if \(w_0=0\), neither
\cref{eq:nl-trapping-condition} nor
\cref{eq:nl-critical-condition} can hold. Thus every
\(\kappa>2\) lies in the non-monotonic alignment case, while
\(0<\kappa\le2\) lies in the monotonic one. This proves
\cref{eq:nl-no-null-trap}.

\paragraph{Upper-edge asymptotics in the alignment basin.}
Laplace's method applied to
\cref{eq:nl-upper-edge-density} gives
\begin{equation}
    M_{\nu_\phi}(q)
    =
    A_+
    e^{\lambda_+q}
    q^{-\beta_+}
    \left(
        1+O(q^{-1})
    \right)
    \label{eq:nl-M-general-asympt}
\end{equation}
and
\begin{equation}
    m_{\nu_\phi}(q)
    =
    \lambda_+
    -
    \frac{\beta_+}{q}
    +
    O(q^{-2}).
    \label{eq:nl-m-general-asympt}
\end{equation}

For \(\kappa\ne1\), the first integral gives
\begin{equation}
    |1-a(t)|
    =
    \frac{|1-\kappa|}{\kappa^2}
    \frac{a(t)^2}{M_{\nu_\phi}(q(t))^2}.
    \label{eq:nl-a-first-integral-use}
\end{equation}
Since \(a(t)\to1\) and \(q(t)\to+\infty\),
\begin{equation}
    |1-a(t)|
    =
    O\!\left(
        q(t)^{2\beta_+}
        e^{-2\lambda_+q(t)}
    \right).
    \label{eq:nl-a-q-asympt}
\end{equation}
Because \(\dot q(t)\to2\), this error is integrable. Hence
\begin{equation}
    q(t)
    =
    2t+q_\infty+o(1),
    \qquad
    q_\infty
    :=
    -2\int_0^\infty(a(s)-1)\,ds.
    \label{eq:nl-q-general-asympt}
\end{equation}
Substituting back into
\cref{eq:nl-a-first-integral-use} yields
\begin{equation}
    |1-a(t)|
    \sim
    K_a^+
    t^{2\beta_+}
    e^{-4\lambda_+t},
    \label{eq:nl-a-general-asympt}
\end{equation}
where
\begin{equation}
    K_a^+
    :=
    \frac{|1-\kappa|}{\kappa^2}
    \frac{2^{2\beta_+}}{A_+^2}
    e^{-2\lambda_+q_\infty}.
    \label{eq:nl-Ka-plus}
\end{equation}
When \(\kappa=1\), \cref{eq:nl-kappa-one} and the equation for
\(q\) give
\begin{equation}
    a(t)\equiv1,
    \qquad
    q(t)=2t.
    \label{eq:nl-a-kappa-one-repeat}
\end{equation}

Using \(r=a\,m_{\nu_\phi}(q)\),
\cref{eq:nl-m-general-asympt,eq:nl-q-general-asympt} give
\begin{equation}
    r(t)
    =
    \lambda_+
    -
    \frac{\beta_+}{2t}
    +
    O(t^{-2})
    +
    O\!\left(
        t^{2\beta_+}e^{-4\lambda_+t}
    \right).
    \label{eq:nl-r-general-asympt}
\end{equation}
The self-consistency identity
\cref{eq:nl-a-limit-selfconsistent} is equivalent to
\begin{equation}
    B(t)
    =
    \frac12
    \left[
        \log\kappa
        +
        \log M_{\nu_\phi}(q(t))
        -
        \log a(t)
    \right].
    \label{eq:nl-B-exact-log}
\end{equation}
Therefore
\begin{equation}
    B(t)
    =
    \lambda_+t
    -
    \frac{\beta_+}{2}\log t
    +
    B_\infty
    +
    O(t^{-1}),
    \label{eq:nl-B-general-asympt}
\end{equation}
where
\begin{equation}
    B_\infty
    :=
    \frac{\lambda_+q_\infty}{2}
    +
    \frac12
    \log\left(
        \frac{\kappa A_+}{2^{\beta_+}}
    \right).
    \label{eq:nl-Binfty}
\end{equation}
For \(\kappa\ne1\),
\begin{equation}
    K_a^+
    =
    |1-\kappa|e^{-4B_\infty}.
    \label{eq:nl-Ka-Binfty-relation}
\end{equation}

Finally,
\begin{equation}
    F_{\mathrm{pop}}(t)
    =
    -\frac12
    +
    \frac12(a(t)-1)^2
    \label{eq:nl-Fpop-square}
\end{equation}
gives
\begin{equation}
    F_{\mathrm{pop}}(t)
    -
    F_{\mathrm{pop}}(\infty)
    \sim
    \frac12(K_a^+)^2
    t^{4\beta_+}
    e^{-8\lambda_+t},
    \qquad
    F_{\mathrm{pop}}(\infty)=-\frac12,
    \label{eq:nl-pop-large-general}
\end{equation}
for \(\kappa\ne1\). Likewise,
\begin{equation}
    F_{\mathrm{tr}}(t)
    -
    F_{\mathrm{tr}}(\infty)
    =
    \frac{\lambda_+-r(t)}{2}
    +
    \frac{r(t)}{2}(a(t)-1)
    \label{eq:nl-train-large-decomp}
\end{equation}
yields
\begin{equation}
    F_{\mathrm{tr}}(t)
    -
    F_{\mathrm{tr}}(\infty)
    =
    \frac{\beta_+}{4t}
    +
    O(t^{-2})
    +
    O\!\left(
        t^{2\beta_+}e^{-4\lambda_+t}
    \right),
    \qquad
    F_{\mathrm{tr}}(\infty)
    =
    -\frac{\lambda_+}{2}.
    \label{eq:nl-train-large-general}
\end{equation}
These formulas imply the alignment summary
\cref{eq:nl-alignment-summary}.

\paragraph{Lower-edge asymptotics in the strict trapped branch.}
Set
\begin{equation}
    s(t):=-q(t),
    \qquad
    \Delta(s):=M_{\nu_\phi}(-s)-w_0.
    \label{eq:nl-lower-s-Delta}
\end{equation}
Laplace's method at the lower edge gives
\begin{equation}
    \Delta(s)
    =
    A_-
    e^{-\lambda_-s}
    s^{-\beta_-}
    \left(
        1+O(s^{-1})
    \right)
    \label{eq:nl-lower-M-asympt}
\end{equation}
and
\begin{equation}
    m_{\nu_\phi}(-s)
    \sim
    \begin{cases}
        \dfrac{\lambda_-A_-}{w_0}
        e^{-\lambda_-s}s^{-\beta_-},
        & \lambda_->0,
        \\[1.2ex]
        \dfrac{\beta_-A_-}{w_0}
        s^{-(\beta_-+1)},
        & \lambda_-=0.
    \end{cases}
    \label{eq:nl-lower-m-asympt}
\end{equation}

Expanding \cref{eq:nl-F-branch} at \(a_*\) gives
\begin{equation}
    a(t)-a_*
    \sim
    \frac{
        2a_*(a_*-1)
    }{
        (a_*-2)w_0
    }
    \Delta(s(t)).
    \label{eq:nl-a-lower-q-asympt}
\end{equation}
Moreover,
\begin{equation}
    \dot s(t)
    =
    2(a(t)-2)
    \longrightarrow
    v_*,
    \label{eq:nl-trap-s-speed}
\end{equation}
and therefore
\begin{equation}
    s(t)\sim v_*t.
    \label{eq:nl-trap-s-leading}
\end{equation}

Suppose first that \(\lambda_->0\). Then
\cref{eq:nl-a-lower-q-asympt} is exponentially integrable, so
\begin{equation}
    q_-
    :=
    \lim_{t\to\infty}
    \bigl(q(t)+v_*t\bigr)
    =
    -2\int_0^\infty(a(s)-a_*)\,ds
    \label{eq:nl-trap-phase}
\end{equation}
exists and
\begin{equation}
    q(t)
    =
    -v_*t+q_-+o(1).
    \label{eq:nl-trap-q-asympt}
\end{equation}
Define
\begin{equation}
    K_a^-
    :=
    \frac{
        2a_*(a_*-1)
    }{
        a_*-2
    }
    \frac{A_-}{w_0}
    v_*^{-\beta_-}
    e^{\lambda_-q_-}
    \label{eq:nl-Ka-minus}
\end{equation}
and
\begin{equation}
    K_r^-
    :=
    \frac{
        a_*\lambda_-A_-
    }{
        w_0
    }
    v_*^{-\beta_-}
    e^{\lambda_-q_-}.
    \label{eq:nl-Kr-minus}
\end{equation}
Then
\begin{equation}
    a(t)-a_*
    \sim
    K_a^-
    t^{-\beta_-}
    e^{-\lambda_-v_*t},
    \label{eq:nl-trap-a-positive-edge}
\end{equation}
and
\begin{equation}
    r(t)
    \sim
    K_r^-
    t^{-\beta_-}
    e^{-\lambda_-v_*t}.
    \label{eq:nl-trap-r-positive-edge}
\end{equation}

If instead \(\lambda_-=0\), define
\begin{equation}
    K_{a,0}^-
    :=
    \frac{
        2a_*(a_*-1)
    }{
        a_*-2
    }
    \frac{A_-}{w_0}
    v_*^{-\beta_-}
    \label{eq:nl-Ka-minus-zero}
\end{equation}
and
\begin{equation}
    K_{r,0}^-
    :=
    \frac{
        a_*\beta_-A_-
    }{
        w_0
    }
    v_*^{-(\beta_-+1)}.
    \label{eq:nl-Kr-minus-zero}
\end{equation}
Then
\begin{equation}
    a(t)-a_*
    \sim
    K_{a,0}^-
    t^{-\beta_-},
    \label{eq:nl-trap-a-zero-edge}
\end{equation}
and
\begin{equation}
    r(t)
    \sim
    K_{r,0}^-
    t^{-(\beta_-+1)}.
    \label{eq:nl-trap-r-zero-edge}
\end{equation}
Since
\begin{equation}
    \frac{d}{dt}
    \bigl(q(t)+v_*t\bigr)
    =
    -2(a(t)-a_*),
    \label{eq:nl-trap-clock-correction}
\end{equation}
integration gives the more precise refinement
\begin{equation}
    q(t)+v_*t
    =
    \begin{cases}
        -\dfrac{2K_{a,0}^-}{1-\beta_-}
        t^{1-\beta_-}
        +o(t^{1-\beta_-}),
        & 0<\beta_-<1,
        \\[1.2ex]
        -2K_{a,0}^-\log t+o(\log t),
        & \beta_-=1,
        \\[1.2ex]
        q_-
        +
        \dfrac{2K_{a,0}^-}{\beta_--1}
        t^{1-\beta_-}
        +o(t^{1-\beta_-}),
        & \beta_->1,
    \end{cases}
    \label{eq:nl-trap-q-zero-refined}
\end{equation}
where, in the last case,
\begin{equation}
    q_-
    =
    -2\int_0^\infty(a(s)-a_*)\,ds.
    \label{eq:nl-trap-q-zero-phase}
\end{equation}

The exact identity \cref{eq:nl-B-exact-log} gives
\begin{equation}
    B(t)
    \longrightarrow
    B_*
    :=
    \frac12
    \log\left(
        \frac{\kappa w_0}{a_*}
    \right).
    \label{eq:nl-trap-B-limit}
\end{equation}
Equivalently,
\begin{equation}
    B_*-B(t)
    =
    \int_t^\infty r(u)\,du.
    \label{eq:nl-trap-B-tail-integral}
\end{equation}
Using the preceding expressions for \(r\),
\begin{equation}
    B_*-B(t)
    \sim
    \begin{cases}
        \dfrac{K_r^-}{\lambda_-v_*}
        t^{-\beta_-}
        e^{-\lambda_-v_*t},
        & \lambda_->0,
        \\[1.2ex]
        \dfrac{a_*A_-}{w_0}
        v_*^{-(\beta_-+1)}
        t^{-\beta_-},
        & \lambda_-=0.
    \end{cases}
    \label{eq:nl-trap-B-asympt}
\end{equation}

Taylor expansion of \(F_{\mathrm{pop}}\) at \(a_*\) gives
\begin{equation}
    F_{\mathrm{pop}}(t)
    -
    F_{\mathrm{pop}}(\infty)
    \sim
    \begin{cases}
        (a_*-1)K_a^-
        t^{-\beta_-}
        e^{-\lambda_-v_*t},
        & \lambda_->0,
        \\[1.2ex]
        (a_*-1)K_{a,0}^-
        t^{-\beta_-},
        & \lambda_-=0,
    \end{cases}
    \label{eq:nl-trap-pop-asympt}
\end{equation}
where
\begin{equation}
    F_{\mathrm{pop}}(\infty)
    =
    a_*
    \left(
        \frac{a_*}{2}-1
    \right).
    \label{eq:nl-trap-pop-limit-proof}
\end{equation}
Moreover,
\begin{equation}
    F_{\mathrm{tr}}(t)
    =
    \frac{a(t)-2}{2}r(t),
    \label{eq:nl-trap-Ftr-identity}
\end{equation}
so
\begin{equation}
    F_{\mathrm{tr}}(t)
    \sim
    \begin{cases}
        \dfrac{a_*-2}{2}
        K_r^-
        t^{-\beta_-}
        e^{-\lambda_-v_*t},
        & \lambda_->0,
        \\[1.2ex]
        \dfrac{a_*-2}{2}
        K_{r,0}^-
        t^{-(\beta_-+1)},
        & \lambda_-=0.
    \end{cases}
    \label{eq:nl-trap-train-asympt}
\end{equation}
These formulas imply
\cref{eq:nl-trap-r-asympt,eq:nl-trap-loss-summary}.

\paragraph{Critical boundary.}
Write
\begin{equation}
    \delta(t):=a(t)-2>0,
    \qquad
    s(t):=-q(t).
    \label{eq:nl-critical-delta-s}
\end{equation}
Since \(C_\kappa w_0^2=4\), the first integral becomes
\begin{equation}
    \frac{\delta(t)^2}{1+\delta(t)}
    =
    \frac{8\Delta(s(t))}{w_0}
    +
    \frac{4\Delta(s(t))^2}{w_0^2},
    \label{eq:nl-critical-invariant}
\end{equation}
and hence
\begin{equation}
    \delta(t)
    \sim
    \left(
        \frac{8\Delta(s(t))}{w_0}
    \right)^{1/2}.
    \label{eq:nl-critical-delta-Delta}
\end{equation}

Suppose first that \(\lambda_->0\). From
\cref{eq:nl-lower-M-asympt,eq:nl-critical-delta-Delta},
\begin{equation}
    \delta(t)
    \sim
    d_{\mathrm c}
    e^{-\lambda_-s(t)/2}
    s(t)^{-\beta_-/2},
    \qquad
    d_{\mathrm c}
    :=
    \left(
        \frac{8A_-}{w_0}
    \right)^{1/2}.
    \label{eq:nl-critical-delta-positive}
\end{equation}
Since
\begin{equation}
    \dot s(t)=2\delta(t),
    \label{eq:nl-critical-sdot}
\end{equation}
integration of the inverse equation gives
\begin{equation}
    t
    =
    \frac{1}{\lambda_-d_{\mathrm c}}
    e^{\lambda_-s(t)/2}
    s(t)^{\beta_-/2}
    \left(
        1+O(s(t)^{-1})
    \right).
    \label{eq:nl-critical-inverse-clock}
\end{equation}
Taking logarithms and inverting yields
\begin{equation}
    q(t)
    =
    -\frac{2}{\lambda_-}\log t
    +
    \frac{\beta_-}{\lambda_-}\log\log t
    +
    q_{\mathrm c}
    +
    o(1),
    \label{eq:nl-critical-clock}
\end{equation}
where
\begin{equation}
    q_{\mathrm c}
    :=
    \frac{1}{\lambda_-}
    \log\left[
        \frac{w_0}{8A_-\lambda_-^2}
        \left(
            \frac{2}{\lambda_-}
        \right)^{\beta_-}
    \right].
    \label{eq:nl-critical-clock-constant}
\end{equation}

Furthermore,
\begin{equation}
    m_{\nu_\phi}(q(t))
    \sim
    \frac{\lambda_-}{8}\delta(t)^2,
    \label{eq:nl-critical-m-delta-positive}
\end{equation}
and hence
\begin{equation}
    \dot\delta(t)
    =
    -\lambda_-\delta(t)^2
    \left(
        1+o(1)
    \right).
    \label{eq:nl-critical-delta-ode-positive}
\end{equation}
Therefore
\begin{equation}
    a(t)-2
    \sim
    \frac{1}{\lambda_-t},
    \qquad
    m_{\nu_\phi}(q(t))
    \sim
    \frac{1}{8\lambda_-t^2},
    \qquad
    r(t)
    \sim
    \frac{1}{4\lambda_-t^2}.
    \label{eq:nl-critical-state-positive}
\end{equation}
The loss identities then give
\begin{equation}
    F_{\mathrm{pop}}(t)
    \sim
    \frac{1}{\lambda_-t},
    \qquad
    F_{\mathrm{tr}}(t)
    \sim
    \frac{1}{8\lambda_-^2t^3}.
    \label{eq:nl-critical-losses}
\end{equation}
Finally, \cref{eq:nl-B-exact-log} gives
\begin{equation}
    B(t)
    \longrightarrow
    B_{\mathrm c}
    :=
    \frac12
    \log\left(
        \frac{\kappa w_0}{2}
    \right)
    =
    \frac14\log(\kappa-1),
    \label{eq:nl-critical-B-limit}
\end{equation}
and
\begin{equation}
    B_{\mathrm c}-B(t)
    \sim
    \frac{1}{4\lambda_-t}.
    \label{eq:nl-critical-B}
\end{equation}
This proves
\cref{eq:nl-critical-positive-summary}.

Now suppose that \(\lambda_-=0\). Define
\begin{equation}
    d_0
    :=
    \left(
        \frac{8A_-}{w_0}
    \right)^{1/2},
    \label{eq:nl-critical-zero-d}
\end{equation}
so that
\begin{equation}
    \delta(t)
    \sim
    d_0s(t)^{-\beta_-/2}.
    \label{eq:nl-critical-zero-delta-s}
\end{equation}
Together with \cref{eq:nl-critical-sdot}, this gives
\begin{equation}
    s(t)^{(\beta_-+2)/2}
    \sim
    (\beta_-+2)d_0\,t.
    \label{eq:nl-critical-zero-s-integrated}
\end{equation}
Set
\begin{equation}
    Q_0
    :=
    \left[
        (\beta_-+2)d_0
    \right]^{2/(\beta_-+2)}.
    \label{eq:nl-critical-zero-Q}
\end{equation}
Then
\begin{equation}
    -q(t)
    \sim
    Q_0
    t^{2/(\beta_-+2)}
    \label{eq:nl-critical-zero-clock}
\end{equation}
and
\begin{equation}
    a(t)-2
    \sim
    \frac{Q_0}{\beta_-+2}
    t^{-\beta_-/(\beta_-+2)}.
    \label{eq:nl-critical-zero-a}
\end{equation}

From \cref{eq:nl-lower-m-asympt},
\begin{equation}
    m_{\nu_\phi}(-s)
    \sim
    \frac{\beta_-A_-}{w_0}
    s^{-(\beta_-+1)}.
    \label{eq:nl-critical-zero-m-s}
\end{equation}
Since
\begin{equation}
    Q_0^{\beta_-+2}
    =
    (\beta_-+2)^2
    \frac{8A_-}{w_0},
    \label{eq:nl-critical-Q-identity}
\end{equation}
we obtain
\begin{equation}
    m_{\nu_\phi}(q(t))
    \sim
    \frac{
        \beta_-Q_0
    }{
        8(\beta_-+2)^2
    }
    t^{-2(\beta_-+1)/(\beta_-+2)}
    \label{eq:nl-critical-zero-m}
\end{equation}
and
\begin{equation}
    r(t)
    \sim
    \frac{
        \beta_-Q_0
    }{
        4(\beta_-+2)^2
    }
    t^{-2(\beta_-+1)/(\beta_-+2)}.
    \label{eq:nl-critical-zero-r}
\end{equation}

Using
\begin{equation}
    F_{\mathrm{pop}}(t)
    =
    \delta(t)
    +
    \frac12\delta(t)^2
    \label{eq:nl-critical-Fpop-delta}
\end{equation}
and
\begin{equation}
    F_{\mathrm{tr}}(t)
    =
    \frac12r(t)\delta(t),
    \label{eq:nl-critical-Ftr-delta}
\end{equation}
gives
\begin{equation}
    F_{\mathrm{pop}}(t)
    \sim
    \frac{Q_0}{\beta_-+2}
    t^{-\beta_-/(\beta_-+2)}
    \label{eq:nl-critical-zero-pop}
\end{equation}
and
\begin{equation}
    F_{\mathrm{tr}}(t)
    \sim
    \frac{
        \beta_-Q_0^2
    }{
        8(\beta_-+2)^3
    }
    t^{-(3\beta_-+2)/(\beta_-+2)}.
    \label{eq:nl-critical-zero-train}
\end{equation}
The same limit \(B_{\mathrm c}\) as in
\cref{eq:nl-critical-B-limit} holds, and integration of
\cref{eq:nl-critical-zero-r} gives
\begin{equation}
    B_{\mathrm c}-B(t)
    \sim
    \frac{Q_0}{4(\beta_-+2)}
    t^{-\beta_-/(\beta_-+2)}.
    \label{eq:nl-critical-zero-B}
\end{equation}
These formulas prove
\cref{eq:nl-critical-zero-summary}.
\end{proof}

\paragraph{Marchenko--Pastur specialization.}

For isotropic Gaussian training data with
\begin{equation}
\frac{p}{m}
\longrightarrow
\phi\in(0,\infty),
\label{eq:nl-MP-ratio}
\end{equation}
the limiting spectral measure is the Marchenko--Pastur law
\begin{equation}
d\mu_\phi(\lambda)
=
\left(
1-\frac1\phi
\right)_+
\delta_0(d\lambda)
+
\frac{
\sqrt{
(\lambda_+-\lambda)(\lambda-\lambda_-)
}
}{
2\pi\phi\lambda
}
\mathbf 1_{[\lambda_-,\lambda_+]}(\lambda)
\,d\lambda,
\label{eq:nl-MP-density}
\end{equation}
where
\begin{equation}
\lambda_\pm
=
(1\pm\sqrt\phi)^2,
\qquad
w_0
=
\left(
1-\frac1\phi
\right)_+.
\label{eq:nl-MP-edges}
\end{equation}

At the upper edge,
\begin{equation}
\beta_+
=
\frac32,
\qquad
A_+
=
\frac{
1
}{
2\sqrt\pi\,
\phi^{3/4}
\lambda_+
}.
\label{eq:nl-MP-upper-constants}
\end{equation}
Consequently, throughout the alignment basin,
\begin{equation}
q(t)
=
2t+q_\infty+o(1),
\qquad
r(t)
=
\lambda_+
-
\frac{3}{4t}
+
O(t^{-2})
+
O(t^3e^{-4\lambda_+t}),
\label{eq:nl-MP-q-r}
\end{equation}
and, for \(\kappa\ne1\),
\begin{equation}
|1-a(t)|
\sim
K_a^+
t^3e^{-4\lambda_+t}.
\label{eq:nl-MP-a}
\end{equation}
The loss corrections satisfy
\begin{equation}
F_{\mathrm{pop}}(t)
-
F_{\mathrm{pop}}(\infty)
\sim
\frac12
(K_a^+)^2
t^6
e^{-8\lambda_+t}
\label{eq:nl-MP-pop-large}
\end{equation}
and
\begin{equation}
F_{\mathrm{tr}}(t)
-
F_{\mathrm{tr}}(\infty)
=
\frac{3}{8t}
+
O(t^{-2})
+
O(t^3e^{-4\lambda_+t}).
\label{eq:nl-MP-train-large}
\end{equation}

When \(\phi>1\), the zero atom has mass
\begin{equation}
w_0
=
1-\frac1\phi,
\label{eq:nl-MP-zero-atom}
\end{equation}
and the nonzero lower edge is strictly positive:
\begin{equation}
\lambda_-
=
(\sqrt\phi-1)^2>0.
\label{eq:nl-MP-lower-edge}
\end{equation}
The lower-edge constants are
\begin{equation}
\beta_-
=
\frac32,
\qquad
A_-
=
\frac{
1
}{
2\sqrt\pi\,
\phi^{3/4}
\lambda_-
}.
\label{eq:nl-MP-lower-constants}
\end{equation}

Thus, for \(\kappa>2\), alignment occurs precisely when
\begin{equation}
\frac{\kappa^2}{\kappa-1}
\left(
1-\frac1\phi
\right)^2
<4.
\label{eq:nl-MP-alignment-basin}
\end{equation}
Under the reverse strict inequality,
\begin{equation}
\frac{\kappa^2}{\kappa-1}
\left(
1-\frac1\phi
\right)^2
>4,
\label{eq:nl-MP-trapping-basin}
\end{equation}
define \(a_*\), \(v_*\), and \(q_-\) by
\cref{eq:nl-astar-explicit},
\cref{eq:nl-trap-speed}, and
\cref{eq:nl-trap-phase}.  Then
\begin{equation}
q(t)
=
-v_*t+q_-+o(1)
\label{eq:nl-MP-trapped-q}
\end{equation}
and
\begin{equation}
a(t)-a_*
\sim
K_{a,\phi}^-
t^{-3/2}
e^{-\lambda_-v_*t},
\label{eq:nl-MP-trapped-a}
\end{equation}
where
\begin{equation}
K_{a,\phi}^-
:=
\frac{
a_*(a_*-1)
}{
(a_*-2)
\sqrt\pi\,
\phi^{3/4}
\lambda_-w_0
}
v_*^{-3/2}
e^{\lambda_-q_-}.
\label{eq:nl-MP-Ka-minus}
\end{equation}
Likewise,
\begin{equation}
r(t)
\sim
K_{r,\phi}^-
t^{-3/2}
e^{-\lambda_-v_*t},
\label{eq:nl-MP-trapped-r}
\end{equation}
where
\begin{equation}
K_{r,\phi}^-
:=
\frac{
a_*
}{
2\sqrt\pi\,
\phi^{3/4}
w_0
}
v_*^{-3/2}
e^{\lambda_-q_-}.
\label{eq:nl-MP-Kr-minus}
\end{equation}
Consequently,
\begin{equation}
F_{\mathrm{pop}}(t)
-
F_{\mathrm{pop}}(\infty)
\sim
(a_*-1)K_{a,\phi}^-
t^{-3/2}
e^{-\lambda_-v_*t}
\label{eq:nl-MP-trapped-pop}
\end{equation}
and
\begin{equation}
F_{\mathrm{tr}}(t)
\sim
\frac{a_*-2}{2}
K_{r,\phi}^-
t^{-3/2}
e^{-\lambda_-v_*t}.
\label{eq:nl-MP-trapped-train}
\end{equation}
In particular,
\begin{equation}
a(t)-a_*
\asymp
r(t)
\asymp
t^{-3/2}
e^{-2\lambda_-(a_*-2)t}.
\label{eq:nl-MP-trapped-rate}
\end{equation}

At the critical boundary
\begin{equation}
\frac{\kappa^2}{\kappa-1}
\left(
1-\frac1\phi
\right)^2
=
4,
\label{eq:nl-MP-critical-basin}
\end{equation}
one obtains
\begin{equation}
a(t)-2
\sim
\frac{1}{\lambda_-t},
\qquad
r(t)
\sim
\frac{1}{4\lambda_-t^2},
\label{eq:nl-MP-critical-state}
\end{equation}
\begin{equation}
q(t)
=
-\frac{2}{\lambda_-}\log t
+
\frac{3}{2\lambda_-}\log\log t
+
q_{\mathrm c}
+
o(1),
\label{eq:nl-MP-critical-clock}
\end{equation}
and
\begin{equation}
F_{\mathrm{pop}}(t)
\sim
\frac{1}{\lambda_-t},
\qquad
F_{\mathrm{tr}}(t)
\sim
\frac{1}{8\lambda_-^2t^3}.
\label{eq:nl-MP-critical-losses}
\end{equation}

Although the Marchenko--Pastur lower edge satisfies
\(\lambda_-=0\) at \(\phi=1\), one then has \(w_0=0\).
Thus neither trapping nor the critical boundary is possible at that
point.  The \(\lambda_-=0\) trapped and critical asymptotics in
\cref{prop:nl-large-time-alternatives} instead describe more
general limiting spectral measures that simultaneously have a zero atom
and continuous spectrum accumulating at zero.

\subsubsection{Landscape interpretation of null-space trapping}
\label{app:narrow-landscape}

The large-time analysis above shows that, in the strict trapped branch,
the limiting system satisfies
\begin{equation}
    a(t)\longrightarrow a_*>2,
    \qquad
    r(t)\longrightarrow0,
\end{equation}
while at the critical boundary \(a(t)\to2\).
We now relate these values to the finite-dimensional landscape of the
empirical loss.  The statement below is deterministic and does not involve
a large-dimensional limit.

\begin{prop}[Null-space critical points of the empirical loss]
\label{prop:nl-nullspace-landscape}
Let $\mS\in\mathbb R^{p\times p}$ be symmetric positive semidefinite
and nonzero, and suppose $\mU\mS=0$.  Write
\begin{equation}
    \mG:=\mU\mU^\top.
\end{equation}
Then $\mU$ is a critical point of $\widehat L$. Moreover:
\begin{enumerate}
    \item If $\lambda_{\min}(\mG)>2$, then $\mU$ is a nonglobal
    local minimum.

    \item If $\lambda_{\min}(\mG)<2$, then $\mU$ is a strict saddle:
    the Hessian has a negative direction.

    \item If $\lambda_{\min}(\mG)=2$, then $\mU$ is a degenerate
    saddle. Its Hessian is positive semidefinite, but there exist
    arbitrarily small perturbations both increasing and decreasing
    $\widehat L$; the decreasing variation may first appear at
    fourth order.
\end{enumerate}
Consequently, among null-space critical points, the local minima are
precisely those satisfying
\begin{equation}
    \mU\mU^\top\succ2\mI_n.
\end{equation}
\end{prop}

\begin{proof}
Set
\begin{equation}
    \mG:=\mU\mU^\top,
    \qquad
    \mH:=\mU\mS\mU^\top.
    \label{eq:nl-nullspace-GH}
\end{equation}
By the exact train-loss identity
\cref{eq:nl-train-exact-GH},
\begin{equation}
\widehat L(\mU)
=
\frac{1}{2p}\Tr\mS
+
\frac{1}{2p}
\Tr\!\left[
    (\mG-2\mI_n)\mH
\right].
\label{eq:nl-nullspace-loss-identity}
\end{equation}
Under the null-space condition
\begin{equation}
    \mU\mS=0,
\label{eq:nl-nullspace-condition-proof}
\end{equation}
one has \(\mH=0\), and therefore
\begin{equation}
    \widehat L(\mU)
    =
    \frac{1}{2p}\Tr\mS.
\label{eq:nl-nullspace-loss-value}
\end{equation}
Moreover, \(\mU\) is critical.  Indeed, by the common gradient identity,
\begin{equation}
    \frac{\partial\widehat L}{\partial\mU}
    =
    -\frac1p
    \mU
    \left[
        \mS(\mI_p-\mU^\top\mU)
        +
        (\mI_p-\mU^\top\mU)\mS
    \right],
\label{eq:nl-nullspace-gradient}
\end{equation}
and both terms vanish under
\cref{eq:nl-nullspace-condition-proof}.

We first record the local expansion that will be used in the two
non-minimum cases.  For an arbitrary
\(\mDelta\in\mathbb R^{n\times p}\), let
\begin{equation}
    \mU_\epsilon
    :=
    \mU+\epsilon\mDelta.
\label{eq:nl-nullspace-linear-perturbation}
\end{equation}
Since \(\mU\mS=0\) and \(\mS\) is symmetric,
\begin{equation}
    \mU_\epsilon\mS\mU_\epsilon^\top
    =
    \epsilon^2
    \mDelta\mS\mDelta^\top.
\label{eq:nl-nullspace-H-expansion}
\end{equation}
Also,
\begin{equation}
\mU_\epsilon\mU_\epsilon^\top
=
\mG
+
\epsilon
\left(
    \mU\mDelta^\top+\mDelta\mU^\top
\right)
+
\epsilon^2
\mDelta\mDelta^\top.
\label{eq:nl-nullspace-G-expansion}
\end{equation}
Substituting
\cref{eq:nl-nullspace-H-expansion}--%
\cref{eq:nl-nullspace-G-expansion}
into \cref{eq:nl-nullspace-loss-identity} gives
\begin{align}
    \widehat L(\mU_\epsilon)-\widehat L(\mU)
    &=
    \frac{\epsilon^2}{2p}
    \Tr\!\left[
        (\mG-2\mI_n)
        \mDelta\mS\mDelta^\top
    \right]
    \notag\\
    &\quad+
    \frac{\epsilon^3}{2p}
    \Tr\!\left[
        (\mU\mDelta^\top+\mDelta\mU^\top)
        \mDelta\mS\mDelta^\top
    \right]
    \notag\\
    &\quad+
    \frac{\epsilon^4}{2p}
    \Tr\!\left[
        \mDelta\mDelta^\top
        \mDelta\mS\mDelta^\top
    \right].
\label{eq:nl-nullspace-full-expansion}
\end{align}
In particular, the Hessian quadratic form at the null-space critical
point is
\begin{equation}
    \left.
    \frac{d^2}{d\epsilon^2}
    \widehat L(\mU+\epsilon\mDelta)
    \right|_{\epsilon=0}
    =
    \frac1p
    \Tr\!\left[
        (\mG-2\mI_n)
        \mDelta\mS\mDelta^\top
    \right].
\label{eq:nl-nullspace-hessian-form}
\end{equation}

\paragraph{Case 1: \(\lambda_{\min}(\mG)>2\).}

By continuity, there exists a neighborhood of \(\mU\) such that
\begin{equation}
    \mU'\mU'^\top-2\mI_n\succ0
\label{eq:nl-nullspace-positive-neighborhood}
\end{equation}
for every \(\mU'\) in this neighborhood.  Since
\begin{equation}
    \mU'\mS\mU'^\top\succeq0,
\label{eq:nl-nullspace-H-positive}
\end{equation}
the identity \cref{eq:nl-nullspace-loss-identity} yields
\begin{equation}
\widehat L(\mU')-\widehat L(\mU)
=
\frac{1}{2p}
\Tr\!\left[
    (\mU'\mU'^\top-2\mI_n)
    \mU'\mS\mU'^\top
\right]
\ge0.
\label{eq:nl-nullspace-local-min}
\end{equation}
Thus \(\mU\) is a local minimum.

It is generally non-strict, since perturbations remaining in
\(\ker\mS\) may leave the loss unchanged.  It is also nonglobal:
because \(\mS\ne0\), choose a unit eigenvector \(\vx\) satisfying
\begin{equation}
    \mS\vx=\lambda\vx,
    \qquad
    \lambda>0.
\label{eq:nl-nullspace-positive-eigenvector}
\end{equation}
Let \(\widetilde{\mU}\) have first row \(\vx^\top\) and all other rows
zero.  Then
\begin{equation}
    \widetilde{\mU}^\top\widetilde{\mU}
    =
    \vx\vx^\top,
\label{eq:nl-nullspace-comparison-U}
\end{equation}
and hence
\begin{equation}
    \widehat L(\widetilde{\mU})
    =
    \frac{\Tr\mS-\lambda}{2p}
    <
    \frac{\Tr\mS}{2p}
    =
    \widehat L(\mU).
\label{eq:nl-nullspace-nonglobal}
\end{equation}

\paragraph{Case 2: \(\lambda_{\min}(\mG)<2\).}

Choose a unit eigenvector \(\vv\in\mathbb R^n\) such that
\begin{equation}
    \mG\vv=g\vv,
    \qquad
    g<2,
\label{eq:nl-nullspace-below-eigenvector}
\end{equation}
and choose \(\vx\) as in
\cref{eq:nl-nullspace-positive-eigenvector}.
Since \(\mU\mS=0\) and \(\lambda>0\),
\begin{equation}
    \mU\vx=0.
\label{eq:nl-nullspace-Ux-zero}
\end{equation}
Set
\begin{equation}
    \mDelta:=\vv\vx^\top.
\label{eq:nl-nullspace-negative-direction}
\end{equation}
Then
\begin{equation}
    \mDelta\mS\mDelta^\top
    =
    \lambda\vv\vv^\top,
\label{eq:nl-nullspace-negative-H}
\end{equation}
and the Hessian form
\cref{eq:nl-nullspace-hessian-form} becomes
\begin{equation}
    \left.
    \frac{d^2}{d\epsilon^2}
    \widehat L(\mU+\epsilon\vv\vx^\top)
    \right|_{\epsilon=0}
    =
    \frac{\lambda}{p}(g-2)
    <
    0.
\label{eq:nl-nullspace-negative-hessian}
\end{equation}
Thus the Hessian has a negative direction, and \(\mU\) is a strict
saddle in the usual optimization sense.

In fact, because \(\mU\vx=0\), the mixed term in
\cref{eq:nl-nullspace-G-expansion} vanishes, and the loss change is
exactly
\begin{equation}
    \widehat L(\mU+\epsilon\vv\vx^\top)
    -
    \widehat L(\mU)
    =
    \frac{\lambda\epsilon^2}{2p}
    \left(
        g-2+\epsilon^2
    \right),
\label{eq:nl-nullspace-negative-quadratic}
\end{equation}
which is negative for all sufficiently small
\(\epsilon\ne0\).

\paragraph{Case 3: \(\lambda_{\min}(\mG)=2\).}

In this case
\begin{equation}
    \mG-2\mI_n\succeq0.
\label{eq:nl-nullspace-boundary-PSD}
\end{equation}
Since
\begin{equation}
    \mDelta\mS\mDelta^\top\succeq0
\label{eq:nl-nullspace-DeltaSDelta-PSD}
\end{equation}
for every \(\mDelta\), the Hessian formula
\cref{eq:nl-nullspace-hessian-form} shows that
\begin{equation}
    \left.
    \frac{d^2}{d\epsilon^2}
    \widehat L(\mU+\epsilon\mDelta)
    \right|_{\epsilon=0}
    \ge0
\label{eq:nl-nullspace-boundary-hessian}
\end{equation}
for every direction \(\mDelta\).  Thus the Hessian is positive
semidefinite: there is no negative quadratic direction.

We now show that \(\mU\) is nevertheless a saddle.  Choose a unit
vector \(\vv\) such that
\begin{equation}
    \mG\vv=2\vv
\label{eq:nl-nullspace-critical-v}
\end{equation}
and let
\begin{equation}
    \vw:=\mU^\top\vv.
\label{eq:nl-nullspace-critical-w}
\end{equation}
Then
\begin{equation}
    \|\vw\|^2
    =
    \vv^\top\mG\vv
    =
    2,
\label{eq:nl-nullspace-critical-w-norm}
\end{equation}
while
\begin{equation}
    \mS\vw
    =
    \mS\mU^\top\vv
    =
    0.
\label{eq:nl-nullspace-critical-w-kernel}
\end{equation}
Choose again a unit positive-eigenvalue eigenvector \(\vx\) of \(\mS\),
so that
\begin{equation}
    \mS\vx=\lambda\vx,
    \qquad
    \lambda>0,
    \qquad
    \mU\vx=0,
    \qquad
    \vw^\top\vx=0.
\label{eq:nl-nullspace-critical-x}
\end{equation}

First, the simple perturbation
\begin{equation}
    \mU_\epsilon^+
    :=
    \mU+\epsilon\vv\vx^\top
\label{eq:nl-nullspace-critical-positive-path}
\end{equation}
satisfies
\begin{equation}
    \widehat L(\mU_\epsilon^+)
    -
    \widehat L(\mU)
    =
    \frac{\lambda}{2p}\epsilon^4
    >
    0
\label{eq:nl-nullspace-critical-positive}
\end{equation}
for every \(\epsilon\ne0\).

For a decreasing perturbation, define
\begin{equation}
    \alpha_\epsilon
    :=
    \frac{\epsilon^2}{2},
    \qquad
    \vz_\epsilon
    :=
    -\alpha_\epsilon\vw+\epsilon\vx,
\label{eq:nl-nullspace-critical-z}
\end{equation}
and set
\begin{equation}
    \mU_\epsilon^-
    :=
    \mU+\vv\vz_\epsilon^\top.
\label{eq:nl-nullspace-critical-negative-path}
\end{equation}
Using
\begin{equation}
    \mU\vw
    =
    \mU\mU^\top\vv
    =
    2\vv,
    \qquad
    \mU\vx=0,
\label{eq:nl-nullspace-critical-Uz-input}
\end{equation}
we obtain
\begin{equation}
    \mU\vz_\epsilon
    =
    -2\alpha_\epsilon\vv.
\label{eq:nl-nullspace-critical-Uz}
\end{equation}
Moreover,
\begin{equation}
    \|\vz_\epsilon\|^2
    =
    2\alpha_\epsilon^2+\epsilon^2.
\label{eq:nl-nullspace-critical-z-norm}
\end{equation}
Therefore
\begin{equation}
    \mU_\epsilon^-
    (\mU_\epsilon^-)^\top
    =
    \mG
    +
    \left(
        -\epsilon^2+\frac{\epsilon^4}{2}
    \right)
    \vv\vv^\top,
\label{eq:nl-nullspace-critical-G}
\end{equation}
whereas the component of \(\vz_\epsilon\) proportional to \(\vw\)
lies in \(\ker\mS\), so
\begin{equation}
    \mU_\epsilon^-
    \mS
    (\mU_\epsilon^-)^\top
    =
    \lambda\epsilon^2
    \vv\vv^\top.
\label{eq:nl-nullspace-critical-H}
\end{equation}
Substitution into
\cref{eq:nl-nullspace-loss-identity} yields
\begin{equation}
\begin{aligned}
    \widehat L(\mU_\epsilon^-)
    -
    \widehat L(\mU)
    &=
    \frac{\lambda\epsilon^2}{2p}
    \left(
        -\epsilon^2+\frac{\epsilon^4}{2}
    \right)
    \\
    &=
    -\frac{\lambda}{2p}\epsilon^4
    +
    O(\epsilon^6)
    <
    0
\end{aligned}
\label{eq:nl-nullspace-critical-negative}
\end{equation}
for every sufficiently small nonzero \(\epsilon\).

Thus arbitrarily close to \(\mU\) there are points with both larger
and smaller empirical loss, while the Hessian at \(\mU\) is positive
semidefinite.  Hence \(\mU\) is a degenerate saddle; its decreasing
variation first appears at fourth order.

The three cases also show that, among null-space critical points,
the local minima are precisely those satisfying
\begin{equation}
    \mU\mU^\top\succ2\mI_n.
\label{eq:nl-nullspace-minimum-region}
\end{equation}
\end{proof}

\paragraph{Relation to the trapped limiting dynamics.}
\cref{prop:nl-large-time-alternatives} shows that under the
strict trapping condition \cref{eq:nl-trapping-condition},
\begin{equation}
    a(t)\longrightarrow a_*>2,
    \qquad
    q(t)\longrightarrow-\infty,
\end{equation}
while \cref{eq:nl-trap-r-asympt} gives
\begin{equation}
    r(t)\longrightarrow0.
\end{equation}
Together with
\cref{thm:nl-general-nu}, which yields
\begin{equation}
    \mG_N(t)\approx a(t)\mI_n,
    \qquad
    \mH_N(t)\approx r(t)\mI_n
\end{equation}
on every fixed time interval, this identifies the strict trapped
large-time state of the limiting system with the local-minimum region
\begin{equation}
    \mU\mS=0,
    \qquad
    \mU\mU^\top\succ2\mI_n
\end{equation}
of the finite-dimensional tied empirical loss.

At the critical boundary
\cref{eq:nl-critical-condition}, instead,
\(a(t)\to2\).  The corresponding null-space configuration lies on the
degenerate boundary
\begin{equation}
    \lambda_{\min}(\mU\mU^\top)=2
\end{equation}
and, by \cref{prop:nl-nullspace-landscape}, is a degenerate saddle.

We stress that this is a landscape interpretation of the large-time
limit of the deterministic limiting system.  The finite-dimensional
convergence theorem is uniform only on fixed time intervals and does not
assert that a finite-\(N\) trajectory converges as \(t\to\infty\) to an
exact null-space critical point.

\subsubsection{Role of Gaussianity and brief extension to anisotropic data}
\label{app:narrow-anisotropic}

The preceding proof separates three assumptions that are sometimes conflated.

First, Gaussianity of the \emph{training data} is not used in the
quenched approximation
\cref{thm:nl-quenched}.  Conditional on \(\mS_N\), the training
data appear only through the deterministic matrix \(\mS_N\).  The
Gaussian concentration argument is applied to the initialization
\(\mZ\), not to the samples \(\vx_i\).

Second, isotropy of the data is needed only to reduce the population
loss to \cref{eq:nl-pop-exact-G}.  Neither the training flow nor the
scalarization of \(\mG\) and \(\mH\) requires population isotropy.

Third, isotropy alone does not imply a particular limiting empirical
spectral measure.  The passage from \(a_N,r_N\) to \(a,r\) uses the standing spectral
assumptions \cref{eq:common-spectrum-tight}--\cref{eq:common-esd-convergence}.  The limiting spectral measure
may be Marchenko--Pastur, but it need not be.

We briefly describe the anisotropic extension.  Suppose
\begin{equation}
\mSigma_N
=
\E \vx\vx^\top
\ne \mI_p
\label{eq:nl-aniso-Sigma}
\end{equation}
and assume
\begin{equation}
\sup_N
\|\mSigma_N\|_{\mathrm{op}}
<
\infty.
\label{eq:nl-aniso-Sigma-bound}
\end{equation}
The empirical training flow remains the common flow
\cref{eq:common-U-flow} with \(\gamma=1\), since it depends on the observed data only
through \(\mS_N\).  Thus the conclusions for \(\mG_N\), \(\mH_N\), \(a\),
and \(r\) are unchanged whenever the same assumptions on \(\mS_N\)
hold.

The population loss becomes
\begin{equation}
L_{\Sigma}(\mU)
=
\frac1{2p}
\Tr
\left[
(\mI_p-\mU^\top \mU)
\mSigma_N
(\mI_p-\mU^\top \mU)
\right].
\label{eq:nl-aniso-pop-loss}
\end{equation}
Introduce
\begin{equation}
\mK_N(t)
=
\mU_N(t)\mSigma_N\mU_N(t)^\top.
\label{eq:nl-aniso-K}
\end{equation}
For the proxy trajectory,
\begin{equation}
\overline \mK_N(t)
=
\frac1p
\mZ\mC_{\Sigma,N}(t)\mZ^\top,
\label{eq:nl-aniso-Kbar}
\end{equation}
where
\begin{equation}
\mC_{\Sigma,N}(t)
=
\kappa_N
e^{-2B_N(t)}
e^{q_N(t)\mS_N/2}
\mSigma_N
e^{q_N(t)\mS_N/2}.
\label{eq:nl-aniso-Csigma}
\end{equation}
Its normalized trace is
\begin{equation}
\frac1p
\Tr\mC_{\Sigma,N}(t)
=
\kappa_N
e^{-2B_N(t)}
W_N(q_N(t)),
\label{eq:nl-aniso-Csigma-trace}
\end{equation}
where
\begin{equation}
W_N(q)
=
\frac1p
\Tr
\left(
\mSigma_N e^{q\mS_N}
\right).
\label{eq:nl-aniso-WN}
\end{equation}
If, uniformly on compact \(q\)-intervals,
\begin{equation}
W_N(q)
\xrightarrow{\mathbb P}
W(q),
\label{eq:nl-aniso-W-limit}
\end{equation}
then the same Gaussian Gram concentration argument gives
\begin{equation}
\mK_N(t)
=
b(t)\mI_n
+
o_{\mathbb P}(1)
\quad
\text{in operator norm},
\label{eq:nl-aniso-K-scalar}
\end{equation}
where
\begin{equation}
b(t)
=
\kappa
e^{-2B(t)}
W(q(t)).
\label{eq:nl-aniso-b}
\end{equation}
Expanding \cref{eq:nl-aniso-pop-loss},
\begin{equation}
L_{\Sigma,N}
=
\frac1{2p}\Tr\mSigma_N
-
\frac1p\Tr\mK_N
+
\frac1{2p}\Tr(\mG_N\mK_N),
\label{eq:nl-aniso-pop-expand}
\end{equation}
so that
\begin{equation}
\psi_N
\left(
L_{\Sigma,N}(t)
-
\frac1{2p}\Tr\mSigma_N
\right)
\xrightarrow{\mathbb P}
b(t)
\left(
\frac{a(t)}2-1
\right).
\label{eq:nl-aniso-pop-limit}
\end{equation}
In the isotropic case,
\begin{equation}
\mSigma_N=\mI_p
\quad\Longrightarrow\quad
W(q)=M_{\nu_\phi}(q)
\quad\Longrightarrow\quad
b(t)=a(t),
\label{eq:nl-aniso-isotropic-recovery}
\end{equation}
recovering the previous population formula.

The limit in \cref{eq:nl-aniso-W-limit} is an additional assumption and is
not determined, in general, by the limiting empirical spectral measure of
$\mS_N$.  Indeed, if
\begin{equation}
    \mS_N
    =
    \sum_{i=1}^p
    \lambda_i\vv_i\vv_i^\top,
\end{equation}
then
\begin{equation}
    W_N(q)
    =
    \frac1p
    \sum_{i=1}^p
    e^{q\lambda_i}
    \vv_i^\top\mSigma_N\vv_i.
\end{equation}
Hence the eigenvalue distribution of $\mS_N$ determines the locations
$\lambda_i$ but not the weights
$\vv_i^\top\mSigma_N\vv_i$.  These weights record the alignment between the
eigenvectors of the empirical covariance and the population covariance.
Therefore identifying $W$ generally requires additional information on this
joint spectral geometry; knowledge of the limiting law of $\mS_N$ alone is
insufficient.\subsection{Small-data (\cref{sec:small-data})}
\label{app:small-data}

The small-data proof also starts from the common identities of \cref{app:common-preliminaries},
but the natural data normalization is the one in
\cref{eq:common-small-data-normalization}.  We therefore do not repeat the
definitions of \(\mS_N\), the error matrix, the losses, or the full gradient
flow.  The regime-specific dimensions satisfy
\begin{equation}
  p=p_N\longrightarrow\infty,
  \qquad
  n=n_N\longrightarrow\infty,
  \qquad
  m=m_N\longrightarrow\infty,
  \label{eq:sd-dimensions-diverge}
\end{equation}
and
\begin{equation}
  \phi_N:=\frac pm\longrightarrow\infty,
  \qquad
  \psi_N:=\frac pn\longrightarrow\psi\in(0,\infty),
  \qquad
  \rho_N:=n\sigma_N^2\longrightarrow\rho\in(0,\infty).
  \label{eq:sd-scaling}
\end{equation}
In particular,
\begin{equation}
  \frac mn=\frac{\psi_N}{\phi_N}\longrightarrow0.
  \label{eq:sd-m-over-n}
\end{equation}
The training matrix may be random or deterministic.  We assume the common
active-space near-isometry \cref{eq:common-data-Gram-near-isometry}.  In
particular the bounded-spectrum and spectral-limit requirements of
\cref{app:common-preliminaries} hold for the normalized operator
\(\widetilde{\mS}_N=\phi_N^{-1}\mS_N\); no Gaussianity of the training
samples is required.

We use the small-data learning rate \(\eta=m\), so
\begin{equation}
  \gamma_N=\frac{\eta}{p}=\frac1{\phi_N}.
  \label{eq:sd-gamma}
\end{equation}
Thus the full dynamics are exactly the common flow
\cref{eq:common-U-flow} with \(\gamma=\phi_N^{-1}\).  For the population
loss we assume only second-moment isotropy of the population distribution,
\begin{equation}
  \E[\vx\vx^\top]=\mI_p,
  \label{eq:sd-pop-isotropy}
\end{equation}
which, by \cref{eq:common-losses}, gives the common isotropic population
loss \cref{eq:losses_as_E}.  The train-loss identity is likewise already
contained in \cref{eq:common-losses}.

\paragraph{Main theorem.}
Let
\begin{equation}
    \mathrm{Nar}(k,j)
    :=
    \frac{1}{k}
    \binom{k}{j}
    \binom{k}{j-1},
    \qquad
    1\le j\le k,
    \label{eq:sd-Narayana}
\end{equation}
denote the Narayana numbers.

\begin{theorem}[Small-data finite-time limit]
\label{thm:sd-main}
Assume \cref{eq:sd-dimensions-diverge}, \cref{eq:sd-scaling}, and the
common active-space near-isometry condition
\cref{eq:common-data-Gram-near-isometry}, and fix
$T<\infty$.  There is a unique finite-time solution
\begin{equation}
    \left(
        a(t),\mu_1(t),\mu_2(t),\ldots
    \right),
    \qquad
    0\le t\le T,
    \label{eq:sd-limit-solution}
\end{equation}
in the class of moment sequences with uniformly bounded support, satisfying
\begin{equation}
    \dot a
    =
    4a(1-a)-2a\mu_1,
    \qquad
    a(0)=\rho,
    \label{eq:sd-a-ode}
\end{equation}
and, for every $k\ge1$,
\begin{equation}
    \dot\mu_k
    =
    -2\mu_{k+1}
    -2a\sum_{j=0}^{k-1}\mu_j\mu_{k-j}
    +2\mu_1\mu_k,
    \qquad
    \mu_0\equiv1,
    \label{eq:sd-mu-ode}
\end{equation}
with initial conditions
\begin{equation}
    \mu_k(0)
    =
    \rho^k
    \sum_{j=1}^{k}\mathrm{Nar}(k,j)\psi^j,
    \qquad
    k\ge1.
    \label{eq:sd-mu-init}
\end{equation}

More precisely, after rotating coordinates so that the training span is the
first $m$ coordinate directions and writing
\begin{equation}
    \mU_N(t)
    =
    \begin{bmatrix}
        \mA_N(t) & \mB_N(t)
    \end{bmatrix},
    \qquad
    \mM_N(t):=\mB_N(t)\mB_N(t)^\top,
    \label{eq:sd-block-def}
\end{equation}
one has, for every fixed $k\ge0$,
\begin{equation}
    \sup_{0\le t\le T}
    \left\|
        \mA_N(t)^\top \mM_N(t)^k\mA_N(t)
        -
        a(t)\mu_k(t)\mI_m
    \right\|_{\mathrm{op}}
    \xrightarrow{\mathbb P}0,
    \label{eq:sd-matrix-isotropy}
\end{equation}
where $\mu_0\equiv1$.

The train loss converges uniformly on compact time intervals according to
\begin{equation}
    \sup_{0\le t\le T}
    \left|
        \widehat L_N(t)
        -
        \left\{
            \frac12(1-a(t))^2
            +
            \frac12a(t)\mu_1(t)
        \right\}
    \right|
    \xrightarrow{\mathbb P}0.
    \label{eq:sd-train-limit}
\end{equation}

The population loss has the constant leading-order limit
\begin{equation}
    L_\infty^{(0)}
    :=
    \frac12
    \left[
        1-2\rho+(1+\psi)\rho^2
    \right],
    \label{eq:sd-pop-leading-def}
\end{equation}
in the sense that
\begin{equation}
    \sup_{0\le t\le T}
    \left|
        L_N(t)-L_\infty^{(0)}
    \right|
    \xrightarrow{\mathbb P}0.
    \label{eq:sd-pop-leading}
\end{equation}
Moreover, the first nontrivial population correction is obtained without any
further assumption on the data beyond
\cref{eq:common-data-Gram-near-isometry}:
\begin{equation}
    \sup_{0\le t\le T}
    \left|
        \phi_N\bigl(L_N(t)-L_N(0)\bigr)-C(t)
    \right|
    \xrightarrow{\mathbb P}0,
    \label{eq:sd-pop-centered}
\end{equation}
where
\begin{align}
    C(t)
    ={}&
    \frac12(1-a(t))^2
    +a(t)\mu_1(t)
    -\frac12(1-\rho)^2
    -\psi\rho^2
    \nonumber\\
    &\quad
    +2\int_0^t
        a(s)\bigl[\mu_1(s)-\mu_2(s)\bigr]
    \,ds.
    \label{eq:sd-C-def}
\end{align}
In particular, $C(0)=0$.
\end{theorem}

The centered formulation \cref{eq:sd-pop-centered} is convenient because it
does not require a rate of convergence of $\rho_N$ or $\psi_N$.  If these
ratios are fixed to greater accuracy than $1/\phi_N$, it is equivalent to the
usual absolute $1/\phi_N$ expansion.

\paragraph{Proof plan.}
The proof is divided into the following steps.

\begin{enumerate}
    \item We first replace the data by exactly orthogonalized data.  In this
    case the active and inactive weight blocks satisfy an exact reduced
    matrix ODE.

    \item For the orthogonalized system we introduce
    $\mQ_{k,N}=\mA_N^\top \mM_N^k\mA_N$.  These matrices satisfy an exact closed
    infinite hierarchy.  The submanifold $\mQ_{k,N}=c_k\mI_m$ is exactly
    invariant.

    \item We show that the random Gaussian initialization is asymptotically
    scalar in the strong sense
    $\mQ_{k,N}(0)\simeq c_k(0)\mI_m$.  The condition $m/n\to0$ is precisely what
    turns quadratic-form concentration into operator-norm concentration on
    the $m$-dimensional active block.

    \item We place the matrix hierarchy in a weighted Banach algebra and use
    a dimension-independent Gronwall estimate.  This propagates initial
    scalarity uniformly for $0\le t\le T$ and rigorously produces the scalar
    moment hierarchy \cref{eq:sd-a-ode}--\cref{eq:sd-mu-ode}.

    \item The train-loss limit and the $1/\phi_N$ population correction are
    then direct consequences of exact trace identities.

    \item Finally, for a general training matrix satisfying
    \cref{eq:common-data-Gram-near-isometry}, we use the polar decomposition of $\mX_N$.
    The normalized Gram matrix $\bm{\Gamma}_N$ appears as a perturbation of the
    orthogonalized block equations.  A finite-time ODE stability estimate
    transfers all of the preceding limits to the original data.
\end{enumerate}

\subsubsection{Exact orthogonal-data reduction}
\label{sec:sd-orthogonal-reduction}

We first consider the reference model
\begin{equation}
    \mX_N=\sqrt p\,\mQ_N,
    \qquad
    \mQ_N^\top \mQ_N=\mI_m.
    \label{eq:sd-orth-data}
\end{equation}
After a fixed orthogonal change of coordinates we may assume that
\begin{equation}
    \mQ_N
    =
    \begin{pmatrix}
        \mI_m\\
        0
    \end{pmatrix}.
    \label{eq:sd-Q-coordinate}
\end{equation}
The right-orthogonal invariance of the standing initialization in
\cref{eq:common-Gaussian-init} implies that this rotation does not change its law.

For \cref{eq:sd-orth-data}, the empirical covariance is
\begin{equation}
    \mS_N
    =
    \phi_N\mP_m,
    \qquad
    \mP_m
    =
    \begin{pmatrix}
        \mI_m&0\\
        0&0
    \end{pmatrix}.
    \label{eq:sd-S-orth}
\end{equation}
Thus the common flow \cref{eq:common-U-flow}, with
\(\gamma=\phi_N^{-1}\), reduces exactly to
\begin{equation}
    \dot \mU_N
    =
    \mU_N(\mP_m\mE_N+\mE_N\mP_m).
    \label{eq:sd-U-orth-flow}
\end{equation}
With the decomposition \cref{eq:sd-block-def},
\begin{equation}
    \mE_N
    =
    \begin{pmatrix}
        \mI_m-\mA_N^\top \mA_N & -\mA_N^\top \mB_N\\
        -\mB_N^\top \mA_N & \mI_{p-m}-\mB_N^\top \mB_N
    \end{pmatrix}.
    \label{eq:sd-E-block}
\end{equation}
Multiplying the blocks in \cref{eq:sd-U-orth-flow} gives
\begin{equation}
    \dot \mA_N
    =
    2\mA_N(\mI_m-\mA_N^\top \mA_N)-\mM_N\mA_N,
    \label{eq:sd-A-orth}
\end{equation}
and
\begin{equation}
    \dot \mB_N
    =
    -\mA_N\mA_N^\top \mB_N.
    \label{eq:sd-B-orth}
\end{equation}
Consequently,
\begin{equation}
    \dot \mM_N
    =
    -\mA_N\mA_N^\top \mM_N
    -
    \mM_N\mA_N\mA_N^\top.
    \label{eq:sd-M-orth}
\end{equation}
\Crefrange{eq:sd-A-orth}{eq:sd-M-orth} are exact.

\subsubsection{The exact matrix moment hierarchy}
\label{sec:sd-matrix-hierarchy}

For every integer $k\ge0$, define
\begin{equation}
    \mQ_{k,N}(t)
    :=
    \mA_N(t)^\top \mM_N(t)^k\mA_N(t).
    \label{eq:sd-Qk-def}
\end{equation}
In particular,
\begin{equation}
    \mQ_{0,N}=\mA_N^\top \mA_N.
    \label{eq:sd-Q0-def}
\end{equation}
We write the resulting matrix sequence as
\begin{equation}
    \tQ_N(t):=\bigl(\mQ_{0,N}(t),\mQ_{1,N}(t),\ldots\bigr).
    \label{eq:sd-Q-sequence-notation}
\end{equation}

\begin{lemma}[Exact matrix hierarchy]
\label{lem:sd-matrix-hierarchy}
For the orthogonal-data dynamics
\cref{eq:sd-A-orth}--\cref{eq:sd-M-orth},
\begin{align}
    \dot \mQ_{k,N}
    ={}&
    4\mQ_{k,N}
    -2\mQ_{0,N}\mQ_{k,N}
    -2\mQ_{k,N}\mQ_{0,N}
    -2\mQ_{k+1,N}
    \nonumber\\
    &\quad
    -\sum_{j=0}^{k-1}
    \left(
        \mQ_{j,N}\mQ_{k-j,N}
        +
        \mQ_{j+1,N}\mQ_{k-1-j,N}
    \right),
    \qquad k\ge0,
    \label{eq:sd-exact-Q-hierarchy}
\end{align}
where the sum is empty when $k=0$.
\end{lemma}

\begin{proof}
From \cref{eq:sd-A-orth},
\begin{equation}
    \dot \mA_N
    =
    2\mA_N-2\mA_N\mQ_{0,N}-\mM_N\mA_N.
    \label{eq:sd-A-rewritten}
\end{equation}
Hence
\begin{equation}
    \dot \mA_N^\top
    =
    2\mA_N^\top
    -2\mQ_{0,N}\mA_N^\top
    -\mA_N^\top \mM_N.
    \label{eq:sd-AT-rewritten}
\end{equation}
Differentiating \cref{eq:sd-Qk-def}, the two terms in which one
differentiates $\mA_N$ give
\begin{align}
    &\dot \mA_N^\top \mM_N^k\mA_N
    +
    \mA_N^\top \mM_N^k\dot \mA_N
    \nonumber\\
    &\qquad
    =
    4\mQ_{k,N}
    -2\mQ_{0,N}\mQ_{k,N}
    -2\mQ_{k,N}\mQ_{0,N}
    -2\mQ_{k+1,N}.
    \label{eq:sd-Q-A-contribution}
\end{align}
For the remaining contribution,
\begin{equation}
    \frac{d}{dt}\mM_N^k
    =
    \sum_{j=0}^{k-1}
        \mM_N^j\dot{\mM}_N\mM_N^{k-1-j}.
    \label{eq:sd-M-power-derivative}
\end{equation}
Substitution of \cref{eq:sd-M-orth} gives
\begin{align}
    &\mA_N^\top
    \mM_N^j\dot{\mM}_N\mM_N^{k-1-j}
    \mA_N
    \nonumber\\
    &\qquad
    =
    -\mQ_{j,N}\mQ_{k-j,N}
    -
    \mQ_{j+1,N}\mQ_{k-1-j,N}.
    \label{eq:sd-Q-M-contribution}
\end{align}
Summing \cref{eq:sd-Q-M-contribution} over $j$ and adding
\cref{eq:sd-Q-A-contribution} proves
\cref{eq:sd-exact-Q-hierarchy}.
\end{proof}

For $k=0$, Lemma \ref{lem:sd-matrix-hierarchy} reads
\begin{equation}
    \dot \mQ_{0,N}
    =
    4\mQ_{0,N}
    -4\mQ_{0,N}^2
    -2\mQ_{1,N}.
    \label{eq:sd-Q0-ODE}
\end{equation}
Since $\mM_N\succeq0$, we have
\begin{equation}
    \mQ_{1,N}\succeq0.
    \label{eq:sd-Q1-positive}
\end{equation}

The hierarchy \cref{eq:sd-exact-Q-hierarchy} is already closed at the
matrix level.  Thus no moment-closure approximation is needed.  The only
remaining issue is to prove that its random solution becomes scalar in the
active $m$-dimensional coordinates.

\subsubsection{Uniform a priori bounds}
\label{sec:sd-apriori}

The first bound is a contraction property of the inactive block.

\begin{lemma}[Inactive contraction]
\label{lem:sd-M-contraction}
For every $t\ge0$ for which the solution exists,
\begin{equation}
    \|\mB_N(t)\|_{\mathrm{op}}
    \le
    \|\mB_N(0)\|_{\mathrm{op}},
    \label{eq:sd-B-contraction}
\end{equation}
and therefore
\begin{equation}
    \|\mM_N(t)\|_{\mathrm{op}}
    \le
    \|\mM_N(0)\|_{\mathrm{op}}.
    \label{eq:sd-M-contraction}
\end{equation}
\end{lemma}

\begin{proof}
For every deterministic $\vv\in\mathbb R^{p-m}$,
\begin{align}
    \frac{d}{dt}\|\mB_N(t)\vv\|^2
    &=
    2\vv^\top \mB_N^\top\dot \mB_N\vv
    \nonumber\\
    &=
    -2\vv^\top
        \mB_N^\top \mA_N\mA_N^\top \mB_N\vv
    \nonumber\\
    &=
    -2\|\mA_N^\top \mB_N\vv\|^2
    \le0.
    \label{eq:sd-B-vector-contraction}
\end{align}
Taking the supremum over unit $\vv$ gives
\cref{eq:sd-B-contraction}; squaring yields
\cref{eq:sd-M-contraction}.
\end{proof}

The active block is also uniformly bounded.

\begin{lemma}[Active operator-norm bound]
\label{lem:sd-A-bound}
For the orthogonal-data flow,
\begin{equation}
    \|\mA_N(t)\|_{\mathrm{op}}^2
    =
    \lambda_{\max}(\mQ_{0,N}(t))
    \le
    \max\left\{
        1,\|\mA_N(0)\|_{\mathrm{op}}^2
    \right\}.
    \label{eq:sd-A-bound}
\end{equation}
\end{lemma}

\begin{proof}
Let $\lambda_N(t)=\lambda_{\max}(\mQ_{0,N}(t))$.  At any time at which
$\lambda_N$ is differentiable, choose a normalized top eigenvector $\vv$.
Using \cref{eq:sd-Q0-ODE} and \cref{eq:sd-Q1-positive},
\begin{align}
    \dot\lambda_N(t)
    &=
    \vv^\top\dot \mQ_{0,N}\vv
    \nonumber\\
    &=
    4\lambda_N
    -4\lambda_N^2
    -2\vv^\top \mQ_{1,N}\vv
    \nonumber\\
    &\le
    4\lambda_N(1-\lambda_N).
    \label{eq:sd-lambda-logistic}
\end{align}
The same inequality holds for the upper Dini derivative.  Comparison with
the scalar logistic equation gives \cref{eq:sd-A-bound}.
\end{proof}

Combining Lemmas \ref{lem:sd-M-contraction} and \ref{lem:sd-A-bound},
for every $k\ge0$,
\begin{equation}
    \|\mQ_{k,N}(t)\|_{\mathrm{op}}
    \le
    \|\mA_N(t)\|_{\mathrm{op}}^2
    \|\mM_N(t)\|_{\mathrm{op}}^k.
    \label{eq:sd-Qk-geometric-bound}
\end{equation}

\subsubsection{A Banach algebra for the hierarchy}
\label{sec:sd-banach}

Fix $L>0$ and choose $r>0$ such that
\begin{equation}
    rL<1.
    \label{eq:sd-rL}
\end{equation}
For a sequence
$\tR=(\mR_0,\mR_1,\ldots)$ of $m\times m$ matrices, set
\begin{equation}
    \|\tR\|_r
    :=
    \sum_{k=0}^{\infty}
        r^k\|\mR_k\|_{\mathrm{op}}.
    \label{eq:sd-Banach-norm}
\end{equation}
Denote the resulting Banach space by $\mathcal B_r^{(m)}$.

For two sequences define the convolution
\begin{equation}
    (\tR*\tS)_k
    :=
    \sum_{j=0}^{k}\mR_j\mS_{k-j}.
    \label{eq:sd-convolution}
\end{equation}
The norm \cref{eq:sd-Banach-norm} is submultiplicative:
\begin{equation}
    \|\tR*\tS\|_r
    \le
    \|\tR\|_r\|\tS\|_r.
    \label{eq:sd-convolution-bound}
\end{equation}
The left-shift operator
\begin{equation}
    (D\tR)_k:=\mR_{k+1}
    \label{eq:sd-shift-def}
\end{equation}
obeys
\begin{equation}
    \|D\tR\|_r
    \le
    \frac1r\|\tR\|_r.
    \label{eq:sd-shift-bound}
\end{equation}

Let $\mathcal F_m$ denote the vector field given componentwise by the
right-hand side of \cref{eq:sd-exact-Q-hierarchy}.  The preceding estimates
imply that, on every ball
\begin{equation}
    \|\tR\|_r\le R,
    \qquad
    \|\tS\|_r\le R,
    \label{eq:sd-Banach-ball}
\end{equation}
there is a constant $K_{r,R}<\infty$, independent of $m$, such that
\begin{equation}
    \|\mathcal F_m(\tR)-\mathcal F_m(\tS)\|_r
    \le
    K_{r,R}\|\tR-\tS\|_r.
    \label{eq:sd-F-Lipschitz}
\end{equation}
For example, one may take
\begin{equation}
    K_{r,R}
    =
    4+\frac2r+12R.
    \label{eq:sd-Lipschitz-constant}
\end{equation}
The precise numerical constant is immaterial; what matters is that it does
not depend on $m$.

Indeed, the linear terms are controlled by
\cref{eq:sd-shift-bound}, while every quadratic term is controlled by
\cref{eq:sd-convolution-bound} and the elementary inequality
\begin{equation}
    \|\mR\mS-\widetilde{\mR}\widetilde{\mS}\|
    \le
    \|\mR-\widetilde{\mR}\|\|\mS\|
    +
    \|\widetilde{\mR}\|\|\mS-\widetilde{\mS}\|.
    \label{eq:sd-product-difference}
\end{equation}

The scalar subspace is exactly invariant.  More precisely, if
$\mathbf c=(c_0,c_1,\ldots)$ is a scalar sequence and
\begin{equation}
    \iota_m(\mathbf c)
    :=
    (c_0\mI_m,c_1\mI_m,\ldots),
    \label{eq:sd-scalar-embedding}
\end{equation}
then there is a scalar vector field $\mathcal F_{\mathrm{sc}}$ such that
\begin{equation}
    \mathcal F_m(\iota_m(\mathbf c))
    =
    \iota_m\bigl(\mathcal F_{\mathrm{sc}}(\mathbf c)\bigr).
    \label{eq:sd-invariant-manifold}
\end{equation}
Explicitly,
\begin{equation}
    \dot c_0
    =
    4c_0-4c_0^2-2c_1,
    \label{eq:sd-c0-ode}
\end{equation}
and, for $k\ge1$,
\begin{equation}
    \dot c_k
    =
    4c_k
    -4c_0c_k
    -2c_{k+1}
    -2\sum_{j=0}^{k-1}c_jc_{k-j}.
    \label{eq:sd-ck-ode}
\end{equation}

Thus the desired scalar closure is not an approximation at the level of the
limiting vector field: it is an exact invariant manifold of the full matrix
hierarchy.

\subsubsection{Scalarization of the random initial condition}
\label{sec:sd-initial-scalarization}

Under the rotated Gaussian initialization,
\begin{equation}
    \mA_N(0)
    =
    \sqrt{\frac{\rho_N}{n}}\,\mZ_{A,N},
    \qquad
    \mB_N(0)
    =
    \sqrt{\frac{\rho_N}{n}}\,\mZ_{B,N},
    \label{eq:sd-A0-B0}
\end{equation}
where $\mZ_{A,N}\in\mathbb R^{n\times m}$ and
$\mZ_{B,N}\in\mathbb R^{n\times(p-m)}$ have independent standard Gaussian
entries and are independent.

Consequently,
\begin{equation}
    \mM_N(0)
    =
    \frac{\rho_N}{n}
    \mZ_{B,N}\mZ_{B,N}^\top.
    \label{eq:sd-M0-Wishart}
\end{equation}
For $k\ge0$, set
\begin{equation}
    s_{k,N}
    :=
    \frac{\rho_N}{n}
    \Tr\bigl(\mM_N(0)^k\bigr).
    \label{eq:sd-skN}
\end{equation}

\begin{lemma}[Conditional scalarization]
\label{lem:sd-initial-scalarization}
For each fixed $k\ge0$,
\begin{equation}
    \left\|
        \mQ_{k,N}(0)-s_{k,N}\mI_m
    \right\|_{\mathrm{op}}
    \xrightarrow{\mathbb P}0.
    \label{eq:sd-fixed-k-scalarization}
\end{equation}
\end{lemma}

\begin{proof}
Condition on $\mM_N(0)$ and abbreviate it by $\mM$.  From
\cref{eq:sd-A0-B0},
\begin{equation}
    \mQ_{k,N}(0)
    =
    \frac{\rho_N}{n}
    \mZ_{A,N}^\top \mM^k\mZ_{A,N}.
    \label{eq:sd-Qk0-conditioned}
\end{equation}
Let
\begin{equation}
    \mH_{k,N}
    :=
    \frac1n
    \mZ_{A,N}^\top \mM^k\mZ_{A,N}
    -
    \frac1n\Tr(\mM^k)\mI_m.
    \label{eq:sd-HkN}
\end{equation}
For a deterministic unit vector $\vv\in\mathbb R^m$,
$\mZ_{A,N}\vv$ is a standard Gaussian vector $\vg\in\mathbb R^n$, so
\begin{equation}
    \vv^\top \mH_{k,N}\vv
    =
    \frac1n
    \left(
        \vg^\top \mM^k\vg-\Tr(\mM^k)
    \right).
    \label{eq:sd-H-quadratic}
\end{equation}

Fix $L<\infty$ and work on the event
\begin{equation}
    \|\mM\|_{\mathrm{op}}\le L.
    \label{eq:sd-M0-L-event}
\end{equation}
The standard Gaussian quadratic-form estimate implies that, for every
$\varepsilon>0$,
\begin{equation}
    \mathbb P
    \left(
        \left.
        |\vv^\top \mH_{k,N}\vv|>\varepsilon
        \,\right|\,\mM
    \right)
    \le
    2\exp\!\left(
        -c_kn\min\{\varepsilon^2,\varepsilon\}
    \right),
    \label{eq:sd-gaussian-quadratic-tail}
\end{equation}
where $c_k>0$ depends on $k$ and $L$, but not on $n$ or $m$.

Let $\mathcal N_m$ be a $1/4$-net of the Euclidean unit sphere in
$\mathbb R^m$ with
\begin{equation}
    |\mathcal N_m|\le 9^m.
    \label{eq:sd-net-cardinality}
\end{equation}
For every symmetric matrix $\mH$,
\begin{equation}
    \|\mH\|_{\mathrm{op}}
    \le
    2\max_{\vv\in\mathcal N_m}|\vv^\top \mH\vv|.
    \label{eq:sd-net-opnorm}
\end{equation}
A union bound using \cref{eq:sd-gaussian-quadratic-tail} therefore gives
\begin{equation}
    \mathbb P
    \left(
        \left.
        \|\mH_{k,N}\|_{\mathrm{op}}>2\varepsilon
        \,\right|\,\mM
    \right)
    \le
    2\exp
    \left(
        m\log9
        -
        c_kn\min\{\varepsilon^2,\varepsilon\}
    \right).
    \label{eq:sd-net-union}
\end{equation}
By \cref{eq:sd-m-over-n}, the right-hand side tends to zero.

It remains to remove the conditioning event
\cref{eq:sd-M0-L-event}.  Since
\cref{eq:sd-M0-Wishart} is Gaussian Wishart with
\begin{equation}
    \frac{p-m}{n}
    =
    \psi_N
    \left(1-\frac1{\phi_N}\right)
    \longrightarrow\psi,
    \label{eq:sd-psi-perp-limit}
\end{equation}
the largest eigenvalue of $\mM_N(0)$ is tight; in fact,
\begin{equation}
    \|\mM_N(0)\|_{\mathrm{op}}
    \xrightarrow{\mathbb P}
    \rho(1+\sqrt\psi)^2.
    \label{eq:sd-M0-edge}
\end{equation}
Choosing $L>\rho(1+\sqrt\psi)^2$ completes the proof after multiplying
$\mH_{k,N}$ by $\rho_N$.
\end{proof}

The scalar coefficients in Lemma
\ref{lem:sd-initial-scalarization} have deterministic limits.  By
Marchenko--Pastur moment convergence applied to
\cref{eq:sd-M0-Wishart},
\begin{equation}
    \frac1n
    \Tr\bigl(\mM_N(0)^k\bigr)
    \xrightarrow{\mathbb P}
    \rho^k
    \sum_{j=1}^{k}
        \mathrm{Nar}(k,j)\psi^j,
    \qquad
    k\ge1.
    \label{eq:sd-MP-moments-M}
\end{equation}
Hence
\begin{equation}
    s_{0,N}\longrightarrow\rho,
    \label{eq:sd-s0-limit}
\end{equation}
and
\begin{equation}
    s_{k,N}
    \xrightarrow{\mathbb P}
    c_k(0)
    :=
    \rho^{k+1}
    \sum_{j=1}^{k}
        \mathrm{Nar}(k,j)\psi^j,
    \qquad
    k\ge1.
    \label{eq:sd-ck0-limit}
\end{equation}

The convergence can be strengthened from each fixed $k$ to the weighted
Banach norm.

\begin{lemma}[Banach-norm scalarization at initialization]
\label{lem:sd-Banach-initial}
There exists $r>0$ such that
\begin{equation}
    \left\|
        \tQ_N(0)
        -
        \iota_m(\mathbf c(0))
    \right\|_r
    \xrightarrow{\mathbb P}0,
    \label{eq:sd-Banach-initial-convergence}
\end{equation}
where
\begin{equation}
    \mathbf c(0)
    :=
    \bigl(
        c_0(0),c_1(0),\ldots
    \bigr).
    \label{eq:sd-c-sequence}
\end{equation}
\end{lemma}

\begin{proof}
Choose $L>\rho(1+\sqrt\psi)^2$.  By standard Gaussian singular-value
estimates, for some deterministic $C<\infty$,
\begin{equation}
    \mathbb P
    \left(
        \|\mM_N(0)\|_{\mathrm{op}}\le L,
        \quad
        \|\mA_N(0)\|_{\mathrm{op}}^2\le C
    \right)
    \longrightarrow1.
    \label{eq:sd-init-good-event}
\end{equation}
On this event,
\begin{equation}
    \|\mQ_{k,N}(0)\|_{\mathrm{op}}
    \le
    CL^k.
    \label{eq:sd-Qk0-tail-bound}
\end{equation}
The deterministic moments in \cref{eq:sd-ck0-limit} satisfy, after
increasing $L$ if necessary,
\begin{equation}
    |c_k(0)|
    \le
    C'L^k.
    \label{eq:sd-ck0-tail-bound}
\end{equation}
Choose $r>0$ with $rL<1$.  For every $K$,
\begin{align}
    &\sum_{k>K}
        r^k
        \left\|
            \mQ_{k,N}(0)-c_k(0)\mI_m
        \right\|_{\mathrm{op}}
    \nonumber\\
    &\qquad
    \le
    (C+C')
    \sum_{k>K}(rL)^k.
    \label{eq:sd-tail-geometric}
\end{align}
The right-hand side tends to zero as $K\to\infty$, uniformly on the
high-probability event \cref{eq:sd-init-good-event}.  For each fixed $K$,
the contribution of $0\le k\le K$ tends to zero by
Lemma \ref{lem:sd-initial-scalarization} and
\cref{eq:sd-ck0-limit}.  First let $N\to\infty$, then $K\to\infty$.
\end{proof}

\subsubsection{Propagation of scalarity}
\label{sec:sd-propagation}

Let $\mathbf c(t)$ be the maximal solution in the scalar weighted space to
\cref{eq:sd-c0-ode}--\cref{eq:sd-ck-ode} with initial condition
\cref{eq:sd-s0-limit}--\cref{eq:sd-ck0-limit}.  Local existence and
uniqueness follow from the Banach-space Picard--Lindelof theorem and the
local Lipschitz estimate \cref{eq:sd-F-Lipschitz}.

We next prove both global existence on every finite interval and convergence
of the matrix hierarchy to this solution.

\begin{prop}[Finite-time propagation of isotropy]
\label{prop:sd-propagation}
For the $r>0$ chosen in Lemma \ref{lem:sd-Banach-initial}, and every
$T<\infty$,
\begin{equation}
    \sup_{0\le t\le T}
    \left\|
        \tQ_N(t)
        -
        \iota_m(\mathbf c(t))
    \right\|_r
    \xrightarrow{\mathbb P}0.
    \label{eq:sd-Banach-propagation}
\end{equation}
In particular, the scalar Banach-space solution exists uniquely on every
finite time interval.
\end{prop}

\begin{proof}
On the event \cref{eq:sd-init-good-event}, Lemmas
\ref{lem:sd-M-contraction} and \ref{lem:sd-A-bound} imply
\begin{equation}
    \sup_{t\ge0}
    \|\tQ_N(t)\|_r
    \le
    C_A
    \sum_{k=0}^{\infty}(rL)^k
    =
    \frac{C_A}{1-rL},
    \label{eq:sd-Q-uniform-Banach-bound}
\end{equation}
where
\begin{equation}
    C_A:=\max\{1,C\}.
    \label{eq:sd-CA}
\end{equation}
Let $R$ be larger than twice the right-hand side of
\cref{eq:sd-Q-uniform-Banach-bound}.  Up to the first exit time of either
$\tQ_N$ or $\iota_m(\mathbf c)$ from the radius-$R$ ball,
\cref{eq:sd-F-Lipschitz} and the integral form of the two ODEs give
\begin{align}
    &
    \left\|
        \tQ_N(t)
        -
        \iota_m(\mathbf c(t))
    \right\|_r
    \nonumber\\
    &\qquad
    \le
    \left\|
        \tQ_N(0)
        -
        \iota_m(\mathbf c(0))
    \right\|_r
    +
    K_{r,R}
    \int_0^t
    \left\|
        \tQ_N(s)
        -
        \iota_m(\mathbf c(s))
    \right\|_r
    ds.
    \label{eq:sd-Gronwall-pre}
\end{align}
Gronwall's inequality yields
\begin{equation}
    \sup_{0\le s\le t}
    \left\|
        \tQ_N(s)
        -
        \iota_m(\mathbf c(s))
    \right\|_r
    \le
    e^{K_{r,R}t}
    \left\|
        \tQ_N(0)
        -
        \iota_m(\mathbf c(0))
    \right\|_r.
    \label{eq:sd-Gronwall}
\end{equation}
By Lemma \ref{lem:sd-Banach-initial}, the right-hand side tends to zero in
probability for every fixed $t$.

We now remove the stopping time carefully.  Write
\begin{equation}
  B_r:=\frac{C_A}{1-rL},
  \qquad
  d_R:=R-B_r>0.
  \label{eq:sd-continuation-margin}
\end{equation}
Fix a deterministic \(T<\infty\), and set
\begin{equation}
  \varepsilon_T:=\frac{d_R}{2}e^{-K_{r,R}T}>0.
  \label{eq:sd-continuation-epsilon}
\end{equation}
Let \(\mathcal E_N(T)\) be the event on which
\cref{eq:sd-init-good-event} holds and
\begin{equation}
  \|\tQ_N(0)-\iota_m(\mathbf c(0))\|_r
  <\varepsilon_T.
  \label{eq:sd-continuation-event-error}
\end{equation}
By Lemma \ref{lem:sd-Banach-initial},
\begin{equation}
  \mathbb P(\mathcal E_N(T))\longrightarrow1.
  \label{eq:sd-continuation-event-probability}
\end{equation}
In particular, \(\mathcal E_N(T)\) is nonempty for all sufficiently large
\(N\).  On this event the matrix hierarchy obeys
\begin{equation}
  \sup_{t\ge0}\|\tQ_N(t)\|_r\le B_r,
  \label{eq:sd-continuation-Q-bound}
\end{equation}
and, up to the first time at which the deterministic scalar trajectory
\(\iota_m(\mathbf c(t))\) reaches the radius-\(R\) boundary,
\cref{eq:sd-Gronwall} gives
\begin{equation}
  \sup_{0\le s\le t}
  \|\tQ_N(s)-\iota_m(\mathbf c(s))\|_r
  <\frac{d_R}{2},
  \qquad
  0\le t\le T.
  \label{eq:sd-continuation-error-bound}
\end{equation}
If the scalar trajectory reached that boundary at some first time
\(\tau\le T\), then on any realization in \(\mathcal E_N(T)\),
\begin{equation}
\begin{aligned}
  R
  =\|\iota_m(\mathbf c(\tau))\|_r
  &\le
  \|\tQ_N(\tau)\|_r
  +\|\tQ_N(\tau)-\iota_m(\mathbf c(\tau))\|_r\\
  &<B_r+\frac{d_R}{2}<R,
\end{aligned}
  \label{eq:sd-continuation-contradiction}
\end{equation}
a contradiction.  Since the scalar trajectory is deterministic, the
existence of such a realization is enough to rule out the boundary hitting
time altogether.  Thus the scalar solution remains in the radius-\(R\) ball
throughout \([0,T]\); by local Banach-space existence it cannot blow up
before \(T\).  The stopped estimate is therefore valid on all of \([0,T]\),
and Lemma \ref{lem:sd-Banach-initial} now gives
\cref{eq:sd-Banach-propagation}.  Since \(T\) is arbitrary, the scalar
solution exists uniquely on every finite interval.
\end{proof}

Taking any fixed coordinate $k$ in
\cref{eq:sd-Banach-propagation} gives
\begin{equation}
    \sup_{0\le t\le T}
    \left\|
        \mQ_{k,N}(t)-c_k(t)\mI_m
    \right\|_{\mathrm{op}}
    \xrightarrow{\mathbb P}0.
    \label{eq:sd-fixed-k-propagation}
\end{equation}

\subsubsection{Identification of the scalar hierarchy}
\label{sec:sd-identification}

Set
\begin{equation}
    a(t):=c_0(t).
    \label{eq:sd-a-def}
\end{equation}
We first verify that $a(t)$ is strictly positive at every finite time, so
that the normalized moments are well defined.

At finite $N$,
\begin{equation}
    0\preceq
    \mQ_{1,N}(t)
    \preceq
    \|\mM_N(t)\|_{\mathrm{op}}\mQ_{0,N}(t).
    \label{eq:sd-Q1-Q0-order}
\end{equation}
Passing to the scalar limit and using
\cref{eq:sd-M-contraction} gives, for some finite $L$,
\begin{equation}
    0\le c_1(t)\le La(t).
    \label{eq:sd-c1-a-bound}
\end{equation}
From \cref{eq:sd-c0-ode},
\begin{equation}
    \dot a
    =
    4a(1-a)-2c_1
    \ge
    a(4-4a-2L).
    \label{eq:sd-a-positive-comparison}
\end{equation}
Since $a(0)=\rho>0$, scalar comparison implies
\begin{equation}
    a(t)>0
    \qquad
    \text{for every finite }t.
    \label{eq:sd-a-positive}
\end{equation}

We may therefore define
\begin{equation}
    \mu_k(t):=\frac{c_k(t)}{a(t)},
    \qquad
    \mu_0(t)\equiv1.
    \label{eq:sd-mu-def}
\end{equation}
\Cref{eq:sd-c0-ode} immediately becomes
\cref{eq:sd-a-ode}.  For $k\ge1$, substitute
$c_k=a\mu_k$ into \cref{eq:sd-ck-ode}:
\begin{align}
    \dot a\,\mu_k+a\dot\mu_k
    ={}&
    4a\mu_k
    -4a^2\mu_k
    -2a\mu_{k+1}
    \nonumber\\
    &\quad
    -2a^2
    \sum_{j=0}^{k-1}
        \mu_j\mu_{k-j}.
    \label{eq:sd-mu-intermediate}
\end{align}
Subtracting $\dot a\,\mu_k$ and using \cref{eq:sd-a-ode} yields precisely
\cref{eq:sd-mu-ode}.  The initial conditions
\cref{eq:sd-mu-init} follow from
\cref{eq:sd-s0-limit} and \cref{eq:sd-ck0-limit}.

This proves the hierarchy claimed in \cref{thm:sd-main}.
Moreover, if a solution is a moment sequence of probability measures whose
supports are uniformly contained in \([0,L_*]\) on \([0,T]\), then
\begin{equation}
  |\mu_k(t)|\le L_*^k
  \qquad
  \text{for every }k\ge0\text{ and }0\le t\le T.
  \label{eq:sd-bounded-support-moment-bound}
\end{equation}
For every \(r<L_*^{-1}\), the corresponding sequence belongs to the weighted
Banach space used above.  Thus the Banach-space uniqueness in
\cref{prop:sd-propagation} implies uniqueness in the
bounded-support class stated in \cref{thm:sd-main}.

\subsubsection{The active spectral measure and uniqueness}
\label{sec:sd-active-measure}

The moment interpretation can also be made rigorous.  Diagonalize
\begin{equation}
    \mM_N(t)
    =
    \sum_{\alpha=1}^{n}
        \lambda_{\alpha,N}(t)
        \vv_{\alpha,N}(t)\vv_{\alpha,N}(t)^\top.
    \label{eq:sd-M-spectral}
\end{equation}
Let
\begin{equation}
    a_N(t)
    :=
    \frac1m\Tr\mQ_{0,N}(t),
    \label{eq:sd-aN}
\end{equation}
and, whenever $a_N(t)>0$, define
\begin{equation}
    \omega^{\mathrm{act}}_{N,t}
    :=
    \frac{1}{ma_N(t)}
    \sum_{\alpha=1}^{n}
        \|\mA_N(t)^\top \vv_{\alpha,N}(t)\|^2
        \delta_{\lambda_{\alpha,N}(t)}.
    \label{eq:sd-nuN}
\end{equation}
This is a probability measure because
\begin{equation}
    \sum_{\alpha=1}^{n}
        \|\mA_N^\top \vv_{\alpha,N}\|^2
    =
    \Tr(\mA_N^\top \mA_N)
    =
    ma_N.
    \label{eq:sd-nuN-normalization}
\end{equation}
Its moments are exactly
\begin{equation}
    \int\lambda^k\,d\omega^{\mathrm{act}}_{N,t}(\lambda)
    =
    \frac{
        \Tr\mQ_{k,N}(t)
    }{
        ma_N(t)
    }.
    \label{eq:sd-nuN-moments}
\end{equation}

By Lemma \ref{lem:sd-M-contraction}, these measures have uniformly bounded
support with probability tending to one.  By
\cref{eq:sd-fixed-k-propagation},
\begin{equation}
    \frac1m\Tr\mQ_{k,N}(t)
    \xrightarrow{\mathbb P}
    c_k(t),
    \label{eq:sd-trace-Q-limit}
\end{equation}
uniformly on compact time intervals.  More explicitly, from any subsequence
of \(N\) one may extract a further subsequence along which the support bound
and the moment convergences for all integer \(k\ge0\) hold almost surely,
using a diagonal argument.  Compactness of probability measures on the
resulting fixed compact interval then gives a further weakly convergent
subsequence.  Bounded support allows passage of every polynomial moment to
the weak limit, so every such limit \(\omega^{\mathrm{act}}_t\) satisfies
\begin{equation}
    \mu_k(t)
    =
    \int\lambda^k\,d\omega^{\mathrm{act}}_t(\lambda),
    \qquad
    k\ge0.
    \label{eq:sd-active-measure-moments}
\end{equation}
The compactly supported moment problem is determinate, hence all
subsequential limits coincide.  This identifies a unique active spectral
measure \(\omega^{\mathrm{act}}_t\).

Equivalently, for $|z|$ sufficiently small one may introduce
\begin{equation}
    H(t,z)
    :=
    \sum_{k=0}^{\infty}\mu_k(t)z^k
    =
    \int
    \frac{1}{1-z\lambda}
    \,d\omega^{\mathrm{act}}_t(\lambda).
    \label{eq:sd-H-def}
\end{equation}
Summing \cref{eq:sd-mu-ode} gives
\begin{equation}
    \partial_tH
    =
    -2\frac{H-1}{z}
    -2aH(H-1)
    +2\mu_1H,
    \qquad
    \mu_1=\partial_zH(t,0),
    \label{eq:sd-H-PDE}
\end{equation}
where the apparent singularity at $z=0$ is removable.  The local
analytic uniqueness of \cref{eq:sd-H-PDE}, equivalently the
Banach-space uniqueness proved above, supplies a rigorous meaning to the
infinite moment hierarchy.

\subsubsection{Train loss for orthogonal data}
\label{sec:sd-train-orthogonal}

For \cref{eq:sd-orth-data}, the empirical loss satisfies the exact identity
\begin{equation}
    \widehat L_N(t)
    =
    \frac{1}{2m}
    \|\mI_m-\mQ_{0,N}(t)\|_F^2
    +
    \frac{1}{2m}
    \Tr\mQ_{1,N}(t).
    \label{eq:sd-train-exact-orth}
\end{equation}
Indeed, the two terms are respectively the active reconstruction error and
the leakage into the inactive block.

By \cref{eq:sd-fixed-k-propagation},
\begin{equation}
    \sup_{0\le t\le T}
    \left|
        \frac1m
        \Tr(\mI_m-\mQ_{0,N}(t))^2
        -
        (1-a(t))^2
    \right|
    \xrightarrow{\mathbb P}0,
    \label{eq:sd-Q0-loss-convergence}
\end{equation}
and
\begin{equation}
    \sup_{0\le t\le T}
    \left|
        \frac1m\Tr\mQ_{1,N}(t)
        -
        a(t)\mu_1(t)
    \right|
    \xrightarrow{\mathbb P}0.
    \label{eq:sd-Q1-loss-convergence}
\end{equation}
Substitution in \cref{eq:sd-train-exact-orth} proves
\cref{eq:sd-train-limit} for orthogonal data.

\subsubsection{Population loss and its first correction for orthogonal data}
\label{sec:sd-pop-orthogonal}

The population loss admits the exact block decomposition
\begin{align}
    L_N(t)
    ={}&
    \frac{1}{2p}
        \|\mI_m-\mQ_{0,N}(t)\|_F^2
    +
    \frac1p
        \Tr\mQ_{1,N}(t)
    \nonumber\\
    &\quad
    +
    \frac{1}{2p}
        \|\mI_{p-m}-\mB_N(t)^\top \mB_N(t)\|_F^2.
    \label{eq:sd-pop-block}
\end{align}
Define
\begin{equation}
    F_{\perp,N}(t)
    :=
    \frac{1}{2p}
    \|\mI_{p-m}-\mB_N(t)^\top \mB_N(t)\|_F^2
    \label{eq:sd-Fperp-def}
\end{equation}
and
\begin{equation}
    G_N(t)
    :=
    \frac{1}{2m}
        \|\mI_m-\mQ_{0,N}(t)\|_F^2
    +
    \frac1m
        \Tr\mQ_{1,N}(t).
    \label{eq:sd-GN-def}
\end{equation}
Then
\begin{equation}
    L_N(t)
    =
    F_{\perp,N}(t)
    +
    \frac1{\phi_N}G_N(t).
    \label{eq:sd-pop-FG}
\end{equation}

Differentiating \cref{eq:sd-Fperp-def} and using
\cref{eq:sd-B-orth}, one obtains the exact identity
\begin{equation}
    \dot F_{\perp,N}(t)
    =
    \frac2p
    \left[
        \Tr\mQ_{1,N}(t)
        -
        \Tr\mQ_{2,N}(t)
    \right].
    \label{eq:sd-Fperp-derivative}
\end{equation}
Indeed, if
\begin{equation}
    \mC_N:=\mB_N^\top \mB_N,
    \label{eq:sd-CN-def}
\end{equation}
then
\begin{equation}
    \dot \mC_N
    =
    -2\mB_N^\top \mA_N\mA_N^\top \mB_N,
    \label{eq:sd-CN-dot}
\end{equation}
and cyclicity of the trace gives
\begin{align}
    \dot F_{\perp,N}
    &=
    \frac2p
    \Tr
    \left[
        (\mI_{p-m}-\mC_N)
        \mB_N^\top \mA_N\mA_N^\top \mB_N
    \right]
    \nonumber\\
    &=
    \frac2p
    \left[
        \Tr(\mA_N^\top \mM_N\mA_N)
        -
        \Tr(\mA_N^\top \mM_N^2\mA_N)
    \right].
    \label{eq:sd-Fperp-trace-proof}
\end{align}

Integrating \cref{eq:sd-Fperp-derivative} and using
$m/p=1/\phi_N$ gives the exact identity
\begin{align}
    \phi_N\bigl(L_N(t)-L_N(0)\bigr)
    ={}&
    G_N(t)-G_N(0)
    \nonumber\\
    &\quad
    +2
    \int_0^t
    \left[
        \frac1m\Tr\mQ_{1,N}(s)
        -
        \frac1m\Tr\mQ_{2,N}(s)
    \right]ds.
    \label{eq:sd-exact-pop-correction-orth}
\end{align}

By \cref{eq:sd-fixed-k-propagation},
\begin{equation}
    G_N(t)
    \xrightarrow{\mathbb P}
    G(t)
    :=
    \frac12(1-a(t))^2+a(t)\mu_1(t),
    \label{eq:sd-G-limit}
\end{equation}
uniformly for $t\in[0,T]$, and
\begin{equation}
    \frac1m\Tr\mQ_{2,N}(t)
    \xrightarrow{\mathbb P}
    a(t)\mu_2(t).
    \label{eq:sd-Q2-limit}
\end{equation}
At time zero,
\begin{equation}
    G(0)
    =
    \frac12(1-\rho)^2+\psi\rho^2.
    \label{eq:sd-G0-limit}
\end{equation}
Passing to the limit in
\cref{eq:sd-exact-pop-correction-orth} proves
\cref{eq:sd-pop-centered}--\cref{eq:sd-C-def} in the orthogonal-data
model.

The leading population limit follows from the fact that
\cref{eq:sd-exact-pop-correction-orth} implies
\begin{equation}
    \sup_{0\le t\le T}
    |L_N(t)-L_N(0)|
    =
    O_{\mathbb P}(\phi_N^{-1}).
    \label{eq:sd-pop-change-small}
\end{equation}
At initialization,
\begin{equation}
    L_N(0)
    =
    \frac12
    -
    \frac1p
        \Tr(\mU_N(0)^\top \mU_N(0))
    +
    \frac{1}{2p}
        \Tr
        \left[
            (\mU_N(0)^\top \mU_N(0))^2
        \right].
    \label{eq:sd-L0-init}
\end{equation}
Marchenko--Pastur convergence for the Gaussian initialization gives
\begin{equation}
    \frac1p
    \Tr(\mU_N(0)^\top \mU_N(0))
    \xrightarrow{\mathbb P}\rho,
    \label{eq:sd-init-first-moment}
\end{equation}
and
\begin{equation}
    \frac1p
    \Tr
    \left[
        (\mU_N(0)^\top \mU_N(0))^2
    \right]
    \xrightarrow{\mathbb P}
    \rho^2(1+\psi).
    \label{eq:sd-init-second-moment}
\end{equation}
\Crefrange{eq:sd-pop-change-small}{eq:sd-init-second-moment}
prove \cref{eq:sd-pop-leading} for orthogonal data.

\subsubsection{Transfer from orthogonalized to general near-isometric data}
\label{sec:sd-general-data}

We now return to an arbitrary training matrix satisfying
\cref{eq:common-data-Gram-near-isometry}.

On the event that $\mX_N$ has full column rank, write its thin polar
decomposition as
\begin{equation}
    \mX_N
    =
    \sqrt p\,\mQ_N\bm{\Gamma}_N^{1/2},
    \qquad
    \mQ_N^\top \mQ_N=\mI_m.
    \label{eq:sd-polar}
\end{equation}
Since \cref{eq:common-data-Gram-near-isometry} implies
$\lambda_{\min}(\bm{\Gamma}_N)\to1$ in probability, this event has probability
tending to one.

Extend $\mQ_N$ to an orthogonal matrix
\begin{equation}
    \mR_N
    =
    \begin{bmatrix}
        \mQ_N & \mQ_N^\perp
    \end{bmatrix}
    \in O(p),
    \label{eq:sd-R-extension}
\end{equation}
and replace $\mU_N$ by $\mU_N\mR_N$.  Conditional on $\mX_N$, the law of the
rotated initialization is still the one specified in
\cref{eq:common-Gaussian-init}.  Moreover, because this conditional law does not depend
on $\mX_N$, the rotated initialization remains independent of $\bm{\Gamma}_N$.

In the rotated coordinates,
\begin{equation}
    \frac1{\phi_N}\mS_N
    =
    \begin{pmatrix}
        \bm{\Gamma}_N&0\\
        0&0
    \end{pmatrix}.
    \label{eq:sd-S-general-rotated}
\end{equation}
Writing again $\mU_N=[\mA_N\ \mB_N]$, direct block multiplication in the common flow
\cref{eq:common-U-flow} with \(\gamma=\phi_N^{-1}\) gives
\begin{align}
    \dot \mA_N
    ={}&
    \mA_N
    \left[
        \bm{\Gamma}_N(\mI_m-\mQ_{0,N})
        +
        (\mI_m-\mQ_{0,N})\bm{\Gamma}_N
    \right]
    -
    \mM_N\mA_N\bm{\Gamma}_N,
    \label{eq:sd-A-Gamma}
\end{align}
and
\begin{equation}
    \dot \mB_N
    =
    -\mA_N\bm{\Gamma}_N\mA_N^\top \mB_N.
    \label{eq:sd-B-Gamma}
\end{equation}
For $\bm{\Gamma}_N=\mI_m$, these equations reduce exactly to
\cref{eq:sd-A-orth}--\cref{eq:sd-B-orth}.

We now couple the $\bm{\Gamma}_N$-dynamics and the orthogonal-data dynamics using
the same initial matrices $\mA_N(0),\mB_N(0)$.  Denote the two solutions by
\begin{equation}
    (\mA_N^\Gamma,\mB_N^\Gamma)
    \qquad\text{and}\qquad
    (\mA_N^{I},\mB_N^{I}).
    \label{eq:sd-coupled-flows}
\end{equation}

\begin{lemma}[Finite-time stability under Gram perturbations]
\label{lem:sd-Gamma-stability}
For every fixed $T<\infty$, on events whose probability tends to one,
\begin{equation}
    \sup_{0\le t\le T}
    \left(
        \|\mA_N^\Gamma(t)-\mA_N^I(t)\|_{\mathrm{op}}
        +
        \|\mB_N^\Gamma(t)-\mB_N^I(t)\|_{\mathrm{op}}
    \right)
    \le
    C_T
    \|\bm{\Gamma}_N-\mI_m\|_{\mathrm{op}},
    \label{eq:sd-AB-Gamma-stability}
\end{equation}
where $C_T$ is deterministic and independent of $N$.
\end{lemma}

\begin{proof}
By the Gaussian singular-value bounds used above, with probability tending
to one,
\begin{equation}
    \|\mA_N(0)\|_{\mathrm{op}}
    +
    \|\mB_N(0)\|_{\mathrm{op}}
    \le K_0
    \label{eq:sd-init-AB-bound}
\end{equation}
for a deterministic $K_0<\infty$.  The orthogonal-data trajectory then
satisfies a deterministic bound
\begin{equation}
    \sup_{0\le t\le T}
    \left(
        \|\mA_N^I(t)\|_{\mathrm{op}}
        +
        \|\mB_N^I(t)\|_{\mathrm{op}}
    \right)
    \le K_T
    \label{eq:sd-reference-bound}
\end{equation}
by Lemmas \ref{lem:sd-M-contraction} and \ref{lem:sd-A-bound}.

On the set
\begin{equation}
    \|\mA\|_{\mathrm{op}}
    +
    \|\mB\|_{\mathrm{op}}
    \le 2K_T,
    \qquad
    \|\bm{\Gamma}\|_{\mathrm{op}}\le2,
    \label{eq:sd-local-stability-set}
\end{equation}
the polynomial vector field in
\cref{eq:sd-A-Gamma}--\cref{eq:sd-B-Gamma} is uniformly Lipschitz in
$(\mA,\mB)$ and Lipschitz in $\bm{\Gamma}$.  Write $\mathcal F_A$ and
$\mathcal F_B$ for its two block components.  Hence there is a deterministic
$C=C(T,K_T)$ such that
\begin{align}
    &
    \|\mathcal F_A(\mA,\mB;\bm{\Gamma})-\mathcal F_A(\widetilde \mA,\widetilde \mB;\mI_m)\|_{\mathrm{op}}
    \nonumber\\
    &\qquad
    +
    \|\mathcal F_B(\mA,\mB;\bm{\Gamma})-\mathcal F_B(\widetilde \mA,\widetilde \mB;\mI_m)\|_{\mathrm{op}}
    \nonumber\\
    &\qquad\le
    C
    \left(
        \|\mA-\widetilde \mA\|_{\mathrm{op}}
        +
        \|\mB-\widetilde \mB\|_{\mathrm{op}}
        +
        \|\bm{\Gamma}-\mI_m\|_{\mathrm{op}}
    \right).
    \label{eq:sd-vector-field-stability}
\end{align}
Applying \cref{lem:common-ode-stability} up to the first exit of the
$\bm{\Gamma}_N$-trajectory from the radius-$2K_T$ ball yields
\begin{equation}
    \sup_{0\le t\le T}
    \left(
        \|\mA_N^\Gamma(t)-\mA_N^I(t)\|_{\mathrm{op}}
        +
        \|\mB_N^\Gamma(t)-\mB_N^I(t)\|_{\mathrm{op}}
    \right)
    \le
    C_T\|\bm{\Gamma}_N-\mI_m\|_{\mathrm{op}}.
    \label{eq:sd-stability-before-exit}
\end{equation}
For $N$ sufficiently large,
\cref{eq:common-data-Gram-near-isometry} makes the right-hand side smaller than
$K_T$, so the presumed first exit cannot occur.  This proves
\cref{eq:sd-AB-Gamma-stability}.
\end{proof}

Since
\begin{equation}
    \mM_N^\Gamma
    =
    \mB_N^\Gamma(\mB_N^\Gamma)^\top,
    \qquad
    \mM_N^I
    =
    \mB_N^I(\mB_N^I)^\top,
    \label{eq:sd-M-Gamma-I}
\end{equation}
Lemma \ref{lem:sd-Gamma-stability} implies
\begin{equation}
    \sup_{0\le t\le T}
    \|\mM_N^\Gamma(t)-\mM_N^I(t)\|_{\mathrm{op}}
    \xrightarrow{\mathbb P}0.
    \label{eq:sd-M-stability}
\end{equation}
For each fixed $k$,
\begin{equation}
    (\mM_N^\Gamma)^k-(\mM_N^I)^k
    =
    \sum_{j=0}^{k-1}
        (\mM_N^\Gamma)^j
        (\mM_N^\Gamma-\mM_N^I)
        (\mM_N^I)^{k-1-j},
    \label{eq:sd-power-difference}
\end{equation}
so the uniform operator-norm bounds and
\cref{eq:sd-AB-Gamma-stability} imply
\begin{equation}
    \sup_{0\le t\le T}
    \left\|
        (\mA_N^\Gamma)^\top
        (\mM_N^\Gamma)^k
        \mA_N^\Gamma
        -
        (\mA_N^I)^\top
        (\mM_N^I)^k
        \mA_N^I
    \right\|_{\mathrm{op}}
    \xrightarrow{\mathbb P}0.
    \label{eq:sd-Qk-Gamma-stability}
\end{equation}
Combining \cref{eq:sd-Qk-Gamma-stability} with
\cref{eq:sd-fixed-k-propagation} proves
\cref{eq:sd-matrix-isotropy} for the original, non-Gaussian training
matrix.

\subsubsection{Train loss for general near-isometric data}
\label{sec:sd-train-general}

In the rotated coordinates \cref{eq:sd-polar}, the empirical loss has the
exact form
\begin{equation}
    \widehat L_N(t)
    =
    \frac{1}{2m}
    \Tr
    \left\{
        \bm{\Gamma}_N
        \left[
            (\mI_m-\mQ_{0,N}(t))^2+\mQ_{1,N}(t)
        \right]
    \right\}.
    \label{eq:sd-train-Gamma-exact}
\end{equation}
To verify this, note that the upper-left block of $\mE_N^2$ is
\begin{equation}
    (\mE_N^2)_{11}
    =
    (\mI_m-\mQ_{0,N})^2+\mQ_{1,N}.
    \label{eq:sd-E2-upper}
\end{equation}
\Cref{eq:sd-train-Gamma-exact} then follows directly from
\cref{eq:sd-polar} and \cref{eq:common-losses}.

For every matrix $\mR\in\mathbb R^{m\times m}$,
\begin{equation}
    \left|
        \frac1m\Tr
        \bigl[(\bm{\Gamma}_N-\mI_m)\mR\bigr]
    \right|
    \le
    \|\bm{\Gamma}_N-\mI_m\|_{\mathrm{op}}
    \|\mR\|_{\mathrm{op}}.
    \label{eq:sd-normalized-trace-Gamma}
\end{equation}
Therefore \cref{eq:common-data-Gram-near-isometry},
\cref{eq:sd-matrix-isotropy}, and the uniform operator-norm bounds imply
\cref{eq:sd-train-limit}.

\subsubsection{Population correction for general near-isometric data}
\label{sec:sd-pop-general}

The population decomposition \cref{eq:sd-pop-block} is independent of
$\bm{\Gamma}_N$, but the inactive block now evolves according to
\cref{eq:sd-B-Gamma}.  Repeating the differentiation leading to
\cref{eq:sd-Fperp-derivative} gives the exact identity
\begin{equation}
    \dot F_{\perp,N}(t)
    =
    \frac2p
    \Tr
    \left[
        \bm{\Gamma}_N
        \bigl(
            \mQ_{1,N}(t)-\mQ_{2,N}(t)
        \bigr)
    \right].
    \label{eq:sd-Fperp-Gamma}
\end{equation}
Consequently,
\begin{align}
    \phi_N\bigl(L_N(t)-L_N(0)\bigr)
    ={}&
    G_N(t)-G_N(0)
    \nonumber\\
    &\quad+
    2\int_0^t
        \frac1m
        \Tr
        \left[
            \bm{\Gamma}_N
            \bigl(
                \mQ_{1,N}(s)-\mQ_{2,N}(s)
            \bigr)
        \right]ds.
    \label{eq:sd-exact-pop-correction-Gamma}
\end{align}

For $k=1,2$,
\begin{align}
    &
    \left|
        \frac1m
        \Tr(\bm{\Gamma}_N\mQ_{k,N})
        -
        a\mu_k
    \right|
    \nonumber\\
    &\qquad
    \le
    \|\bm{\Gamma}_N\|_{\mathrm{op}}
    \|\mQ_{k,N}-a\mu_k\mI_m\|_{\mathrm{op}}
    +
    |a\mu_k|
    \left|
        \frac1m\Tr\bm{\Gamma}_N-1
    \right|.
    \label{eq:sd-Gamma-Q-limit-bound}
\end{align}
Both terms tend to zero uniformly on $[0,T]$ in probability, by
\cref{eq:common-data-Gram-near-isometry} and
\cref{eq:sd-matrix-isotropy}.  Therefore
\cref{eq:sd-exact-pop-correction-Gamma} converges uniformly to
\cref{eq:sd-C-def}, proving \cref{eq:sd-pop-centered}.

This also explains why an $O(\phi_N^{-1/2})$ operator-norm error in
$\bm{\Gamma}_N$ does not contaminate the $O(\phi_N^{-1})$ population correction.
After multiplication by $\phi_N$, the Gram perturbation enters only through
an $m$-normalized trace.  In particular,
\begin{equation}
    \left|
        \frac1m
        \Tr
        \left[
            (\bm{\Gamma}_N-\mI_m)\mQ_{k,N}
        \right]
    \right|
    \le
    \|\bm{\Gamma}_N-\mI_m\|_{\mathrm{op}}
    \|\mQ_{k,N}\|_{\mathrm{op}}
    =
    o_{\mathbb P}(1).
    \label{eq:sd-no-second-order-needed}
\end{equation}
No second-order expansion of the Gram matrix is required.

Finally, \cref{eq:sd-pop-centered} implies
\begin{equation}
    \sup_{0\le t\le T}
    |L_N(t)-L_N(0)|
    =
    O_{\mathbb P}(\phi_N^{-1}),
    \label{eq:sd-general-pop-change}
\end{equation}
and the initialization limit
\cref{eq:sd-init-first-moment}--\cref{eq:sd-init-second-moment} is
independent of the training data.  Thus
\cref{eq:sd-pop-leading} also holds for general data satisfying
\cref{eq:common-data-Gram-near-isometry}.  This completes the proof of
\cref{thm:sd-main}.

\subsubsection{Expected losses and Gaussian training data as a sufficient condition}
\label{sec:sd-gaussian-instantiation}

We first upgrade the convergence statements of
\cref{thm:sd-main} to $L^1$ under the uniform-integrability condition
used in the main text. Recall from
\cref{eq:common-small-data-normalization} that
$\bm{\Gamma}_N=p^{-1}\mX_N^\top\mX_N$.

\paragraph{Expectation upgrade.}
Assume the hypotheses of \cref{thm:sd-main} and suppose in addition that,
for some $\delta>0$,
\begin{equation}
    \sup_N
    \E\|\bm{\Gamma}_N\|_{\mathrm{op}}^{1+\delta}
    <\infty.
    \label{eq:sd-Gamma-UI}
\end{equation}
Then, for every fixed $T<\infty$,
\begin{equation}
    \E\left[
        \sup_{0\le t\le T}
        \left|
            \widehat L_N(t)
            -
            \left\{
                \frac12(1-a(t))^2
                +
                \frac12a(t)\mu_1(t)
            \right\}
        \right|
    \right]
    \longrightarrow0,
    \label{eq:sd-train-L1}
\end{equation}
and
\begin{equation}
    \E\left[
        \sup_{0\le t\le T}
        \left|
            \phi_N\bigl(L_N(t)-L_N(0)\bigr)
            -
            C(t)
        \right|
    \right]
    \longrightarrow0.
    \label{eq:sd-pop-correction-L1}
\end{equation}
In particular,
\begin{equation}
    \E\left[
        \sup_{0\le t\le T}
        \left|
            L_N(t)-L_\infty^{(0)}
        \right|
    \right]
    \longrightarrow0,
    \label{eq:sd-pop-L1}
\end{equation}
where $L_\infty^{(0)}$ is defined in
\cref{eq:sd-pop-leading-def}.

To prove these statements, define
\begin{equation}
    R_N
    :=
    \max\left\{
        1,\|\mU_N(0)\|_{\mathrm{op}}^2
    \right\}.
    \label{eq:sd-RN-UI}
\end{equation}
By \cref{lem:common-gram-flow}, applied to the common flow with
$\gamma_N=\phi_N^{-1}$,
\begin{equation}
    \sup_{t\ge0}
    \|\mU_N(t)\|_{\mathrm{op}}^2
    \le R_N.
    \label{eq:sd-U-RN-UI}
\end{equation}
After the rotation to the active and inactive blocks, this gives, for
$k=0,1,2$,
\begin{equation}
    \sup_{t\ge0}
    \|\mQ_{k,N}(t)\|_{\mathrm{op}}
    \le R_N^{k+1}.
    \label{eq:sd-Q-RN-UI}
\end{equation}
Indeed, $\|\mA_N\|_{\mathrm{op}}\le\|\mU_N\|_{\mathrm{op}}$ and
$\|\mM_N\|_{\mathrm{op}}=\|\mB_N\|_{\mathrm{op}}^2\le R_N$.
Consequently, the quantity $G_N$ defined in
\cref{eq:sd-GN-def} satisfies
\begin{equation}
    \sup_{t\ge0}|G_N(t)|
    \le
    \frac12(1+R_N)^2+R_N^2.
    \label{eq:sd-G-RN-UI}
\end{equation}
Using the exact identity
\cref{eq:sd-exact-pop-correction-Gamma}, we therefore obtain, for every
fixed $T<\infty$,
\begin{align}
    \sup_{0\le t\le T}
    \left|
        \phi_N\bigl(L_N(t)-L_N(0)\bigr)
    \right|
    &\le
    (1+R_N)^2+2R_N^2
    \nonumber\\
    &\quad
    +2T\|\bm{\Gamma}_N\|_{\mathrm{op}}
    \left(R_N^2+R_N^3\right)
    \nonumber\\
    &\le
    C_T
    \left(1+\|\bm{\Gamma}_N\|_{\mathrm{op}}\right)
    \left(1+R_N^3\right).
    \label{eq:sd-pop-correction-dominator}
\end{align}
Likewise, the exact train-loss identity
\cref{eq:sd-train-Gamma-exact} yields
\begin{equation}
    \sup_{0\le t\le T}
    \widehat L_N(t)
    \le
    \frac12
    \|\bm{\Gamma}_N\|_{\mathrm{op}}
    \left[(1+R_N)^2+R_N^2\right].
    \label{eq:sd-train-dominator}
\end{equation}

By the standing Gaussian initialization
\cref{eq:common-Gaussian-init}, for every fixed $q<\infty$,
\begin{equation}
    \sup_N \E R_N^q<\infty.
    \label{eq:sd-RN-moments}
\end{equation}
Moreover, $R_N$ is independent of $\bm{\Gamma}_N$. Choose
$0<\varepsilon<\delta$. Then
\cref{eq:sd-Gamma-UI,eq:sd-RN-moments} and independence imply that the
right-hand sides of
\cref{eq:sd-pop-correction-dominator,eq:sd-train-dominator} have uniformly
bounded $(1+\varepsilon)$ moments. Hence they are uniformly integrable.
Combining this with the uniform convergence in probability in
\cref{eq:sd-train-limit,eq:sd-pop-centered} proves
\cref{eq:sd-train-L1,eq:sd-pop-correction-L1}.

The standing Gaussian initialization also implies
$L_N(0)\to L_\infty^{(0)}$ in $L^1$: convergence in probability follows
from \cref{eq:sd-L0-init,eq:sd-init-first-moment,eq:sd-init-second-moment},
while Gaussian operator-norm moments give uniform integrability.
Together with \cref{eq:sd-pop-correction-L1} and $\phi_N\to\infty$, this
proves \cref{eq:sd-pop-L1}.

The initialization expectation can in fact be evaluated exactly:
\begin{equation}
    \E L_N(0)
    =
    \frac12
    \left[
        1-2\rho_N
        +
        \left(
            1+\psi_N+\frac1{n_N}
        \right)
        \rho_N^2
    \right].
    \label{eq:sd-pop-initial-expectation}
\end{equation}
Indeed,
\begin{equation}
    \E\left[
        \frac1p
        \Tr\bigl(\mU_N(0)^\top\mU_N(0)\bigr)
    \right]
    =
    \rho_N,
    \label{eq:sd-init-first-expectation}
\end{equation}
while the Gaussian fourth-moment identity gives
\begin{equation}
    \E\left[
        \frac1p
        \Tr\left[
            \bigl(\mU_N(0)^\top\mU_N(0)\bigr)^2
        \right]
    \right]
    =
    \rho_N^2
    \left(
        1+\psi_N+\frac1{n_N}
    \right).
    \label{eq:sd-init-second-expectation}
\end{equation}
Therefore \cref{eq:sd-pop-correction-L1} implies, uniformly for
$0\le t\le T$,
\begin{equation}
    \E L_N(t)
    =
    \frac12
    \left[
        1-2\rho_N
        +
        \left(
            1+\psi_N+\frac1{n_N}
        \right)
        \rho_N^2
    \right]
    +
    \frac1{\phi_N}C(t)
    +
    o(\phi_N^{-1}).
    \label{eq:sd-pop-expectation-general}
\end{equation}
Since
\begin{equation}
    \frac{n_N^{-1}}{\phi_N^{-1}}
    =
    \frac{\psi_N}{m_N}
    \longrightarrow0,
    \label{eq:sd-n-negligible}
\end{equation}
if in addition
\begin{equation}
    \rho_N-\rho=o(\phi_N^{-1}),
    \qquad
    \psi_N-\psi=o(\phi_N^{-1}),
    \label{eq:sd-ratio-rate}
\end{equation}
then \cref{eq:sd-pop-expectation-general} is equivalently
\begin{align}
    \E L_N(t)
    ={}&
    \frac12
    \left[
        1-2\rho+(1+\psi)\rho^2
    \right]
    \nonumber\\
    &+
    \frac1{\phi_N}
    \left\{
        \frac12(1-a(t))^2
        +a(t)\mu_1(t)
        -
        \frac12
        \left[
            1-2\rho+(1+2\psi)\rho^2
        \right]
    \right.
    \nonumber\\
    &\hspace{5em}\left.
        +2\int_0^t
        a(s)\bigl[\mu_1(s)-\mu_2(s)\bigr]\,ds
    \right\}
    +o(\phi_N^{-1}).
    \label{eq:sd-pop-expectation-expansion}
\end{align}
The coefficient of $\phi_N^{-1}$ is precisely the function $C(t)$ in
\cref{eq:sd-C-def}.

\paragraph{Gaussian training data as a sufficient condition.}
We now verify both the active-space near-isometry assumption and
\cref{eq:sd-Gamma-UI} for isotropic Gaussian training data. Suppose
\begin{equation}
    (\mX_N)_{ij}
    \overset{\mathrm{i.i.d.}}{\sim}
    \mathcal N(0,1).
    \label{eq:sd-Gaussian-X}
\end{equation}
The standard Gaussian singular-value inequalities give, for every $t\ge0$,
\begin{equation}
    \mathbb P\left(
        s_{\max}(\mX_N)
        >
        \sqrt p+\sqrt m+t
    \right)
    \le e^{-t^2/2},
    \label{eq:sd-gaussian-smax}
\end{equation}
and
\begin{equation}
    \mathbb P\left(
        s_{\min}(\mX_N)
        <
        \sqrt p-\sqrt m-t
    \right)
    \le e^{-t^2/2}.
    \label{eq:sd-gaussian-smin}
\end{equation}
Since $m/p=\phi_N^{-1}\to0$, these bounds imply
\begin{equation}
    \left\|
        \frac1p\mX_N^\top\mX_N-\mI_m
    \right\|_{\mathrm{op}}
    =
    O_{\mathbb P}
    \left(
        \sqrt{\frac mp}+\frac mp
    \right)
    =
    O_{\mathbb P}(\phi_N^{-1/2}).
    \label{eq:sd-gaussian-Gram-rate}
\end{equation}
Thus the common active-space near-isometry condition
\cref{eq:common-data-Gram-near-isometry} holds. Moreover, integration
of the upper singular-value tail gives, for every fixed $q<\infty$,
\begin{equation}
    \sup_N
    \E\|\bm{\Gamma}_N\|_{\mathrm{op}}^q
    <\infty.
    \label{eq:sd-gaussian-Gamma-moments}
\end{equation}
Hence Gaussian training data also satisfy the additional
uniform-integrability condition \cref{eq:sd-Gamma-UI}.

\paragraph{Universality with respect to the training distribution.}
Gaussianity of the training samples is therefore not required. The
finite-time conclusions of \cref{thm:sd-main} hold for any random or
deterministic sequence of training matrices satisfying
\cref{eq:common-data-Gram-near-isometry}; the $L^1$ and expectation
statements above additionally require \cref{eq:sd-Gamma-UI}.

For example, this includes broad families of isotropic non-Gaussian
distributions with sufficiently strong norm and inner-product concentration.
Isotropy by itself is not sufficient. If
\begin{equation}
    \vx
    =
    \sqrt{\tau}\,\sqrt p\,\vv,
    \qquad
    \vv\sim\operatorname{Unif}(\mathbb S^{p-1}),
    \qquad
    \E\tau=1,
    \label{eq:sd-radial-counterexample}
\end{equation}
and $\tau$ has nonvanishing fluctuations as $p\to\infty$, then
$\E[\vx\vx^\top]=\mI_p$, but
\begin{equation}
    \frac{\|\vx\|^2}{p}
    =
    \tau.
    \label{eq:sd-radial-norm}
\end{equation}
Thus the normalized Gram matrix is approximately diagonal with entries
$\tau_1,\ldots,\tau_m$, rather than approximately $\mI_m$. Different
examples then evolve at different effective rates, and the scalar hierarchy
\cref{eq:sd-a-ode,eq:sd-mu-ode} need not remain valid.

\subsubsection{Remarks on anisotropic data}
\label{sec:sd-anisotropic}

The training-loss part of the argument extends further than the isotropic
population setting.  The empirical dynamics depends on the data, after
rotation to their span, only through the normalized Gram matrix
$\bm{\Gamma}_N$.  Thus if an anisotropic data model is normalized in such a way
that it still satisfies
\begin{equation}
    \|\bm{\Gamma}_N-\mI_m\|_{\mathrm{op}}\longrightarrow0,
    \label{eq:sd-anisotropic-Gram}
\end{equation}
then the same limiting hierarchy and the same train-loss formula continue
to hold.

The population loss is different.  If the population covariance is a
non-scalar matrix $\mSigma_N$, then
\begin{equation}
    L_{\Sigma,N}(U)
    =
    \frac{1}{2p}
    \Tr
    \left[
        \mSigma_N
        (\mI_p-\mU^\top\mU)^2
    \right].
    \label{eq:sd-anisotropic-pop}
\end{equation}
After rotating to the training span, $\mSigma_N$ acquires the block form
\begin{equation}
    \mR_N^\top\mSigma_N\mR_N
    =
    \begin{pmatrix}
        \mSigma_{\parallel\parallel,N}
        &
        \mSigma_{\parallel\perp,N}\\
        \mSigma_{\perp\parallel,N}
        &
        \mSigma_{\perp\perp,N}
    \end{pmatrix}.
    \label{eq:sd-Sigma-block}
\end{equation}
The population correction then depends not only on
$\mA_N^\top \mM_N^k\mA_N$, but also on weighted observables coupling these blocks
to $\mA_N$ and $\mB_N$.  Consequently, the train hierarchy can remain universal
while the population correction changes.

A rigorous anisotropic extension should therefore augment the present
matrix hierarchy by weighted order parameters describing the joint
asymptotic geometry of the training span and $\mSigma_N$.  If, for example,
the compressed covariance becomes asymptotically scalar,
\begin{equation}
    \left\|
        \mQ_N^\top\mSigma_N\mQ_N
        -
        \gamma \mI_m
    \right\|_{\mathrm{op}}
    \longrightarrow0,
    \label{eq:sd-anisotropic-compression}
\end{equation}
and the corresponding active--inactive weighted traces admit deterministic
limits, one expects a finite enlargement of the hierarchy and an
anisotropy-dependent analogue of
\cref{eq:sd-C-def}.  For a fully general deterministic spectral profile
$\mSigma_N$, however, scalar closure should not be expected; a
matrix-valued or measure-valued weighted hierarchy is the natural object.

Finally, all results in this section are finite-time statements: the limit
$N\to\infty$ is uniform on every fixed interval $[0,T]$.  The subsequent
limit $t\to\infty$, and in particular any interchange of the
large-dimension and large-time limits, requires an additional analysis of
the deterministic moment dynamics and is not asserted here.

\section{Experimental details}
\label{app:experiments}

This appendix describes the finite-dimensional experiments used to test the
limiting dynamics derived in \cref{sec:solutions}. We organize the discussion by learning
regime. Several data constructions recur across regimes and are described first.

Unless stated otherwise, the model is initialized according to the same
isotropic Gaussian law as in the theory,
\begin{equation}
    U_{\alpha i}(0) \stackrel{\mathrm{iid}}{\sim} \mathcal N(0,\sigma^2),
\end{equation}
with the regime-specific rescaled initialization parameter held fixed.
Whenever several independent initializations are used, curves show the empirical
mean and error bars show one standard deviation unless specified otherwise.

\subsection{Recurring experimental setups}
\label{app:exp-recurring}

\paragraph{Synthetic Gaussian data.}
For synthetic experiments we sample
\begin{equation}
    \mX\in\mathbb R^{p\times m},
    \qquad
    X_{i\mu}\stackrel{\mathrm{iid}}{\sim}\mathcal N(0,1).
\end{equation}
Thus the population covariance is exactly isotropic,
\begin{equation}
    \mathbb E[\vx\vx^\top]=\mI_p,
\end{equation}
and the empirical covariance
\begin{equation}
    \mS=\frac1m \mX\mX^\top
\end{equation}
has the Marchenko--Pastur spectral limit at fixed
$\phi=p/m$. When $m<p$, its $m$ nonzero eigenvalues are obtained
numerically from the smaller matrix $\mX^\top\mX/m$.

The narrow-latent large-time experiment integrates an exact reduced
finite-dimensional gradient-flow system. In that experiment a single Gaussian
training matrix is fixed and the weight initialization is resampled across
seeds.

\paragraph{Globally whitened CIFAR-10.}
CIFAR-10 images are treated as unlabeled vectors. Each $32\times32$ RGB image
is flattened, giving the original dimension
\begin{equation}
    p_0=32\cdot32\cdot3=3072,
\end{equation}
and pixel intensities are divided by $255$.

Whenever we refer to globally whitened CIFAR-10, the whitening map is fit using
the entire set of
\begin{equation}
    m_{\mathrm{full}}=50\,000
\end{equation}
training images, before any training subset is selected. Let
\begin{equation}
    \bar{\vx}
    =
    \frac1{m_{\mathrm{full}}}
    \sum_{\mu=1}^{m_{\mathrm{full}}}\vx_\mu
\end{equation}
and
\begin{equation}
    \mC_{\mathrm{full}}
    =
    \frac1{m_{\mathrm{full}}}
    \sum_{\mu=1}^{m_{\mathrm{full}}}
    (\vx_\mu-\bar{\vx})(\vx_\mu-\bar{\vx})^\top
    =
    \mV\bm{\Lambda}\mV^\top .
\end{equation}
The whitened coordinates are
\begin{equation}
    \vz=\bm{\Lambda}^{-1/2}\mV^\top(\vx-\bar{\vx}).
\end{equation}
The same training mean and the same whitening transformation are applied to
held-out test images; no test-set statistics enter the preprocessing map.

If only a subset of the training data is used for optimization, that subset is
selected after the global whitening map has been fixed. Thus the covariance of
the complete whitened training split is approximately the identity, while the
covariance $\mS$ of a finite subset remains nontrivial. This construction removes
the large anisotropy of the original CIFAR-10 coordinates while retaining a
nondegenerate finite-sample spectrum.

For the mean-field experiment we retain only the $p=512$ largest-variance PCA
directions before whitening, for computational efficiency. The narrow-latent
and large-data experiments use the full image dimension unless stated
otherwise.

\subsection{Large-data experiments}
\label{app:exp-large-data}

We use all $50\,000$ CIFAR-10 training images and all $10\,000$ test images,
with the global whitening construction described above and $p=3072$. Up to
numerical precision,
\begin{equation}
    \mS_{\mathrm{tr}}=\mI_p.
\end{equation}
We fix
\begin{equation}
    \rho=n\sigma^2=1
\end{equation}
and use
\begin{equation}
    n\in\{2p,p,p/2\}=\{6144,3072,1536\},
    \qquad
    \psi=\frac pn\in\left\{\frac12,1,2\right\}.
\end{equation}
The same Gaussian seed is used for the three widths.

With $\eta=p$, the exact finite-dimensional gradient flow reduces to
\begin{equation}
    \dot{\mU}=2\mU(\mI_p-\mU^\top\mU).
\end{equation}
Writing $\mA=\mU^\top\mU$ gives the exact Gram dynamics
\begin{equation}
    \dot{\mA}=4\mA(\mI_p-\mA).
\end{equation}
Therefore, if
\begin{equation}
    \mA(0)=\mV\diag(a_1(0),\ldots,a_p(0))\mV^\top,
\end{equation}
the eigenvectors remain fixed and
\begin{equation}
    a_j(t)
    =
    \frac{a_j(0)}
    {a_j(0)+(1-a_j(0))e^{-4t}}.
\end{equation}
We use this formula directly, so the finite-width CIFAR-10 curves contain no
time-discretization error. When $n<p$, the $p-n$ exact zero eigenvalues are
retained analytically.

The finite-width training loss is
\begin{equation}
    \widehat L_{\mathrm{tr}}(t)
    =
    \frac1{2p}
    \sum_{j=1}^p
    (1-a_j(t))^2.
\end{equation}
If
\begin{equation}
    \mT
    =
    \frac1{m_{\mathrm{te}}}
    \mX_{\mathrm{te}}\mX_{\mathrm{te}}^\top
\end{equation}
is the empirical covariance of the globally whitened CIFAR-10 test set and
$\vv_j$ is the $j$th eigenvector of $\mA(0)$, we set
\begin{equation}
    w_j=\vv_j^\top \mT \vv_j
\end{equation}
and evaluate
\begin{equation}
    L_{\mathrm{te}}(t)
    =
    \frac1{2p}
    \sum_{j=1}^p
    w_j(1-a_j(t))^2.
\end{equation}

The theoretical comparison is the proportional large-data prediction from
\cref{app:large-data},
\begin{equation}
    L_{\psi,\rho}(t)
    =
    \frac12
    \int
    \bigl(\Theta_t(\rho\lambda)-1\bigr)^2\,d\mu_\psi(\lambda),
    \qquad
    \Theta_t(s)
    =
    \frac{s}{s+(1-s)e^{-4t}},
\end{equation}
with $\mu_\psi$ the Marchenko--Pastur law. The noncritical cases are evaluated
by Gauss--Legendre quadrature; at $\psi=1$ we use the stabilized closed form
described in \cref{app:large-data-rigorous-large-time}. All plotted training,
test, and theoretical curves are divided by their respective initial values.

\subsection{Mean-field experiments}
\label{app:exp-mean-field}

We use a globally whitened CIFAR-10 subset for the train/test comparison shown
in the main text. Global whitening is fit on all $50\,000$ training images,
after which we retain the top
\begin{equation}
    p=512
\end{equation}
PCA coordinates and select a fixed subset of
\begin{equation}
    m=256
\end{equation}
whitened training examples using subset seed $0$. All $10\,000$ test images
are transformed using the same full-training whitening map.

We fix
\begin{equation}
    \rho=0.5
\end{equation}
and repeat the exact finite-width experiment for
\begin{equation}
    \frac np\in\{4,16,64\},
    \qquad
    n\in\{2048,8192,32768\}.
\end{equation}
For each width we use $16$ independent Gaussian initializations,
\begin{equation}
    \mA(0)
    \stackrel{d}{=}
    \frac{\rho}{n}\mZ^\top\mZ,
    \qquad
    Z_{\alpha i}\stackrel{\mathrm{iid}}{\sim}\mathcal N(0,1).
\end{equation}

Let
\begin{equation}
    \mS
    =
    \mQ\diag(\lambda_1,\ldots,\lambda_p)\mQ^\top
\end{equation}
be the covariance of the training subset, and let $\mT$ be the covariance of the
whitened held-out test split. Define
\begin{equation}
    a_t(\lambda)
    =
    \frac{\rho}
    {\rho+(1-\rho)e^{-4\lambda t}}
\end{equation}
and
\begin{equation}
    w_i=\vq_i^\top\mT\vq_i.
\end{equation}
The two mean-field comparison curves are
\begin{equation}
    \widehat L_{\mathrm{MF}}(t)
    =
    \frac1{2p}
    \sum_{i=1}^p
    \lambda_i\bigl(1-a_t(\lambda_i)\bigr)^2,
\end{equation}
and
\begin{equation}
    L_{\mathrm{te,MF}}(t)
    =
    \frac1{2p}
    \sum_{i=1}^p
    w_i\bigl(1-a_t(\lambda_i)\bigr)^2.
\end{equation}
Thus both theoretical curves use the actual finite-sample spectrum of the
training subset; no Marchenko--Pastur approximation is made.

For the finite-width trajectories we evolve the exact Gram dynamics
\begin{equation}
    \dot{\mA}
    =
    2(\mS\mA+\mA\mS)
    -\mS\mA^2
    -\mA^2\mS
    -2\mA\mS\mA.
\end{equation}
Writing $\mE=\mI_p-\mA$ in the $\mS$ eigenbasis gives a semilinear equation
whose linear part is diagonal entrywise. We integrate it with the adaptive
ETD2 scheme using
\begin{equation}
    \mathrm{rtol}=5\times10^{-4},
    \qquad
    \mathrm{atol}=10^{-11},
\end{equation}
and record $161$ equally spaced output times on $0\le t\le4$.
Experimental error bars show one standard deviation over the $16$
initialization seeds. The theory curve is common to the three values of $n/p$;
increasing $n/p$ tests convergence toward the mean-field limit.

\subsection{Narrow-latent experiments}
\label{app:exp-narrow}

The narrow-latent experiments serve two purposes. The CIFAR-10 experiment tests
the finite-time scalar reduction and the initialization-dependent branch
structure. A separate Gaussian experiment tests the derived large-time loss
asymptotics.

\paragraph{Globally whitened CIFAR-10 subset: branch structure.}
We globally whiten CIFAR-10 using the full training split and then choose a
subset of
\begin{equation}
    m=1024
\end{equation}
examples. The experiment uses
\begin{equation}
    p=3072,
    \qquad
    n=1,
    \qquad
    \phi=\frac pm=3,
\end{equation}
and $16$ Gaussian initialization seeds.

Because $n=1$, finite-width row competition is absent and the narrow-latent
order parameters can be followed directly. Let
$\lambda_1,\ldots,\lambda_m$ be the nonzero eigenvalues of the subset
covariance. In its eigenbasis write $C_i$ for the active coordinates of the
single weight row and let $a_\perp$ denote its total squared norm in the data
nullspace. Then
\begin{equation}
    a=\sum_{i=1}^m C_i^2+a_\perp,
    \qquad
    r=\sum_{i=1}^m\lambda_i C_i^2,
\end{equation}
and the original finite-dimensional gradient flow is exactly
\begin{equation}
    \dot C_i
    =
    C_i(2\lambda_i-r-a\lambda_i),
    \qquad
    \dot a_\perp=-2ra_\perp.
\end{equation}
We additionally integrate
\begin{equation}
    \dot q=2(2-a),
    \qquad
    q(0)=0,
\end{equation}
which allows the finite-width spectral clock to be compared directly with the
limiting one.

The finite-width scaled loss corrections are
\begin{equation}
    \widehat F_N(t)
    =
    \psi
    \left(
        \widehat L(t)-\frac{\Tr\mS}{2p}
    \right),
\end{equation}
and
\begin{equation}
    F_N(t)
    =
    \psi
    \left(
        L(t)-\frac12
    \right),
    \qquad
    \psi=\frac pn.
\end{equation}
For $n=1$ these reduce exactly to
\begin{equation}
    \widehat F_N
    =
    r\left(\frac a2-1\right),
    \qquad
    F_N
    =
    a\left(\frac a2-1\right).
\end{equation}
The population observable here is the isotropic population loss from the
theorem, not reconstruction loss on the CIFAR-10 test split. This choice
isolates the train/population distinction predicted by the theory from
additional anisotropy of the empirical CIFAR-10 test covariance.

We consider four initialization scenarios:
\begin{equation}
\begin{array}{c|c}
\text{scenario} & \kappa=p\sigma^2 \\ \hline
\text{monotonic alignment} & 0.5 \\
\text{nonmonotonic alignment} & 4 \\
\text{trapped} & 10 \\
\text{critical} & \kappa_c
\end{array}
\end{equation}
Writing
\begin{equation}
    w_0=\frac{p-\rank(\mS)}p,
\end{equation}
the critical value is the solution $\kappa_c>2$ of
\begin{equation}
    \frac{\kappa_c^2}{\kappa_c-1}w_0^2=4.
\end{equation}
For $p=3072$ and a full-rank subset of size $1024$,
$w_0=2/3$ and $\kappa_c\simeq7.854$.

For the limiting comparison we use the actual empirical subset spectrum.
Specifically,
\begin{equation}
    a_{\perp,\mS}(q)
    =
    \frac1p
    \left[
        p-m+\sum_{i=1}^m e^{q\lambda_i}
    \right],
\end{equation}
\begin{equation}
    m_{\mS}(q)
    =
    \frac{\sum_i\lambda_i e^{q\lambda_i}}
    {p-m+\sum_i e^{q\lambda_i}},
\end{equation}
and solve
\begin{equation}
    \dot a=4a\,m_{\mS}(q)(1-a),
    \qquad
    \dot q=2(2-a),
    \qquad
    r=a\,m_{\mS}(q).
\end{equation}
The associated limiting loss curves are
\begin{equation}
    \widehat F(t)
    =
    r(t)\left(\frac{a(t)}2-1\right),
    \qquad
    F(t)
    =
    a(t)\left(\frac{a(t)}2-1\right).
\end{equation}

The training-loss plot also shows the finite-sample rank-one PCA optimum,
\begin{equation}
    \widehat F_{\mathrm{PCA}}
    =
    -\frac{\lambda_{\max}(\mS)}2.
\end{equation}
This makes the optimization failure in the trapped branch visible directly:
the limiting trapped trajectory approaches $\widehat F=0$ rather than the PCA
optimum. The plots of $a(t)$, $q(t)$, and $r(t)$ additionally display the
nonmonotonic-alignment mechanism: for $\kappa>2$, $q$ first decreases, reaches
its minimum when $a\simeq2$, and subsequently reverses direction in the
alignment branch.

We integrate the exact $n=1$ system using RK4 with the step size scaled by the
empirical upper spectral edge,
\begin{equation}
    \Delta t
    =
    \frac{0.01}{\lambda_{\max}(\mS)},
\end{equation}
up to $t=30$.

\paragraph{Gaussian data: large-time tails.}
To test the narrow-latent large-time formulas without residual
finite-sample anisotropy from CIFAR-10, we use a larger Gaussian experiment with
\begin{equation}
    p=3072\times8=24576,
    \qquad
    m=1024\times8=8192,
    \qquad
    n=1,
\end{equation}
so again $\phi=3$. A single Gaussian training matrix is generated with data
seed $12345$, and we average over eight independent weight initializations.

The experiment evaluates alignment, critical, and trapped initializations,
\begin{equation}
    \kappa_{\mathrm{align}}=4,
    \qquad
    \kappa_{\mathrm{crit}}=\kappa_c,
    \qquad
    \kappa_{\mathrm{trap}}=9.
\end{equation}
The exact finite-width active/null-space dynamics are integrated with RK4 using
\begin{equation}
    \Delta t
    =
    \frac{0.02}{\lambda_{\max}^{\mathrm{emp}}},
    \qquad
    0\le t\le10.
\end{equation}
At $\phi=3$ the asymptotic comparison uses the Marchenko--Pastur edges
\begin{equation}
    \lambda_\pm=(1\pm\sqrt3)^2.
\end{equation}
For alignment, the derived loss residuals are
\begin{equation}
    \widehat F(t)-\widehat F(\infty)
    \sim
    \frac{3}{8t},
\end{equation}
and
\begin{equation}
    F(t)-F(\infty)
    \sim
    \frac12 K_a^2 t^6 e^{-8\lambda_+t}.
\end{equation}

\subsection{Small-data experiments}
\label{app:exp-small-data}

We validate the small-data prediction of \cref{app:small-data} using globally whitened
CIFAR-10. The experiment is designed to separate the two geometric requirements
involved in the comparison. The whitening transformation is fit globally,
using the complete CIFAR-10 training split, so that held-out test images can be
evaluated in the same approximately isotropic coordinate system. The much
smaller subset actually used for optimization is then orthogonalized
separately, so that the active-space near-isometry assumed by the small-data
theory holds exactly at finite dimension.

\paragraph{Global whitening and dimensionality reduction.}
Each $32\times32$ RGB CIFAR-10 image is first flattened into
$\mathbb R^{3072}$ and rescaled by $1/255$. We compute the empirical mean and
covariance using all $50\,000$ training images,
\begin{equation}
    \bar{\vx}
    =
    \frac1{m_{\mathrm{full}}}
    \sum_{\mu=1}^{m_{\mathrm{full}}}\vx_\mu,
    \qquad
    \mC_{\mathrm{full}}
    =
    \frac1{m_{\mathrm{full}}}
    \sum_{\mu=1}^{m_{\mathrm{full}}}
    (\vx_\mu-\bar{\vx})(\vx_\mu-\bar{\vx})^\top.
\end{equation}
Writing
\begin{equation}
    \mC_{\mathrm{full}}
    =
    \mV\bm{\Lambda}\mV^\top,
\end{equation}
we retain the $p=1024$ principal directions with the largest eigenvalues and
apply the whitening transformation
\begin{equation}
    \vz
    =
    \bm{\Lambda}_p^{-1/2}
    \mV_p^\top(\vx-\bar{\vx}).
\end{equation}
The inverse square root is formed only on numerically stable directions:
eigenvalues below $10^{-8}$ times the largest eigenvalue are discarded before
the top $p$ directions are selected.

The same mean $\bar{\vx}$ and the same whitening transformation are applied to
all $10\,000$ CIFAR-10 test images. Thus no test-set statistics enter either
preprocessing or training. Let
\begin{equation}
    \mT_{\mathrm{te}}
    =
    \frac1{m_{\mathrm{te}}}
    \mX_{\mathrm{te}}\mX_{\mathrm{te}}^\top
\end{equation}
denote the empirical second-moment matrix of the resulting globally whitened
test data.

\paragraph{Orthogonalized training subsets.}
After the global whitening map has been fixed, we randomly select $128$
whitened training examples using subset seed $0$. Arrange them as the columns
of
\begin{equation}
    \mX_{\mathrm{sel}}\in\mathbb R^{p\times128}
\end{equation}
and compute a thin QR factorization
\begin{equation}
    \mX_{\mathrm{sel}}=\mQ\mR,
    \qquad
    \mQ^\top\mQ=\mI_{128}.
\end{equation}
We replace the selected training vectors by
\begin{equation}
    \mX_{\mathrm{orth}}=\sqrt p\,\mQ.
\end{equation}
Consequently,
\begin{equation}
    \frac1p
    \mX_{\mathrm{orth}}^\top\mX_{\mathrm{orth}}
    =
    \mI_{128}
\end{equation}
up to floating-point precision.

We consider the nested subset sizes
\begin{equation}
    m\in\{128,32,8\},
\end{equation}
where, for each $m$, the training matrix $\mX_m$ consists of the first $m$
columns of $\mX_{\mathrm{orth}}$. Thus
\begin{equation}
    \frac1p\mX_m^\top\mX_m=\mI_m
\end{equation}
for every experiment. Only the small training subset is orthogonalized. The
held-out test images are merely globally whitened and are otherwise left
unchanged.

We take
\begin{equation}
    p=n=1024,
    \qquad
    \psi=\frac pn=1,
    \qquad
    \rho=n\sigma^2=1.
\end{equation}
The same Gaussian initialization is used for all three values of $m$,
\begin{equation}
    U_{\alpha i}(0)
    \stackrel{\mathrm{iid}}{\sim}
    \mathcal N\!\left(0,\frac{\rho}{n}\right),
\end{equation}
with initialization seed $1$. Sharing the initialization isolates the effect
of changing the amount of training data.

\paragraph{Exact finite-width dynamics.}
For each $m$, we train on the orthogonalized matrix $\mX_m$ using the small-data
learning-rate scaling
\begin{equation}
    \eta=m.
\end{equation}
Rather than forming the $p\times p$ empirical covariance explicitly, we
evaluate the original weight dynamics through matrix products with $\mX_m$.
Writing
\begin{equation}
    \mE_X
    =
    \mX_m-\mU^\top\mU\mX_m,
\end{equation}
the gradient flow is evaluated as
\begin{equation}
    \dot{\mU}
    =
    \frac1p
    \left[
        (\mU\mX_m)\mE_X^\top
        +
        (\mU\mE_X)\mX_m^\top
    \right].
\end{equation}
This is an algebraic reassociation of the original tied-autoencoder gradient
flow and introduces no dynamical approximation.

We integrate this equation by fourth-order Runge--Kutta with
\begin{equation}
    \Delta t=5\times10^{-3},
    \qquad
    0\le t\le1,
\end{equation}
and record the trajectory every two integration steps.

The training loss is evaluated exactly as
\begin{equation}
    \widehat L_m(t)
    =
    \frac1{2pm}
    \left\|
        \mX_m-\mU(t)^\top\mU(t)\mX_m
    \right\|_F^2.
\end{equation}
For the globally whitened test covariance $\mT_{\mathrm{te}}$, we compute the
held-out reconstruction loss
\begin{equation}
    L_{\mathrm{te}}(t)
    =
    \frac1{2p}
    \Tr\!\left[
        \mT_{\mathrm{te}}
        \bigl(\mI_p-\mU(t)^\top\mU(t)\bigr)^2
    \right].
\end{equation}
This expression is algebraically identical to averaging the reconstruction loss
over all $10\,000$ globally whitened test images. In the numerical
implementation it is evaluated through $\mT_{\mathrm{te}}$, avoiding repeated
passes through the full test set.

\paragraph{Small-data theoretical prediction.}
For each $m$ we solve the limiting hierarchy of \cref{app:small-data},
\begin{equation}
    \dot a
    =
    4a(1-a)-2a\mu_1,
\end{equation}
and
\begin{equation}
    \dot\mu_k
    =
    -2\mu_{k+1}
    -2a\sum_{j=0}^{k-1}\mu_j\mu_{k-j}
    +2\mu_1\mu_k,
    \qquad
    k\ge1,
    \qquad
    \mu_0=1.
\end{equation}
To account for the finite dimension of the inactive subspace, the initial
moments use
\begin{equation}
    \psi_\perp=\frac{p-m}{n}
\end{equation}
instead of its limiting value $\psi$. Explicitly,
\begin{equation}
    \mu_k(0)
    =
    \rho^k
    \sum_{j=1}^k
    \mathrm{Nar}(k,j)\psi_\perp^j,
\end{equation}
where $\mathrm{Nar}(k,j)$ are the Narayana numbers.

We truncate the hierarchy at
\begin{equation}
    K=32,
    \qquad
    \mu_{K+1}=0,
\end{equation}
and integrate it with DOP853 using relative and absolute tolerances
$10^{-9}$ and $10^{-11}$, respectively.

For the training loss we use the finite-$m$ correction
\begin{equation}
    \widehat L_m^{\mathrm{th}}(t)
    =
    \frac12(1-a(t))^2
    +
    \frac12 a(t)\mu_1(t)
    +
    \frac{\rho^2(m+1)}{2n}.
\end{equation}

For the population prediction, define
\begin{equation}
    J(t)
    =
    2\int_0^t
    a(s)\bigl(\mu_1(s)-\mu_2(s)\bigr)\,ds
\end{equation}
and
\begin{equation}
    \epsilon=\frac mp.
\end{equation}
The finite-size isotropic prediction used in the experiment is
\begin{equation}
    L_{\mathrm{iso}}^{\mathrm{th}}(t)
    =
    L_0
    +
    \epsilon
    \left[
        \frac12(1-a(t))^2
        +a(t)\mu_1(t)
        -C_0
        +J(t)
    \right],
\end{equation}
where
\begin{equation}
    L_0
    =
    \frac12
    \left[
        1-2\rho
        +
        \rho^2
        \left(
            1+\frac{p+1}{n}
        \right)
    \right]
\end{equation}
and
\begin{equation}
    C_0
    =
    \frac12
    \left[
        1-2\rho+(1+2\psi)\rho^2
    \right].
\end{equation}

\cref{app:small-data} derives this expression for an isotropic population
covariance, whereas the finite globally whitened CIFAR-10 test covariance
$\mT_{\mathrm{te}}$ is not exactly scalar. We therefore define
\begin{equation}
    c_{\mathrm{te}}
    =
    \frac1p\Tr\mT_{\mathrm{te}}
\end{equation}
and use
\begin{equation}
    L_{\mathrm{te}}^{\mathrm{th}}(t)
    =
    c_{\mathrm{te}}
    L_{\mathrm{iso}}^{\mathrm{th}}(t)
\end{equation}
as the theoretical comparison curve. This formula would be exact if
\begin{equation}
    \mT_{\mathrm{te}}
    =
    c_{\mathrm{te}}\mI_p.
\end{equation}

We additionally report
\begin{equation}
    \lambda_{\min}(\mT_{\mathrm{te}}),
    \qquad
    \lambda_{\max}(\mT_{\mathrm{te}}),
    \qquad
    \frac{
        \|\mT_{\mathrm{te}}-c_{\mathrm{te}}\mI_p\|_{\mathrm{op}}
    }{
        c_{\mathrm{te}}
    }
\end{equation}
as diagnostics of the residual test-set anisotropy. Thus the training
comparison directly tests the small-data theorem, while the held-out population
comparison should be interpreted as testing its isotropic prediction after
global whitening, with the remaining discrepancy also reflecting finite
test-covariance anisotropy.

The resulting figures show, separately, the training and held-out population
losses for
\begin{equation}
    m\in\{128,32,8\},
\end{equation}
with solid curves denoting the exact finite-width dynamics and dashed curves
denoting the corresponding small-data predictions.\end{document}